\pdfoutput=1
\documentclass{article}

\usepackage[final]{neurips_2022}

\usepackage{times}
\usepackage{epsfig}
\usepackage{graphicx}
\usepackage{float}
\usepackage{wrapfig}
\usepackage{amsmath,amssymb,amsthm}
\usepackage{algorithm,algorithmicx,algpseudocode}
\usepackage{bm,xspace}
\usepackage{comment}
\usepackage{verbatim}
\usepackage{multirow}
\usepackage{balance}
\usepackage{url}
\usepackage{booktabs}
\usepackage{etoolbox,siunitx}
\usepackage{calc}
\usepackage{pifont,hologo}
\usepackage{color}
\usepackage{ulem}

\newcolumntype{P}[1]{>{\centering\arraybackslash}p{#1}}
\newcolumntype{M}[1]{>{\centering\arraybackslash}m{#1}}

\usepackage[super]{nth}
\usepackage{nicefrac}
\robustify\bfseries

\DeclareMathSymbol{@}{\mathord}{letters}{"3B}

\def\latex/{\LaTeX}
\def\bibtex/{\hologo{BibTeX}}

\makeatletter
\DeclareRobustCommand\onedot{\futurelet\@let@token\@onedot}
\def\@onedot{\ifx\@let@token.\else.\null\fi\xspace}

\usepackage[table]{xcolor}
\definecolor{blue}{HTML}{0055cc}
\definecolor{red}{HTML}{cc1100}
\definecolor{orange}{HTML}{cc7700}
\usepackage{float}
\usepackage{morefloats}
\usepackage[utf8]{inputenc} % allow utf-8 input
\usepackage[T1]{fontenc}    % use 8-bit T1 fonts
\usepackage[pagebackref,breaklinks,colorlinks,linkcolor=red,citecolor=blue]{hyperref}       % hyperlinks
\usepackage{url}            % simple URL typesetting
\usepackage{booktabs}       % professional-quality tables
\usepackage{amsfonts}       % blackboard math symbols
\usepackage{nicefrac}       % compact symbols for 1/2, etc.
\usepackage{microtype}      % microtypography
\usepackage{xcolor}         % colors
\usepackage{amsmath}        % ams math
\usepackage{amssymb}        % ams symbol
\usepackage{lipsum}         % dummy sentences
\usepackage{bbding}         % checkmark
\usepackage{array}          % table column width
\usepackage{subcaption}     % subtable
\usepackage{cleveref}
\crefname{figure}{Fig.}{Figs.}
\crefname{section}{Sec.}{Secs.}
\usepackage{ulem}
\usepackage{CJKutf8}

\usepackage{caption}
\title{Gemini \includegraphics[scale=0.08]{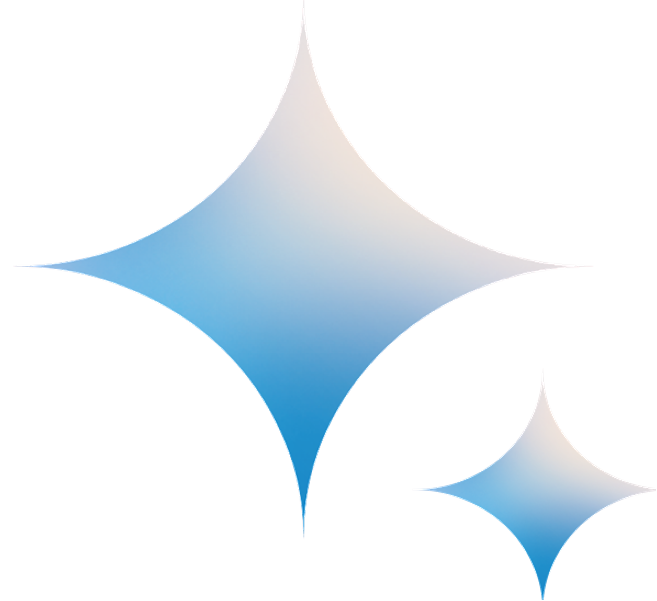} vs GPT-4V \includegraphics[scale=0.08]{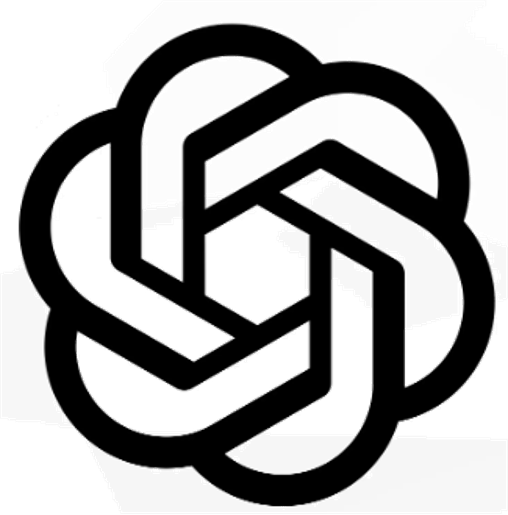}: A Preliminary Comparison and Combination of Vision-Language Models Through Qualitative Cases}

\author{
Zhangyang Qi\textsuperscript{1, 7}\footnotemark[1] \quad 
Ye Fang\textsuperscript{2, 7}\footnotemark[1] \quad 
Mengchen Zhang\textsuperscript{3, 7}\footnotemark[1] \quad
Zeyi Sun\textsuperscript{4, 7}\footnotemark[1] \quad  \\
\textbf{Tong Wu}\textsuperscript{5, 7} \quad 
\textbf{Ziwei Liu}\textsuperscript{6, 7} \quad
\textbf{Dahua Lin}\textsuperscript{5, 7} \quad 
\textbf{Jiaqi Wang}\textsuperscript{7}\footnotemark[2] \quad 
\textbf{Hengshuang Zhao}\textsuperscript{1}\footnotemark[2] \vspace{2mm} \\
\normalsize{$^*$ Equal contribution}\quad  $\dagger$ Corresponding author \vspace{2mm}\\
\textsuperscript{1}The University of Hong Kong \quad 
\textsuperscript{2}Fudan University \quad 
\textsuperscript{3}Zhejiang University \quad \\
\textsuperscript{4}Shanghai Jiao Tong University \quad
\textsuperscript{5}The Chinese University of Hong Kong \quad \\
\textsuperscript{6}Nanyang Technological University \quad
\textsuperscript{7}Shanghai AI Laboratory \vspace{1mm} \\
{\tt\small \{zyqi, hszhao\}@cs.hku.hk, wangjiaqi@pjlab.org.cn} \\
\small \url{https://github.com/Qi-Zhangyang/Gemini-vs-GPT4V}
}

\begin{document}
\begin{CJK}{UTF8}{gbsn} % Begin chinese
\maketitle

%%%%%%%%% ABSTRACT %%%%%%%%%
\hypersetup{
    citecolor=green
}
\begin{abstract}
  The rapidly evolving sector of Multi-modal Large Language Models (MLLMs) is at the forefront of integrating linguistic and visual processing in artificial intelligence. This paper presents an in-depth comparative study of two pioneering models: Google's Gemini and OpenAI's GPT-4V(ision). Our study involves a multi-faceted evaluation of both models across key dimensions such as Vision-Language Capability, Interaction with Humans, Temporal Understanding, and assessments in both Intelligence and Emotional Quotients.
The core of our analysis delves into the distinct visual comprehension abilities of each model. We conducted a series of structured experiments to evaluate their performance in various industrial application scenarios, offering a comprehensive perspective on their practical utility. We not only involve direct performance comparisons but also include adjustments in prompts and scenarios to ensure a balanced and fair analysis.
Our findings illuminate the unique strengths and niches of both models. GPT-4V distinguishes itself with its precision and succinctness in responses, while Gemini excels in providing detailed, expansive answers accompanied by relevant imagery and links.
These understandings not only shed light on the comparative merits of Gemini and GPT-4V but also underscores the evolving landscape of multimodal foundation models, paving the way for future advancements in this area.
After the comparision, we attempted to achieve better results by combining the two models.
Finally, We would like to express our profound gratitude to the teams behind GPT-4V~\cite{openai2023gpt4} and Gemini~\cite{geminiteam2023gemini} for their pioneering contributions to the field. Our acknowledgments are also extended to the comprehensive qualitative analysis presented in `Dawn' by Yang \textit{et al.}~\cite{yang2023dawn}. This work, with its extensive collection of image samples, prompts, and GPT-4V-related results, provided a foundational basis for our analysis.
\end{abstract}

\newpage

\hypersetup{
    linkcolor=black,
}
\tableofcontents
\listoffigures

\hypersetup{
    linkcolor=red,
    citecolor=green
}

%%%%%%%%% BODY %%%%%%%%%
\newpage
\section{Introduction}
\label{Sec.1 Introduction}
\subsection{Motivation and Overview}
The evolution of artificial intelligence has seen the significant rise of Large Language Models (LLMs)~\cite{brown2020language, GPT4, Palm, PaLM2, LLaMA, Training}, which have revolutionized the way machines process and understand textual data. Building upon this, the advent of Multi-modal Large Language Models (MLLMs)~\cite{mllmsurvey, huang2023language, Palm-e, Multimodal-gpt, OpenFlamingo, Minigpt-4, LLaVA, InstructBLIP, mPLUG-Owl, Qwen-VL,zhang2023internlmxcomposer,chen2023sharegpt4v} marks a pivotal advancement in AI, extending capabilities to comprehend and interact with not just text, but also images, 3D models~\cite{GPT4Point}, and video content~\cite{VideoLLM}. Among these modalities, the integration of text and image has emerged as particularly powerful, largely due to the rich and informative nature of text-image pairs. The following, unless otherwise specified, all refer to the MLLMs in the context of images.

The landscape of Multi-modal Large Language Models (MLLMs) is currently divided into two broad categories: closed-source models with their proprietary advancements, and open-source deployable models like LLaVA~\cite{LLaVA}, MiniGPT-4~\cite{Minigpt-4}and InstructBLIP~\cite{InstructBLIP} which are more accessible but often less advanced. Among these, the state-of-the-art in open-source models is GPT-4V~\cite{GPT-4V} from OpenAI, which has established a dominant position in terms of versatility and general applicability. Recently, Google has introduced their own large model, Gemini~\cite{Gemini}, which also boasts high generalization capabilities. This release poses a significant challenge to GPT-4V's leading status. Gemini's entry into the arena of MLLMs brings a new dimension to the field, potentially reshaping the landscape of what is achievable with open-source AI technology, particularly in terms of multimodal understanding and application. Therefore, our paper undertakes a comprehensive comparison of these two models across multiple dimensions and domains. 

It should be noted that the image samples, prompts, and results related to GPT-4V used in our paper are referenced from the study "The Dawn of LMMs: Preliminary Explorations with GPT-4V(ision).~\cite{dawn}" Our work can be seen as a continuation and expansion of this previous research. Since GPT-4V has already been extensively discussed in the original paper, our focus in this report will primarily be on emphasizing and exploring the unique characteristics and capabilities of Gemini.

\Cref{Sec.2 Image Recognition and Understanding} to \cref{Sec.6 Integrated Image and Text Understanding} divide the multimodal evaluation into five aspects. The first level involves basic recognition of images and the text within them. The second level goes beyond recognition to require further inference and reasoning. The third level encompasses multimodal comprehension and inference involving multiple images. We have divided them into the following five sections.

\begin{itemize}
\item \textbf{Image Recognition and Understanding}: \cref{Sec.2 Image Recognition and Understanding} addresses the fundamental recognition and comprehension of image content without involving further inference, including tasks such as identifying landmarks, foods, logos, abstract images, autonomous driving scenes, misinformation detection, spotting differences, and object counting.
\item \textbf{Text Recognition and Understanding in Images}: \cref{Sec.3 Text Recognition and Understanding in Images} concentrates on text recognition (including OCR) within images, such as scene text, mathematical formulas, and chart \& table text recognition. Similarly, no further inference of text content is performed here.
\item \textbf{Image Inference Abilities}: Beyond basic image recognition, \cref{Sec.4 Image Reasoning Abilities} involves more advanced reasoning. This includes understanding humor and scientific concepts, as well as logical reasoning abilities like detective work, image combinations, look for patterns in intelligence tests (IQ Tests), and emotional understanding and expression (EQ Tests).
\item \textbf{Textual Inference in Images}: Building on the text recognition, \cref{Sec.5 Textual Reasoning in Images} involves further reasoning beyond text recognition, including mathematical problem-solving, chart \& table information reasoning, and document comprehension like paper, report and Graphic Design.
\item \textbf{Integrated Image and Text Understanding}: \cref{Sec.6 Integrated Image and Text Understanding} evaluates the collective understanding and reasoning abilities involving both image and text. For instance, tasks include settling items from a supermarket shopping cart, as well as guiding and modifying image generation.
\end{itemize}

\Cref{Sec.7 Object Localization} to \cref{Sec.9 Multilingual Capabilities} evaluate performance in three specialized tasks, namely, object localization, temporal understanding, and multilingual comprehension.

\begin{itemize}
\item \textbf{Object Localization}: \cref{Sec.7 Object Localization} highlights object localization capabilities, tasking the models with providing relative coordinates for specified objects. This includes a focus on outdoor objects like cars in parking lots and abstract image localization.
\item \textbf{Temporal Video Understanding}: \\cref{Sec.8 Temporal Video Understanding} evaluates the models' comprehension of temporality using key frames. This section includes two tasks: one involving the understanding of video sequences and the other focusing on sorting key frames.
\item \textbf{Multilingual Capabilities}: \cref{Sec.9 Multilingual Capabilities} thoroughly assesses capabilities in recognizing, understanding, and producing content in multiple languages. This includes the ability to recognize non-English content within images and express information in other languages.
\end{itemize}

\cref{Sec.10 Industry Application} presents various application scenarios for multimodal large models. We aim to showcase more possibilities to the industry, providing innovative ideas. There is potential to customize multimodal large models for unique domains. Here, we demonstrate seven sub-domains:

\begin{itemize}
\item \textbf{Industry: Defect Detection}: This task involves the detection of defects in products on industrial assembly lines, including textiles, metal components, pharmaceuticals and more.
\item \textbf{Industry: Grocery Checkout}: This refers to an autonomous checkout system in supermarkets, aimed at identifying all items in a shopping cart for billing. The goal is to achieve comprehensive recognition of all items within the shopping cart.
\item \textbf{Industry: Auto Insurance}: This task involves evaluating the extent of damage in car accidents and providing approximate repair costs, as well as offering repair recommendations.
\item \textbf{Industry: Customized Captioner}: The aim is to identify the relative positions of various objects within a scene, with object names provided as condition and prompts in advance.
\item \textbf{Industry: Evaluation Image Generation}: This involves assessing the alignment between generated images and given text prompts, evaluating the quality of the generation model.
\item \textbf{Industry: Embodied Agent}: This application involves deploying the model in embodied intelligence and smart home systems, offering thoughts and decisions for indoor scenarios.
\item \textbf{Industry: GUI Navigation}: This task focuses on guiding users through PC/Mobile GUI interfaces, assisting with information reception, online searches, and shopping tasks.
\end{itemize}

Finally, in \cref{Sec.11 Integrated Use of GPT-4V and Gemini}, we explore how to combine both SOTA models to leverage their respective strengths and mitigate their weaknesses. In summary, GPT-4V provides more accurate results, while Gemini excels in providing more detailed responses, along with image and link outputs.

\subsection{Gemini's Input Modes}
Our goal is to clarify the input modality of Gemini. GPT-4V's input modality supports the continuous ingestion of multiple images as context, thereby possessing enhanced memory capabilities. However, for Gemini, its unique attributes are manifested in several aspects, as follows:

\begin{itemize}
\item \textbf{Single Image Input}: Gemini is limited to inputting a single image at a time. Additionally, it cannot process independent images; instead, it requires accompanying textual instructions.
\item \textbf{Limited Memory Capacity}: Unlike GPT-4V, Gemini's multimodal module lacks the ability to retain memory of past image inputs and outputs. Therefore, when dealing with multiple images, our approach requires combining all the images into a single image input. This integrated input mode will be used unless explicitly stated otherwise.
\item \textbf{Sensitive Information Masking}: Gemini exhibits some degree of obfuscation when processing images containing explicit facial or medical information, making it unable to recognize these images. This may impose certain limitations on its generalization ability.
\item \textbf{Image and Link Output}: Unlike GPT-4V, which is limited to generating textual outputs, Gemini has the ability to create images related to the content and provide corresponding links. This establishes a higher level of association similar to search engine functionality.
\item \textbf{Video Input and Comprehension}: Gemini demonstrates the capability to understand videos and requires a YouTube link as a video input. It's important to note that it can effectively process videos accompanied by accurate subtitle files. However, its comprehension ability may be limited when dealing with single, simple, and information-scarce videos.
\end{itemize}

\subsection{Prompt Techniques}
Prompt Engineering holds significant importance for both unimodal language models~\cite{ouyang2022training, multitask-prompted-training, mishra2022cross, brown2020language, wei2022finetuned, tsimpoukelli2021multimodal}and multimodal large-scale models~\cite{alayrac2022flamingo, chen2023llava-interactive, InstructBLIP}. The prompt design under consideration is tailored for GPT-4V, and direct input into Gemini may yield unsatisfactory responses. In such cases, adjustments to Gemini's prompt are made to align with the input requirements of its architecture.

\subsection{Sample Collection}
All our data is sourced from "The Dawn of LMMs: Preliminary Explorations with GPT-4V(ision)"~\cite{dawn} (except for the images in Section 11, which are sourced from the internet). We have utilized their images, GPT-4V's prompts, and corresponding results. Our work can be seen as a continuation of theirs. Our dataset is diverse and maintains privacy protections. We extend our gratitude to the authors of that work. The raw data of the images is available on the \href{https://github.com/Qi-Zhangyang/Gemini-vs-GPT4V}{project page}.

\subsection{Takeaways (Conclusion)}
We have conducted a comprehensive comparison of GPT-4V and Gemini's multimodal understanding and reasoning abilities across multiple aspects and have reached the following conclusions:

\begin{itemize}
\item \textbf{Image Recognition and Understanding}: In basic image recognition tasks, both models show comparable performance and are capable of completing the tasks effectively.
\item \textbf{Text Recognition and Understanding in Images}: Both models excel in extracting and recognizing text from images. However, improvements are needed in complex formula and dashboard recognition. Gemini performs better in reading table information.
\item \textbf{Image Inference Abilities}: In image reasoning, both models excel in common-sense understanding. Gemini slightly lags in look-for-pattern compared (IQ Tests) to GPT-4V. In EQ tests, both understand emotions and have aesthetic judgment.
\item \textbf{Textual Inference in Images}: In the field of text reasoning, Gemini shows relatively lower performance levels when dealing with complex table-based reasoning and mathematical problem-solving tasks. Furthermore, Gemini tends to offer more detailed outputs.
\item \textbf{Integrated Image and Text Understanding}: In tasks involving complex text and images, Gemini falls behind GPT-4V due to its inability to input multiple images at once, although it performs similarly to GPT-4V in textual reasoning with single images.
\item \textbf{Object Localization}: Both models perform similarly in real-world object localization, with Gemini being slightly less adept at abstract image (tangram) localization.
\item \textbf{Temporal Video Understanding}: 
In understanding temporality, Gemini's single-image input mode falls short compared to GPT-4V, especially in the comprehension of sequence.
\item \textbf{Multilingual Capabilities}: Both models exhibit good multilingual recognition, understanding, and output capabilities, effectively completing the multilingual tasks.
\end{itemize}

In industrial applications, Gemini is outperformed by GPT-4V in \textbf{Embodied Agent} and \textbf{GUI Navigation}, which is also attributed to Gemini's single-image, non-memory input mode. Combining two large models can leverage their respective strengths.

Overall, both Gemini and GPT-4V are powerful and impressive multimodal large models. In terms of overall performance, GPT-4V is slightly stronger than Gemini Pro. This aligns with the results reported by Gemini. We look forward to the release of Gemini Ultra and GPT-4.5, which are expected to bring more possibilities to the field of visual multimodal applications.

\newpage
\section{Image Recognition and Understanding}
\label{Sec.2 Image Recognition and Understanding}
In this section, we primarily discuss the fundamental understanding of images. This task is the most basic, requiring only the identification of objects in an image and their characteristics. It does not involve text-related tasks or further inference. \cref{Sec.2.1_Basic_object_Recognition}, \cref{Sec.2.2 Landmark Recognition},  \cref{Sec.2.3 Food Recognition} and \cref{Sec.2.4 Logo Recognition} focus on the recognition of basic objects, landmarks, food, logos, and abstract images. \cref{Sec.2.6 Scene Understanding} addresses scene understanding in outdoor autonomous driving scenarios. \cref{Sec.2.7 Counterfactual Examples} tests the model's ability to recognize fabricated objects created using text, gauging its discernment of real versus fictitious elements. \cref{Sec.2.8 Object Counting} assesses the model's object counting capabilities, while the final \cref{Sec.2.9 Spot the Difference} explores the model's proficiency in spotting differences, examining its ability to identify subtle details.

\subsection{Basic object Recognition}
\label{Sec.2.1_Basic_object_Recognition}
\begin{figure}[h]
    \centering
    \includegraphics[width=0.99\linewidth]{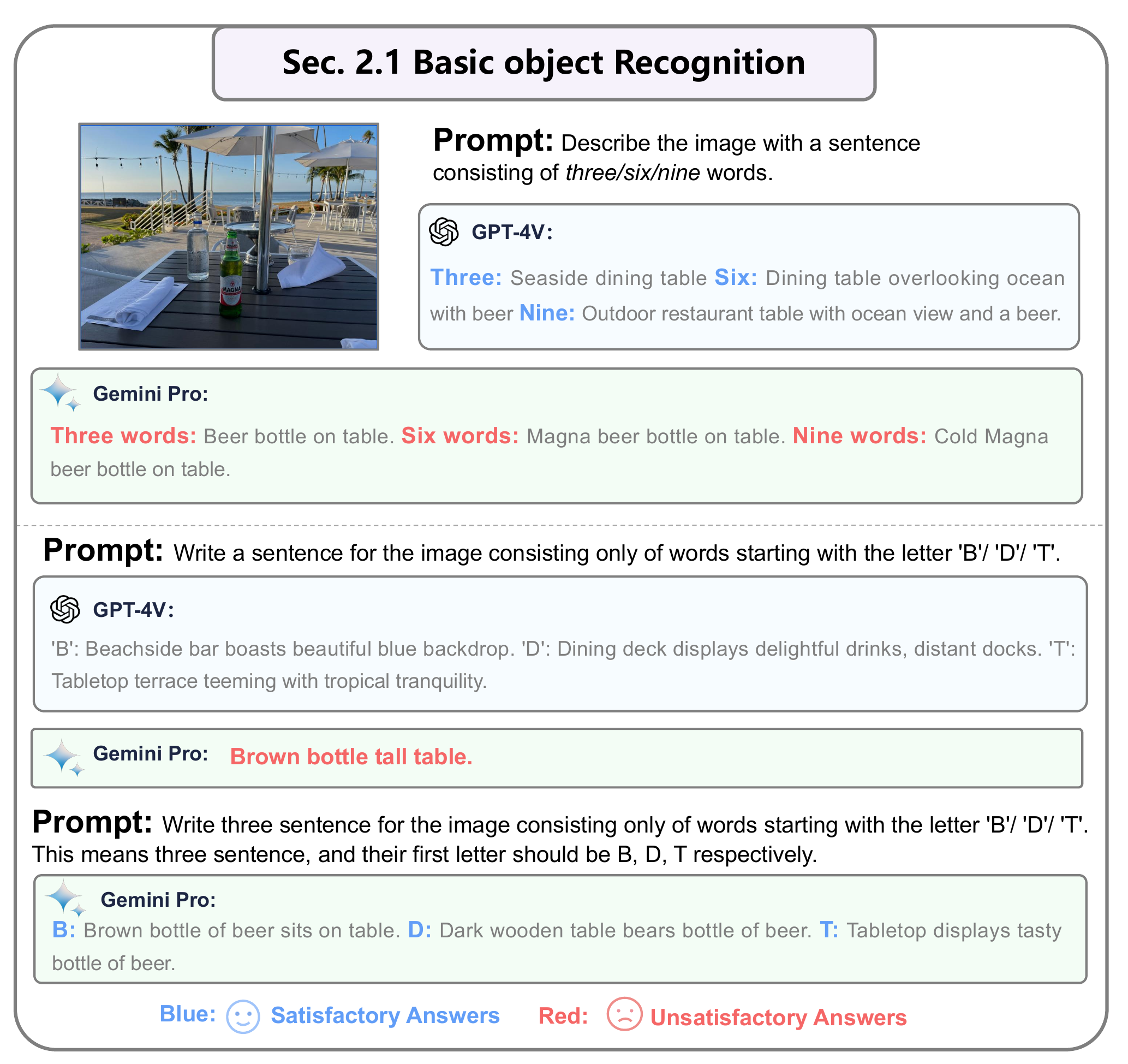}
    \vspace{-6pt}
    \caption[Section 2.1 Basic Object Recognition]{\textbf{Basic object recognition.} GPT-4V and Gemini are both capable of accurately recognizing basic objects and providing textual descriptions as required.}
    \label{fig:Sec.2.1_1}
    \vspace{-10pt}
\end{figure}

\cref{fig:Sec.2.1_1} refers to the recognition of the entire image and corresponding description, using a fixed number of words (three, six and nine words) or an overall description starting with fixed letters (B/D/T in this case). After adjusting the prompts, both GPT-4V and Gemini are able to provide satisfactory responses. indicating the ability to comprehend images and respond according to instructions.

\subsection{Landmark Recognition}
\label{Sec.2.2 Landmark Recognition}
\begin{figure}[H]
    \centering
    \includegraphics[width=0.95\linewidth]{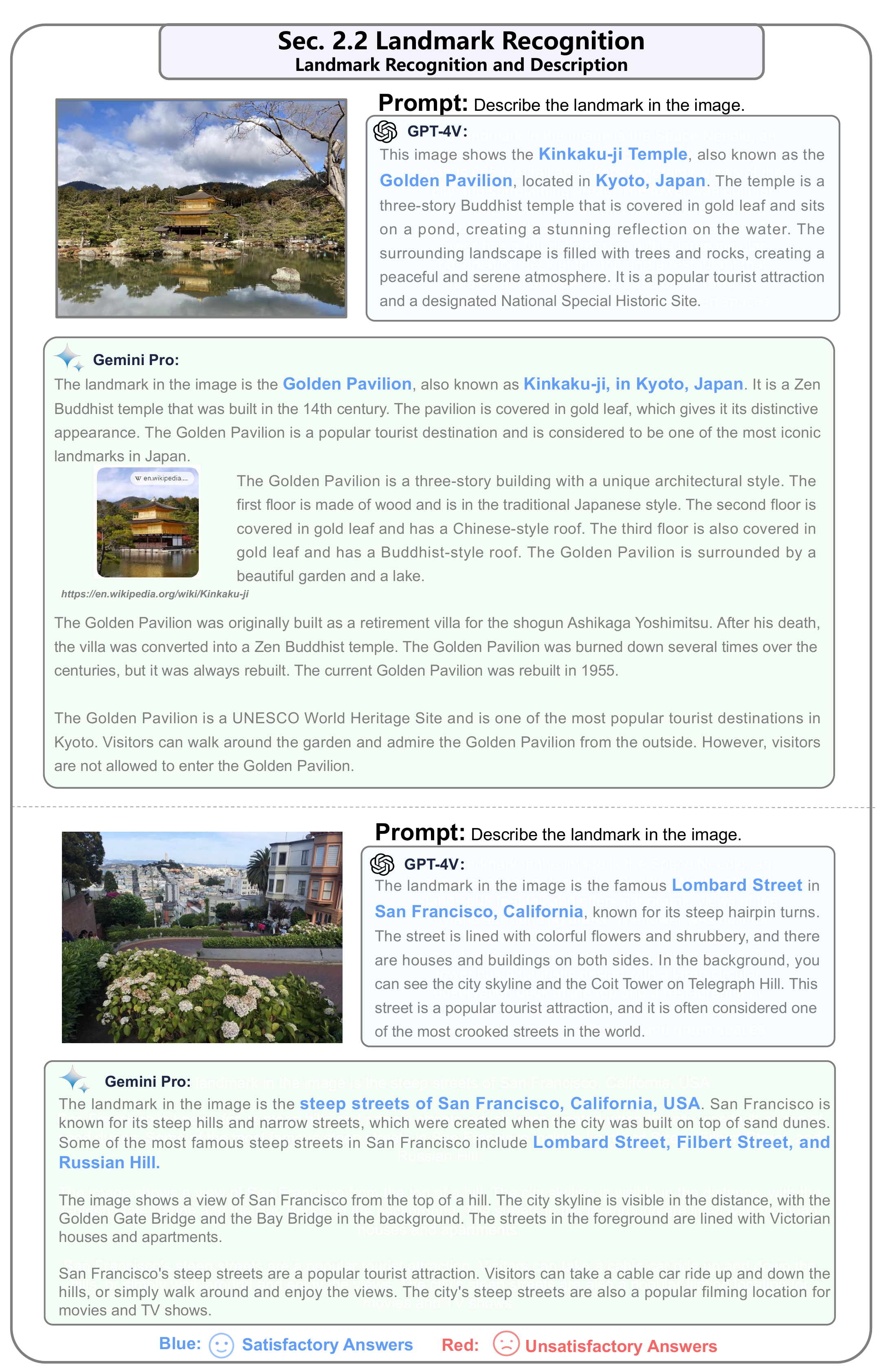}
    \vspace{-6pt}
    \caption[Section 2.2 Landmark Recognition (1)]{\textbf{Landmark recognition and description (1/2).} While both models are proficient in accurately identifying landmarks, GPT-4V generates concise descriptions, whereas Gemini tends to provide more detailed narratives even indoors at tourist attractions..}
    \label{fig:Sec.2.2_1}
    \vspace{-10pt}
\end{figure}

\begin{figure}[H]
    \centering
    \vspace{2cm}
    \includegraphics[width=0.99\linewidth]{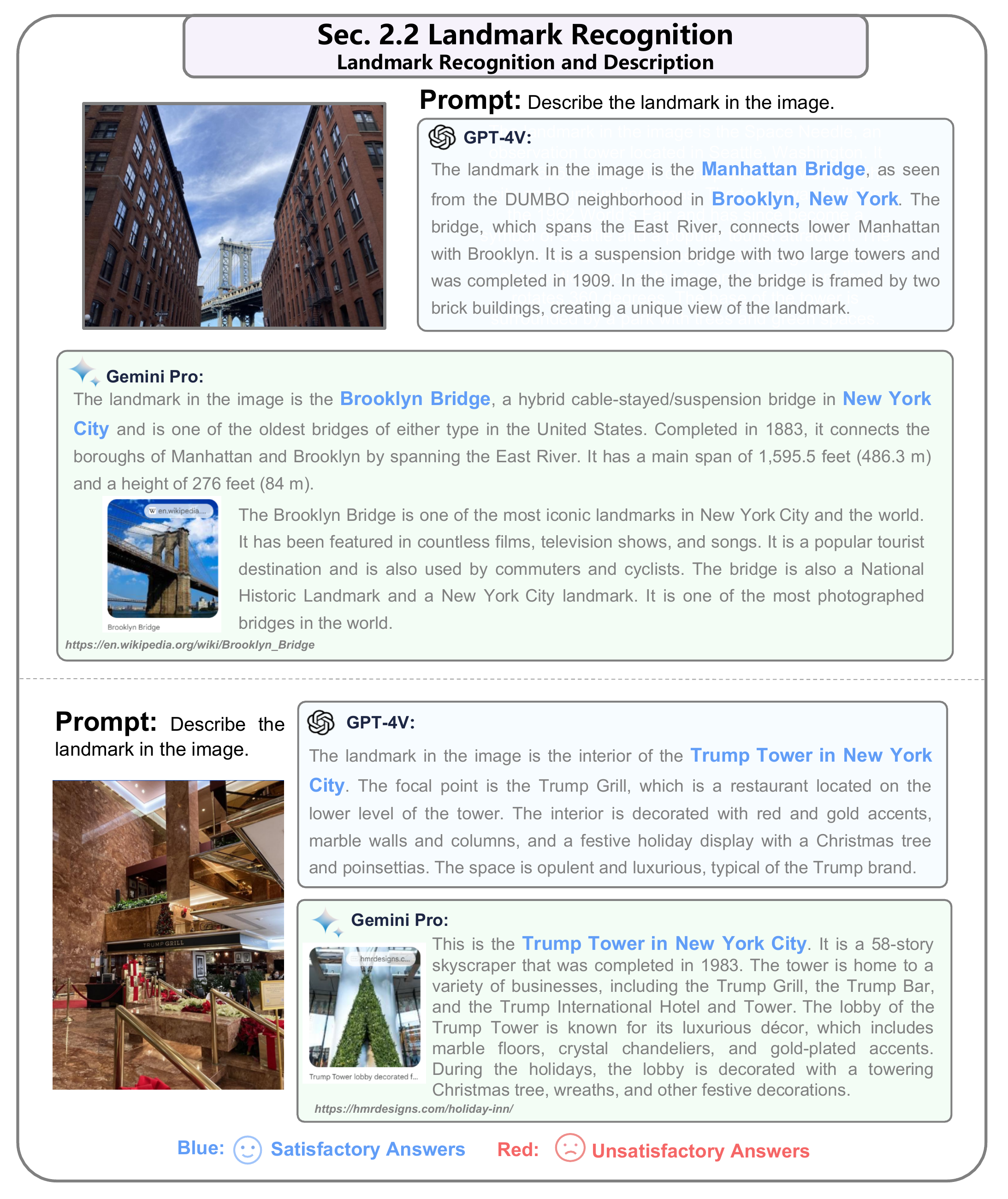}
    \vspace{-6pt}
    \caption[Section 2.2 Landmark Recognition (2)]{\textbf{Landmark recognition and description (2/2).} Both models excel at accurately identifying landmarks, producing vivid and detailed descriptions. Even for the interior of Trump Tower, both models are able to successfully identify it.}
    \label{fig:Sec.2.2_2}
    \vspace{-10pt}
\end{figure}

\cref{fig:Sec.2.2_1} and \cref{fig:Sec.2.2_2} together showcase four famous landmarks, namely Kinkaku-ji Temple, Lombard Street, Manhattan Bridge, and Trump Tower. Here, both GPT-4V and Gemini perform well, with Gemini being able to provide additional related introductions to the scenery. Even for the interior of Trump Tower, both models are able to successfully identify it. Gemini can displays other images and links related to the landmark.

\subsection{Food Recognition}
\label{Sec.2.3 Food Recognition}
\begin{figure}[H]
    \vspace{0.8cm}
    \centering
    \includegraphics[width=0.99\linewidth]{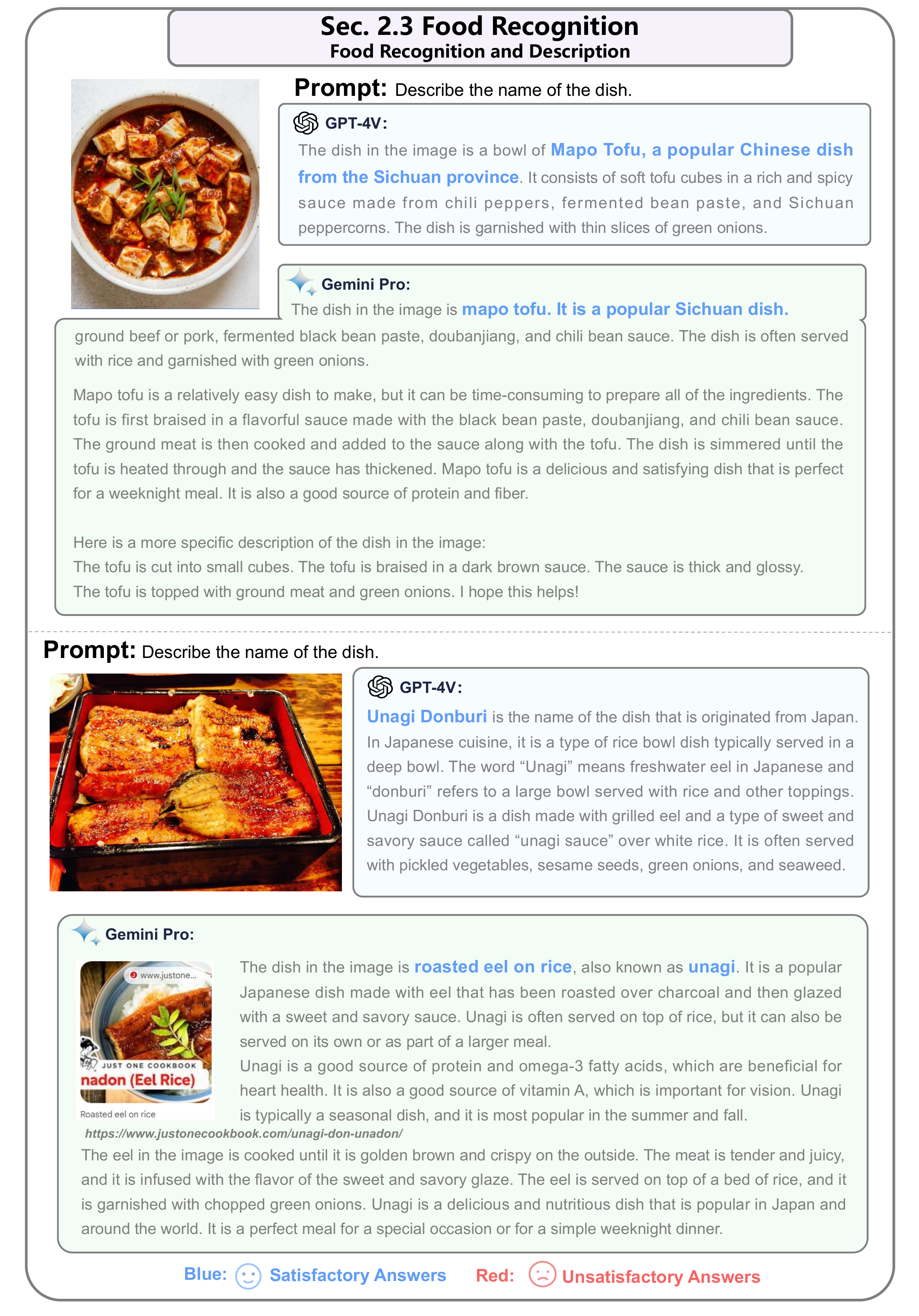}
    \vspace{-6pt}
    \caption[Section 2.3 Food Recognition (1)]{\textbf{Food recognition and description (1/2).} Both models exhibit the ability to recognize a broad spectrum of dishes, extending their identification abilities to minute details like ingredients, garnishes, and cooking techniques depicted within an image of a dish.}
    \label{fig:Sec.2.3_1}
    \vspace{-10pt}
\end{figure}

\begin{figure}[H]
    \centering
    \vspace{3cm}
    \includegraphics[width=0.99\linewidth]{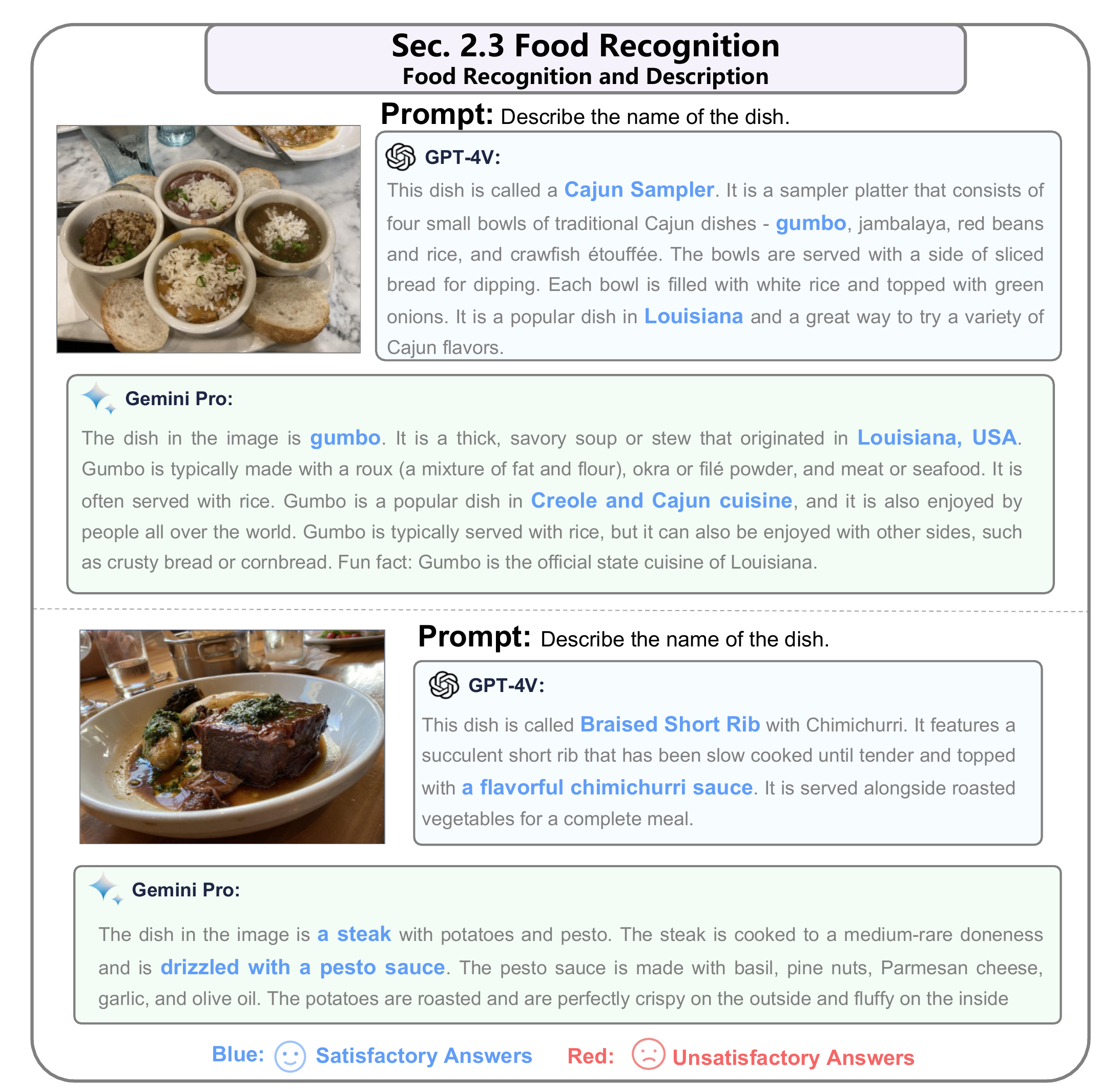}
    \vspace{-6pt}
    \caption[Section 2.3 Food Recognition (2)]{\textbf{Food recognition and description (2/2).} Both models exhibit remarkable capacities in identifying a vast array of culinary dishes, further extending their recognition abilities to specific elements like ingredients, garnishes, and cooking techniques represented within a dish's image.}
    \label{fig:Sec.2.3_2}
    \vspace{-10pt}
\end{figure}

\cref{fig:Sec.2.3_1} and \cref{fig:Sec.2.3_2} pertain to the identification of food, showcasing Chinese cuisine, Japanese cuisine, Western cuisine, and specialties from minority tribes in North America, demonstrating the MLLMs' knowledge range from multiple dimensions, where both models perform well. Similarly, Gemini tends to provide more detailed descriptions and links, such as links to recipes.

\subsection{Logo Recognition}
\label{Sec.2.4 Logo Recognition}
\begin{figure}[H]
    \vspace{1cm}
    \centering
    \includegraphics[width=0.99\linewidth]{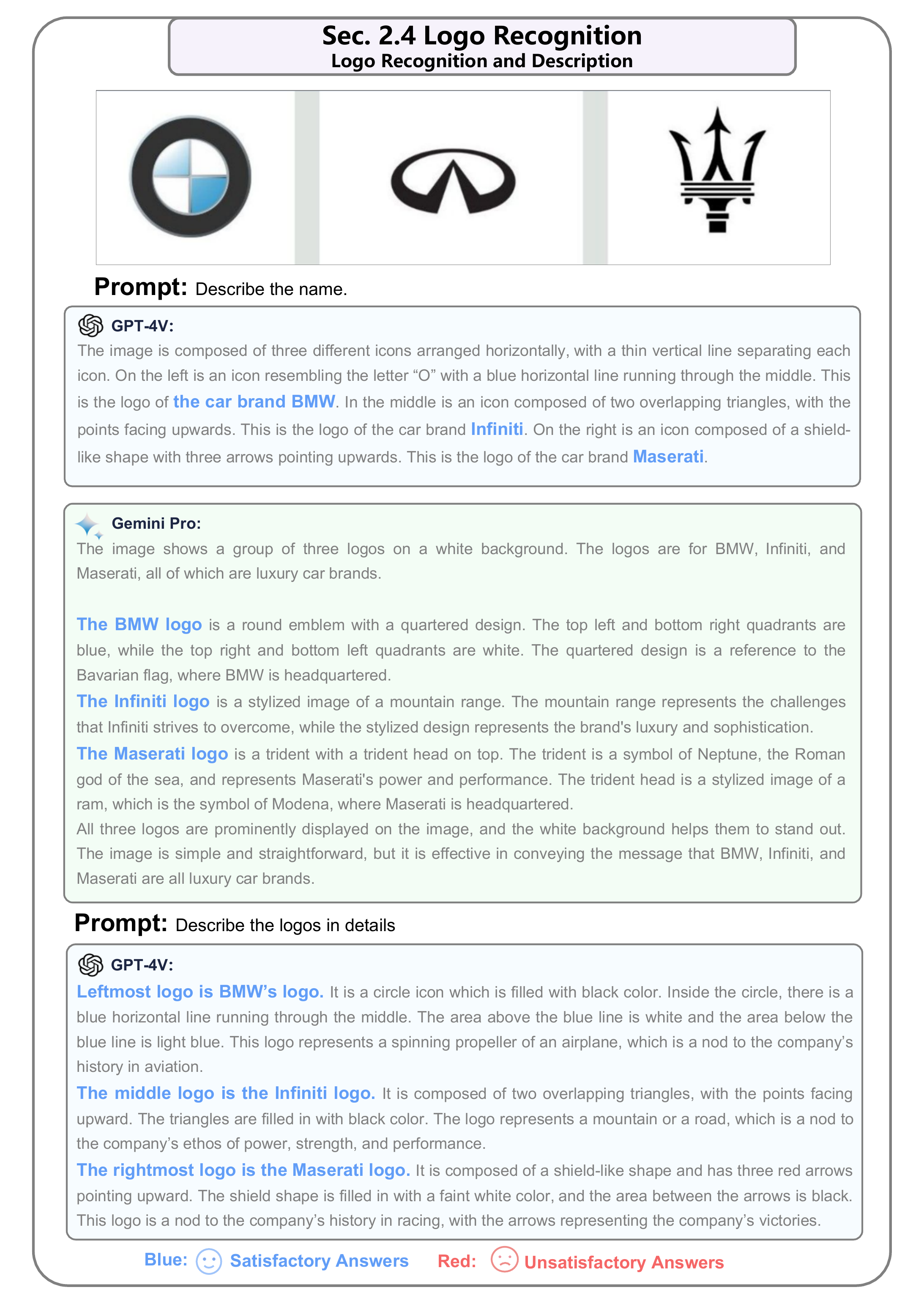}
    \vspace{-6pt}
    \caption[Section 2.4 Logo Recognition (1)]{\textbf{Logo recognition and description.} Both models successfully identify logos and provide comprehensive descriptions, capturing various aspects like design, color, shape, and symbolic representations.}
    \label{fig:Sec.2.4_1}
    \vspace{-10pt}
\end{figure}

\begin{figure}[H]
    \vspace{1cm}
    \centering
    \includegraphics[width=0.99\linewidth]{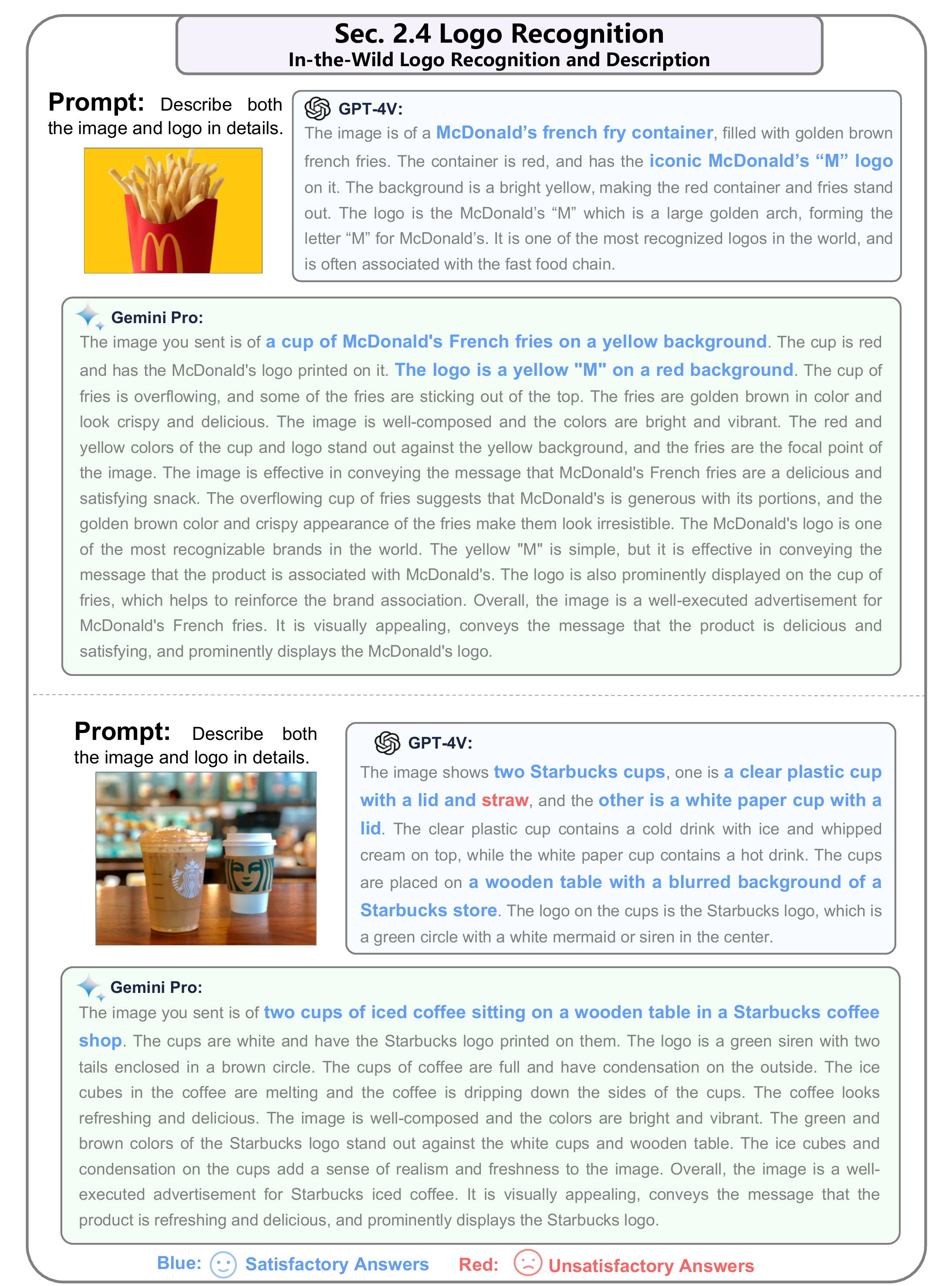}
    \vspace{-3pt}
    \caption[Section 2.4 Logo Recognition (2)]{\textbf{In-the-wild logo recognition and description (1/2).} Both models exhibit a robust capability of identifying logos in various scenarios, accounting for occlusions, lighting conditions, and orientations, while Gemini tends to provide more detailed descriptions. However, in the second case, GPT-4V's description shows minor instances of hallucination.}
    \label{fig:Sec.2.4_2}
    \vspace{-10pt}
\end{figure}

\begin{figure}[H]
    \centering
    \includegraphics[width=0.99\linewidth]{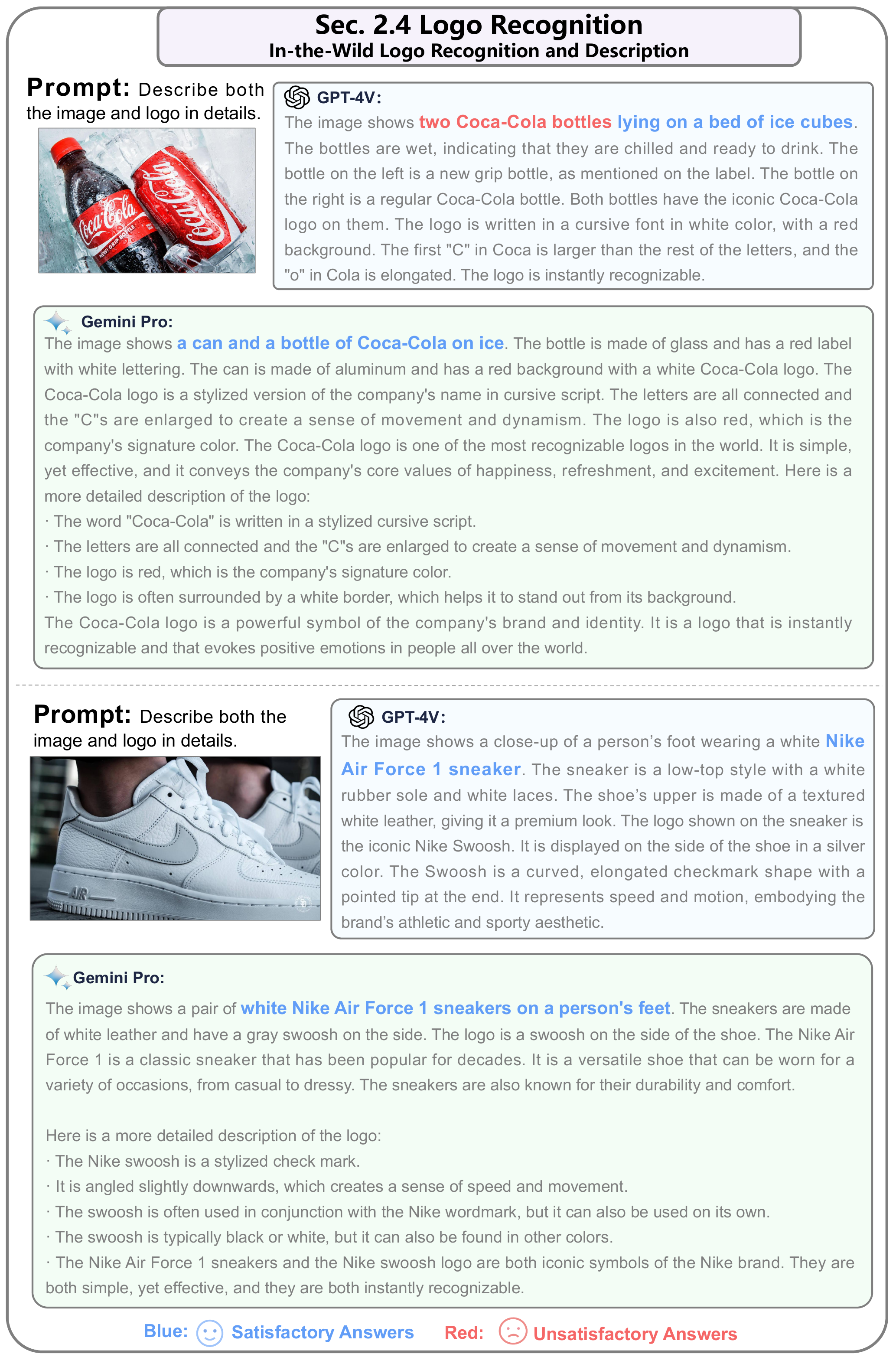}
    \vspace{-6pt}
    \caption[Section 2.4 Logo Recognition (3)]{\textbf{In-the-wild logo recognition and description (2/2).} Both models manifest robust capabilities in recognizing logos in diverse situations, including occlusions, different lighting conditions, and varied orientations, while Gemini tends to provide more detailed descriptions. GPT-4V's description in the first case exhibits a slight error.}
    \label{fig:Sec.2.4_3}
    \vspace{-10pt}
\end{figure}

\cref{fig:Sec.2.4_1}, \cref{fig:Sec.2.4_2} and \cref{fig:Sec.2.4_3} are about logo recognition, including the logo itself and recognition of logos in-the-wild scenarios. Both models generally do not make significant errors, with Gemini providing more detailed introductions. Here we can observe that in response to simple prompts, GPT-4V also tends to provide concise answers, only giving detailed responses when specifically requested to 'in detail'. Furthermore, in 'in-the-wild scenarios,' GPT-4V may excessively focus on objects and provide incorrect answers related to objects, such as mistaking a can for a bottle or inventing the presence of a straw in a coffee cup.

\subsection{Abstract Image Recognition}
\label{Sec.2.5 Abstract Image Recognition}
\begin{figure}[h]
    \vspace{2.5cm}
    \centering
    \includegraphics[width=0.99\linewidth]{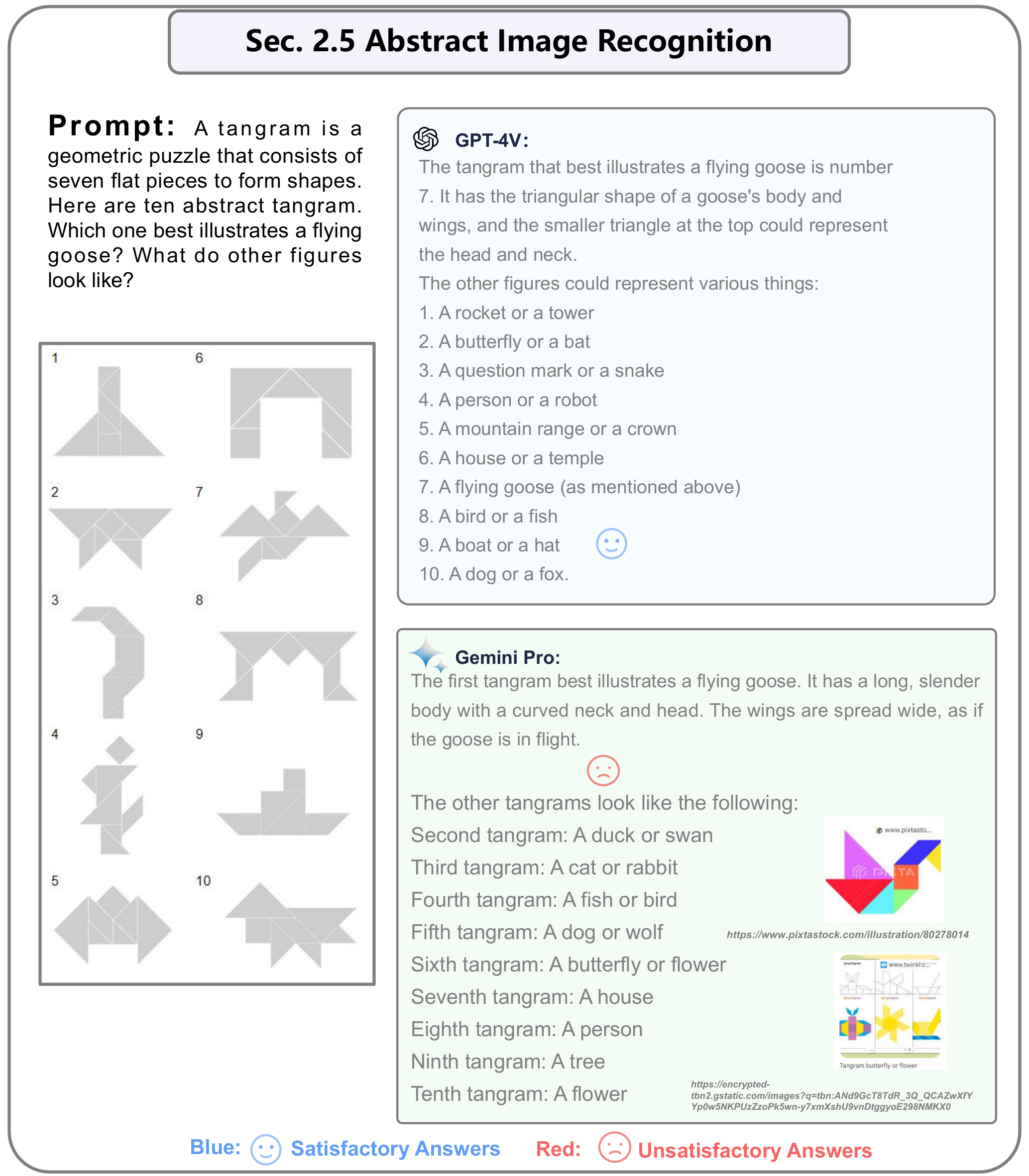}
    \vspace{-3pt}
    \caption[Section 2.5 Abstract Image Recognition]{\textbf{Abstract image recognition. }While GPT-4V provides reasonable associations, the associations given by Gemini do not appear as logical.}
    \label{fig:Sec.2.5_1}
    \vspace{-10pt}
\end{figure}

\cref{fig:Sec.2.5_1} is about the recognition of abstract images, specifically recognizing various shapes composed of tangram pieces. Overall, GPT-4V tends to provide more accurate responses, largely because Gemini struggles with recognizing large images composed of multiple smaller images. This indicates Gemini's limited ability to recognize more abstract objects. Secondly, it's possible that combining multiple images into a single input image may have resulted in a decrease in Gemini's performance.

\subsection{Scene Understanding}
\label{Sec.2.6 Scene Understanding}
\begin{figure}[H]
    \centering
    \includegraphics[width=0.91\linewidth]{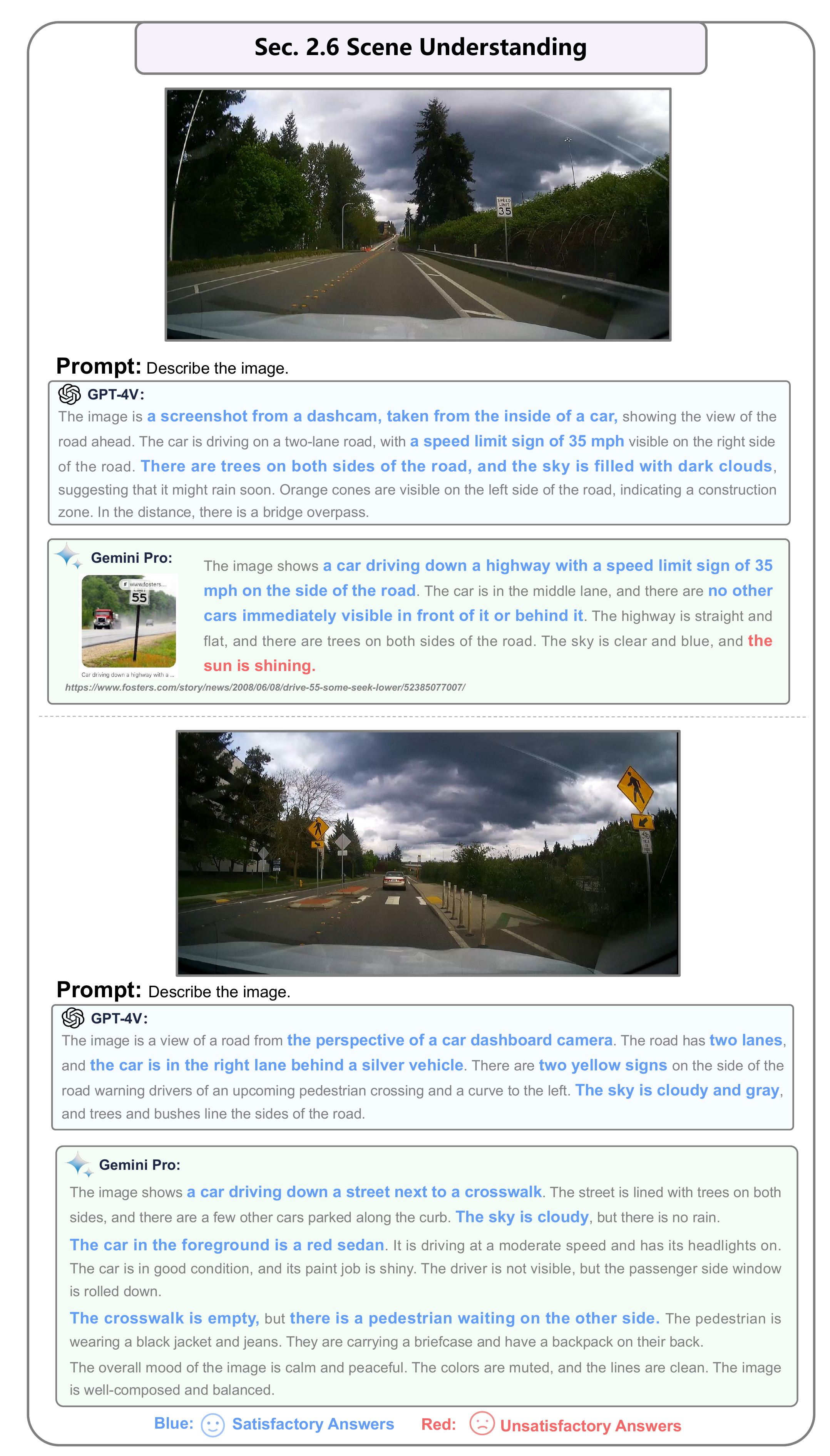}
    \vspace{-6pt}
    \caption[Section 2.6 Scene Understanding]{\textbf{Scene understanding.} Both models are proficient in providing detailed descriptions of the scenes and objects. However, in the first case, Gemini's description showcases a minor error.}
    \label{fig:Sec.2.6_1}
    \vspace{-10pt}
\end{figure}

\cref{fig:Sec.2.6_1} presents an outdoor autonomous driving scene. Cars driving on the road can see pedestrians, traffic signs, and other vehicles. Both models show good capabilities here. However, GPT-4V's responses are more accurate, while Gemini's responses are more detailed. Here, we observe that Gemini has some discrepancies in understanding weather conditions, but overall, the performance of both models is quite comparable.

\subsection{Counterfactual Examples}
\label{Sec.2.7 Counterfactual Examples}
\begin{figure}[H]
    \vspace{3cm}
    \centering
    \includegraphics[width=0.99\linewidth]{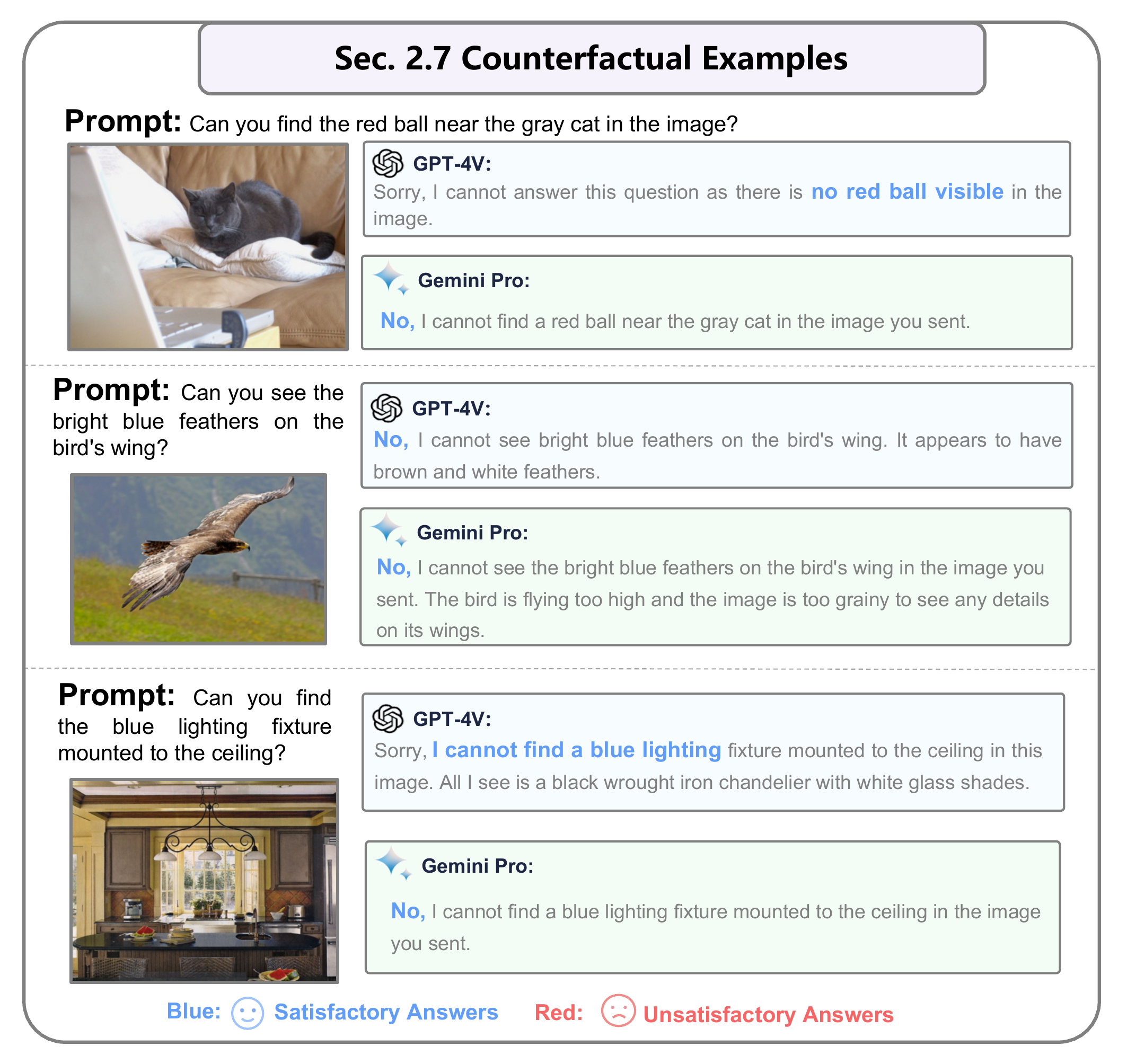}
    \vspace{-6pt}
    \caption[Section 2.7 Counterfactual Examples]{\textbf{Counterfactual judgement.} Both models are capable of providing factual descriptions related to the scenes and objects depicted in the images.}
    \label{fig:Sec.2.7_1}
    \vspace{-10pt}
\end{figure}

\cref{fig:Sec.2.7_1} shows an example where we present an image and fabricate an object that doesn't exist in the picture, then ask about the location of this fabricated object. This is to prevent the models from making purely imaginative connections, rather than truly understanding the image. The results show that both models can identify that the fabricated object does not exist.

\subsection{Object Counting}
\label{Sec.2.8 Object Counting}
\begin{figure}[H]
    \vspace{1.8cm}
    \centering
    \includegraphics[width=0.99\linewidth]{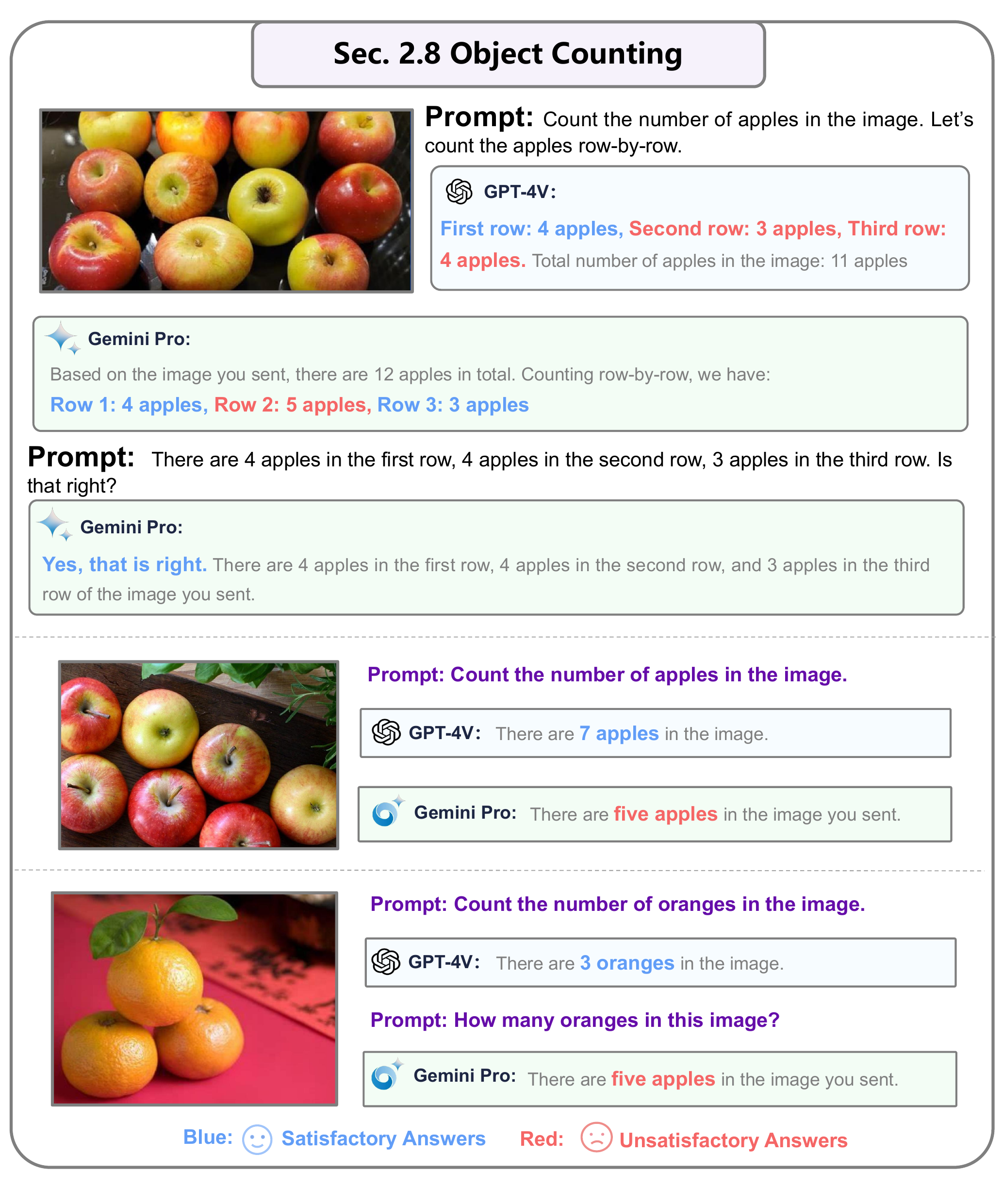}
    \vspace{-6pt}
    \caption[Section 2.8 Object Counting]{\textbf{Object Counting.} The task is to count the number of fruits in the pictures. Gemini's counting ability is relatively weaker compared to GPT-4V. Gemini, due to its relatively weaker multimodal memory performance, cannot be guided to the correct answers.}
    \label{fig:Sec.2.8_1}
    \vspace{-10pt}
\end{figure}

\cref{fig:Sec.2.8_1} demonstrates the models' counting abilities. Here, we present three examples, all of which involve counting the quantity of the same type of fruit in the image. In this context, we observe that Gemini's counting ability is not as strong as GPT-4V. While GPT-4V can arrive at the correct answers through memorization and guidance from prompts, Gemini, due to its relatively weaker multimodal memory performance, cannot be guided to the correct answers.

\subsection{Spot the Difference}
\label{Sec.2.9 Spot the Difference}
\begin{figure}[H]
    \vspace{3.5cm}
    \centering
    \includegraphics[width=0.99\linewidth]{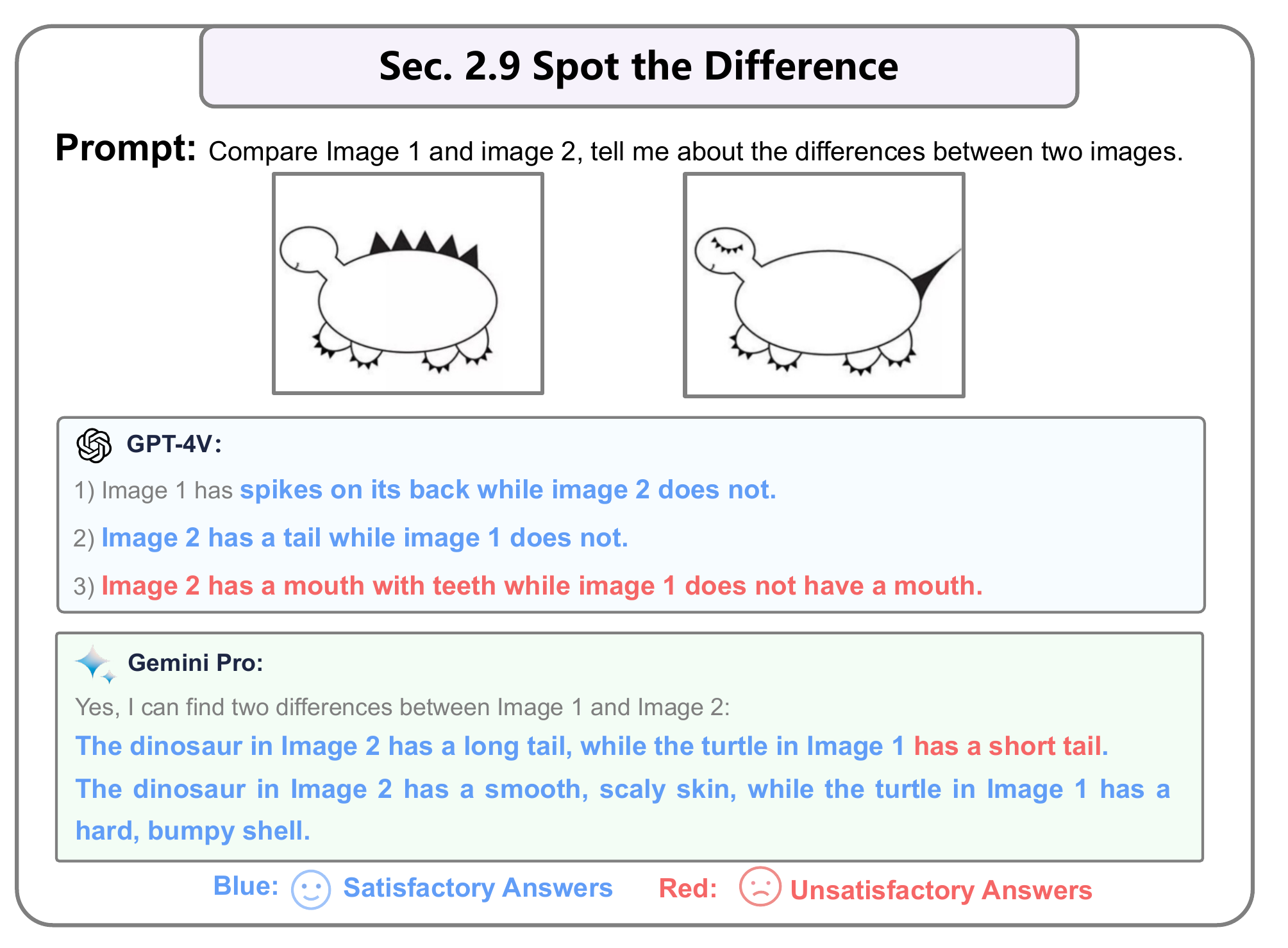}
    \vspace{-3pt}
    \caption[Section 2.9 Spot the Differences]{\textbf{Spot differences.} While GPT-4V successfully spots all differences but confuses an eye for a mouth, Gemini only discerns two out of three differences and incorrectly infers that the right image should have a short tail due to mistaking it for a turtle. Therefore, neither GPT-4V nor Gemini delivers completely satisfactory results in this task.}
    \label{fig:Sec.2.9_1}
    \vspace{-10pt}
\end{figure}

\cref{fig:Sec.2.9_1} demonstrates the models' ability to spot differences. Given two similar stick figure drawings of dinosaurs, the task is to find differences. Neither model was able to identify all the differences, indicating that there is room for improvement in this area. 

\newpage

\newpage

\newpage

\newpage

\newpage

\newpage

\newpage

\newpage

\newpage

% 2.7
\newpage

% 2.8 Object Counting
\newpage

% 2.9
\newpage

\newpage
\section{Text Recognition and Understanding in Images}
\label{Sec.3 Text Recognition and Understanding in Images}
This section mainly discusses the ability to recognize text in images. In contrast to the previous section, this section focuses solely on text extraction and basic understanding, without involving deep logical reasoning, or understanding utilizing knowledge bases. \cref{Sec.3.1 Scene Text Recognition} demonstrates the ability to extract text from scenes, including various scenarios such as billboards. \cref{Sec.3.2 Equation Recognition} involves the extraction ability for mathematical equations. \cref{Sec.3.3 Chart Text Recognition} deals with the extraction and simple understanding of text in charts and tables.

\subsection{Scene Text Recognition}
\label{Sec.3.1 Scene Text Recognition}
\begin{figure}[H]
    \vspace{1.5cm}
    \centering
    \includegraphics[width=0.99\linewidth]{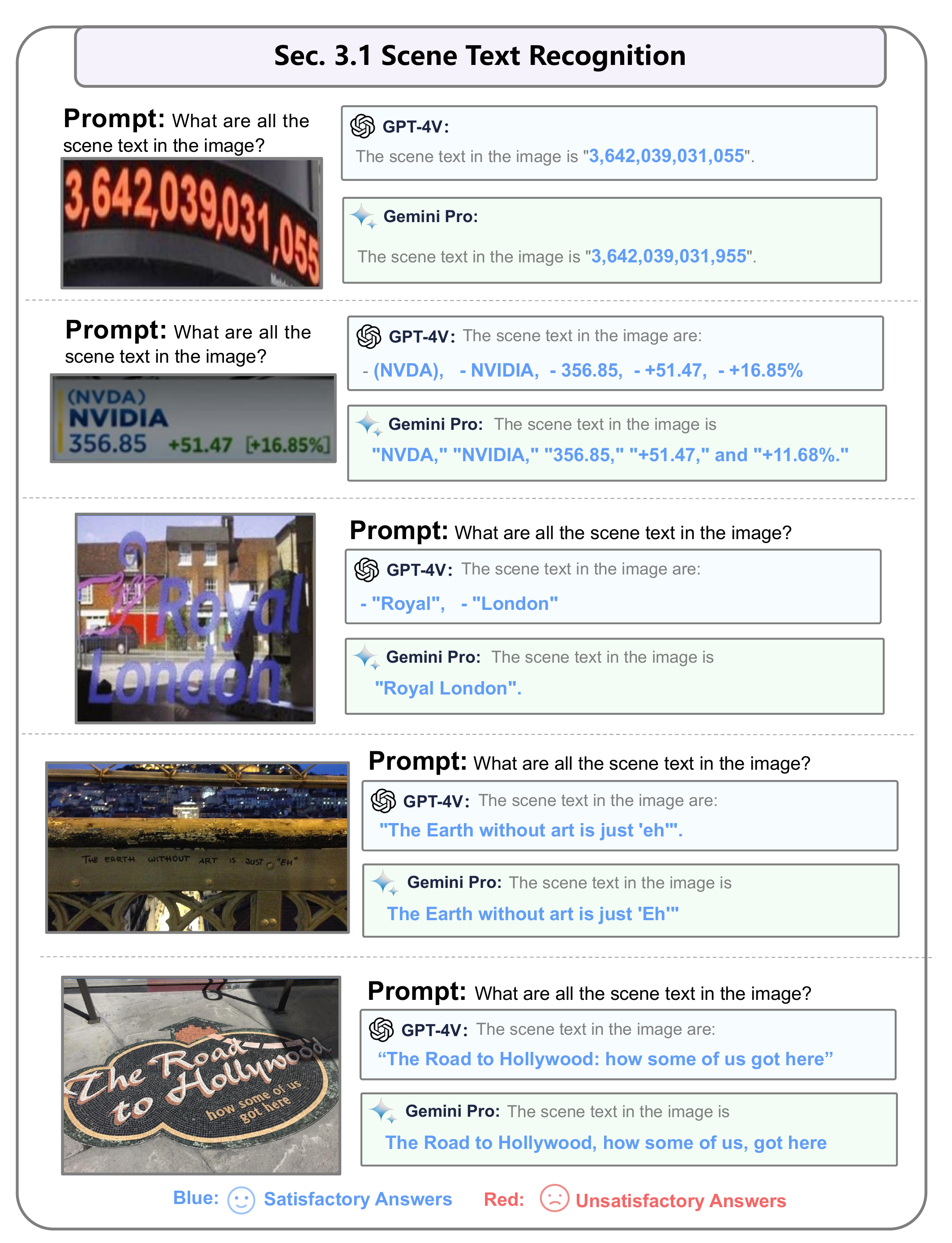}
    \vspace{-6pt}
    \caption[Section 3.1 Scene Text Recognition (1)]{\textbf{Scene text recognition (1/2).} Both models demonstrate proficiency in recognizing scene text under many challenging circumstances.}
    \label{fig:Sec.3.1_1}
    \vspace{-10pt}
\end{figure}

\begin{figure}[H]
    \vspace{1.3cm}
    \centering
    \includegraphics[width=0.99\linewidth]{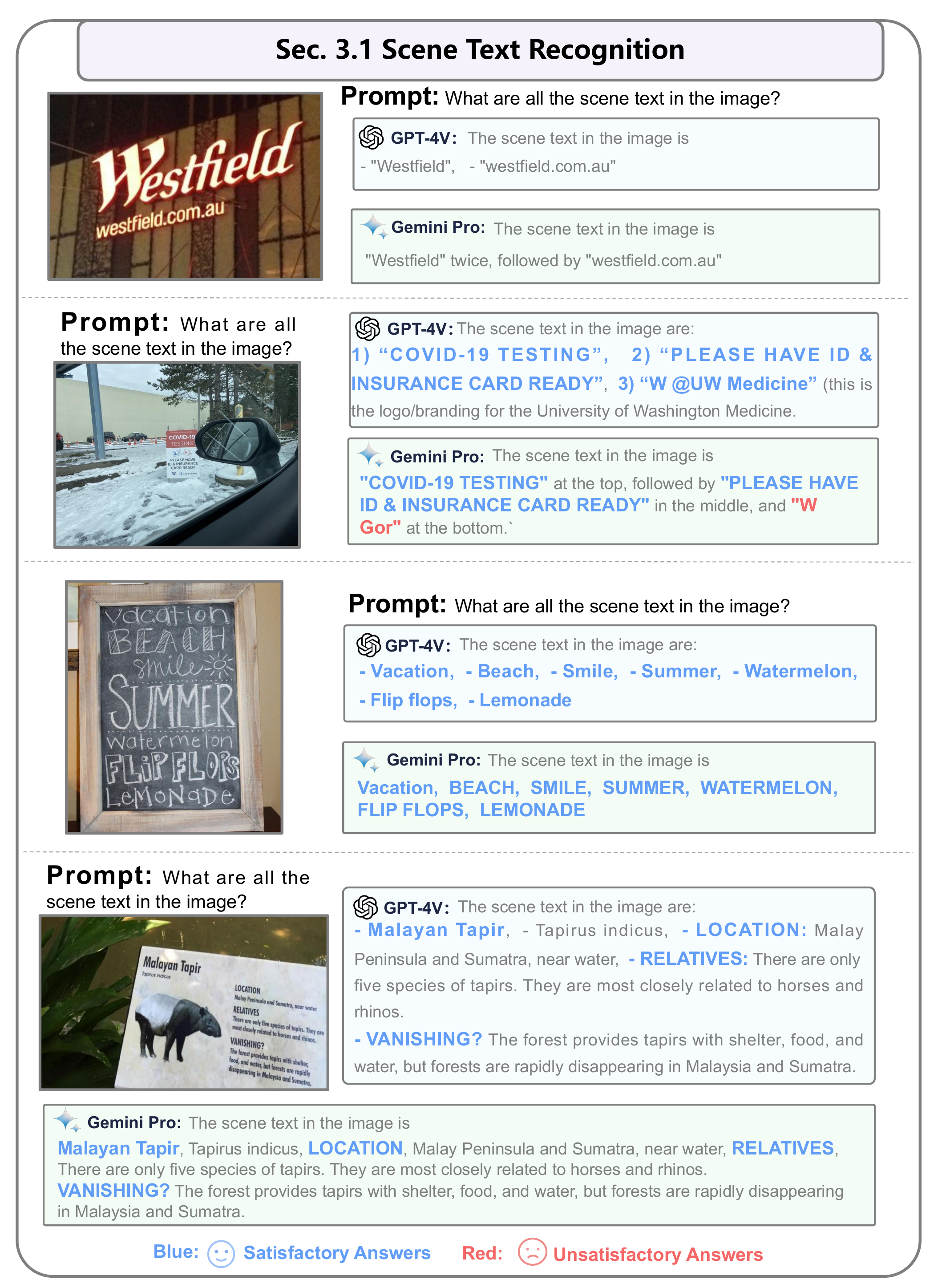}
    \vspace{-6pt}
    \caption[Section 3.1 Scene Text Recognition (2)]{\textbf{Scene text recognition (2/2).} Both models show competence in recognizing scene text across a myriad of challenging scenarios. However, Gemini displays minor errors in the second image's text recognition.}
    \label{fig:Sec.3.1_2}
    \vspace{-10pt}
\end{figure}

\begin{figure}[H]
    \centering
    \includegraphics[width=0.99\linewidth]{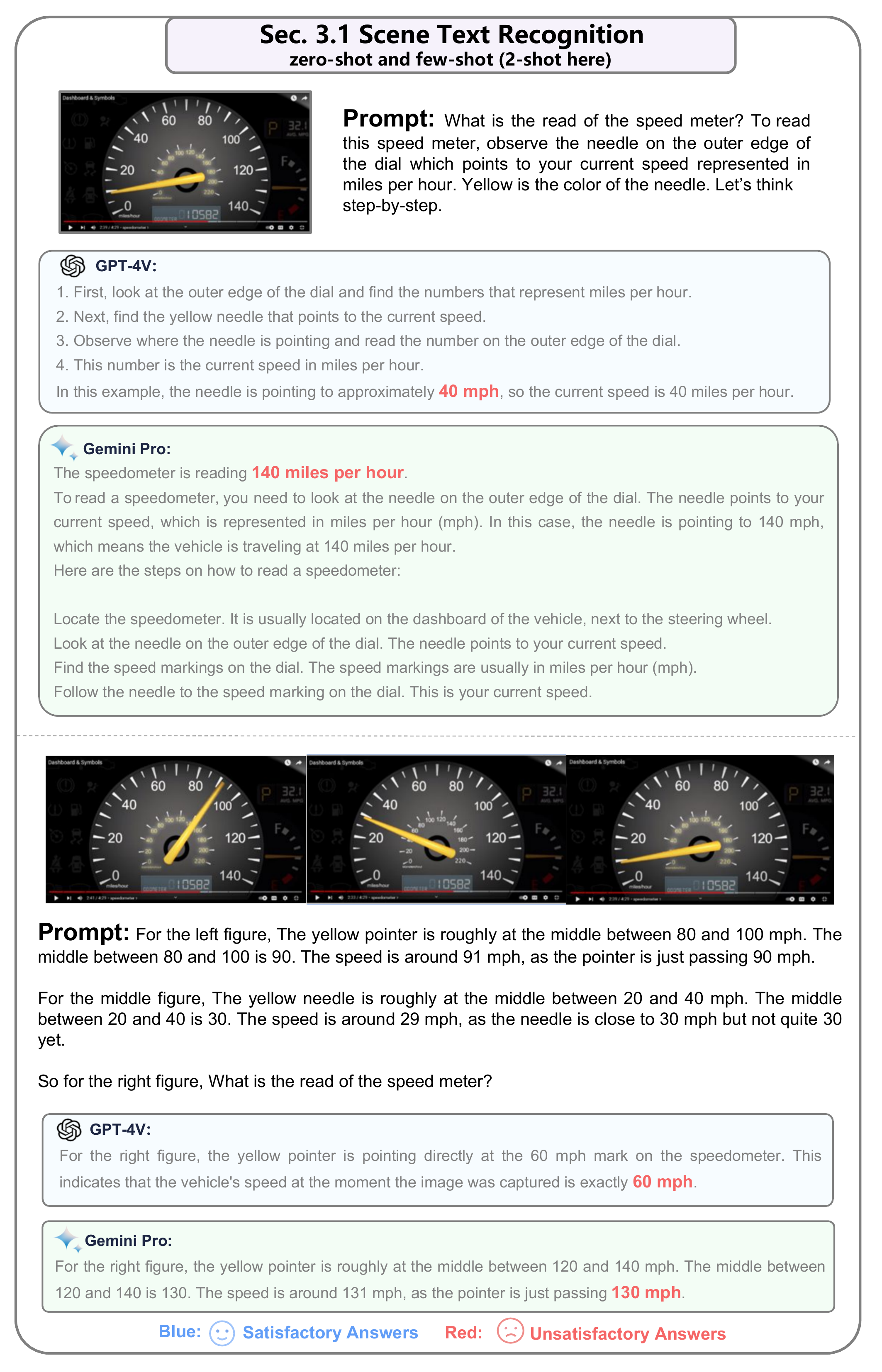}
    \vspace{-6pt}
    \caption[Section 3.1 Scene Text Recognition (3)]{\textbf{Scene text recognition in zero-shot and few-shot scenarios.} In zero-shot scenarios, both models fail to comprehend the dashboard. Even when provided with two examples in few-shot scenarios, neither GPT-4V nor Gemini succeed in correctly identifying the pattern from the samples.}
    \label{fig:Sec3.1_3}
    \vspace{-10pt}
\end{figure}

\cref{fig:Sec.3.1_1} and \cref{fig:Sec.3.1_2} show the extraction of text from images in various scenes, including stock market trading information and shop signs. Both models exhibit good performance. \cref{fig:Sec3.1_3} is about reading and estimating numbers on instrument panels, demonstrating the models' zero-shot and few-shot capabilities. It was found that Gemini's performance is inferior due to the lack of a memory mechanism. This can also be attributed to Gemini's limited memory capabilities, as GPT-4V can process multiple images sequentially, while Gemini can only input images after combining them into a single input.

\subsection{Eqation Recognition}
\label{Sec.3.2 Equation Recognition}
\begin{figure}[H]
    \centering
    \includegraphics[width=0.95\linewidth]{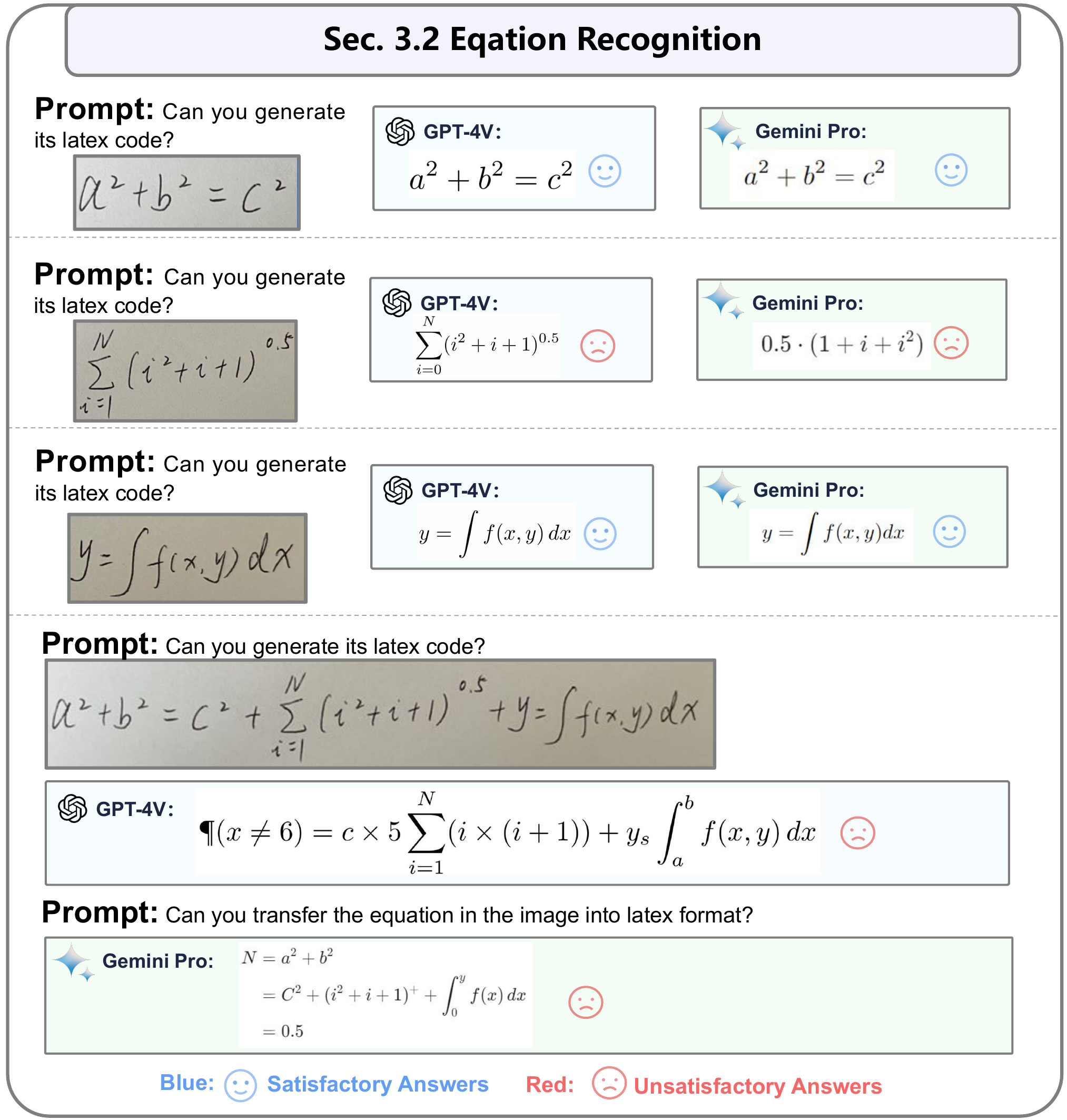}
    \vspace{-2pt}
    \caption[Section 3.2 Equation Recognition]{\textbf{Equation recognition.} Both models are capable of generating LaTeX codes from hand-written inputs. They succeed in simple cases, but fail to accurately generate complex equations. GPT-4V's results are slightly closer to the actual equation, as Gemini is unable to recognize the summation symbol.}
    \label{fig:Sec.3.2_1}
    \vspace{-10pt}
\end{figure}

\cref{fig:Sec.3.2_1} shows the capability of understanding mathematical formulas, where the task is to convert formulas in images into LaTeX format. Both models perform well with simple equations but struggle to accurately recognize complex ones. Both models need improvement in understanding mathematical symbols. We found that Gemini, in particular, fails to recognize the summation symbol.

\subsection{Chart Text Recognition}
\label{Sec.3.3 Chart Text Recognition}
\begin{figure}[H]
    \centering
    \includegraphics[width=0.95\linewidth]{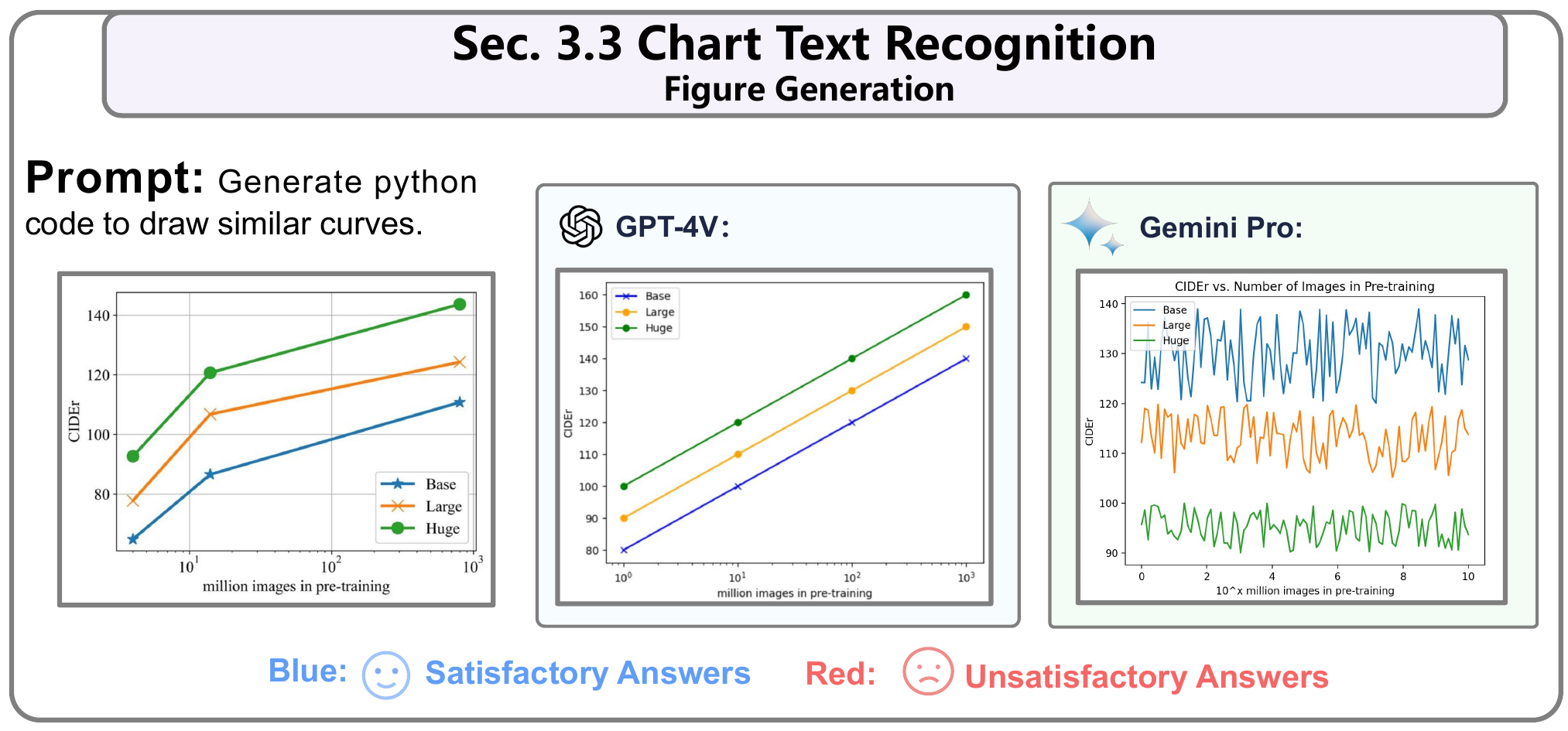}
    \vspace{-3pt}
    \caption[Section 3.3 Chart Text Recognition (1)]{\textbf{Chart text recognition in zero-shot scenario.} GPT-4V generates Python code with a figure that is more consistent with the original code than what Gemini produces.}
    \label{fig:Sec.3.3_1}
    \vspace{-10pt}
\end{figure}

\begin{figure}[H]
    \centering
    \includegraphics[width=0.99\linewidth]{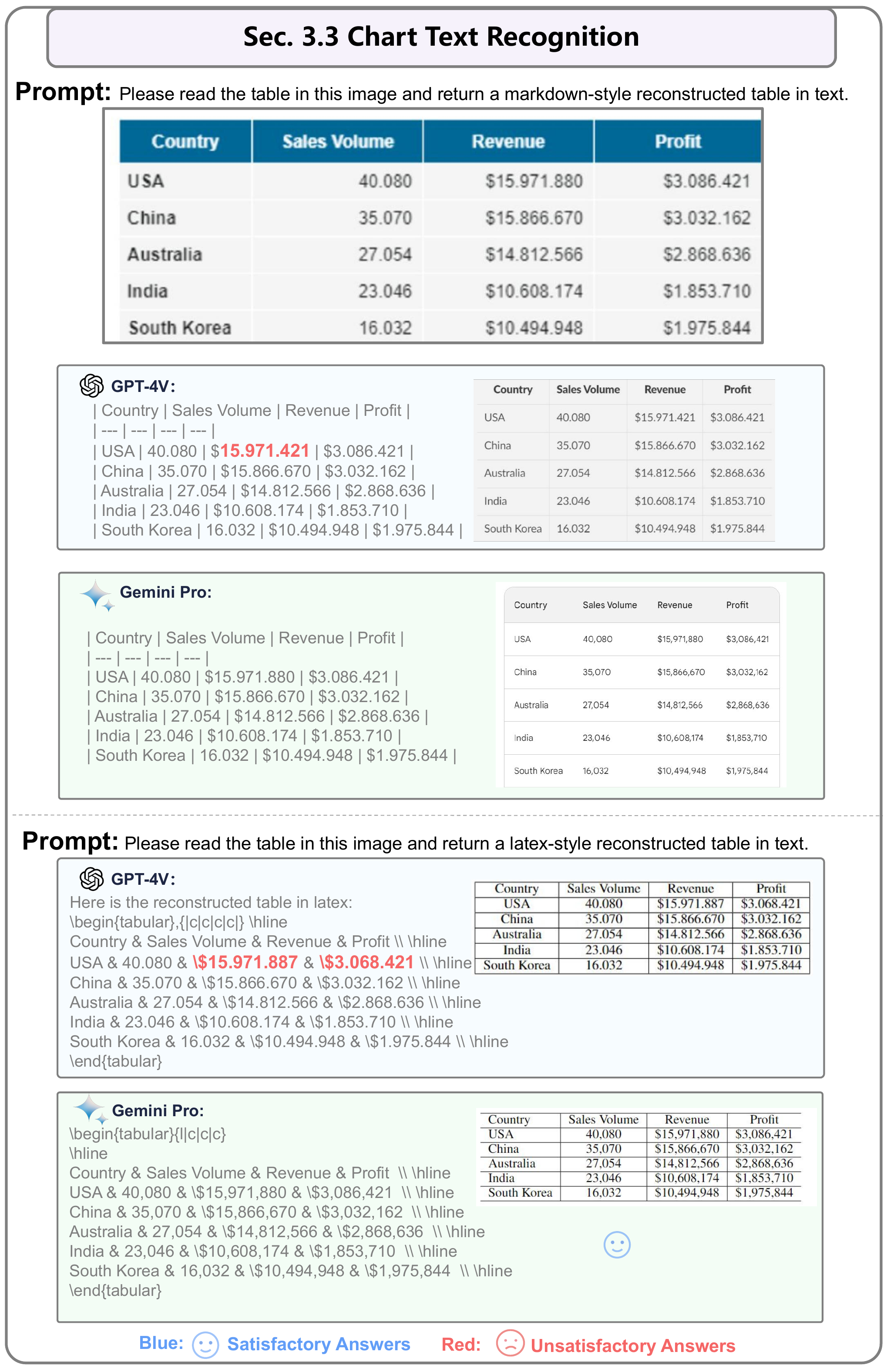}
    \vspace{-6pt}
    \caption[Section 3.3 Chart Text Recognition (2)]{\textbf{Chart text recognition and figure code generation.} Gemini outperforms GPT-4V in generating the correct table based on a given picture. GPT-4V exhibits errors in number recognition.}
    \label{fig:Sec.3.3_2}
    \vspace{-10pt}
\end{figure}

\begin{figure}[H]
    \centering
    \vspace{1.1cm}
    \includegraphics[width=0.99\linewidth]{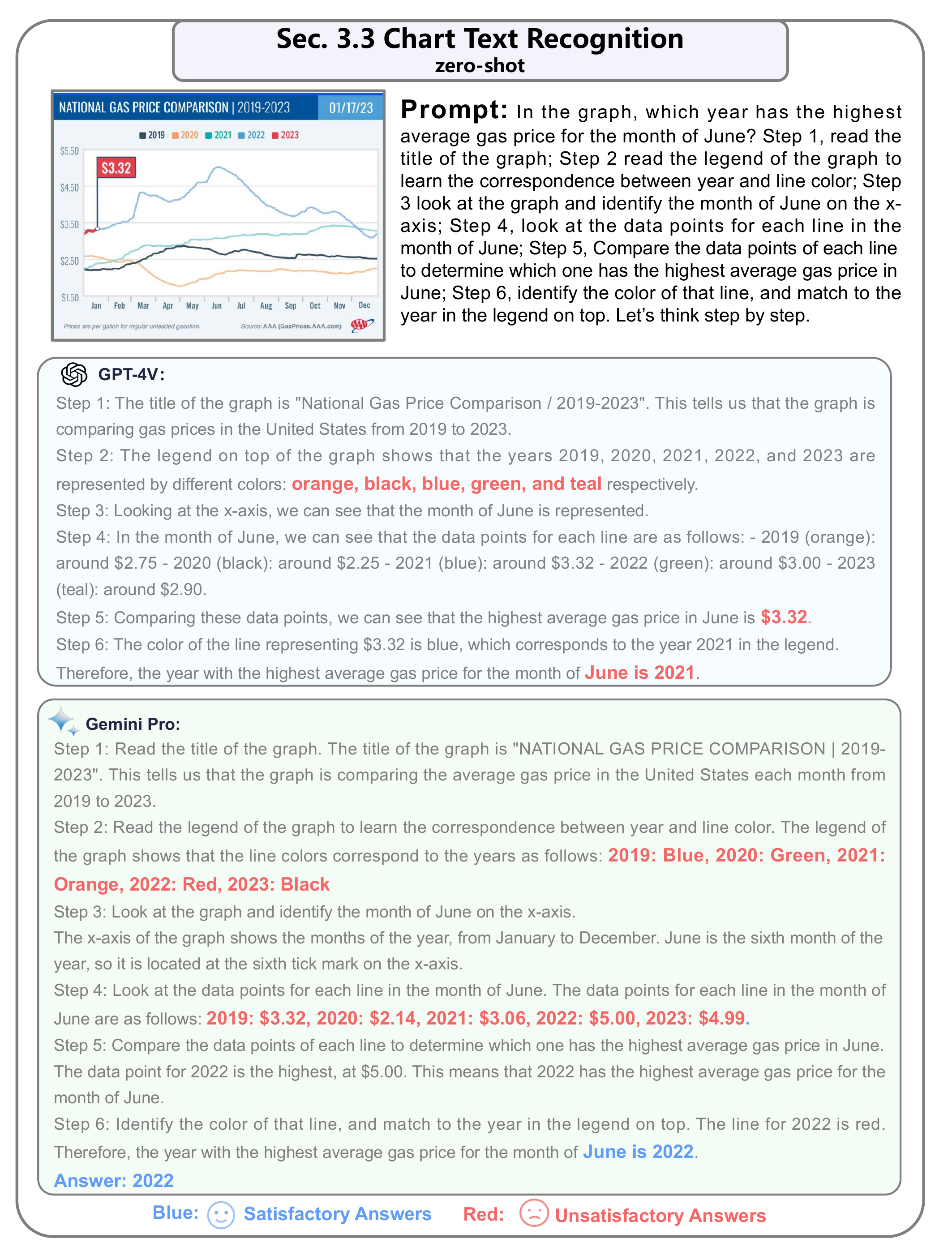}
    \vspace{-6pt}
    \caption[Section 3.3 Chart Text Recognition (3)]{This image exhibits the results of chart text recognition tasks carried out by GPT-4V and Gemini. Both models show poor performance in recognizing charts, misjudging colors and values. Gemini's capability is slightly superior as it provides correct results on Steps 5 and 6, despite errors in the intermediate process details.}
    \label{fig:Sec.3.3_3}
    \vspace{-10pt}
\end{figure}

\cref{fig:Sec.3.3_1} shows the task of reading and reconstructing a Python-generated line chart through Python code, where GPT-4V's reconstruction is more similar to the original chart. Here, GPT-4V excels in simulating the growth trend of the original image, while Gemini provides a more chaotic shape.
\cref{fig:Sec.3.3_2} demonstrates the models' ability to read text from tables, where Gemini performs well without errors. Therefore, once again, it is confirmed that for single images, Gemini performs better when the image resolution is high.
\cref{fig:Sec.3.3_3} presents reading a trend line graph, identifying the highest point and its x and y-axis values, with Gemini showing good performance in correctly choosing the right curve and accurately reading the highest point information. GPT-4V and Gemini both made errors when extracting hidden information from the images. This already involves some basic inference. In \cref{Sec.5 Textual Reasoning in Images}, we will further investigate the multimodal textual reasoning abilities of both models.

% 3.1 Scene Text Recognition
\newpage

\newpage

\newpage

% 3.2 Eqation Recognition
\newpage

% 3.3 Chart Text Recognition

\newpage

\newpage

\newpage
\section{Image Reasoning Abilities}
\label{Sec.4 Image Reasoning Abilities}
This section primarily showcases the model's ability for further inference from image information. Unlike \cref{Sec.2 Image Recognition and Understanding}, this section involves not only simple recognition of the model but also further reasoning based on the information in the images. \cref{Sec.4.1 Humorous Image Understanding} demonstrates the model's understanding of humorous images. \cref{Sec.4.2 Multimodal Knowledge and Commonsense} presents multimodal knowledge and commonsense, testing the model's ability to apply multimodal knowledge and commonsense in interpreting scientific concepts and details in images. \cref{Sec.4.3 Detective Reasoning Ability} explores the detective reasoning skills of the model, focusing on inferring a person's characteristics or profession based on environmental clues. \cref{Sec.4.4 Association of Parts and Objects} assesses the model's ability to combine multiple images into a whole, challenging its capacity to form connections between different visual elements. \cref{Sec.4.5 Intelligence Tests} involves intelligence tests, evaluating the model's IQ through pattern recognition and logical reasoning. \cref{Sec.4.6 Emotional Intelligence Tests} examines the model's emotional intelligence (EQ), focusing on their understanding and expression of emotions and aesthetic judgment.

\subsection{Humorous Image Understanding}
\label{Sec.4.1 Humorous Image Understanding}
\begin{figure}[H]
    \centering
    \includegraphics[width=0.83\linewidth]{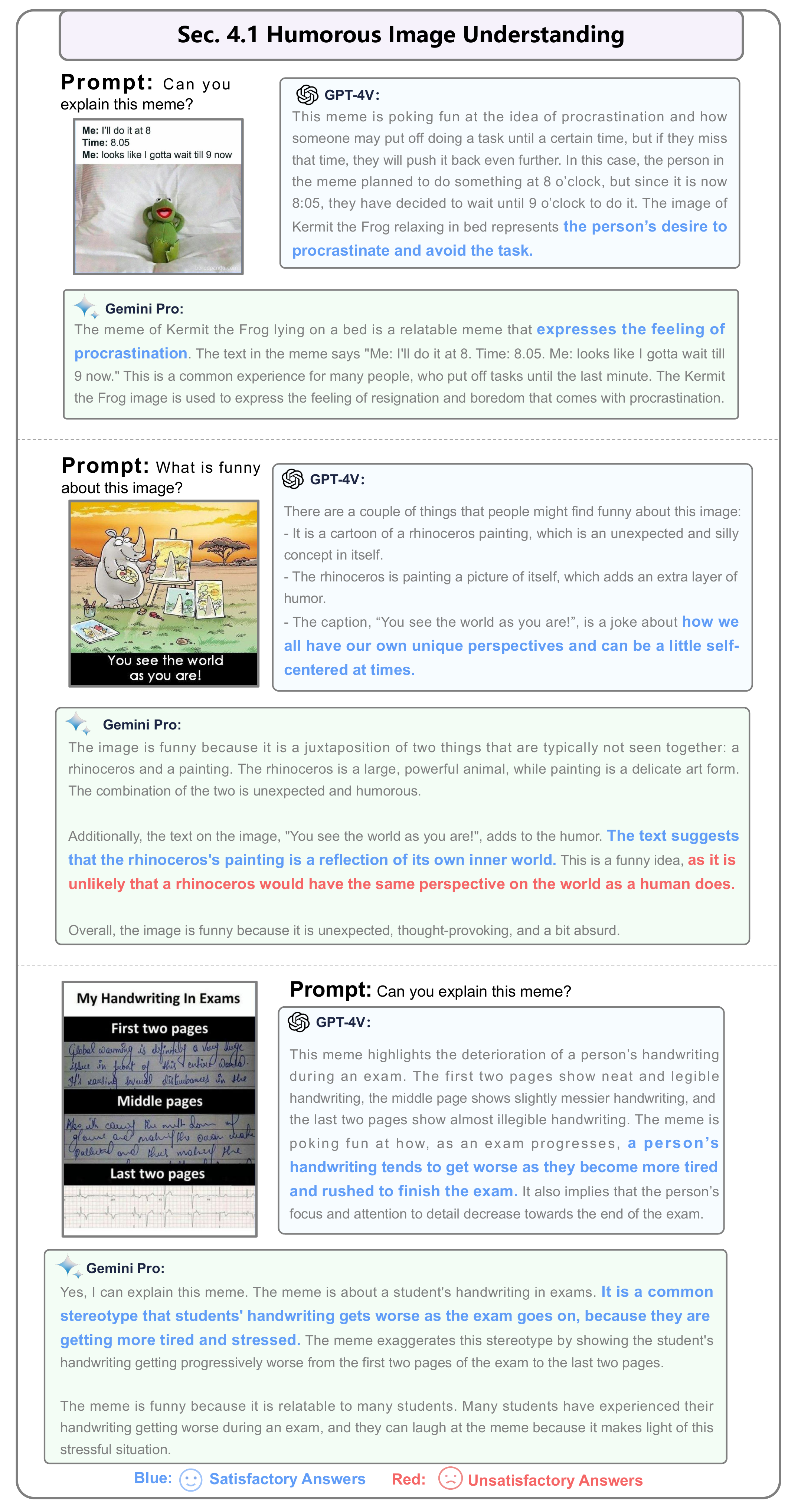}
    \vspace{-4pt}
    \caption[Section 4.1 Humorous Image Understanding]{\textbf{Humorous meme understanding.} Both GPT-4V and Gemini demonstrate the impressive capability to comprehend the humor embedded within memes.}
    \label{fig:Sec.4.1_1}
    \vspace{-10pt}
\end{figure}

\cref{fig:Sec.4.1_1} is about understanding humorous images in the form of jokes. The models need to understand where the punchline is located in the image and the simple text. This section primarily focuses on image understanding, so we classify it as reasoning for images. In the three examples here, both models displayed a correct understanding of the two joke images, demonstrating an ability to comprehend human humor.

\subsection{Multimodal Knowledge and Commonsense}
\label{Sec.4.2 Multimodal Knowledge and Commonsense}
\begin{figure}[H]
    \vspace{.6cm}
    \centering
    \includegraphics[width=0.99\linewidth]{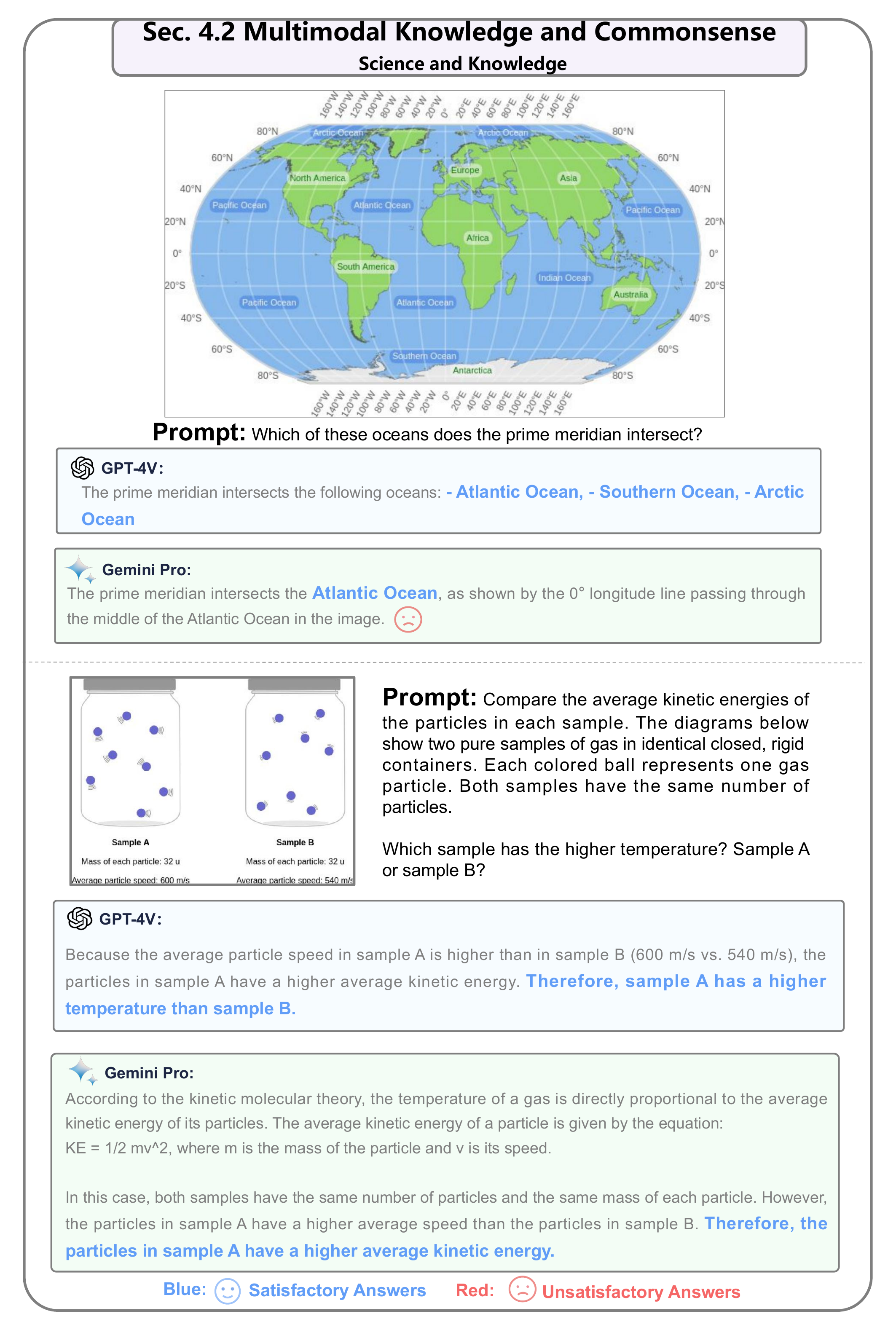}
    \vspace{0pt}
    \caption[Section 4.2 Multimodal Knowledge and Commonsense (1)]{\textbf{Answering science and knowledge questions (1/3).} Both GPT-4V and Gemini can understand the question textually and visually, and gather necessary information to answer the question. In the first case, there are omissions in Gemini’s description.}
    \label{fig:Sec.4.2_1}
    \vspace{-10pt}
\end{figure}

\begin{figure}[H]
    \vspace{1cm}
    \centering
    \includegraphics[width=1.00\linewidth]{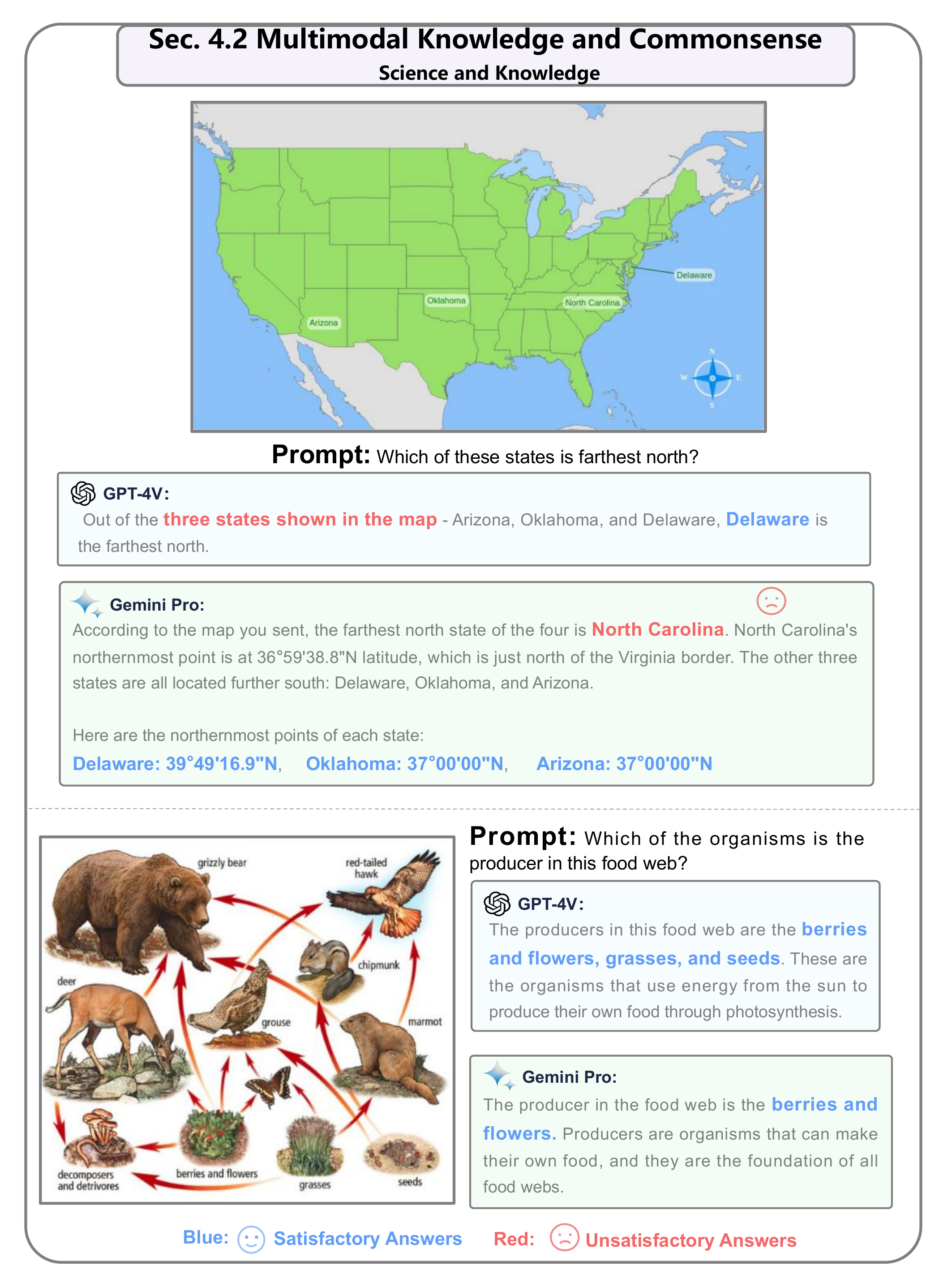}
    \vspace{0pt}
    \caption[Section 4.2 Multimodal Knowledge and Commonsense (2)]{\textbf{Answering science and knowledge questions (2/3).} Both GPT-4V and Gemini can understand the question textually and visually, and gather necessary information to answer the question. In the first case, there are errors in both GPT-4V's and Gemini’s description.}
    \label{fig:Sec.4.2_2}
    \vspace{-10pt}
\end{figure}

\begin{figure}[H]
    \centering
    \vspace{1cm}
    \includegraphics[width=0.99\linewidth]{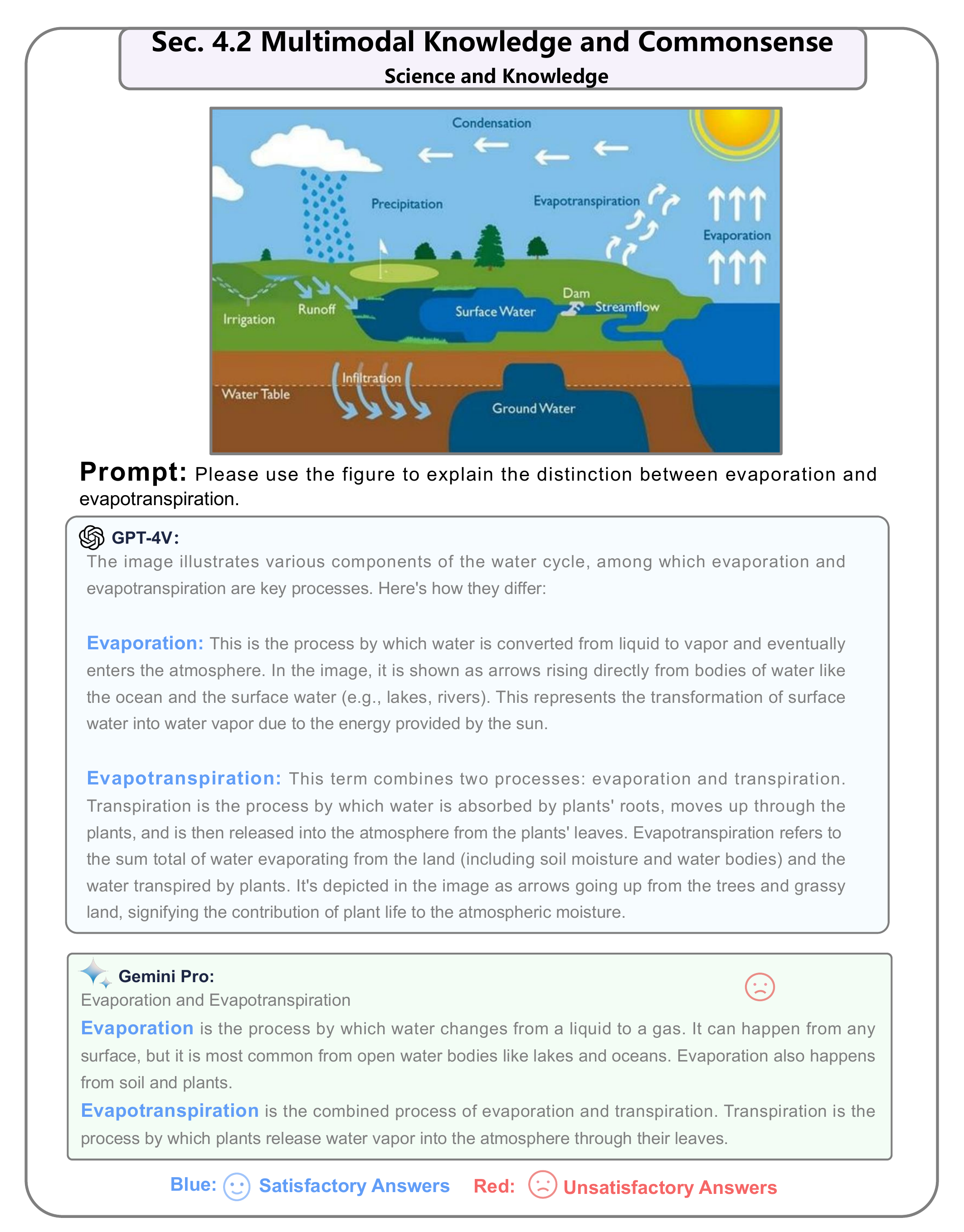}
    \vspace{0pt}
    \caption[Section 4.2 Multimodal Knowledge and Commonsense (3)]{\textbf{Answering science and knowledge questions (3/3).} Both GPT-4V and Gemini can understand the question textually and visually, and gather necessary information to answer the question. In this example, GPT-4V generates better answer than Gemini.}
    \label{fig:Sec.4.2_3}
    \vspace{-10pt}
\end{figure}

The next three images focus on multimodal commonsense understanding, emphasizing scientific knowledge. \cref{fig:Sec.4.2_1} deals with world map comprehension; due to image resolution, Gemini's answer is incomplete, but it excels in general knowledge questions. \cref{fig:Sec.4.2_2} involves the Americas map and food chain questions, where Gemini's answers are also incomplete. \cref{fig:Sec.4.2_3} illustrates the rainfall cycle. In summary, both models possess strong knowledge for answering general scientific queries, yet Gemini struggles with small text extraction.

\subsection{Detective Reasoning Ability}
\label{Sec.4.3 Detective Reasoning Ability}
\begin{figure}[H]
    \vspace{2cm}
    \centering
    \includegraphics[width=0.99\linewidth]{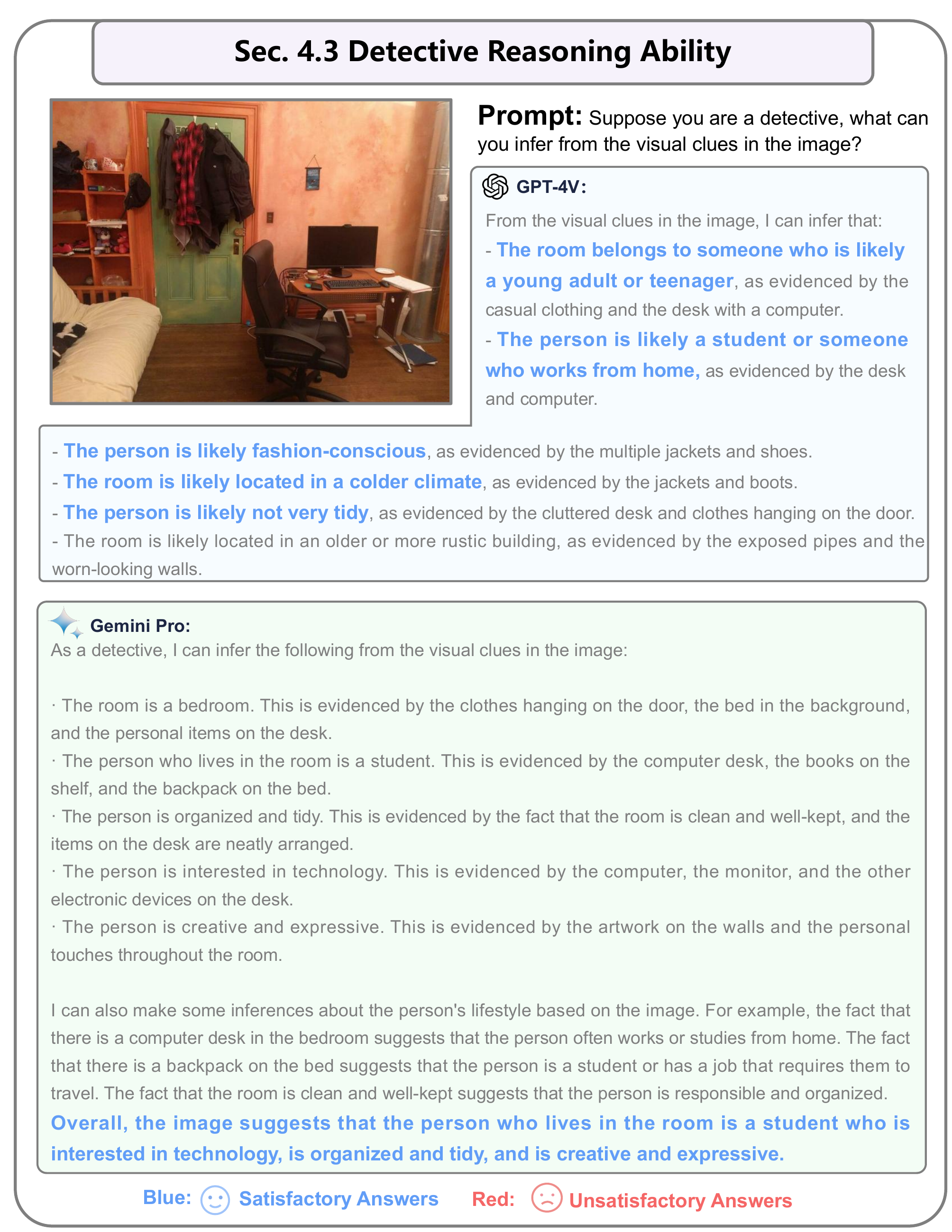}
    \vspace{0pt}
    \caption[Section 4.3 Detective Reasoning Ability]{ \textbf{Multimodal commonsense reasoning.} Both GPT-4V and Gemini can infer images well based on common sense.}
    \label{fig:Sec.4.3_1}
    \vspace{-10pt}
\end{figure}

\cref{fig:Sec.4.3_1} showcases the model's detective and logical reasoning abilities. When given a person's room to deduce their profession and personality, both models were able to make reasonable judgments, with Gemini providing a more detailed thought process. In contrast, GPT-4V could provide concise yet accurate answers.

\subsection{Association of Parts and Objects}
\label{Sec.4.4 Association of Parts and Objects}
\begin{figure}[H]
    \vspace{3cm}
    \centering
    \includegraphics[width=0.99\linewidth]{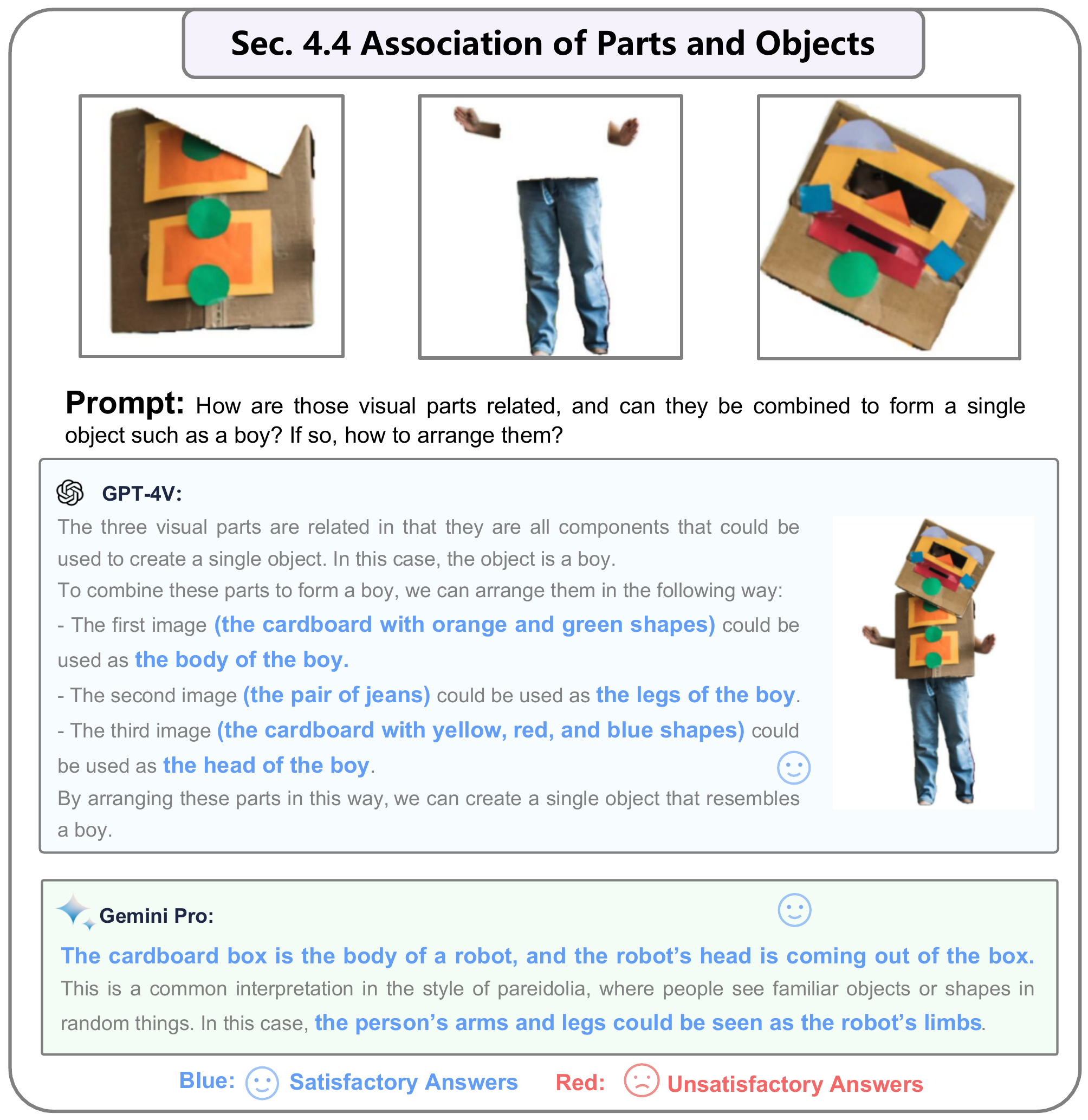}
    \vspace{0pt}
    \caption[Section 4.4 Association of Parts and Objects]{\textbf{Understanding part-object association in real image.} Both GPT4-V and Gemini understand how to combine the three part in a reasonable way. Of course, here GPT-4V provides a more satisfactory answer.}
    \label{fig:Sec.4.4_1}
    \vspace{-10pt}
\end{figure}

\cref{fig:Sec.4.4_1} challenges the model to combine multiple images into a whole. Here, the task involved forming a person using a bag as the head and body. Although both models provided reasonable judgments, GPT-4V offered a more accurate and detailed answer.

\subsection{Intelligence Tests}
\label{Sec.4.5 Intelligence Tests}
\begin{figure}[H]
    \vspace{3cm}
    \centering
    \includegraphics[width=0.99\linewidth]{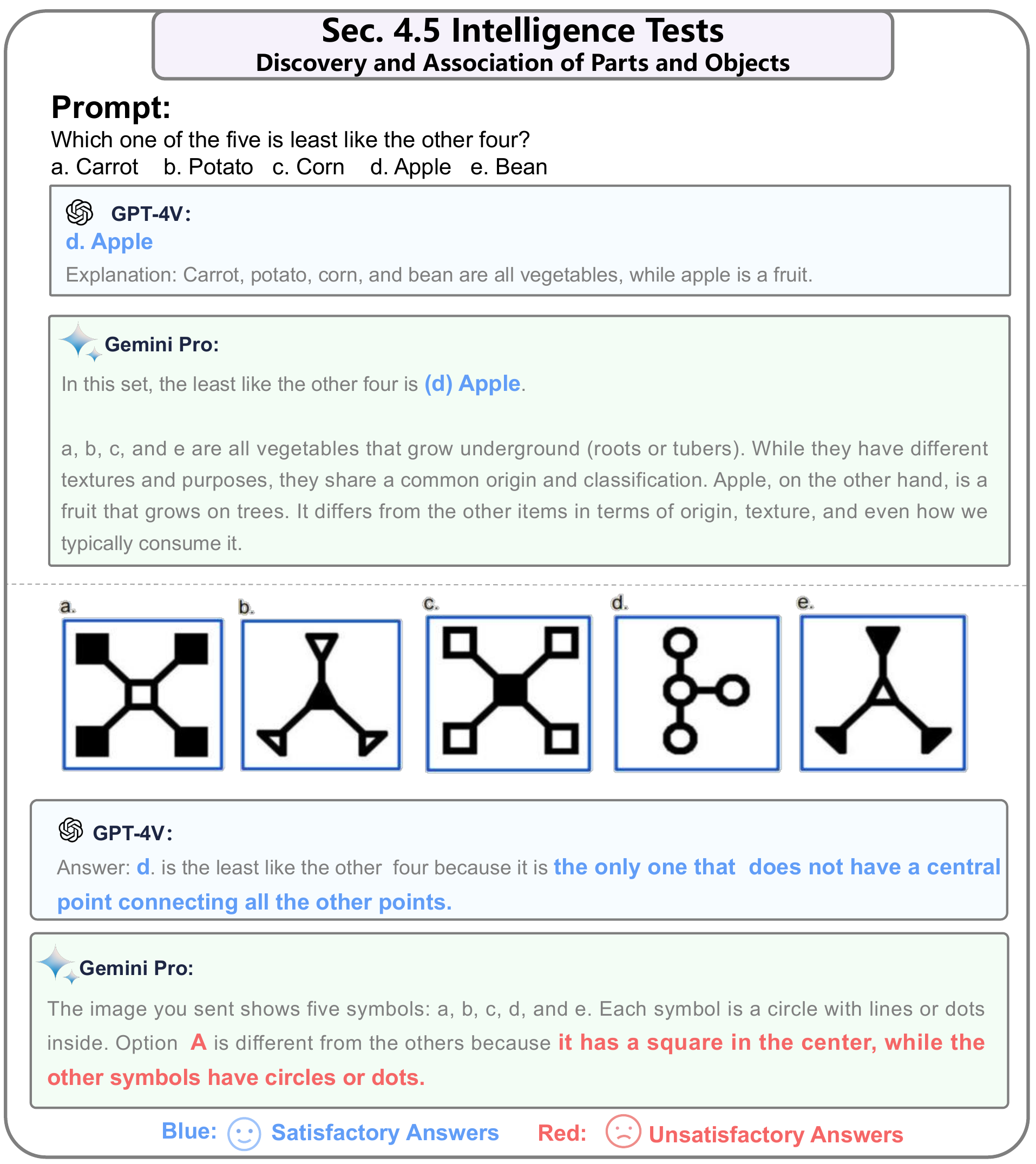}
    \vspace{0pt}
    \caption[Section 4.5 Intelligence Tests (1)]{\textbf{Discovery and association of parts and objects.} In response to example questions from the Wechsler Adult Intelligence Scale (WAIS)~\cite{wechsler1981wechsler}, Gemini performed poorly. This suggests that its ability to recognize and compare multiple images might be limited.}
    \label{fig:Sec.4.5_1}
    \vspace{-10pt}
\end{figure}

\begin{figure}[H]
    \centering
    \includegraphics[width=0.99\linewidth]{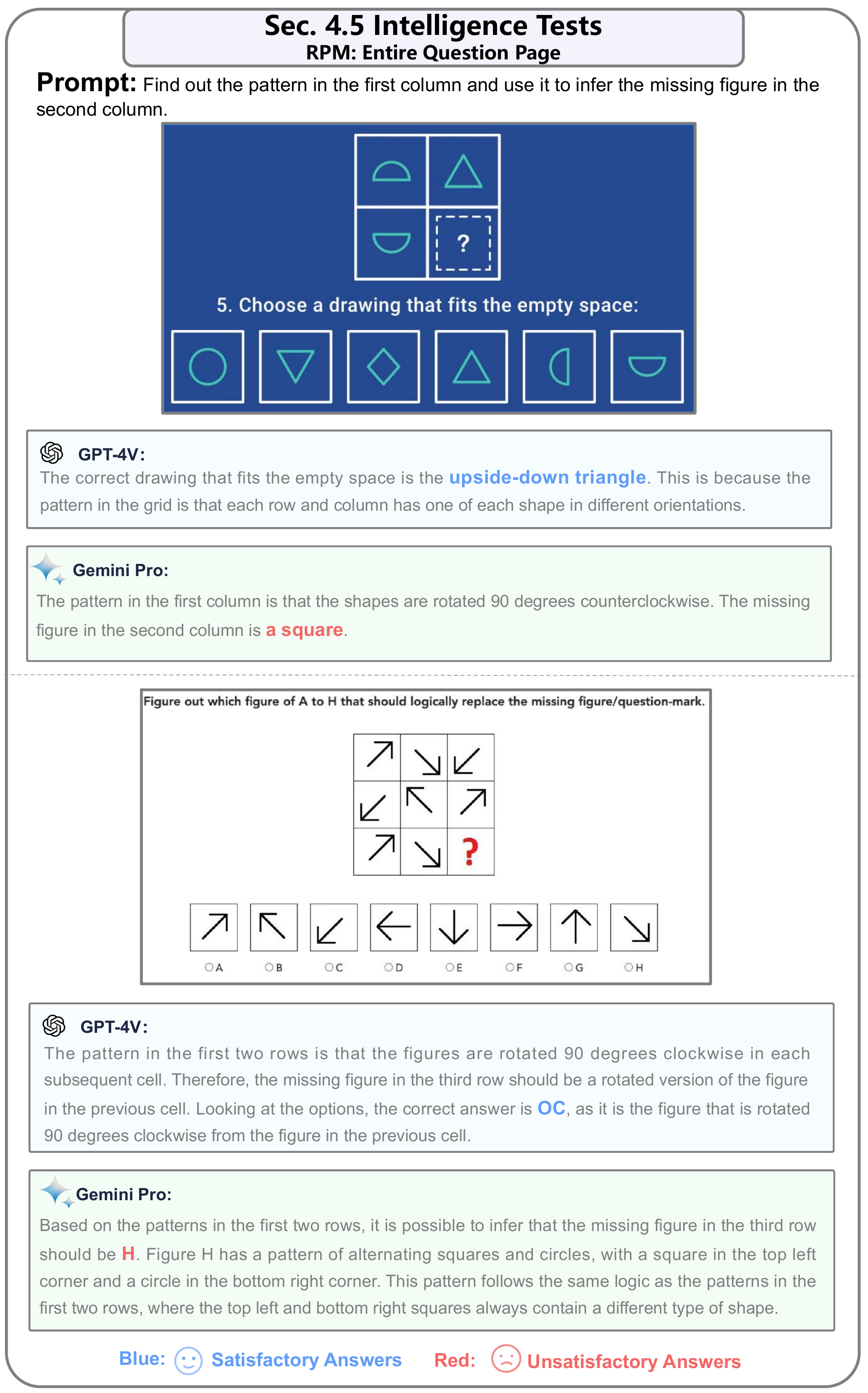}
    \vspace{-4pt}
    \caption[Section 4.5 Intelligence Tests (2)]{\textbf{RPM: entire question page (1/2).} In response to example questions from the Raven’s Progressive Matrices~\cite{raven2003raven}, Gemini performed poorly. This suggests that its ability to recognize, compare multiple images, and discern their inherent connections might be limited.}
    \label{fig:Sec.4.5_2}
    \vspace{-10pt}
\end{figure}

\begin{figure}[H]
    \vspace{3cm}
    \centering
    \includegraphics[width=0.99\linewidth]{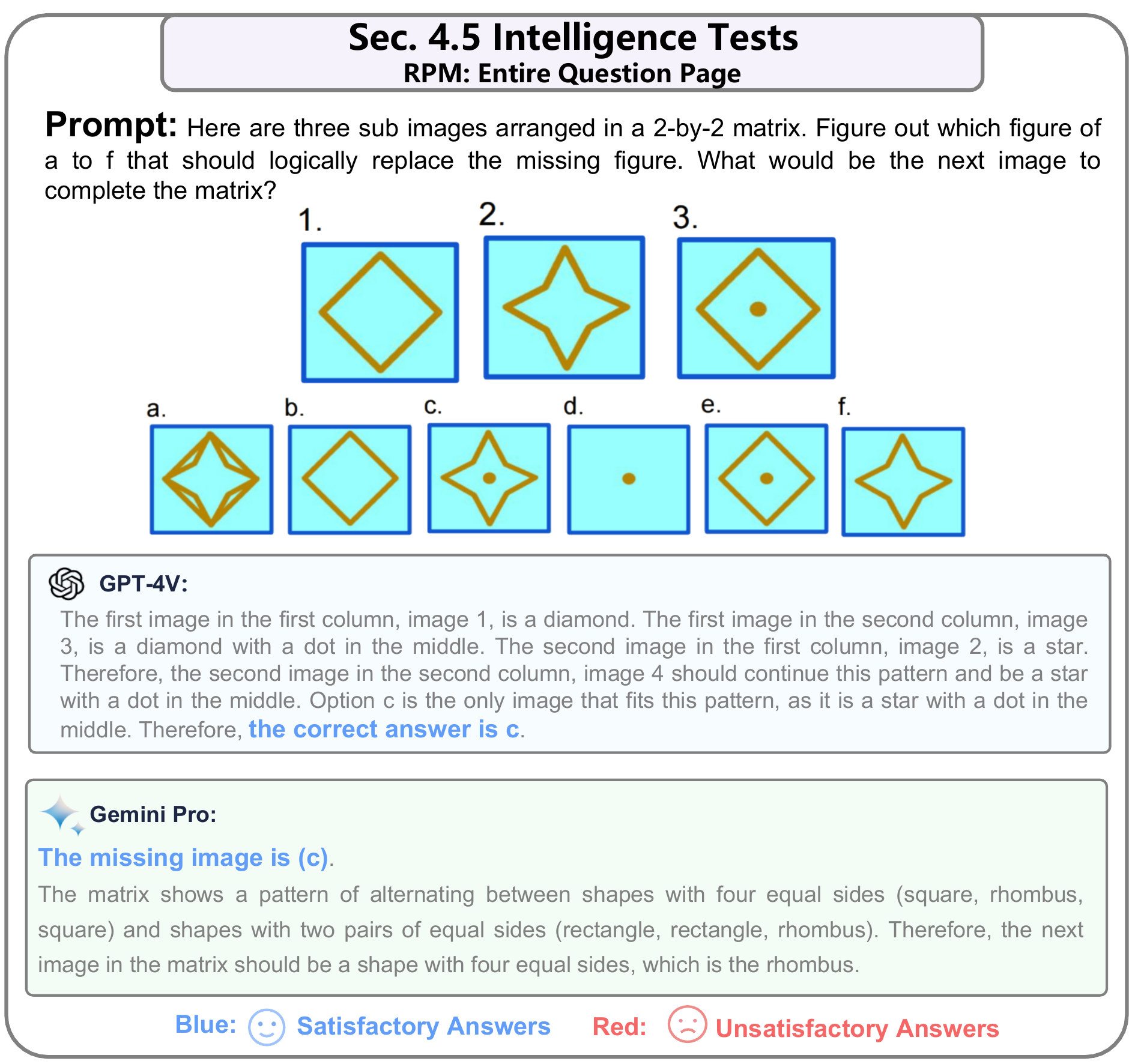}
    \vspace{0pt}
    \caption[Section 4.5 Intelligence Tests (3)]{\textbf{RPM: entire question page (2/2).} In response to example questions from the Raven’s Progressive Matrices~\cite{raven2003raven}, both GPT-4V and Gemini generated the correct results. However, Gemini failed to provide a reasonable explanation. This suggests that its ability to recognize, compare multiple images, and discern their inherent connections might be limited.}
    \label{fig:Sec.4.5_3}
    \vspace{-10pt}
\end{figure}

\begin{figure}[H]
    \vspace{2.8cm}
    \centering
    \includegraphics[width=0.99\linewidth]{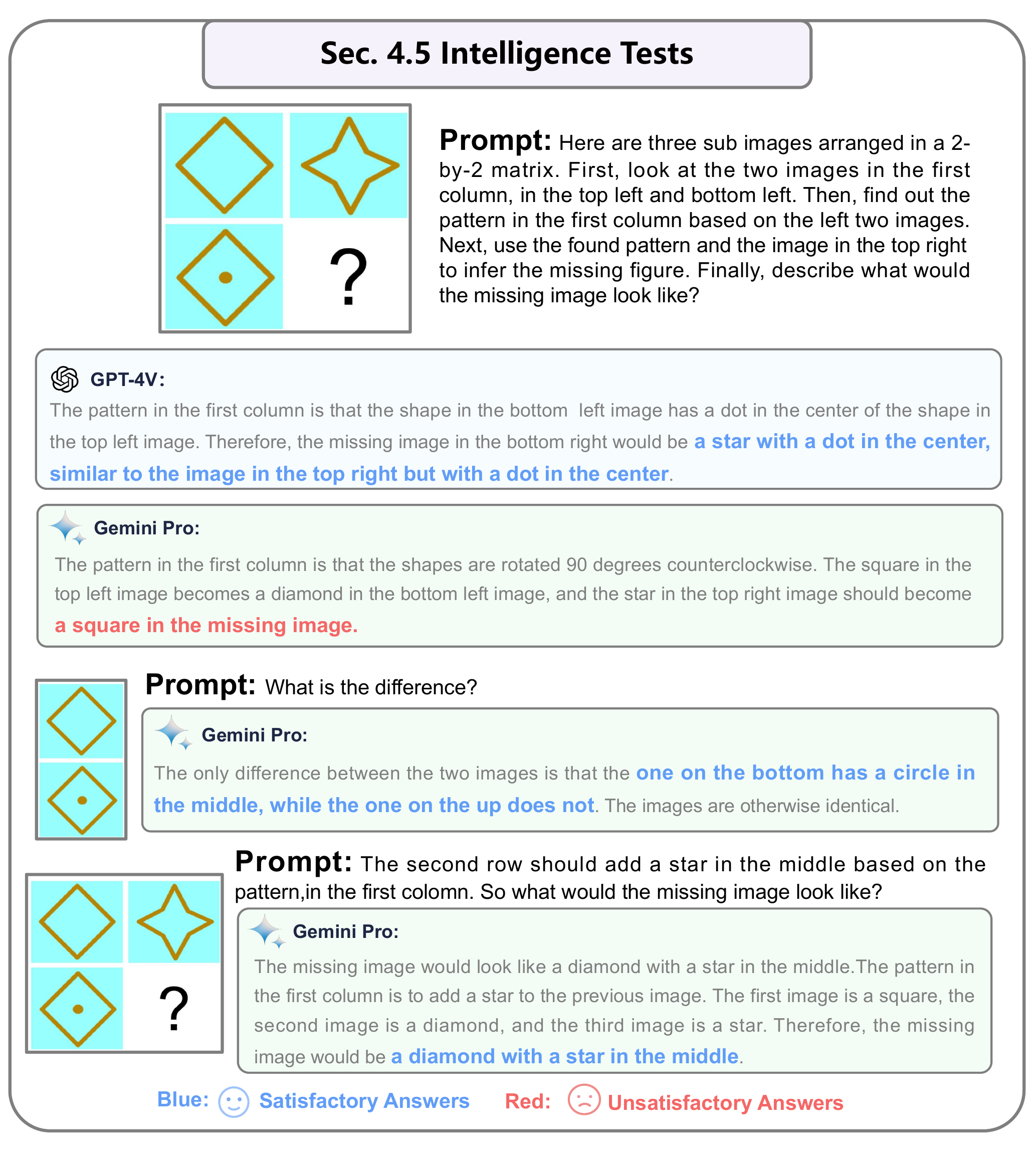}
    \vspace{0pt}
    \caption[Section 4.5 Intelligence Tests (4)]{\textbf{tackling entire question page step-by-step.} n response to the example question from the Raven’s Progressive Matrices~\cite{raven2003raven}, which directly asks for pattern completion, Gemini performed poorly. However, when guided step-by-step, Gemini correctly answered the question. This may indicates that Gemini has the ability to compare differences between images, but it is not very familiar with the pattern of RPM tests}
    \label{fig:Sec.4.5_4}
    \vspace{-10pt}
\end{figure}

\cref{fig:Sec.4.5_1}-\cref{fig:Sec.4.5_4} demonstrate the models' IQ capabilities, mainly featuring intelligence tests with pattern-finding graphical problems. Most of the pattern-finding tasks here are relatively simple. Overall, GPT-4V is able to provide more accurate answers. This could also be attributed to Gemini's inability to recognize multiple images.

\subsection{Emotional Intelligence Tests}
\label{Sec.4.6 Emotional Intelligence Tests}
\begin{figure}[H]
    \vspace{3cm}
    \centering
    \includegraphics[width=0.99\linewidth]{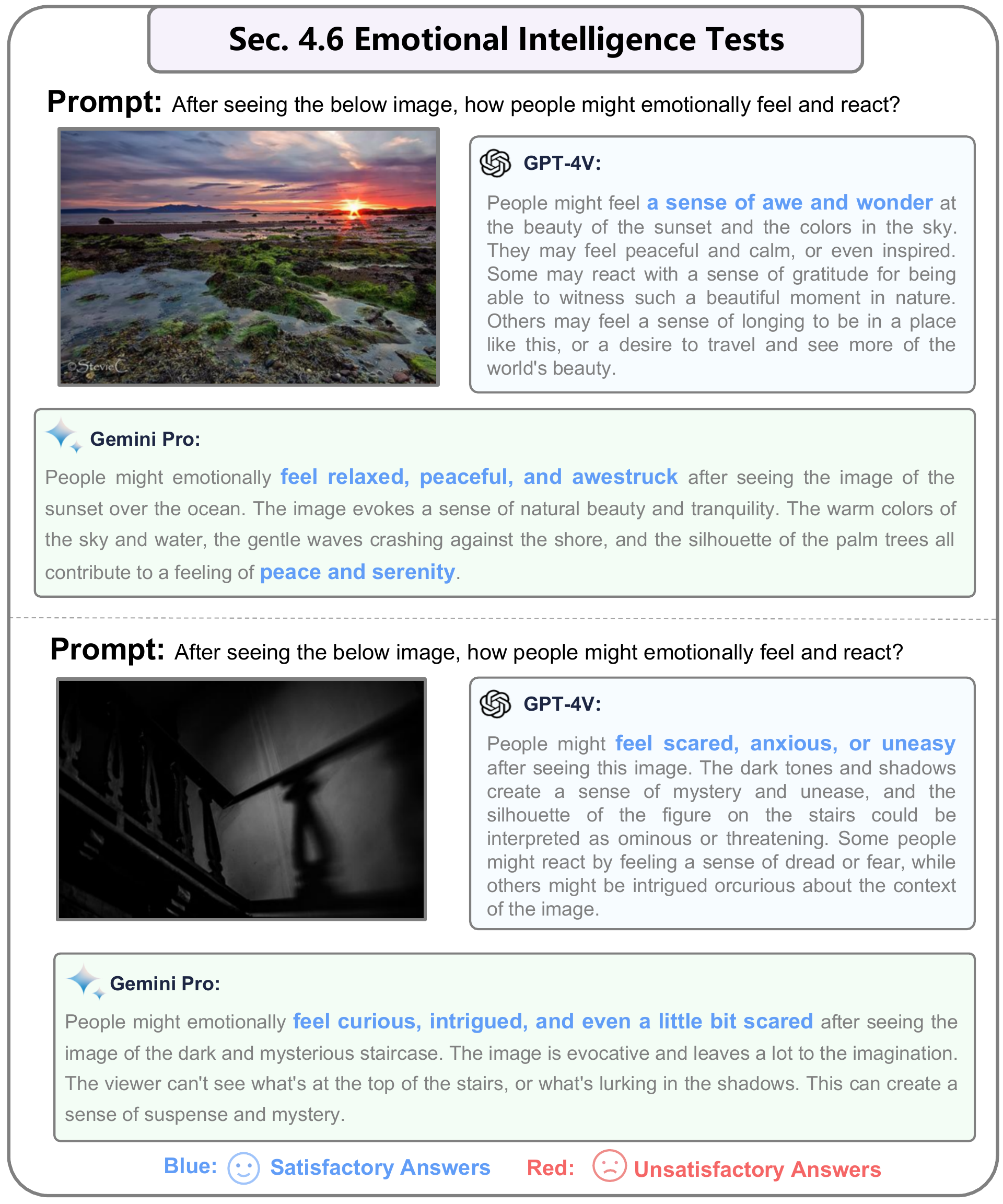}
    \vspace{0pt}
    \caption[Section 4.6 Emotional Intelligence Tests (1)]{\textbf{Emotional intelligence tests.} Both GPT-4V and Gemini understands how different visual contents may arouse human emotions.}
    \label{fig:Sec.4.6_1}
    \vspace{-10pt}
\end{figure}

\begin{figure}[H]
    \vspace{3cm}
    \centering
    \includegraphics[width=0.99\linewidth]{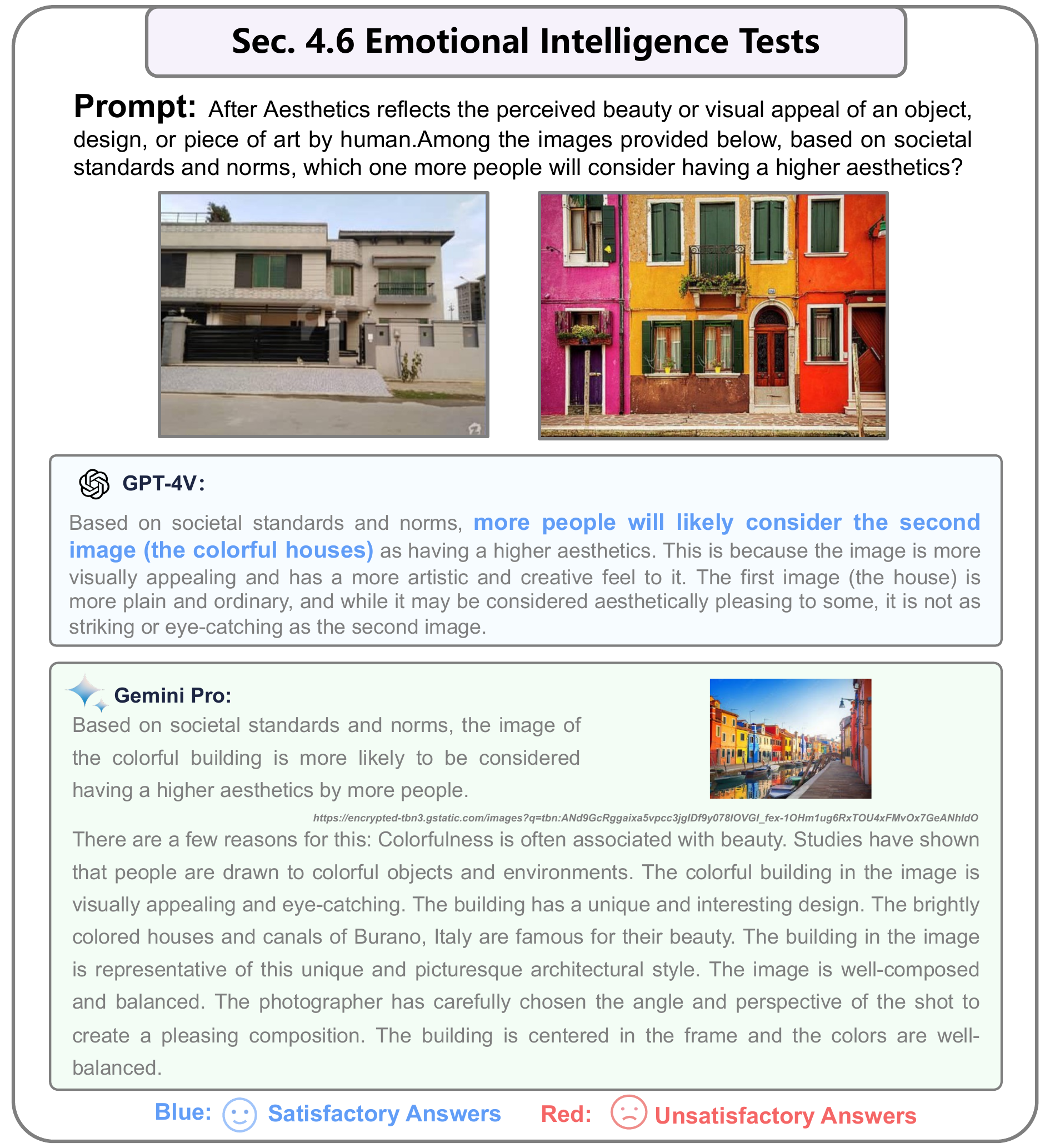}
    \vspace{0pt}
    \caption[Section 4.6 Emotional Intelligence Tests (2)]{\textbf{Aesthetics intelligence tests (1/2).} Both GPT-4V and Gemini can judges image aesthetics based on societal standards and norms.}
    \label{fig:Sec.4.6_2}
    \vspace{-10pt}
\end{figure}

\begin{figure}[H]
    \vspace{3cm}
    \centering
    \includegraphics[width=0.99\linewidth]{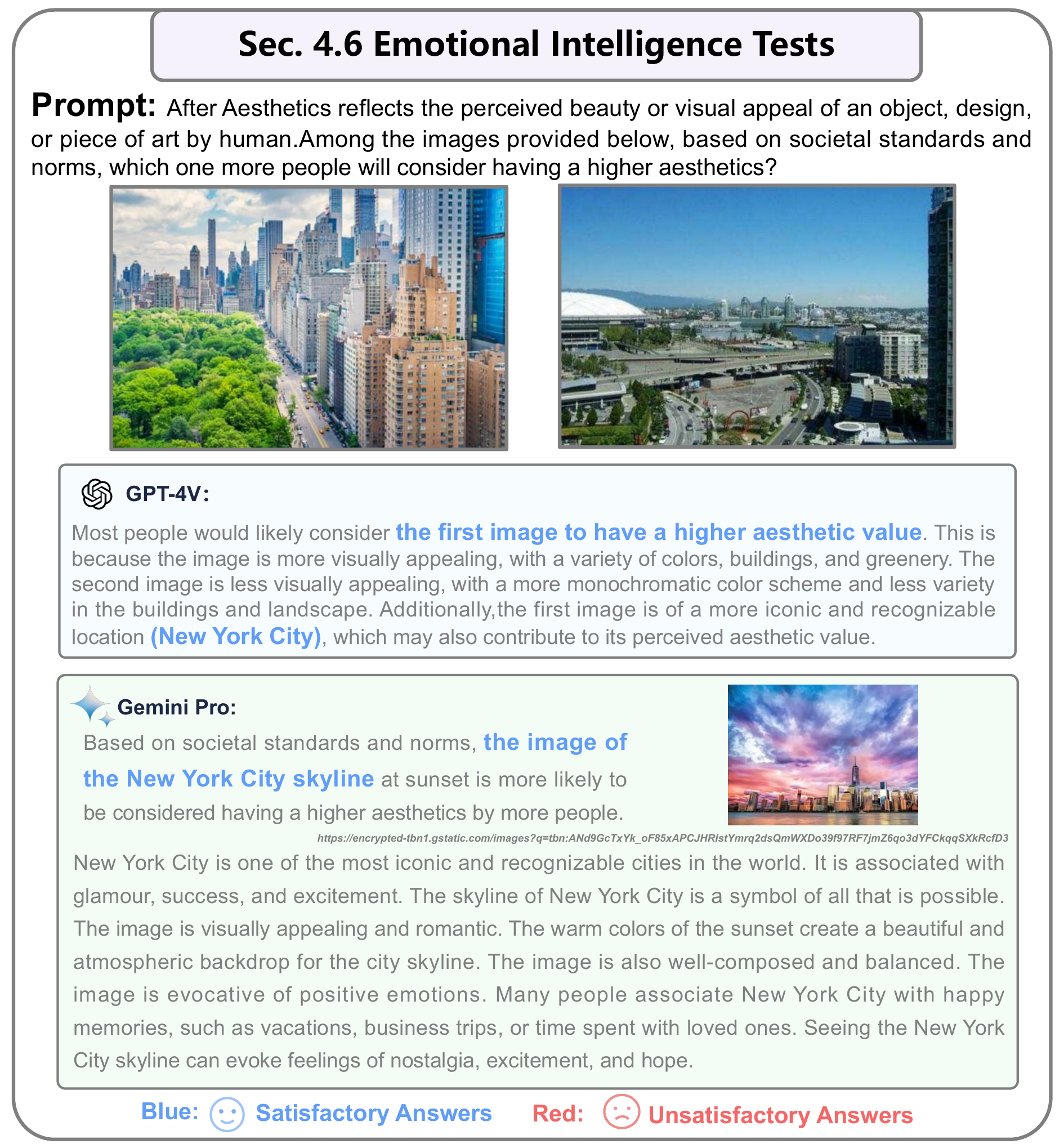}
    \vspace{0pt}
    \caption[Section 4.6 Emotional Intelligence Tests (3)]{\textbf{Aesthetics intelligence tests (2/2).} Both GPT-4V and Gemini can judges image aesthetics based on societal standards and norms.}
    \label{fig:Sec.4.6_3}
    \vspace{-10pt}
\end{figure}

\begin{figure}[H]
    \centering
    \vspace{2cm}
    \includegraphics[width=0.99\linewidth]{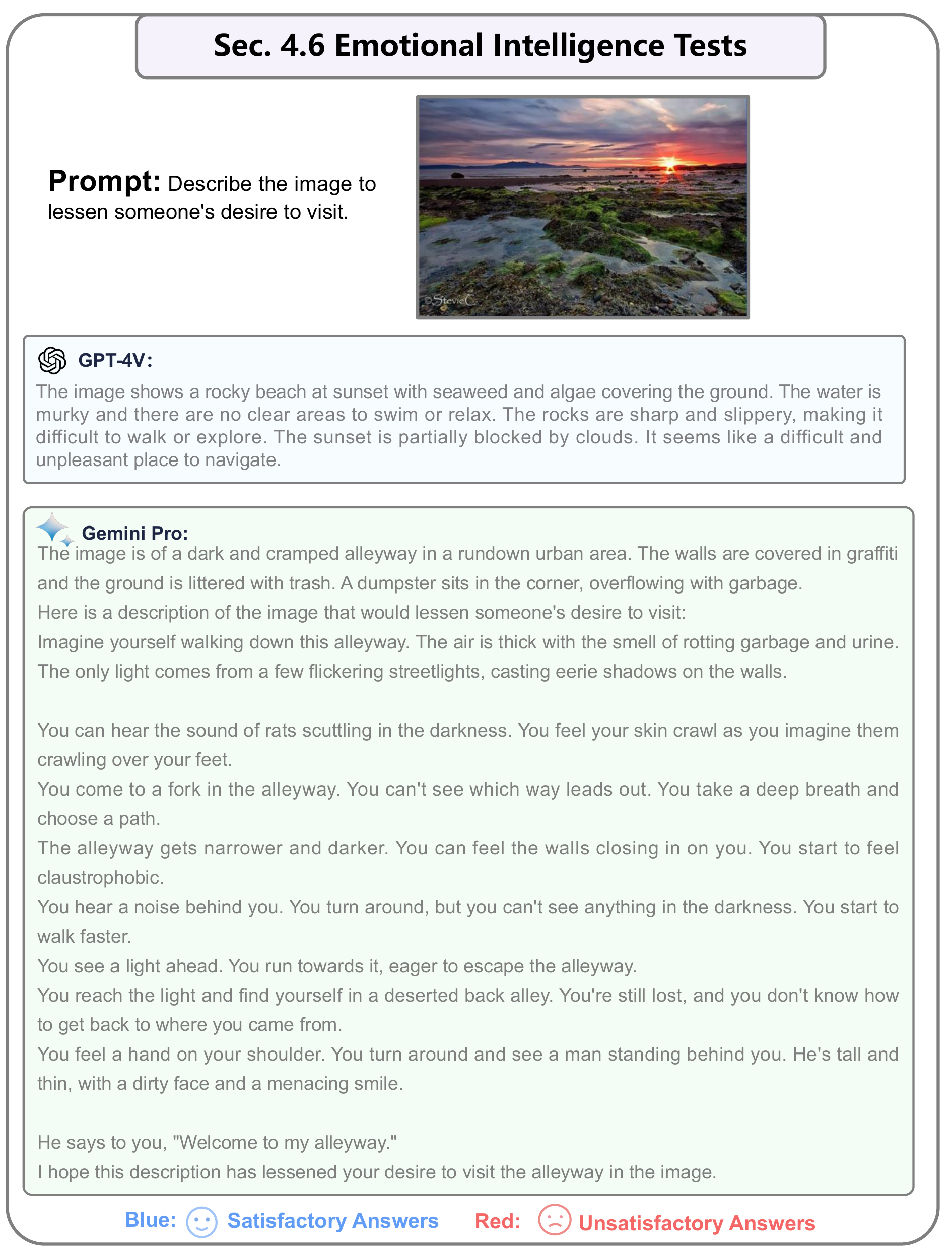}
    \vspace{0pt}
    \caption[Section 4.6 Emotional Intelligence Tests (4)]{\textbf{Lessen human desires.} Gemini can generate better text based on the perceived or desired emotions than GPT4-V.}
    \label{fig:Sec.4.6_4}
    \vspace{-10pt}
\end{figure}

\begin{figure}[H]
    \centering
    \vspace{3cm}
    \includegraphics[width=0.99\linewidth]{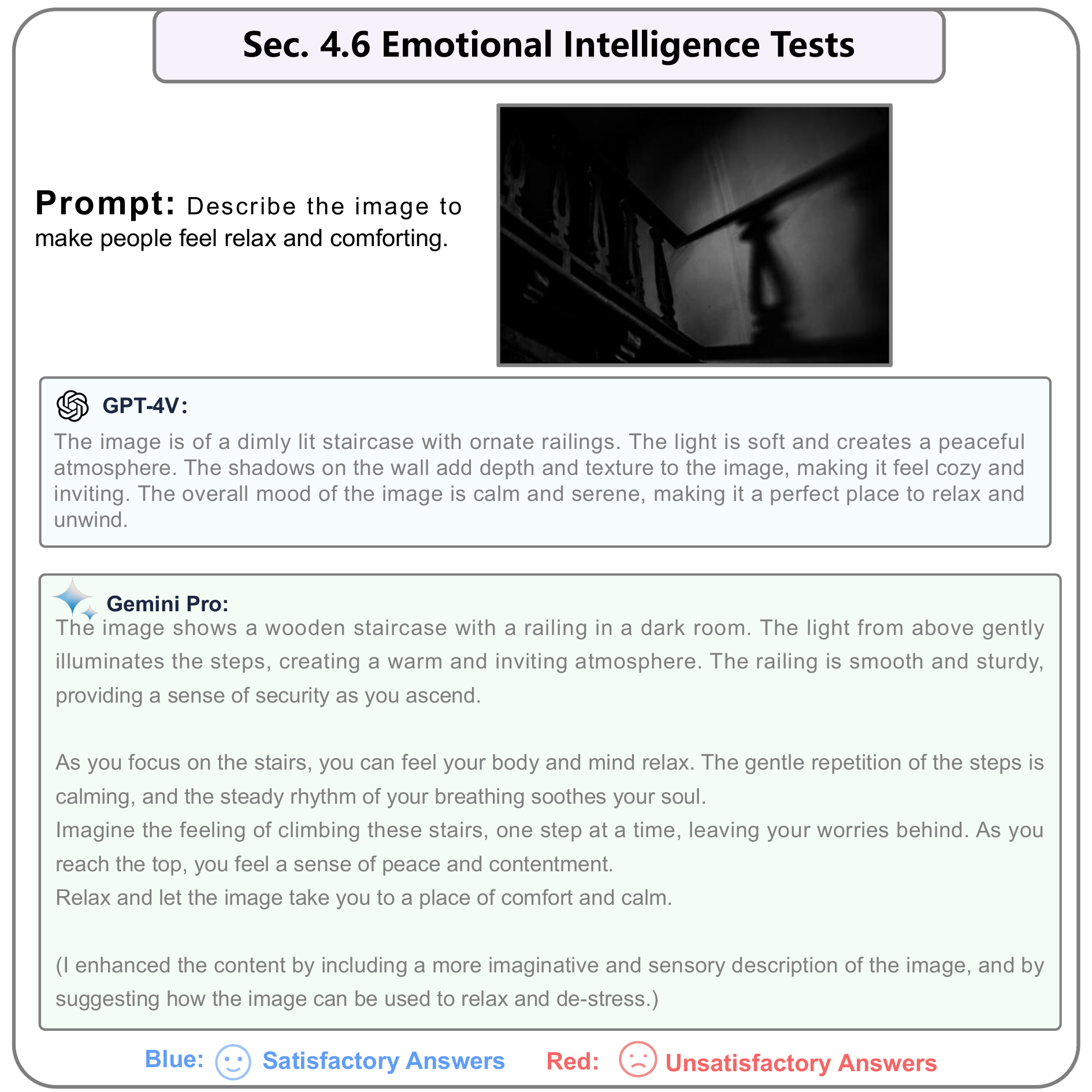}
    \vspace{0pt}
    \caption[Section 4.6 Emotional Intelligence Tests (5)]{\textbf{Soothing human emotions.} Gemini can generate better text based on the perceived or desired emotions than GPT4-V.}
    \label{fig:Sec.4.6_5}
    \vspace{-10pt}
\end{figure}

\cref{fig:Sec.4.6_1}-\cref{fig:Sec.4.6_5} showcase the models' EQ capabilities, focusing on understanding and outputting emotions, including aesthetic judgment. It was found that both models have a fairly accurate grasp of emotions and aesthetic judgment and are capable of producing different emotional outputs according to user requirements.

% 4.1 Humorous Image Understanding
\newpage

% 4.2 Multimodal Knowledge and Commonsense
\newpage

\newpage

\newpage

% 4.3 Detective Reasoning Ability
\newpage

% 4.4 Association of Parts and Objects
\newpage

% 4.5 Intelligence Tests
\newpage

\newpage

\newpage

\newpage

% 4.6 Emotional Intelligence Tests
\newpage

\newpage

\newpage

\newpage

\newpage

\newpage
\section{Textual Reasoning in Images}
\label{Sec.5 Textual Reasoning in Images}
This section is dedicated to showcasing the model's inferential prowess in interpreting textual elements within images. It serves as an extension of the discourse initiated in \cref{Sec.3 Text Recognition and Understanding in Images}, progressing from the fundamental recognition of text to more nuanced inferential analysis. In \cref{Sec.5.1 Visual Math Ability}, the focus is on the model's visual mathematical capabilities, demonstrating its proficiency in identifying texts and mathematical symbols in images and performing elementary computations. Both models under scrutiny exhibit competent responses to straightforward mathematical problems.
\cref{Sec.5.2 Table & Chart Understanding and Reasoning} explores the realm of graphical interpretation and reasoning, encompassing the analysis of flowcharts, bar charts, and other graphical data representations. While the models generally perform well on a broad spectrum of charts, the Gemini model shows a reduced accuracy in interpreting complex experimental tables.
Lastly, \cref{Sec.5.3 Document Understanding and Reasoning} assesses the models' ability to comprehend and reason with a variety of documents, including posters, architectural blueprints, academic papers, and webpages. Both models display parallel levels of performance, with Gemini providing more elaborate responses, albeit with a slight compromise in accuracy.
Pointing to a specific spatial location is anessential capability in human-computer interaction with multimodal systems, such as conducting visually grounded dialogues. research have try to find ways for this visual referring in many different ways~\cite{zhang2023gpt4roi,sun2023alpha}. In large Vision-Language model like GPT-4V and Gemini. It may be enough to simply draw illustration in original RGB pixel space~\cite{shtedritski2023does}. We thus intersperse a series of interactive prompts to evaluate the user-friendliness of the two models.

% 5.1 Visual Math Ability
\subsection{Visual Math Ability}
\label{Sec.5.1 Visual Math Ability}
\begin{figure}[H]
    \centering
    \includegraphics[width=0.99\linewidth]{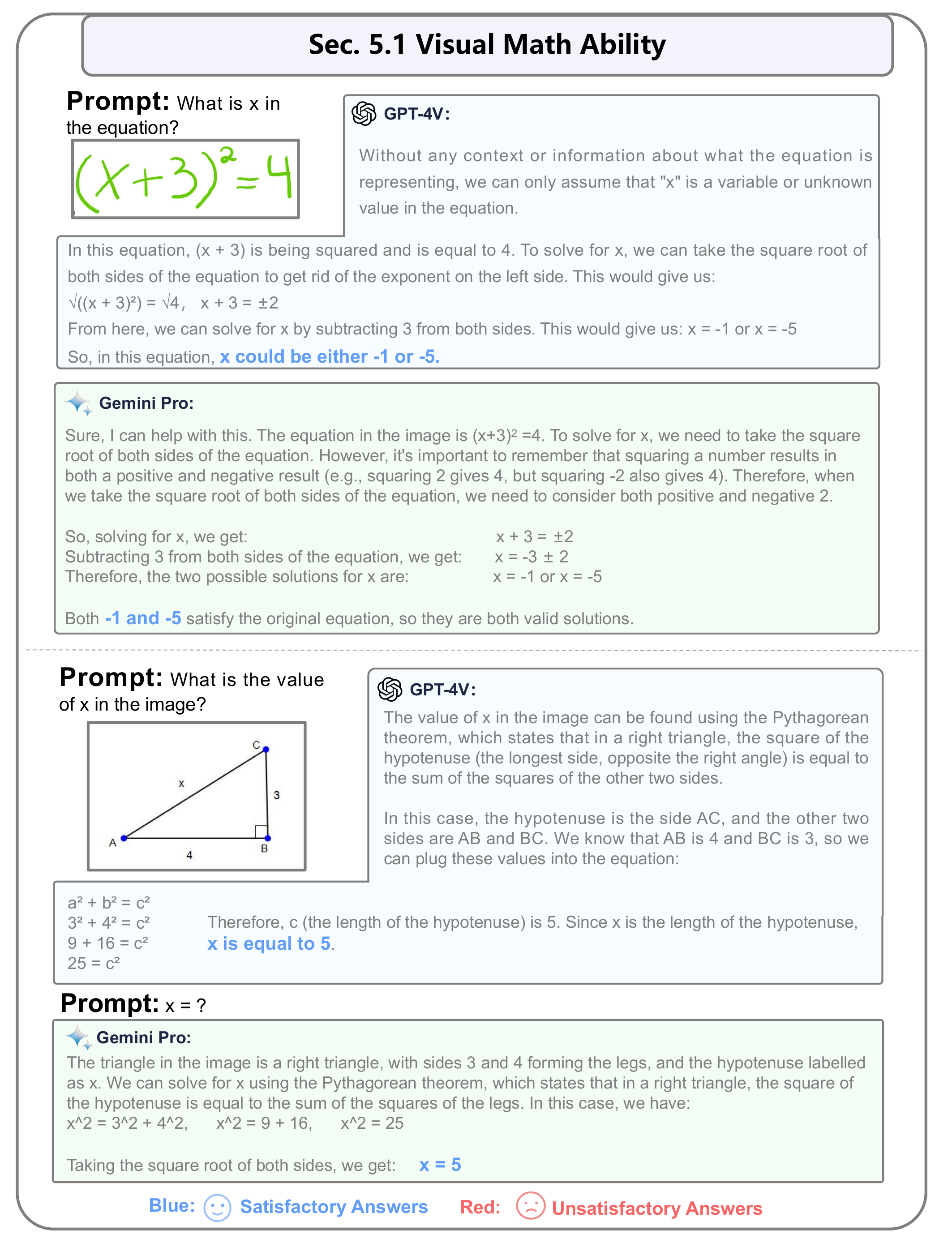}
    \vspace{0pt}
    \caption[Section 5.1 Visual Math Ability]{\textbf{Answering math problems.} Both GPT-4V and Gemini can understand the question textually and visually, and gather necessary information to answer the question. In the first case, there are errors in both GPT-4V's and Gemini’s description.}
    \label{fig:Sec.5.1_1}
    \vspace{-10pt}
\end{figure}

\cref{fig:Sec.5.1_1} demonstrates the model's mathematical abilities, highlighting not only its capacity to recognize textual and mathematical symbols within images but also its aptitude for performing calculations. We observed that for relatively simple problems, both models provided satisfactory solutions.

\subsection{Table \& Chart Understanding and Reasoning}
\label{Sec.5.2 Table & Chart Understanding and Reasoning}
\begin{figure}[H]
    \centering
    \includegraphics[width=0.90\linewidth]{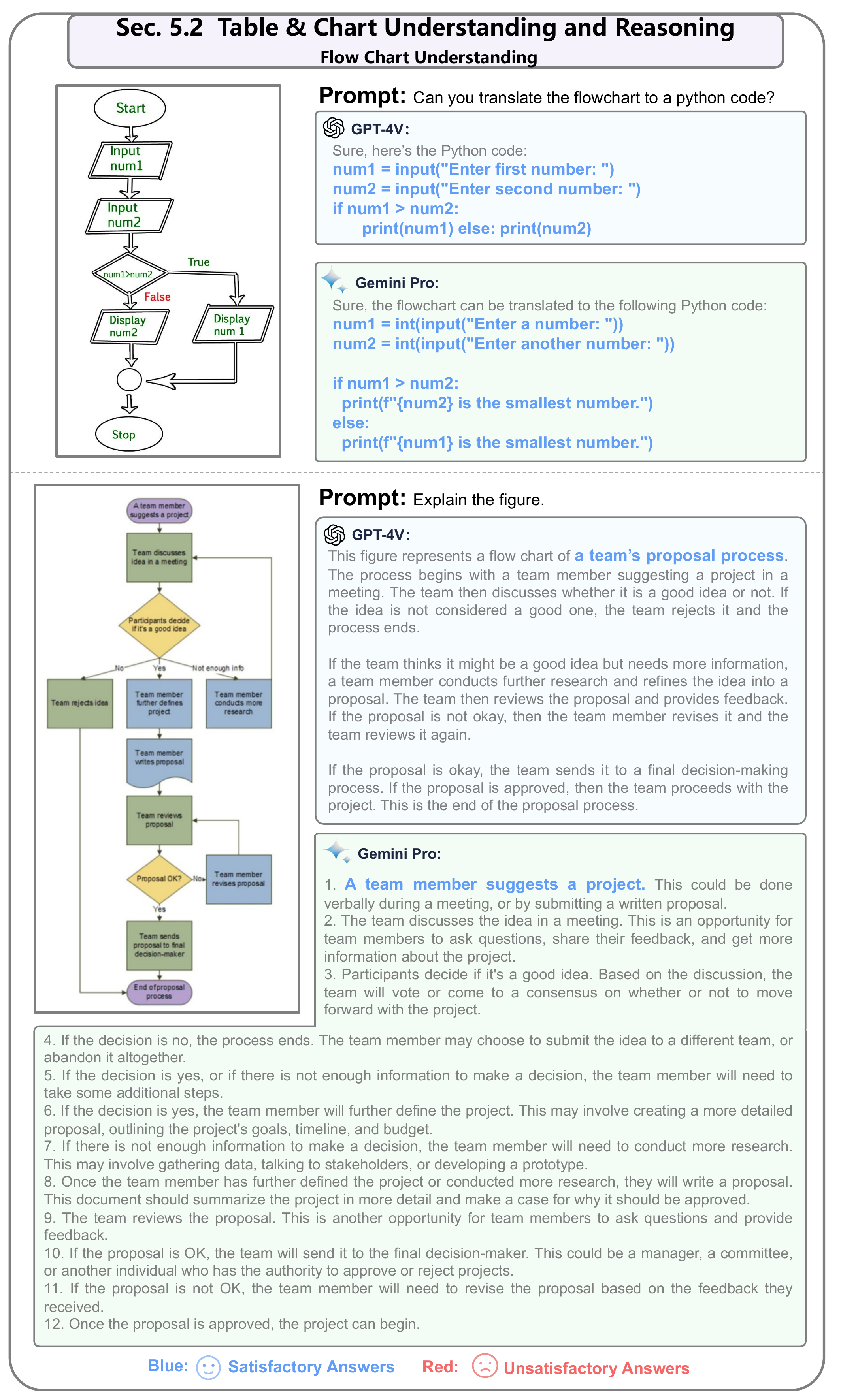}
    \vspace{-2pt}
    \caption[Section 5.2 Table \& Chart Understanding and Reasoning (1)]{\textbf{Flow chart understanding}. Both GPT-4V and Gemini correctly describe the proposal process in details.}
    \label{fig:Sec.5.2_1}
    \vspace{-10pt}
\end{figure}

\begin{figure}[H]
    \centering
    \includegraphics[width=0.99\linewidth]{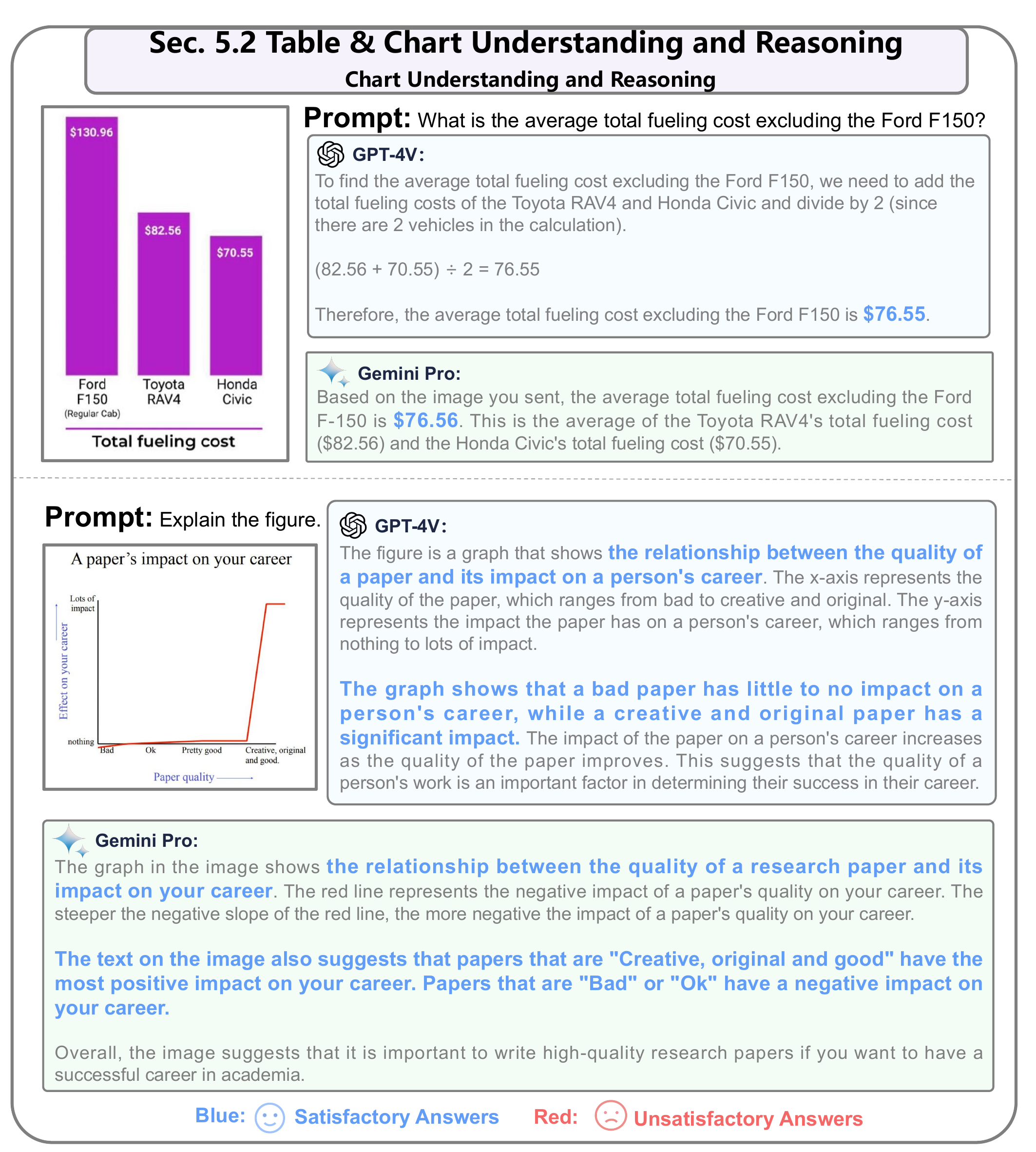}
    \vspace{0pt}
    \caption[Section 5.2 Table \& Chart Understanding and Reasoning (2)]{\textbf{Chart understanding and reasoning.} Both GPT-4V and Gemini show promising results in understanding the details in the chart, as well as in accurately responding to related questions.}
    \label{fig:Sec.5.2_2}
    \vspace{-10pt}
\end{figure}

\begin{figure}[H]
    \centering
    \includegraphics[width=0.99\linewidth]{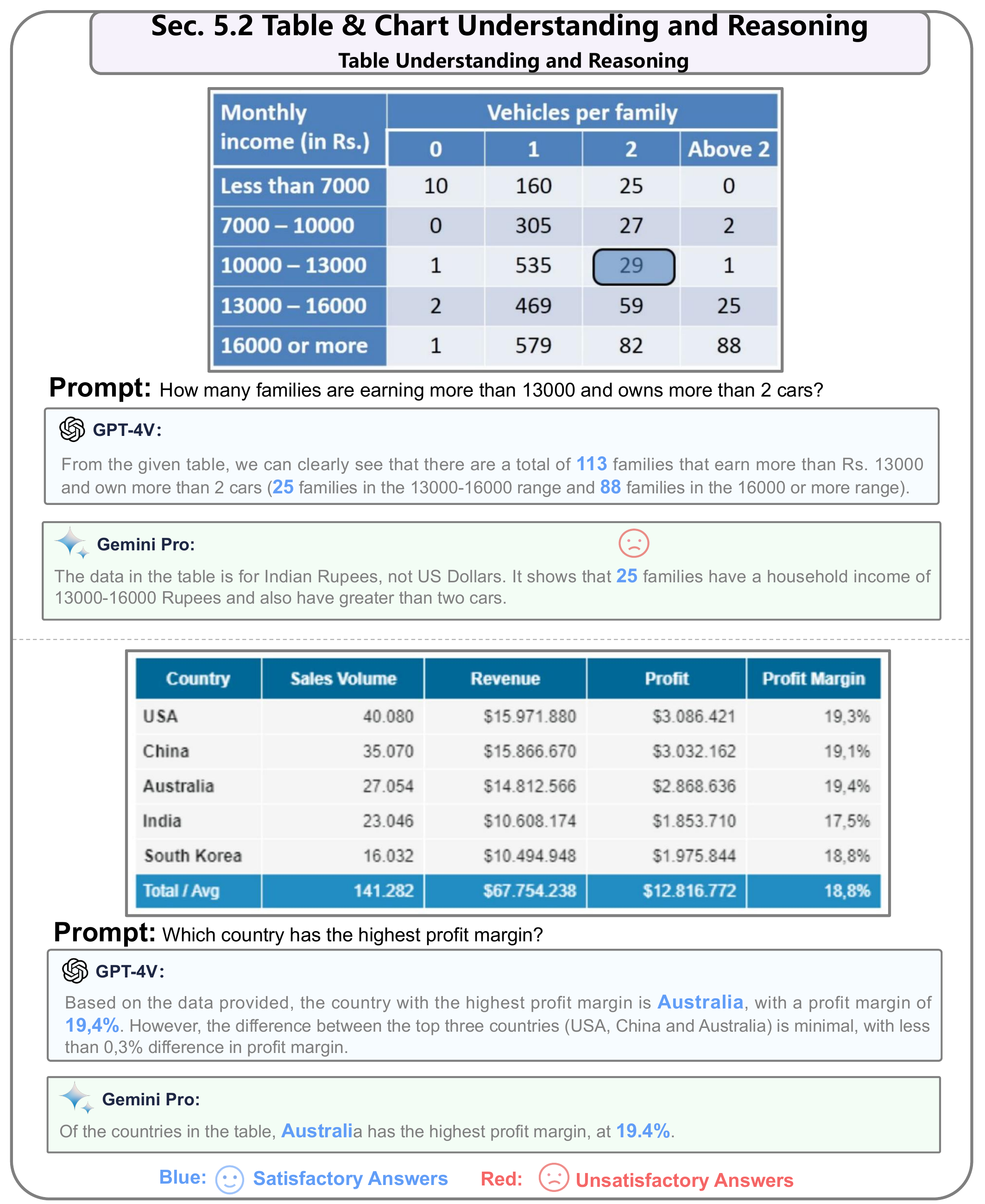}
    \vspace{0pt}
    \caption[Section 5.2 Table \& Chart Understanding and Reasoning (3)]{\textbf{Table understanding and reasoning.} Both GPT-4V and Gemini can understand the details in the table, and answer related questions.}
    \label{fig:Sec.5.2_3}
    \vspace{-10pt}
\end{figure}

\cref{fig:Sec.5.2_1}-\cref{fig:Sec.5.2_3} delve into the models' reasoning capabilities with respect to tables and charts. In these cases, beyond mere text recognition, a certain level of reasoning is necessitated. Our observations indicate that both models exhibit commendable performance in interpreting flowcharts and bar graphs. However, in the context of some experimental tables, the Gemini model struggles to provide accurate answers. This may also be attributed to image resolution issues, as the images are composed of multiple stitched pictures, making it challenging to accurately recognize text within the images.

\subsection{Document Understanding and Reasoning}
\label{Sec.5.3 Document Understanding and Reasoning}
\begin{figure}[H]
    \centering
    \includegraphics[width=0.99\linewidth]{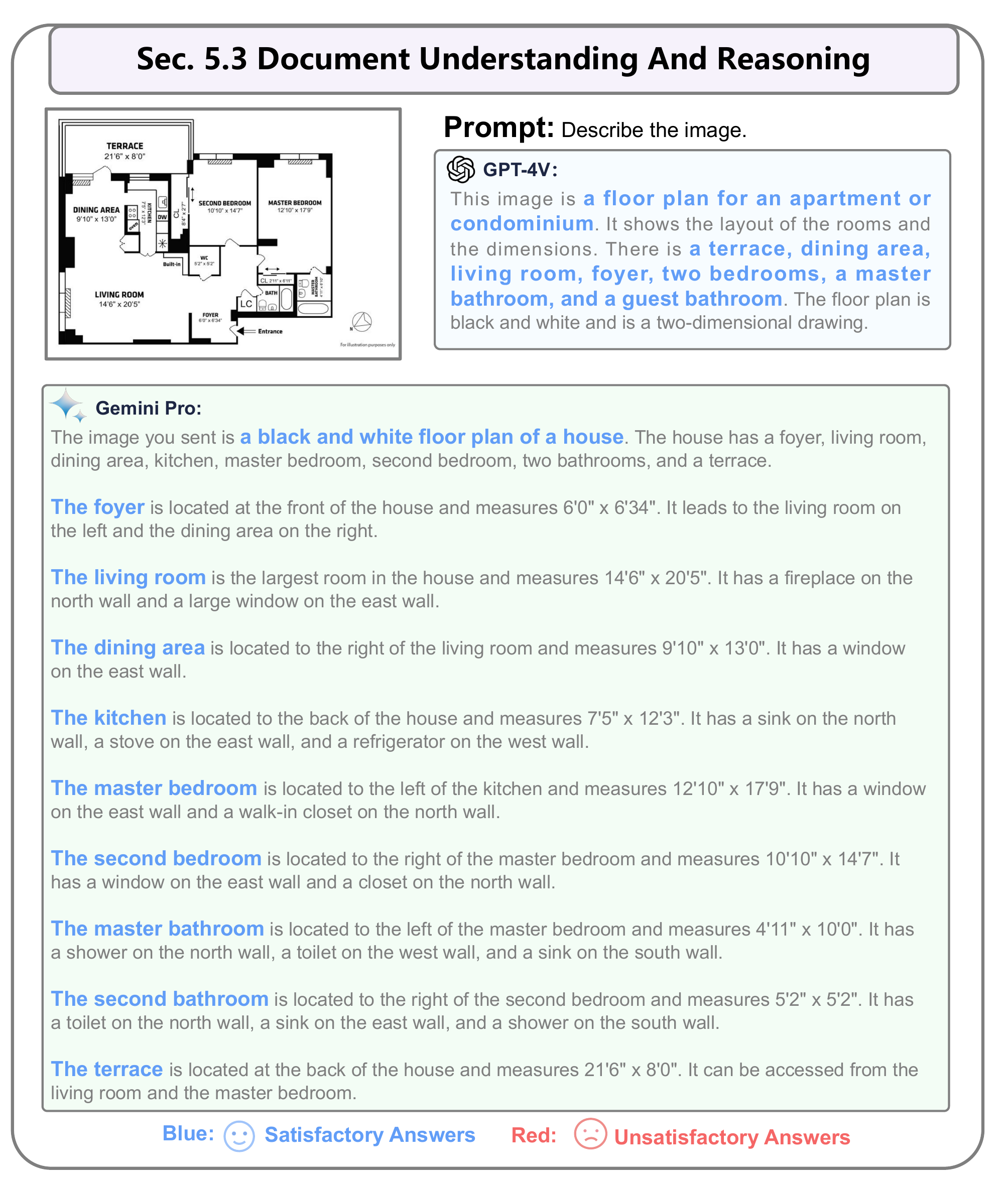}
    \vspace{0pt}
    \caption[Section 5.3 Document Understanding and Reasoning (1)]{\textbf{Document understanding and reasoning (1/3).} Both GPT-4V and Gemini can accurately identify the content on the document and answer questions.}
    \label{fig:Sec.5.3_1}
    \vspace{-10pt}
\end{figure}

\begin{figure}[H]
    \centering
    \includegraphics[width=0.99\linewidth]{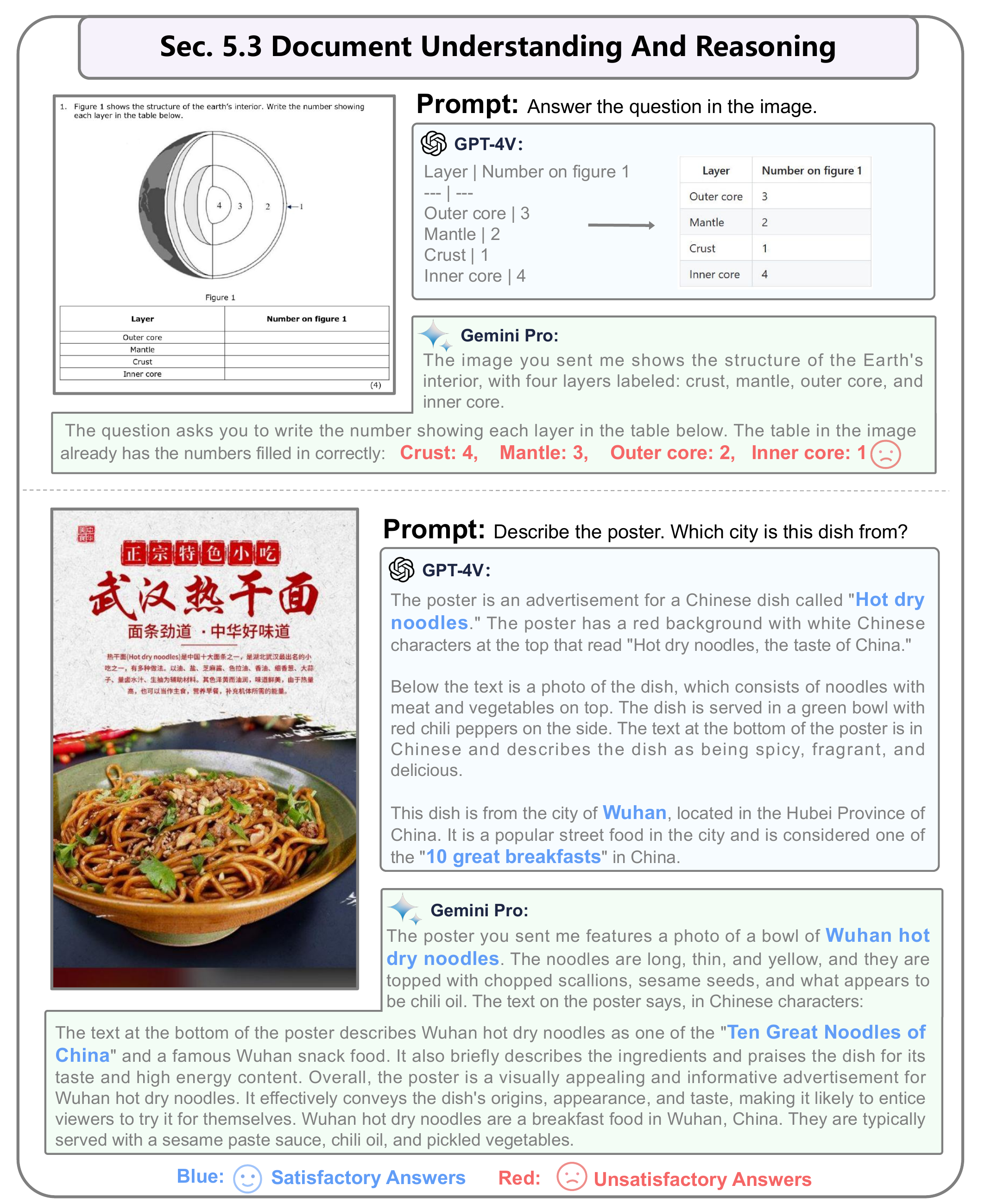}
    \vspace{0pt}
    \caption[Section 5.3 Document Understanding and Reasoning (2)]{\textbf{Document understanding and reasoning (2/3).} Both GPT-4V and Gemini can accurately identify the content on different types of document and answer questions.}
    \label{fig:Sec.5.3_2}
    \vspace{-10pt}
\end{figure}

\begin{figure}[H]
    \centering
    \includegraphics[width=0.99\linewidth]{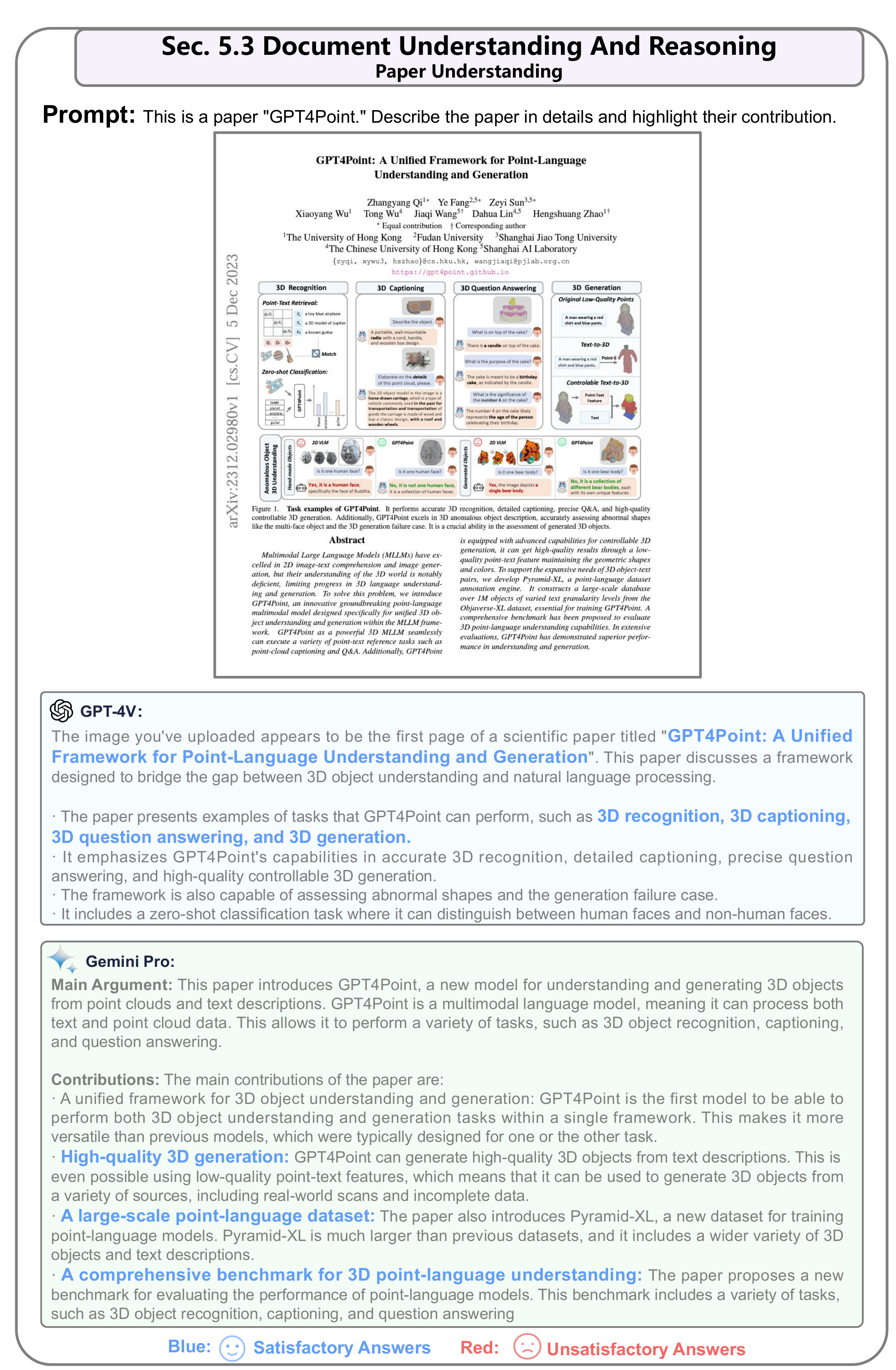}
    \vspace{0pt}
    \caption[Section 5.3 Document Understanding and Reasoning (3)]{\textbf{Document understanding and reasoning (3/3).} Both GPT-4V and Gemini can read the page of a technical report and provide a summary of the content.}
    \label{fig:Sec.5.3_3}
    \vspace{-10pt}
\end{figure}

\begin{figure}[H]
    \centering
    \includegraphics[width=0.99\linewidth]{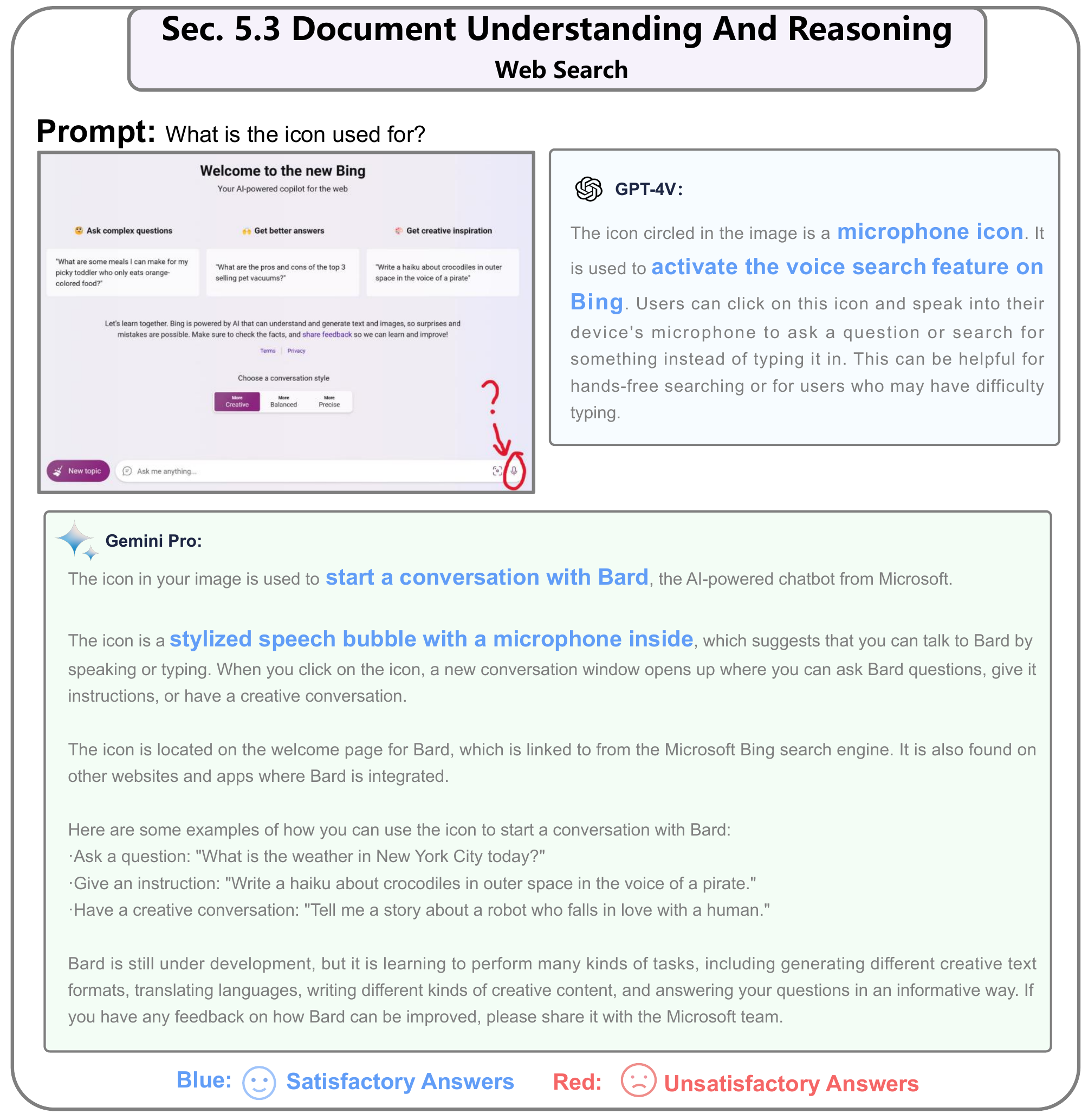}
    \vspace{0pt}
    \caption[Section 5.3 Document Understanding and Reasoning (4)]{\textbf{Visual referring prompting in computer GUI.} Both Gemini and GPT-4V success in helping user to understand the web page.}
    \label{fig:Sec.5.3_4}
    \vspace{-10pt}
\end{figure}

\cref{fig:Sec.5.3_1}-\cref{fig:Sec.5.3_4} examine the models' capability to infer from a diverse range of documents, encompassing materials such as posters, architectural layouts, scholarly articles, and web pages. Both models exhibit comparable efficacy in these tasks. Notably, Gemini tends to offer more elaborate responses, yet it falls short in terms of precision, underscoring an area for potential improvement.

\newpage
\vspace*{\fill}
\vfill

% 5.2 Table & Chart Understanding and Reasoning
\newpage
\vspace*{\fill}
\vfill

\newpage
\vspace*{\fill}
\vfill

\newpage
\vspace*{\fill}
\vfill

\newpage
\vspace*{\fill}
\begin{figure}[H]
    \centering
    \includegraphics[width=0.99\linewidth]{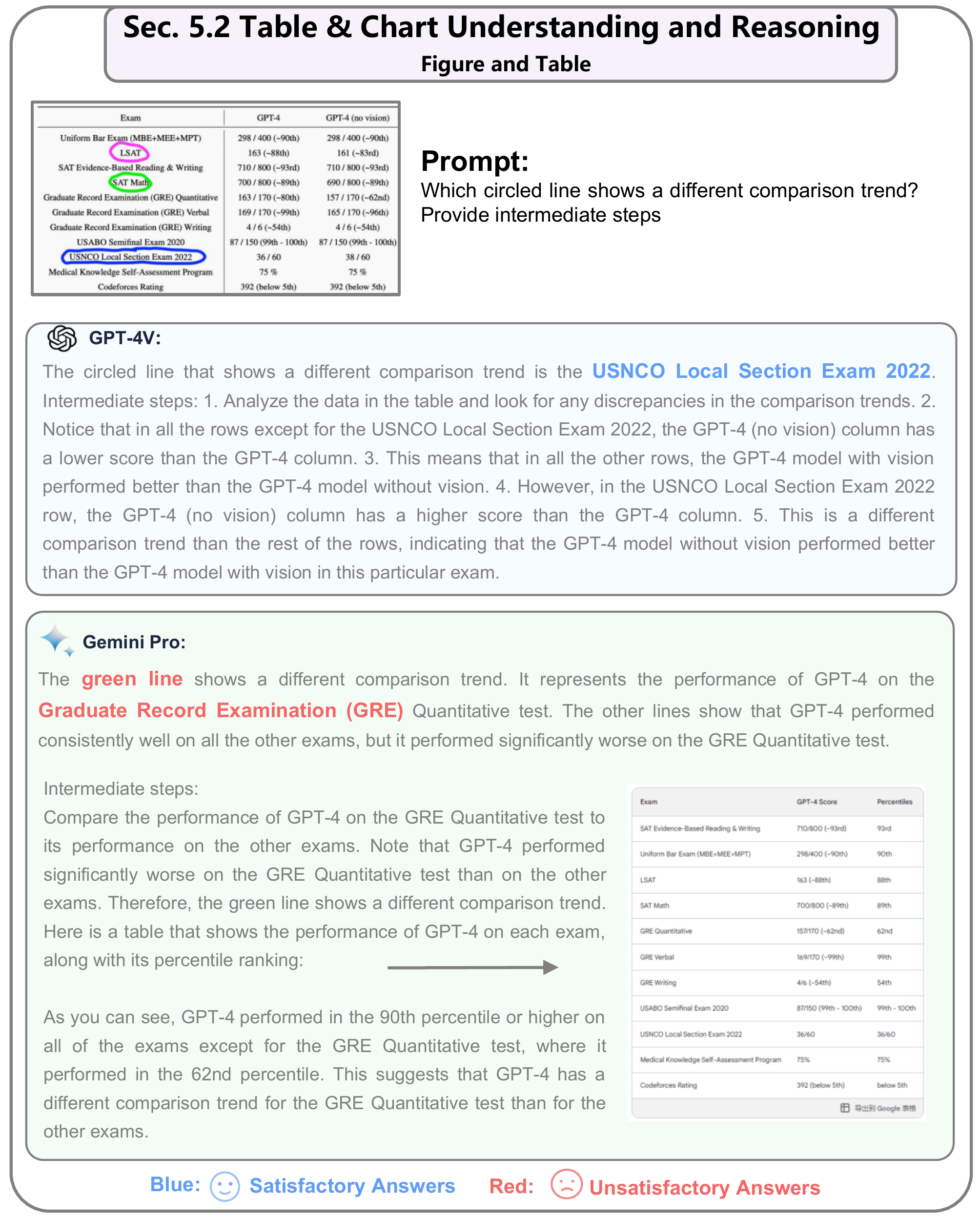}
    \vspace{0pt}
    \caption[Section 5.2 Table \& Chart Understanding and Reasoning (4)]{\textbf{Visual referring prompting.} When highlighting lines in documents and tables with user drawn circle, GPT-4V generate better result than Gemini.}
    \label{fig:Sec.5.2_4}
    \vspace{-10pt}
\end{figure}
\vfill

\newpage
\vspace*{\fill}
\begin{figure}[H]
    \centering
    \includegraphics[width=0.99\linewidth]{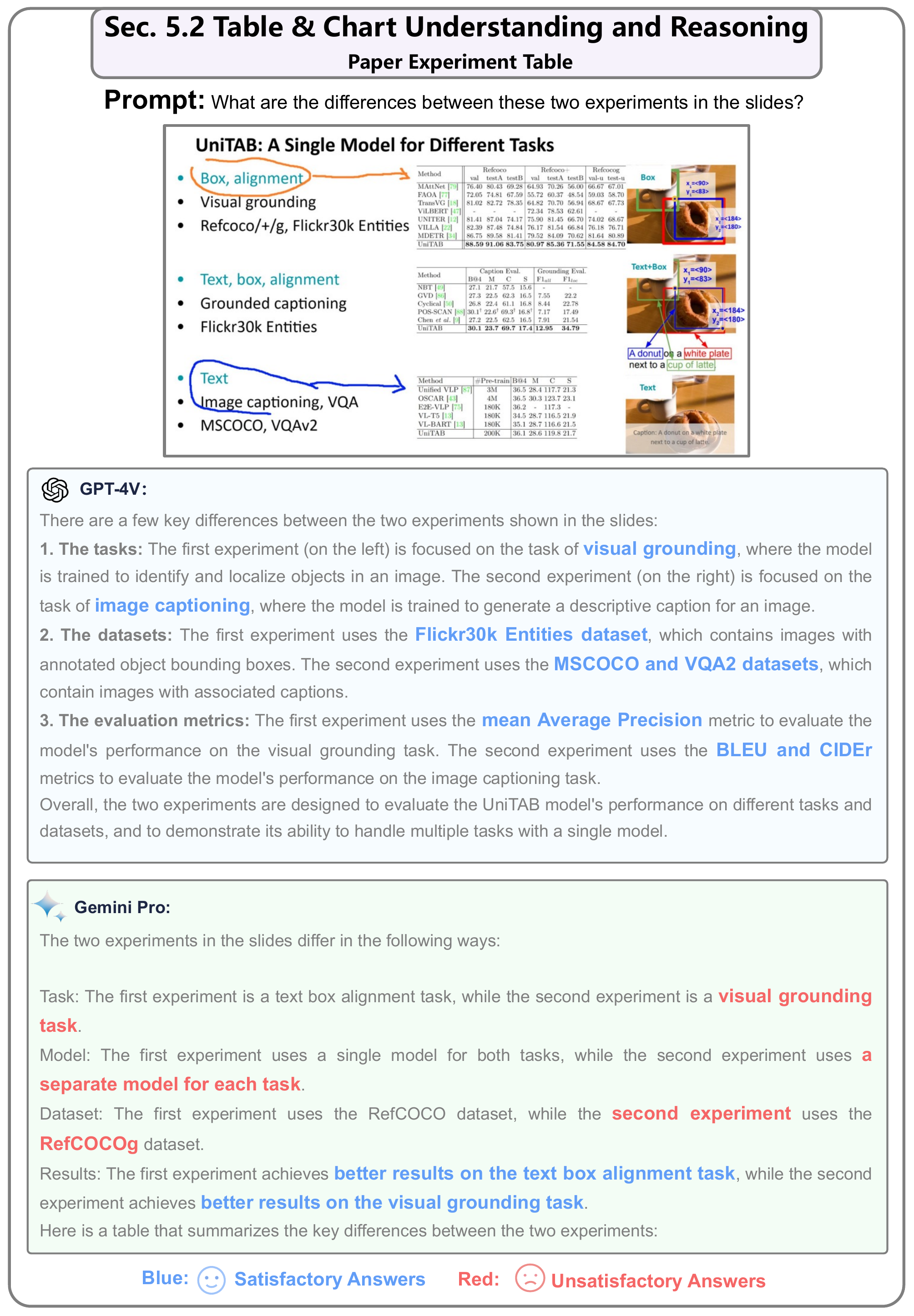}
    \vspace{0pt}
    \caption[Section 5.2 Table \& Chart Understanding and Reasoning (5)]{\textbf{Visual referring prompting in computer slides.} GPT-4V can better understand user mark than Gemini to help user to understand scientific confernce slides.}
    \label{fig:Sec.5.2_5}
    \vspace{-10pt}
\end{figure}
\vfill

% 5.3 Document Understanding and Reasoning
\newpage
\vspace*{\fill}
\vfill

\newpage
\vspace*{\fill}
\vfill

\newpage
\vspace*{\fill}
\vfill

\newpage
\vspace*{\fill}
\vfill

\newpage
\section{Integrated Image and Text Understanding}
\label{Sec.6 Integrated Image and Text Understanding}
This section encompasses tasks that require simultaneous recognition and understanding of both text and images, involving multifaceted analysis of multiple figures. In \cref{Sec.6.1 Interleaved Image-text Inputs}, we explore interleaved image-text inputs, showcasing a task that involves processing three bills to compute a total tax amount. GPT-4V demonstrates its capability to sequentially process each of the three images and provide accurate answers. In contrast, Gemini encounters comprehension challenges when all three images are inputted simultaneously, although it exhibits improved understanding when processing each image individually. The second task involves combining information from two images to calculate a total price, yielding results consistent with the previous task.
\cref{Sec.6.2 Text-to-Image Generation Guidance} is about text-to-image generation guidance, involving interactions and modifications with a text-to-image generative model. Gemini shows promising results from the initial stages, evidencing its proficiency in interacting with generative tools.

\subsection{Interleaved Image-text Inputs}
\label{Sec.6.1 Interleaved Image-text Inputs}
\begin{figure}[H]
    \centering
    \includegraphics[width=0.99\linewidth]{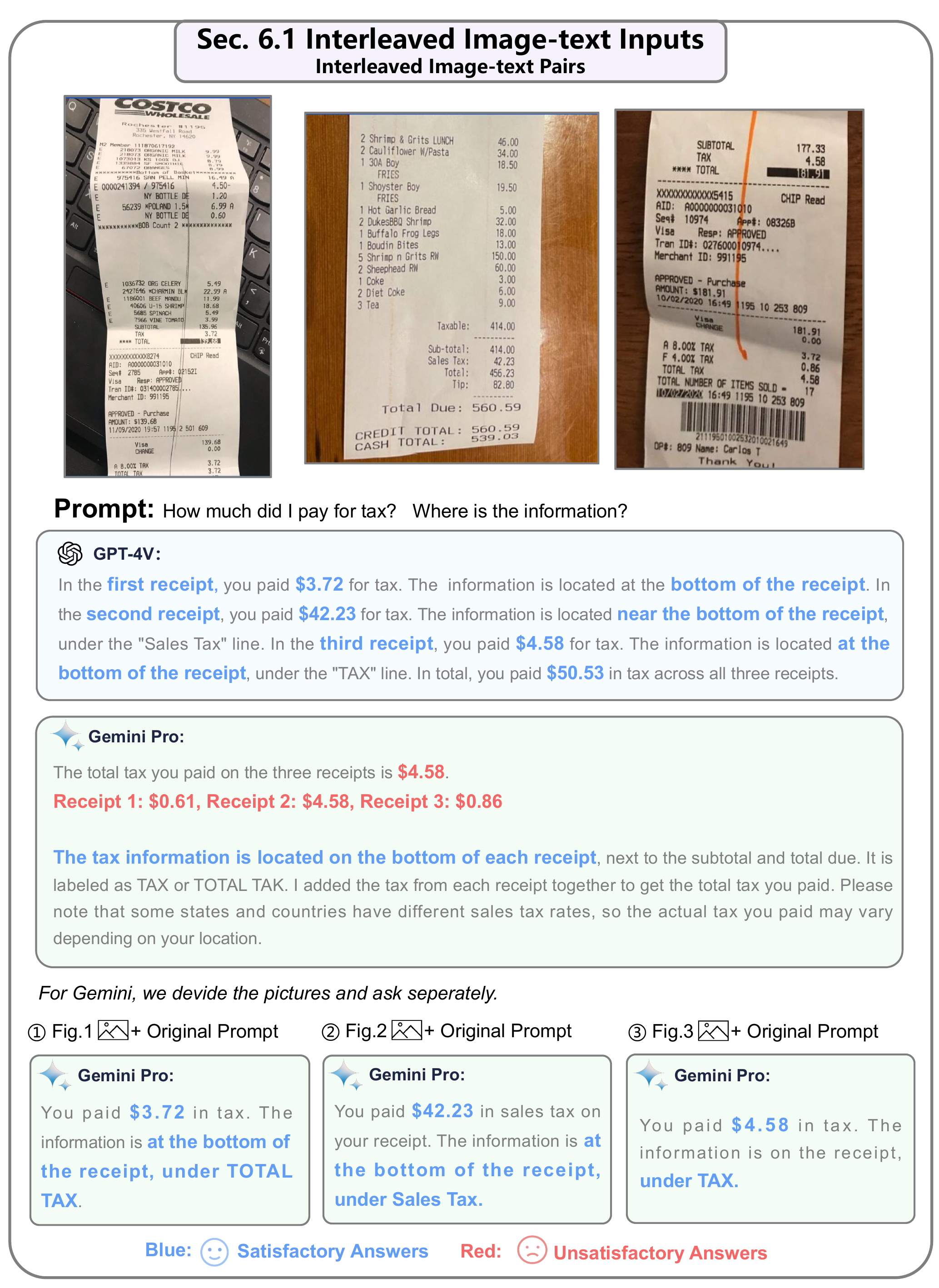}
    \vspace{-2pt}
    \caption[Section 6.1 Interleaved Image-text Inputs (1)]{
    \textbf{Interleaved image-text pairs.} This image involves inputting multiple bills and calculating the taxes within them. GPT-4V can accurately provide output, while Gemini's responses to the concatenated image are not satisfactory. However, it performs well in providing individual answers for each image.}
    \label{fig:Sec.6.1_1}
    \vspace{-10pt}
\end{figure}

\begin{figure}[H]
    \centering
    \includegraphics[width=0.99\linewidth]{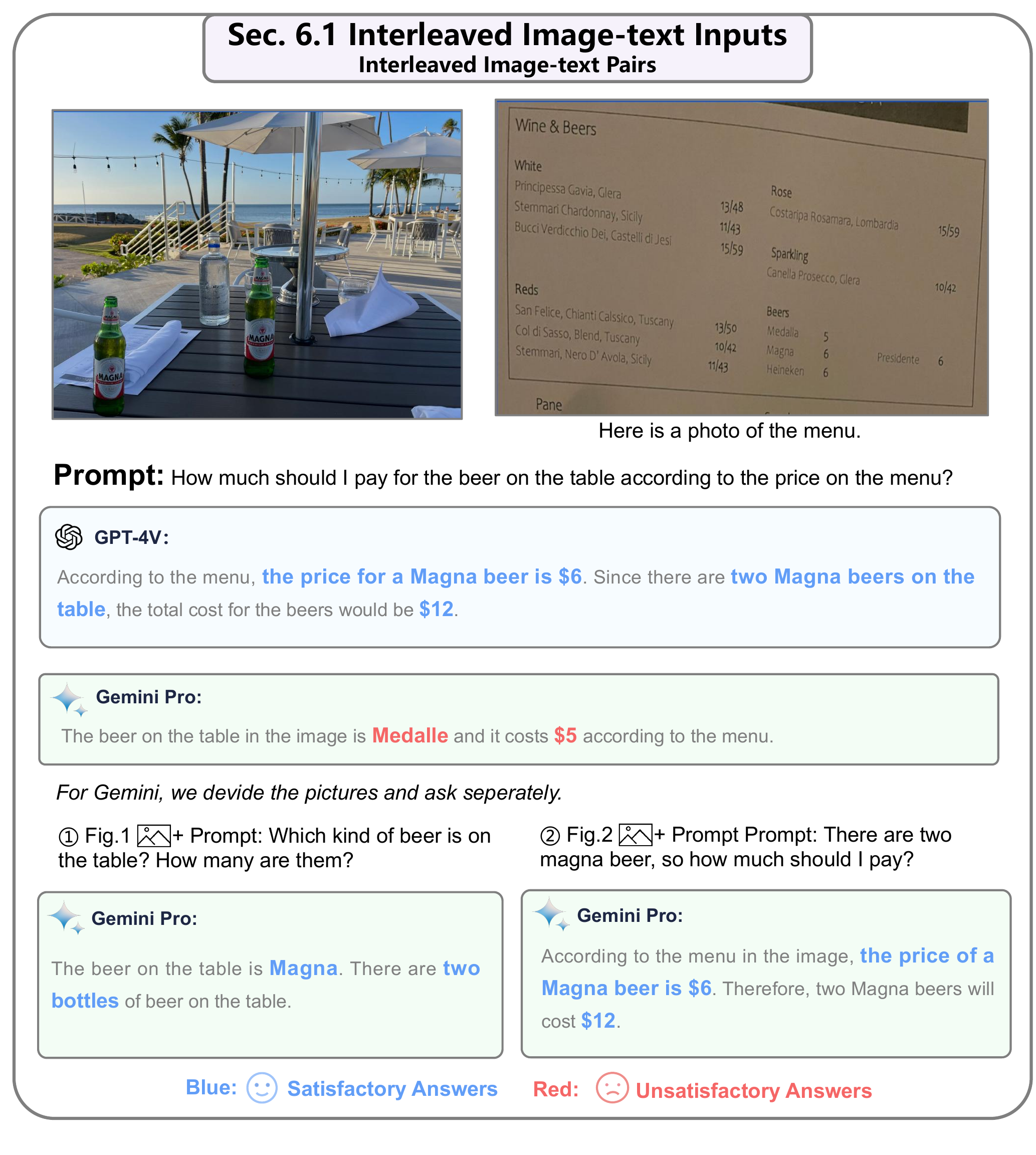}
    \vspace{-10pt}
    \caption[Section 6.1 Interleaved Image-text Inputs (2)]{\textbf{Interleaved image-text pairs.} This image involves inputting tables with wine and beers and calculating the the cost of them on a menu. GPT-4V can accurately provide output, while Gemini's responses to the concatenated image are not satisfactory. However, it performs well in providing individual answers for each image.}
    \label{fig:Sec.6.1_2}
    \vspace{-10pt}
\end{figure}

\cref{fig:Sec.6.1_1} presents a task involving three bills, where the objective is to compute the cumulative tax amount of the bills. It was observed that GPT-4V could effectively process each bill sequentially, yielding precise responses. Conversely, when all three bills were inputted simultaneously into Gemini, it failed to comprehend accurately. However, Gemini demonstrated improved understanding when each bill was inputted and processed individually.
\cref{fig:Sec.6.1_2} focuses on a task that requires calculating the total cost by combining the price of beer from the first image with the menu items from the second image. The findings mirror those from the previous figure, further corroborating the models' distinct capabilities in handling image-based input.

\subsection{Text-to-Image Generation Guidance}
\label{Sec.6.2 Text-to-Image Generation Guidance}
\begin{figure}[H]
    \centering
    \includegraphics[width=0.99\linewidth]{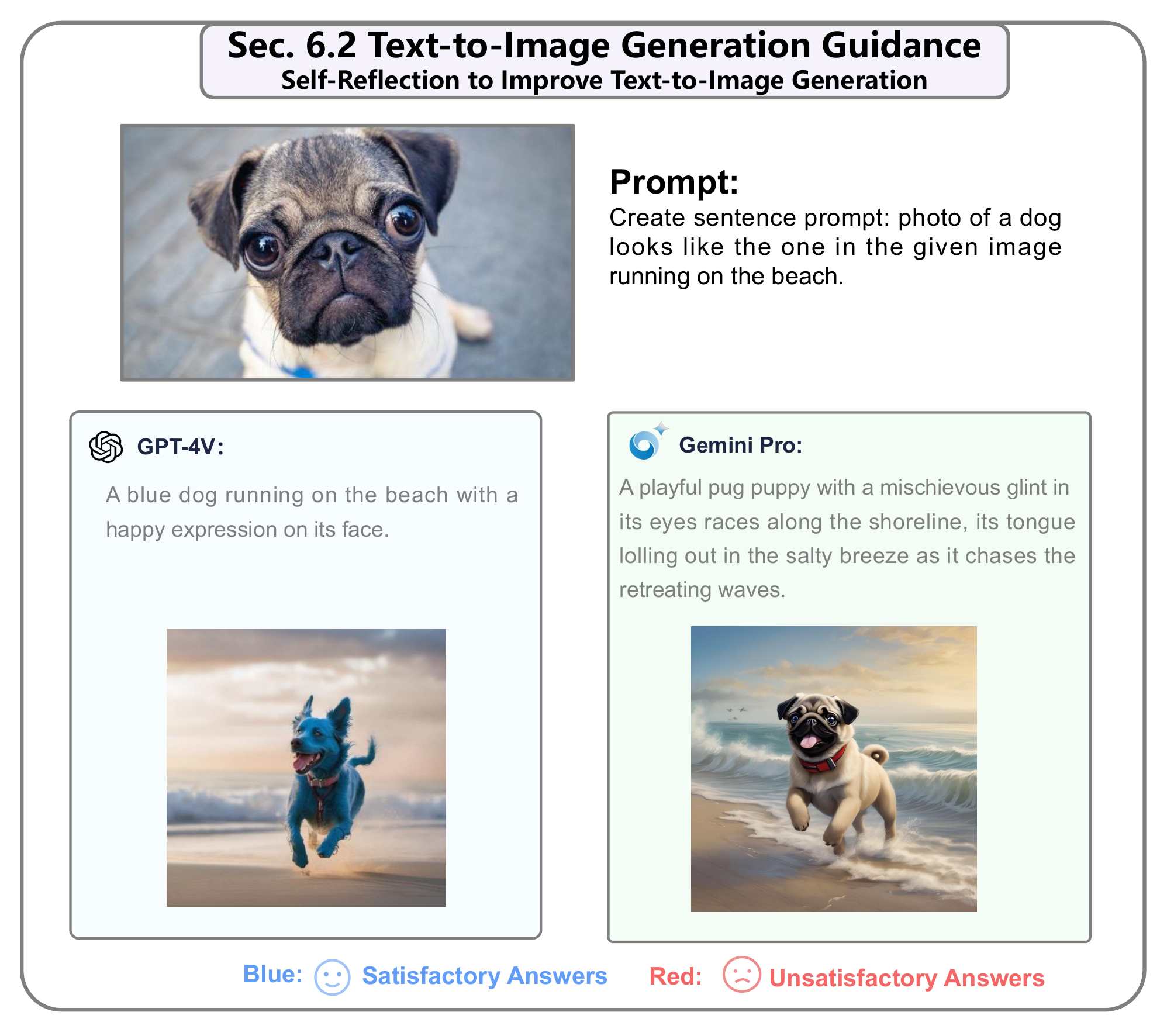}
    \vspace{-6pt}
    \caption[Section 6.2 Text-to-Image Generation Guidance (1)]{\textbf{Self-reflection to improve text-to-image generation (1/2).} Using self-reflection to improve the generated text prompts for a text-to-image model SDXL\cite{SDXL}, GPT-4V reflects the error in the initial prompt that it does not mention the dog’s breed, and makes the correct revision, while Gemini generate better one with breed mentioned.}
    \label{fig:Sec.6.2_1}
    \vspace{-10pt}
\end{figure}

\begin{figure}[H]
    \centering
    \includegraphics[width=0.8\linewidth]{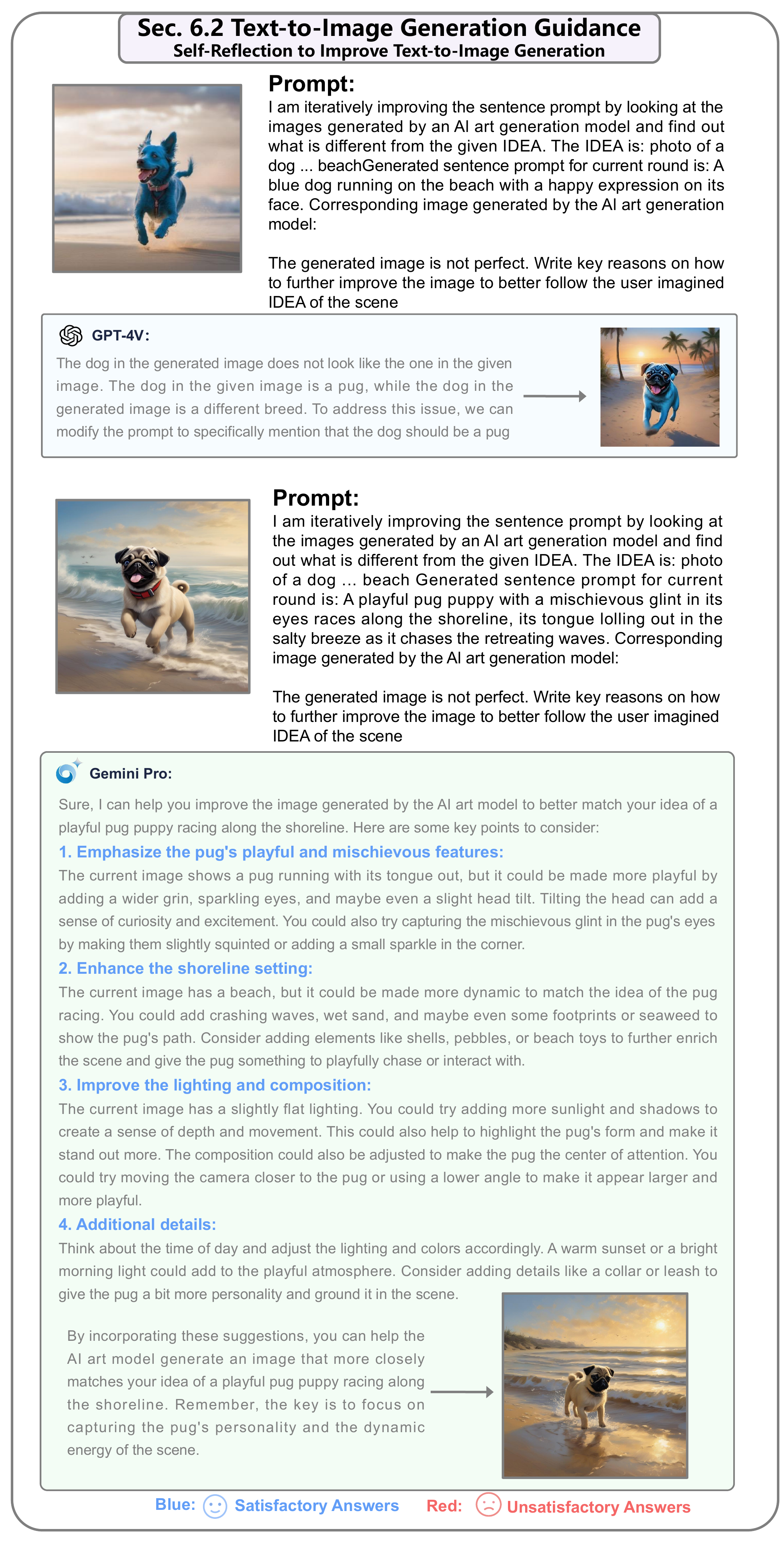}
    \vspace{-6pt}
    \caption[Section 6.2 Text-to-Image Generation Guidance (2)]{\textbf{Self-reflection to improve text-to-image generation (2/2).} We persistently optimize the prompts to enable both Gemini and GPT-4V to generate images that closely align with ground truth representations.}
    \label{fig:Sec.6.2_2}
    \vspace{-10pt}
\end{figure}

In the illustrations of \cref{fig:Sec.6.2_1}-\cref{fig:Sec.6.2_2}, the focus shifts to tasks involving interaction and modification with a text-to-image generative model. Notably, Gemini achieves satisfactory outcomes from the outset. This section highlights Gemini's superior capability in engaging with generative tools, as evidenced by the initial interactions and the quality of the resultant images. The figures detail the nuanced processes and outcomes of these interactions, underscoring Gemini's adeptness in understanding and manipulating generative image models.

% 6.1 Interleaved Image-text Inputs
\newpage
\vspace*{\fill}
\vfill

\newpage
\vspace*{\fill}
\vfill

% 6.2 Text-to-Image Generation Guidance
\newpage
\vspace*{\fill}
\vfill

\newpage

\newpage
\begin{figure}[H]
    \vspace{-20pt}
    \centering
    \includegraphics[width=0.75\linewidth]{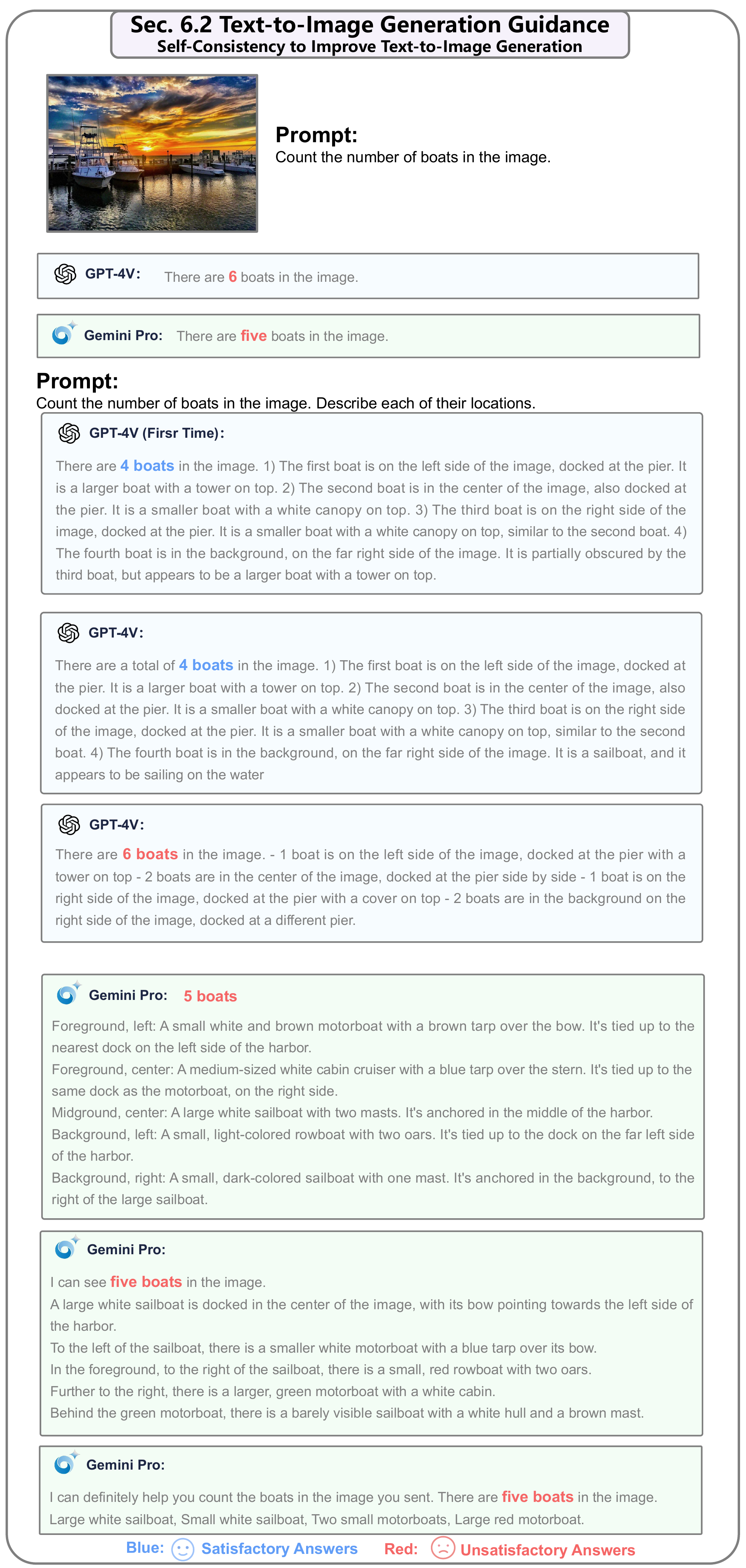}
    \vspace{-3pt}
    \caption[Section 6.2 Text-to-Image Generation Guidance (3)]{\textbf{Counting reliability with self-consistency.} This task aggregates multiple counting results repeated on the same image. GPT-4V's four responses provided three different answers, even though two of them were correct; in contrast, all four of Gemini's responses were consistent, albeit incorrect. This suggests that Gemini exhibits stronger consistency in its results.}
    \label{fig:Sec.6.2_3}
    \vspace{-10pt}
\end{figure}

\newpage
\section{Object Localization}
\label{Sec.7 Object Localization}
\vspace{-0.5mm}
This section is primarily focused on the localization of specific objects within images, requiring the models to provide the relative percentage coordinates of the objects. \cref{Sec.7.1 Object localization in real-world} discusses object localization in real-world scenarios: In tasks involving object localization in real-world settings, both models demonstrate commendable performance. \cref{Sec.7.2 Abstract Image Localization} addresses abstract image localization: In tasks that involve localizing specific sections within abstract images, such as identifying particular parts in Tangram figures, GPT-4V exhibits better accuracy in providing bounding box coordinates.

\subsection{Object localization in real-world}
\label{Sec.7.1 Object localization in real-world}
\begin{figure}[h]
    \centering
    \vspace{-6pt}
    \includegraphics[width=0.86\linewidth]{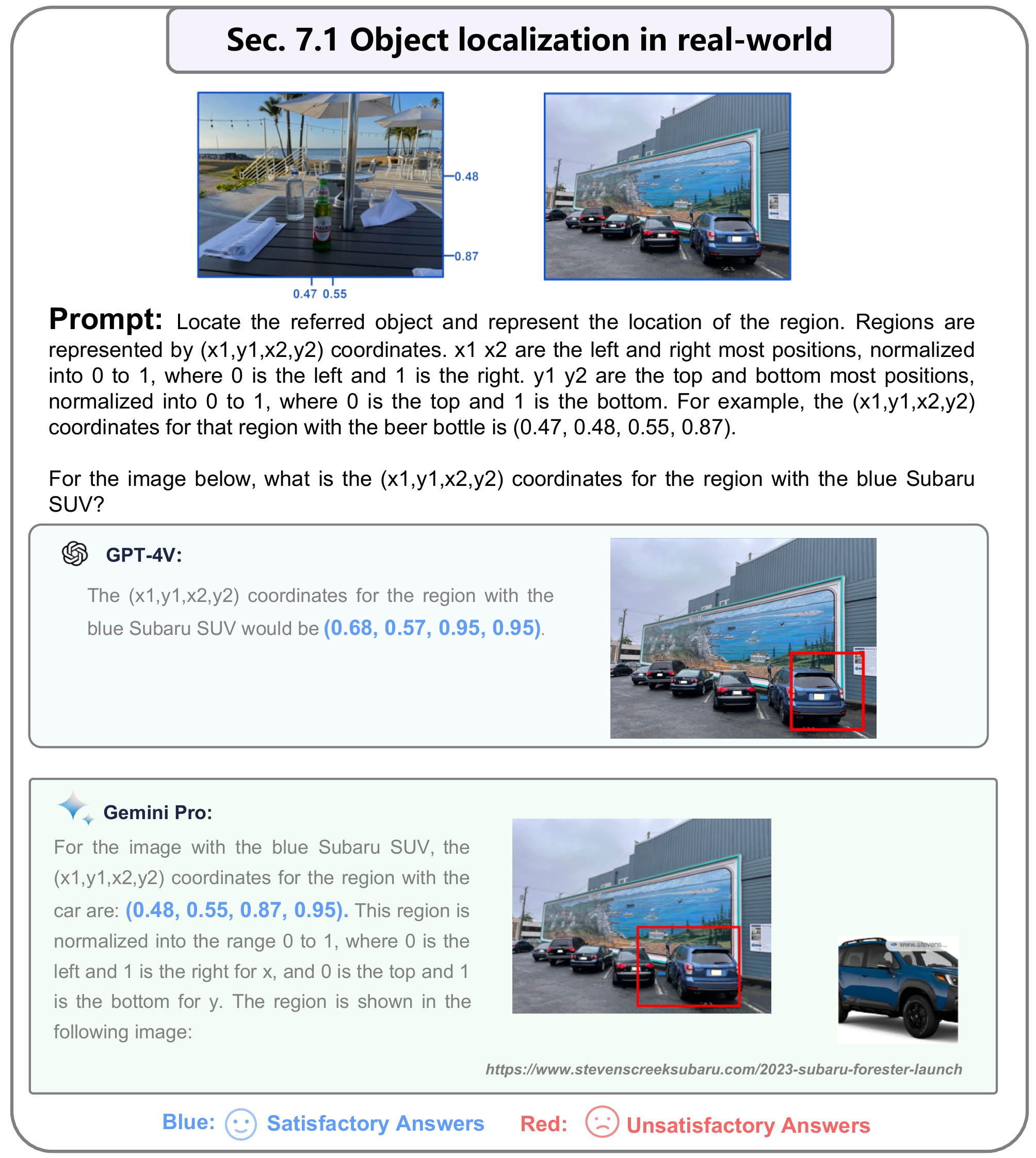}
    \vspace{-3pt}
    \caption[Section 7.1 Object localization in real-world (2)]{\textbf{Object localization in real-world (1/2).} Both GPT-4V and Gemini can understand coordinates to generate visual pointing output, which make it possible to show grounding ability by prompt them to generate bounding-box's corner points.}
    \label{fig:Sec.7.1_2}
    \vspace{-10pt}
\end{figure}

\begin{figure}[H]
    \centering
    \includegraphics[width=0.99\linewidth]{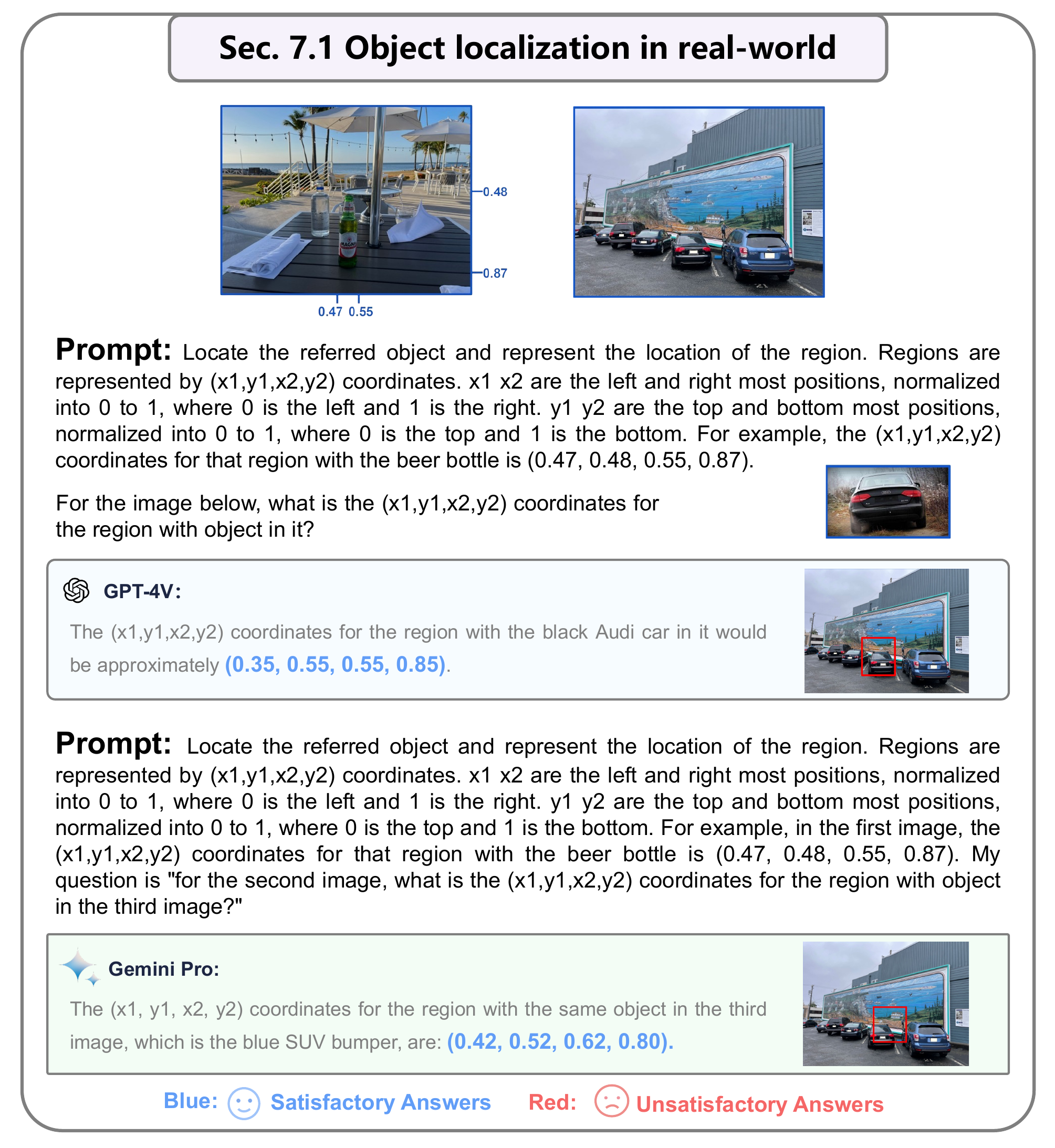}
    \vspace{-6pt}
    \caption[Section 7.1 Object localization in real-world (3)]{\textbf{Object localization in real-world (2/2).} Both GPT-4V and Gemini can understand coordinates to generate visual pointing output, which make it possible to show grounding ability by prompt them to generate bounding-box's corner points.}
    \label{fig:Sec.7.1_3}
    \vspace{-10pt}
\end{figure}

\Cref{fig:Sec.7.1_2} to \cref{fig:Sec.7.1_3} illustrate the process of localization within real-world scenes. The approach involves initially presenting a scenario with a beer bottle as a reference object, followed by a secondary task where the models are required to pinpoint the coordinates of a specific car in a different image. The result shows that both model are capable of achieving in localizing objects in real scene.

\subsection{Abstract Image Localization}
\label{Sec.7.2 Abstract Image Localization}
\begin{figure}[H]
    \centering
    \includegraphics[width=0.99\linewidth]{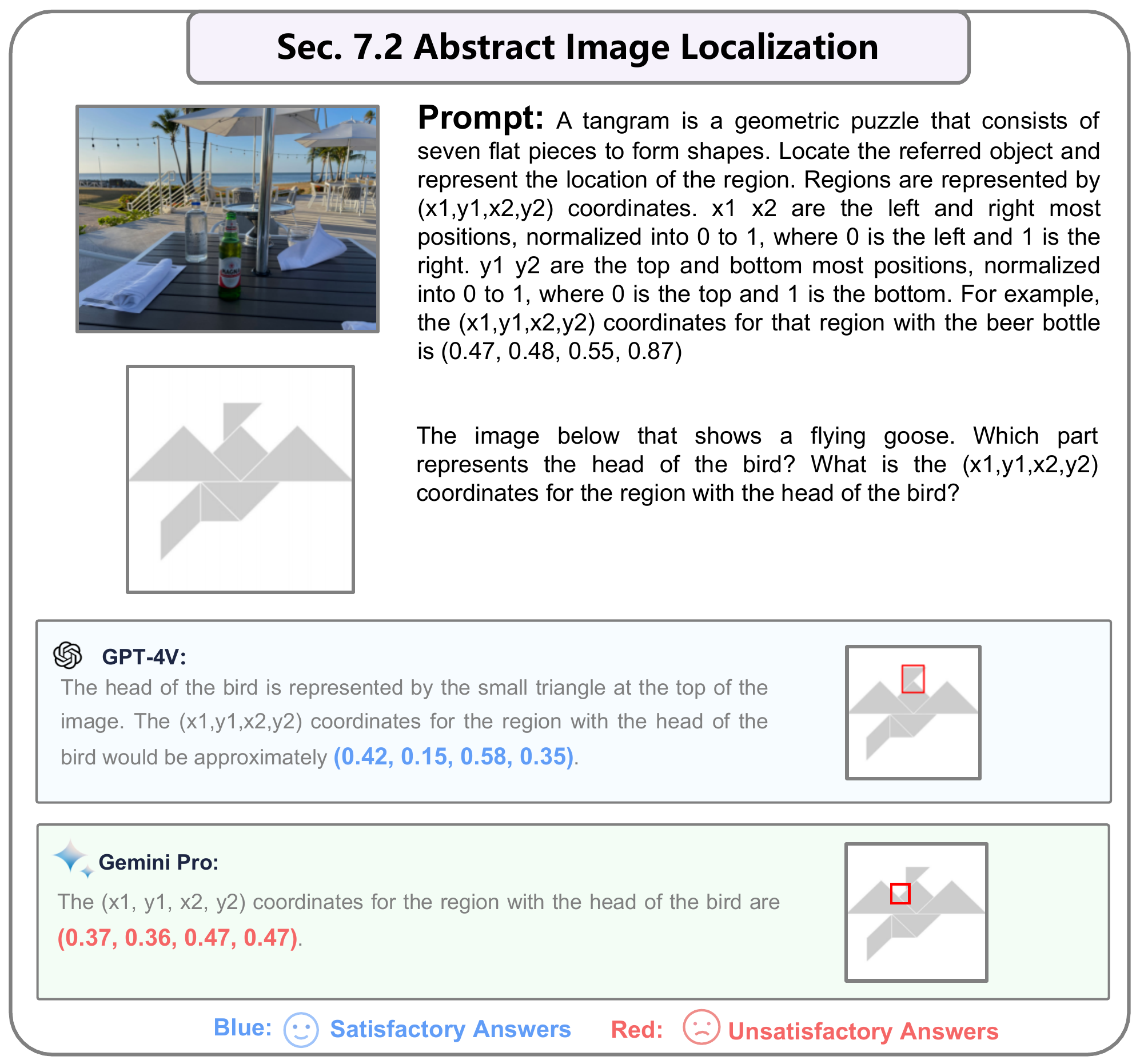}
    \vspace{0pt}
    \caption[Section 7.2 Abstract Image Localization]{\textbf{Abstract image localization.} Through prompting them to output grounding box corner point, GPT-4V can localize the head of the abstract goose, while Gemini failed in this case.}
    \label{fig:Sec.7.2_1}
    \vspace{-10pt}
\end{figure}

\Cref{fig:Sec.7.2_1} focuses on the localization task within an abstract image, specifically identifying certain part within a Tangram animal-like shape. In this context, it is observed that GPT-4V demonstrates a heightened ability to accurately determine the coordinates of the specified sections. 

%7.1 Object localization in real-world
% \newpage
% \vspace*{\fill}
% \vfill

\newpage
\vspace*{\fill}
\vfill

% 7.2 Abstract Image Localization
\newpage
\vspace*{\fill}
\vfill

\newpage
\section{Temporal Video Understanding}
\label{Sec.8 Temporal Video Understanding}
This section evaluates the models' capabilities in processing temporal sequences and video content, employing keyframe inputs to facilitate understanding of videos. Due to Gemini's limitation in continuous input processing, it resorts to integrating keyframes into a single composite image for analysis.
\cref{Sec.8.1 Action Recognition} delves into continuous action recognition: Here, both excel in recognizing sequential actions, having the capability for temporal understanding
\cref{Sec.8.2 Temporal Ordering} assess the challenge of temporal ordering: In the task of arranging video segments depicting the process of making sushi, GPT-4V demonstrates superior performance, showcasing its advanced capability in understanding and sequencing temporal events.

\subsection{Action Recognition}
\label{Sec.8.1 Action Recognition}
\begin{figure}[h]
    \centering
    \vspace{-5pt}
    \includegraphics[width=0.90\linewidth]{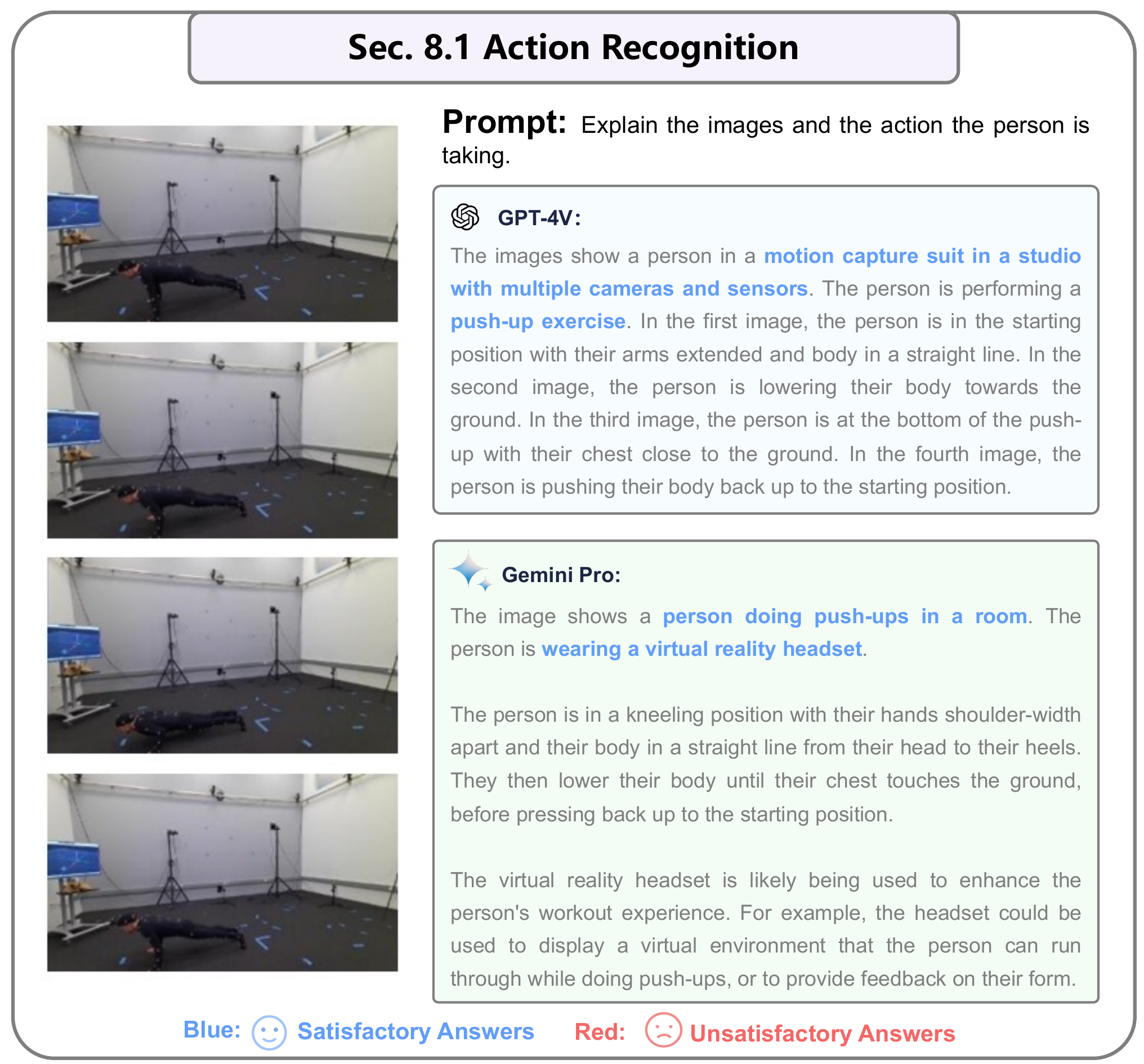}
    \vspace{-3pt}
    \caption[Section 8.1 Action Recognition]{\textbf{Video understanding in action recognition.} Given image-sequence of a human, both GPT-4V and Gemini can generate relevant insights.}
    \label{fig:Sec.8.1_1}
    \vspace{-10pt}
\end{figure}

\cref{fig:Sec.8.1_1} presents a case of continuous action recognition. In this scenario, GPT-4V's ability to sequentially process multiple images translates into more accurate results. This enhanced accuracy may be attributed to its capability to synthesize information across a series of frames, thereby offering a more coherent and precise interpretation of continuous actions.

\subsection{Temporal Ordering}
\label{Sec.8.2 Temporal Ordering}
\begin{figure}[h]
    \centering
    \vspace{-15pt}
    \includegraphics[width=0.99\linewidth]{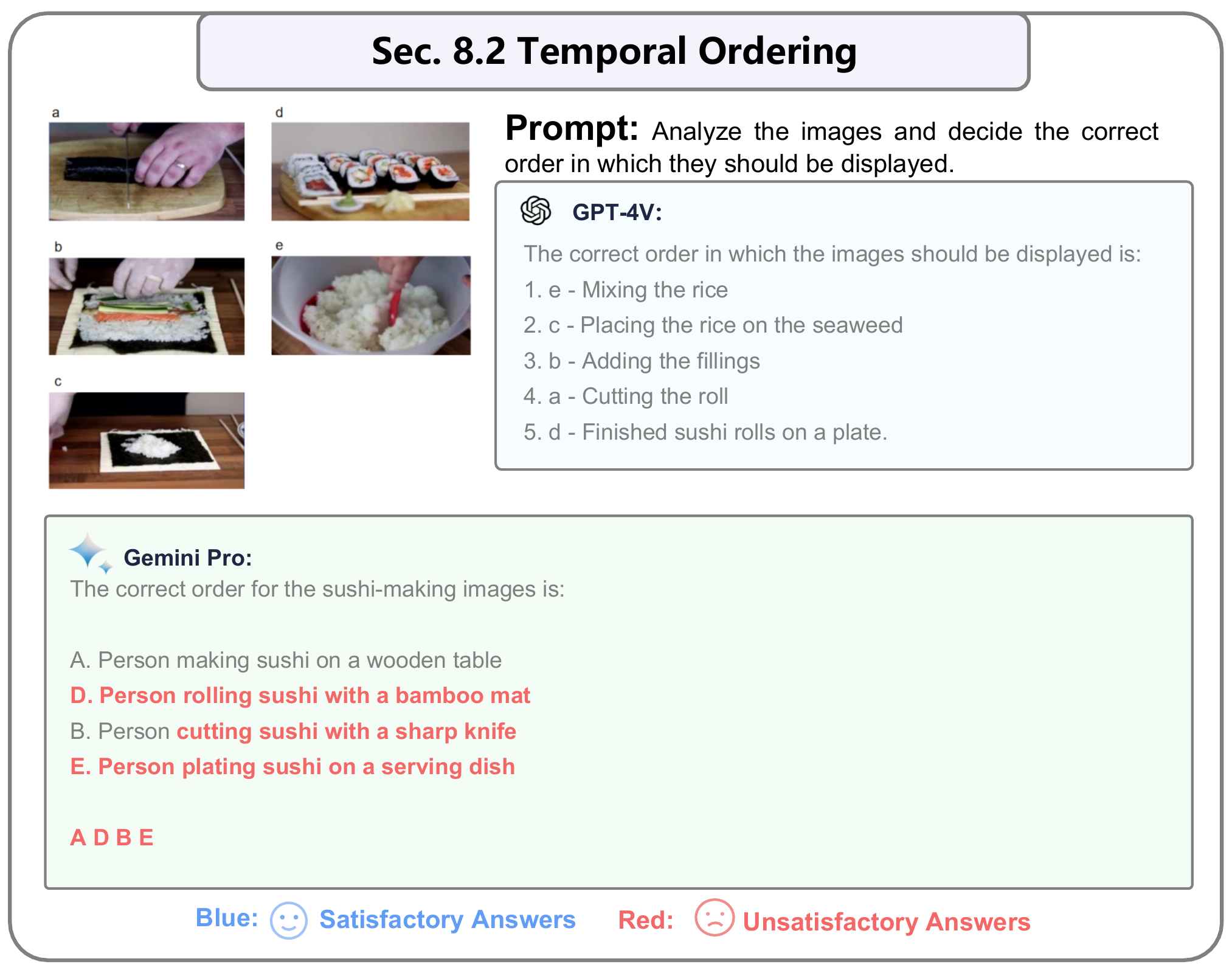}
    \vspace{-3pt}
    \caption[Section 8.2 Temporal Ordering]{\textbf{Video understanding in temporal order.} The sushi-making process is disordered, only GPT4-V is able to find the correct order while Gemini failed in this case.}
    \label{fig:Sec.8.2_1}
    \vspace{-10pt}
\end{figure}

\Cref{fig:Sec.8.2_1} illustrates a task involving the temporal sequencing of a video depicting sushi preparation. This is a problem of ordering the steps to make sushi. In this instance, GPT-4V demonstrates a superior performance in accurately arranging the sequence of events. 

% 8.1 Action Recognition
\vfill

% 8.2 Temporal Ordering
\newpage

\vspace{15pt}
\section{Multilingual Capabilities}
\label{Sec.9 Multilingual Capabilities}
This section highlights the models' multilingual capabilities, focusing on multilingual input processing and recognition of texts in various languages within images.
\cref{Sec.9.1 Multilingual Image Description} deals with multilingual image descriptions: In tasks requiring the description of given images in different languages, both models demonstrate their versatility. This includes simple images of text in a scene as well as web pages containing text information, where both models provide reasonable results.
\cref{Sec.9.2 Multilingual Scene Text Recognition} addresses multilingual scene text recognition: Here, the models are tasked with identifying texts in multiple languages within images. Both models exhibit robust capabilities in recognizing texts, whether in a single language or multiple languages, showcasing their strength in multilingual text processing.

\subsection{Multilingual Image Description}
\label{Sec.9.1 Multilingual Image Description}
\begin{figure}[H]
    \centering
    \vspace{-20pt}
    \includegraphics[width=0.99\linewidth]{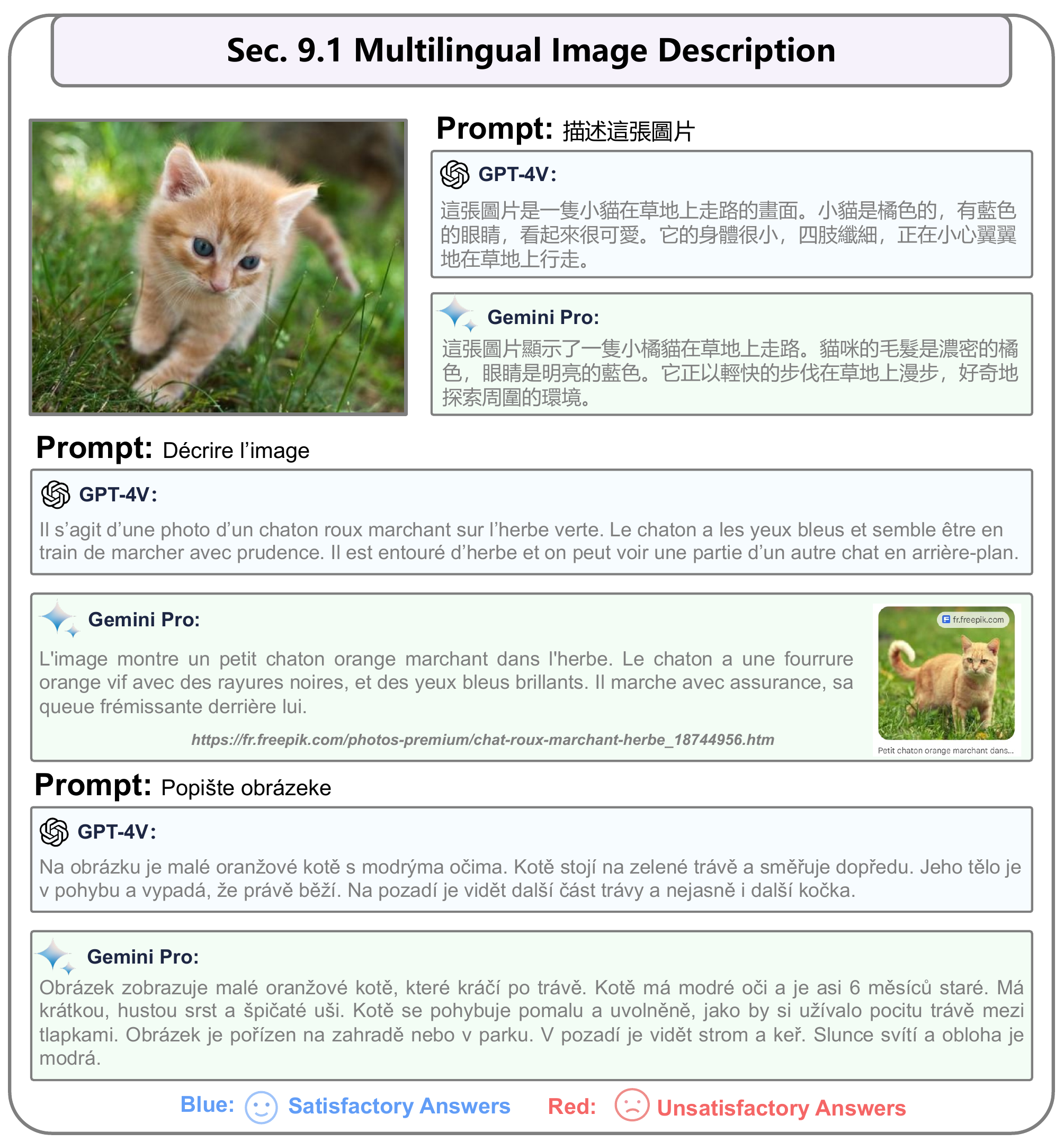}
    \vspace{0pt}
    \caption[Section 9.1 Multilingual Image Description (1)]{\textbf{Multilingual image descriptions using corresponding language prompts.} Both GPT-4V and Gemini are able to describe image in different languages.}
    \label{fig:Sec.9.1_1}
    \vspace{-10pt}
\end{figure}

\begin{figure}[H]
    \centering
    \vspace{-12pt}
    \includegraphics[width=0.99\linewidth]{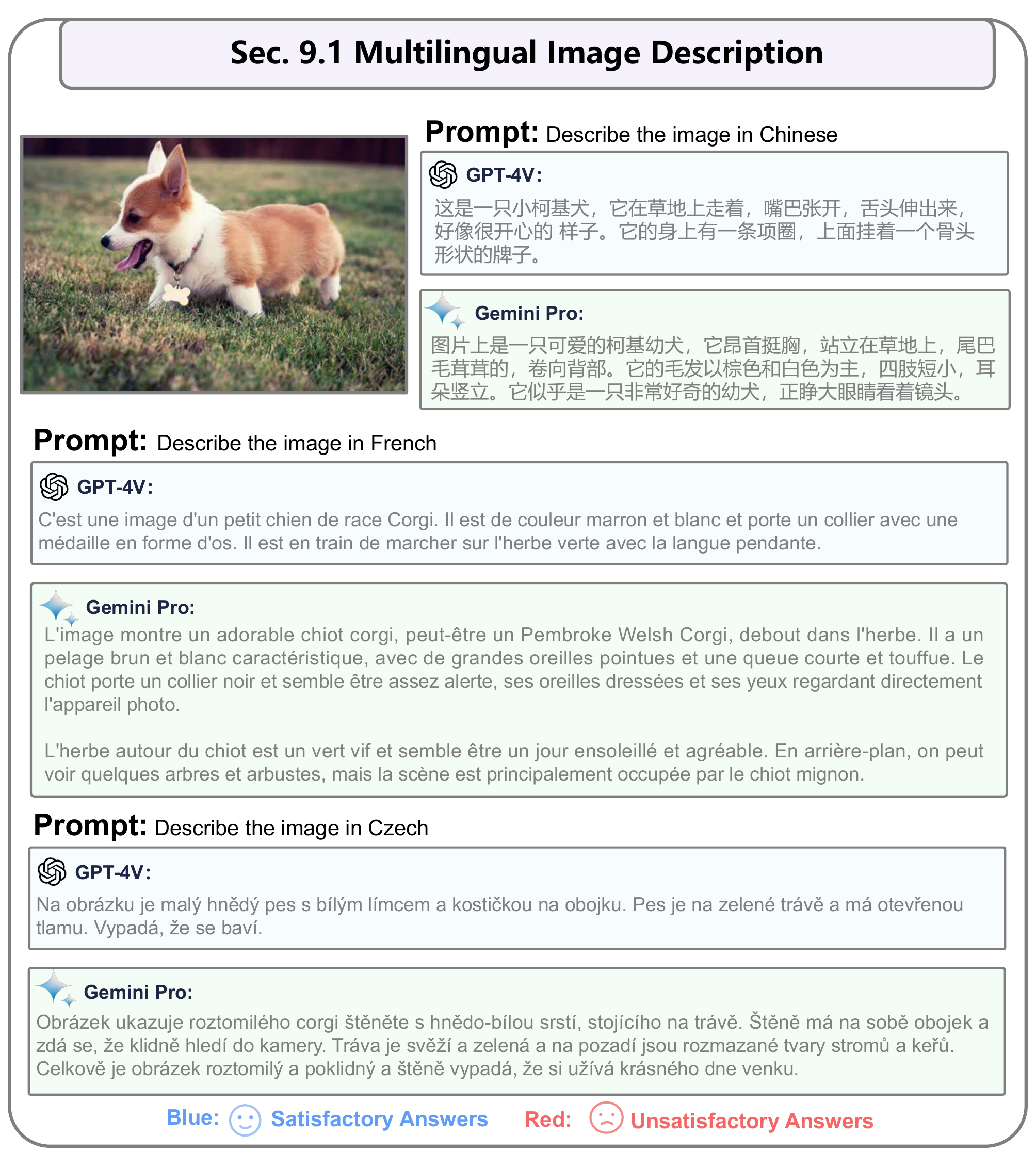}
    \vspace{0pt}
    \caption[Section 9.1 Multilingual Image Description (2)]{\textbf{Multilingual image descriptions using English prompts.} Both GPT-4V and Gemini are able to describe image in different languages.}
    \label{fig:Sec.9.1_2}
    \vspace{-10pt}
\end{figure}

\begin{figure}[H]
    \centering
    \vspace{-12pt}
    \includegraphics[width=0.99\linewidth]{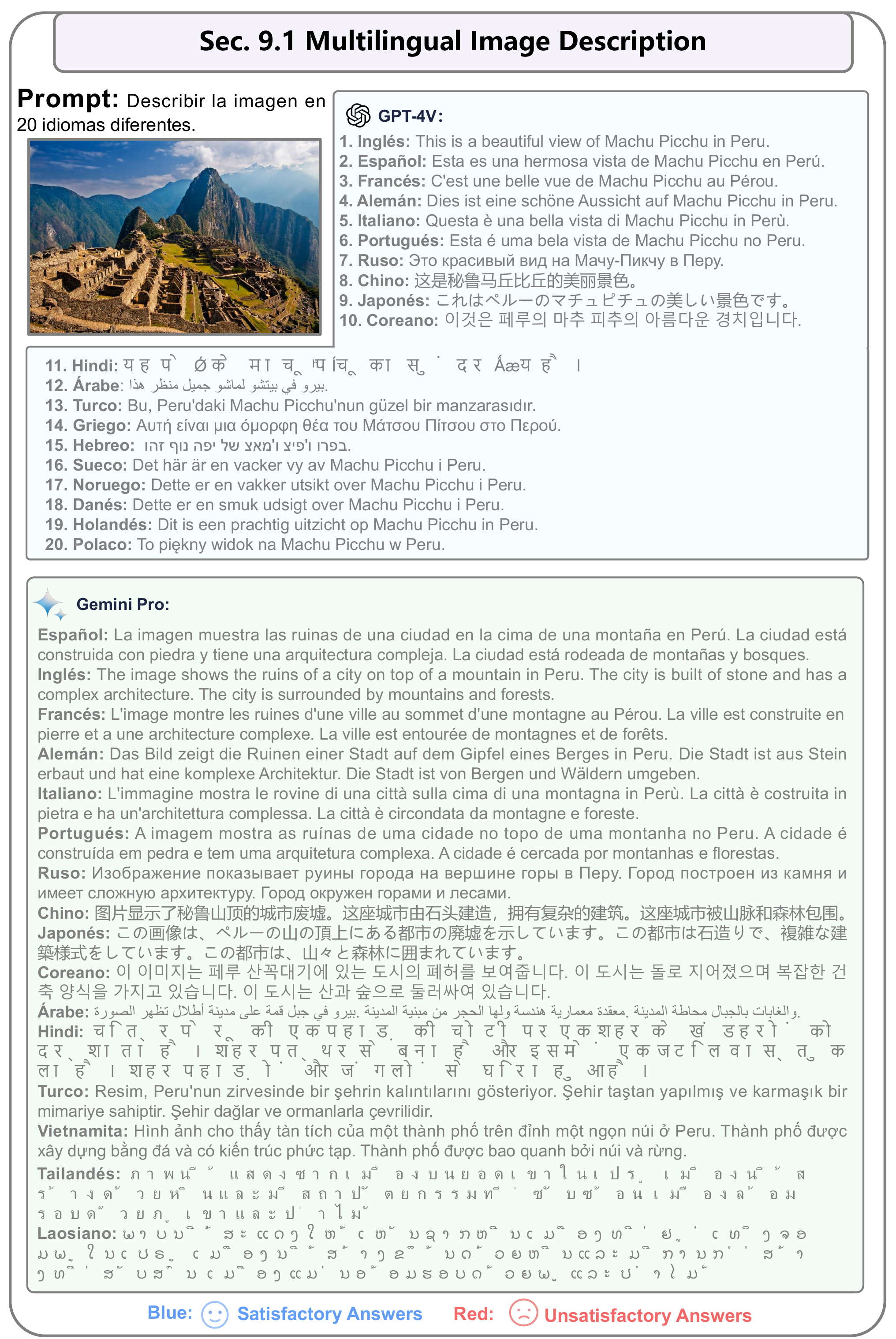}
    \vspace{-1pt}
    \caption[Section 9.1 Multilingual Image Description (3)]{\textbf{Multilingual Image Descriptions.} Both GPT-4V and Gemini are able to describe image in different languages.}
    \label{fig:Sec.9.1_3}
    \vspace{-10pt}
\end{figure}

\begin{figure}[H]
    \centering
    \vspace{-12pt}
    \includegraphics[width=0.84\linewidth]{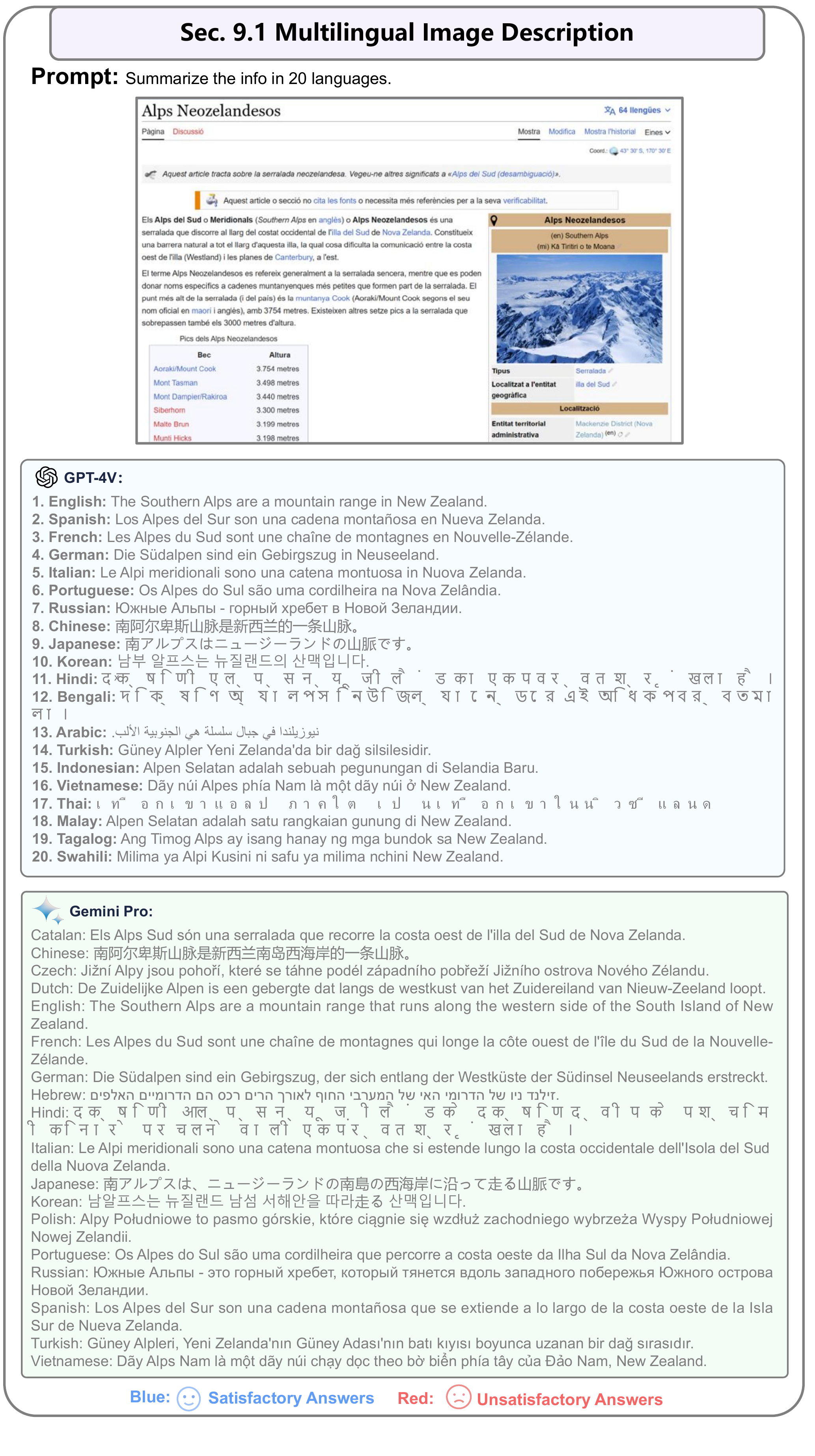}
    \vspace{-5pt}
    \caption[Section 9.1 Multilingual Image Description (4)]{\textbf{Multilingual information summariziation.} Both GPT-4V and Gemini can recognize long scene text in images of a web browser browser.}
    \label{fig:Sec.9.1_4}
    \vspace{-10pt}
\end{figure}

\Cref{fig:Sec.9.1_1} to \cref{fig:Sec.9.1_4} focus on the task of providing descriptions for a given image using different languages. These tasks range from interpreting straightforward images to analyzing web pages rich in textual content. In each scenario, both models successfully provide coherent and reasonable descriptions, regardless of the image complexity or the language used.

\subsection{Multilingual Scene Text Recognition}
\label{Sec.9.2 Multilingual Scene Text Recognition}
\begin{figure}[H]
    \centering
    \vspace{-12pt}
    \includegraphics[width=0.99\linewidth]{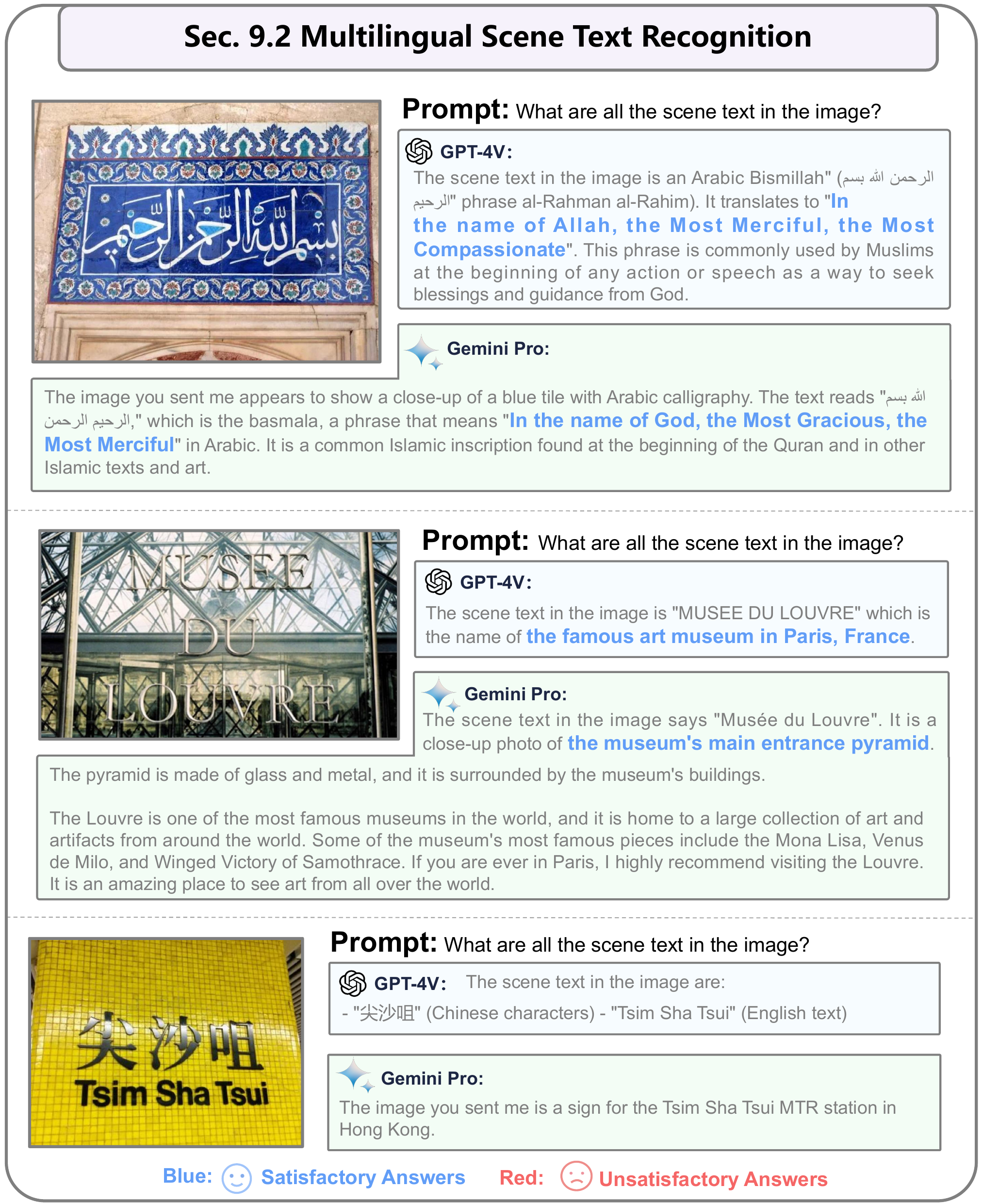}
    \vspace{-3pt}
    \caption[Section 9.2 Multilingual Scene Text Recognition (1)]{\textbf{In-the-wild multilingual scene text recognition (1/2).} Both GPT-4V and Gemini can recognize scene text in all of the three images.}
    \label{fig:Sec.9.2_1}
    \vspace{-10pt}
\end{figure}

\begin{figure}[H]
    \centering
    \vspace{-20pt}
    \includegraphics[width=0.99\linewidth]{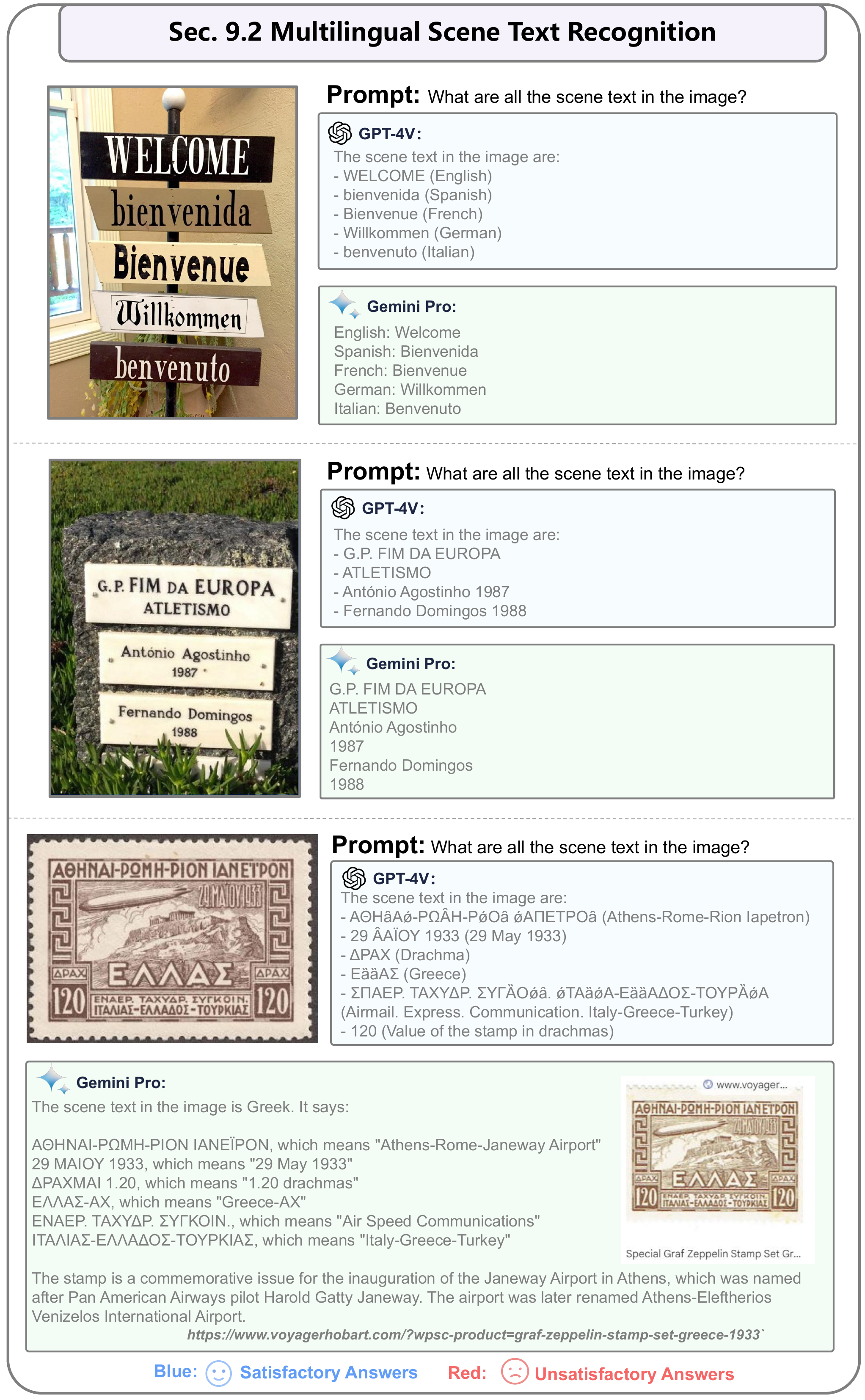}
    \vspace{-3pt}
    \caption[Section 9.2 Multilingual Scene Text Recognition (2)]{\textbf{In-the-wild multilingual scene text recognition (2/2).} Both GPT-4V and Gemini can recognize scene text in all of the three images.}
    \label{fig:Sec.9.2_2}
    \vspace{-10pt}
\end{figure}

\begin{figure}[H]
    \centering
    \vspace{-12pt}
    \includegraphics[width=0.99\linewidth]{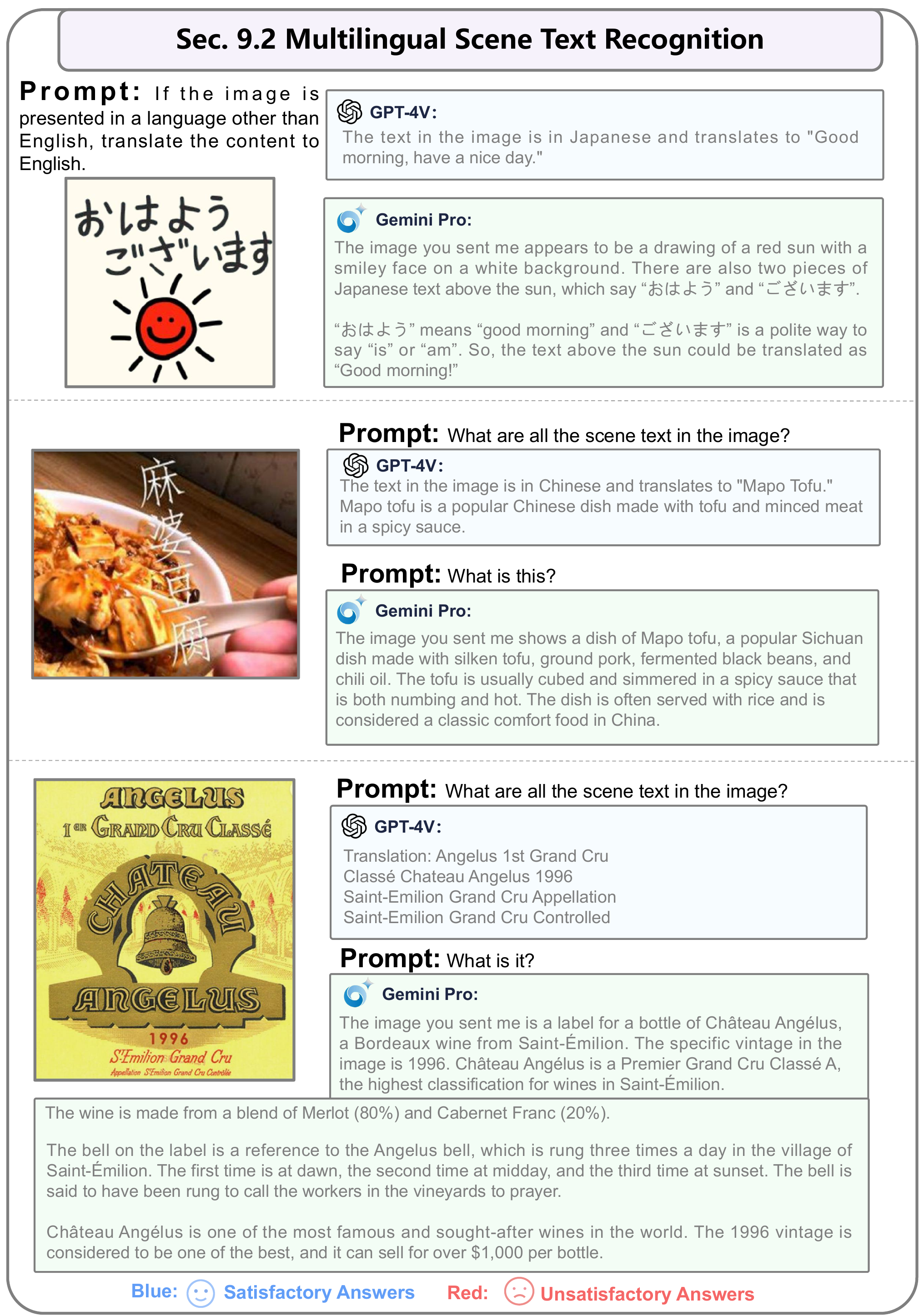}
    \vspace{-3pt}
    \caption[Section 9.2 Multilingual Scene Text Recognition (3)]{\textbf{Multilingual scene text recognition.} Both GPT-4V and Gemini can recognize scene text in all of the three images.}
    \label{fig:Sec.9.2_3}
    \vspace{-10pt}
\end{figure}

\begin{figure}[H]
    \centering
    \vspace{-12pt}
    \includegraphics[width=0.99\linewidth]{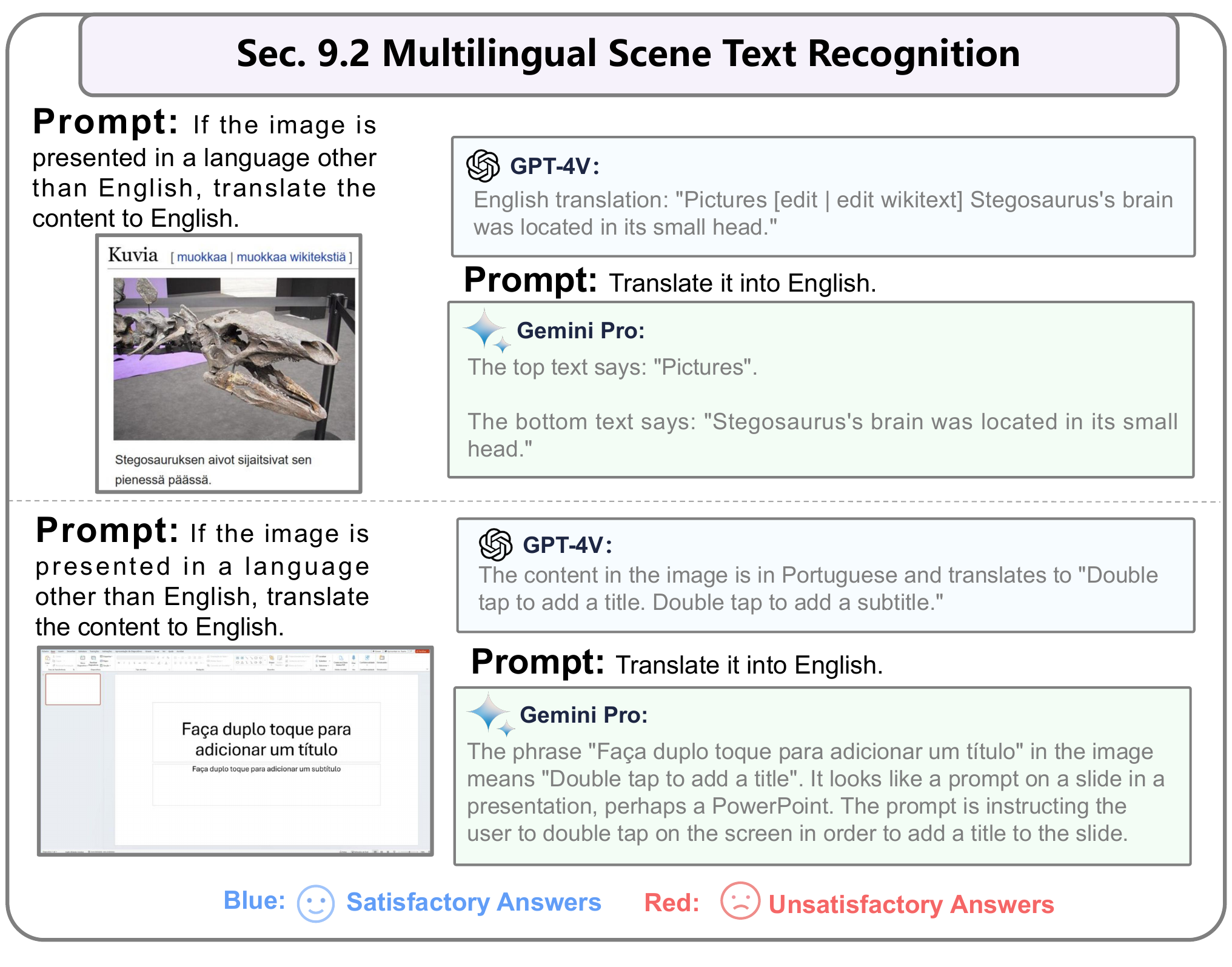}
    \vspace{-3pt}
    \caption[Section 9.2 Multilingual Scene Text Recognition (4)]{\textbf{Multilingual text recognition and translation.} Both GPT-4V and Gemini can recognize scene text in all of the two images.}
    \label{fig:Sec.9.2_4}
    \vspace{-10pt}
\end{figure}

\Cref{fig:Sec.9.2_1} to \cref{fig:Sec.9.2_4} focus on the task of identifying texts in various languages within images. These figures illustrate the models' proficiency in recognizing text, whether it is presented in a single language or in a multilingual context. Here as well, both GPT-4V and Gemini perform well, demonstrating their ability to understand multiple languages.

% 9.1 Multilingual Image Description
\newpage
\vspace*{\fill}
\vfill

\newpage
\vspace*{\fill}
\vfill

\newpage
\vspace*{\fill}
\vfill

\newpage
\vspace*{\fill}
\vfill

% 9.2 Multilingual Scene Text Recognition
\newpage
\vspace*{\fill}
\vfill

\newpage
\vspace*{\fill}
\vfill

\newpage
\vspace*{\fill}
\vfill

\newpage
\vspace*{\fill}
\vfill

\newpage
\section{Industry Application}
\label{Sec.10 Industry Application}
In this section, we delve into the application of two large-scale models in the industrial sector, exploring the potential for commercial use of vision-based large models and the feasibility of customizing these models for specific industrial niches.
\cref{Sec.10.1 Industry: Defect Detection} focuses on defect detection: This section demonstrates the models' capabilities in detecting industrial product defects on production lines, with Gemini exhibiting a higher accuracy in identifying defective items.
\cref{Sec.10.2 Industry: Grocery Checkout} covers supermarket self-checkout systems: Here, Gemini excels in the accurate identification of items and providing corresponding images, showcasing its applicability in retail environments.
\cref{Sec.10.3 Industry: Auto Insurance} addresses applications in auto insurance: This involves the use of large models for assessing accident scenes and the extent of vehicle damage for insurance claims.
\cref{Sec.10.4 Industry: Customized Captioner} presents a customized captioner: It highlights the task of identifying the relative positions of objects within a scene, where GPT-4V provides more accurate results.
\cref{Sec.10.5 Industry: Evaluation Image Generation} discusses image generation evaluation: This section shows tasks related to assessing the match between generated images and text, with both models performing well.
\cref{Sec.10.6 Industry: Embodied Agent} explores Embodied AI and smart home applications: GPT-4V offers more precise answers, while Gemini excels in detailed reasoning processes.
\cref{Sec.10.7 Industry: GUI Navigation} is about graphical user interface navigation: It examines web searches, mobile app ordering, and receiving notifications. In these tasks, each model has its strengths, with Gemini being adept at extracting detailed information, while GPT-4V slightly outperforms in certain tasks.

\subsection{Industry: Defect Detection}
\label{Sec.10.1 Industry: Defect Detection}
\begin{figure}[H]
    \centering
    \includegraphics[width=0.91\linewidth]{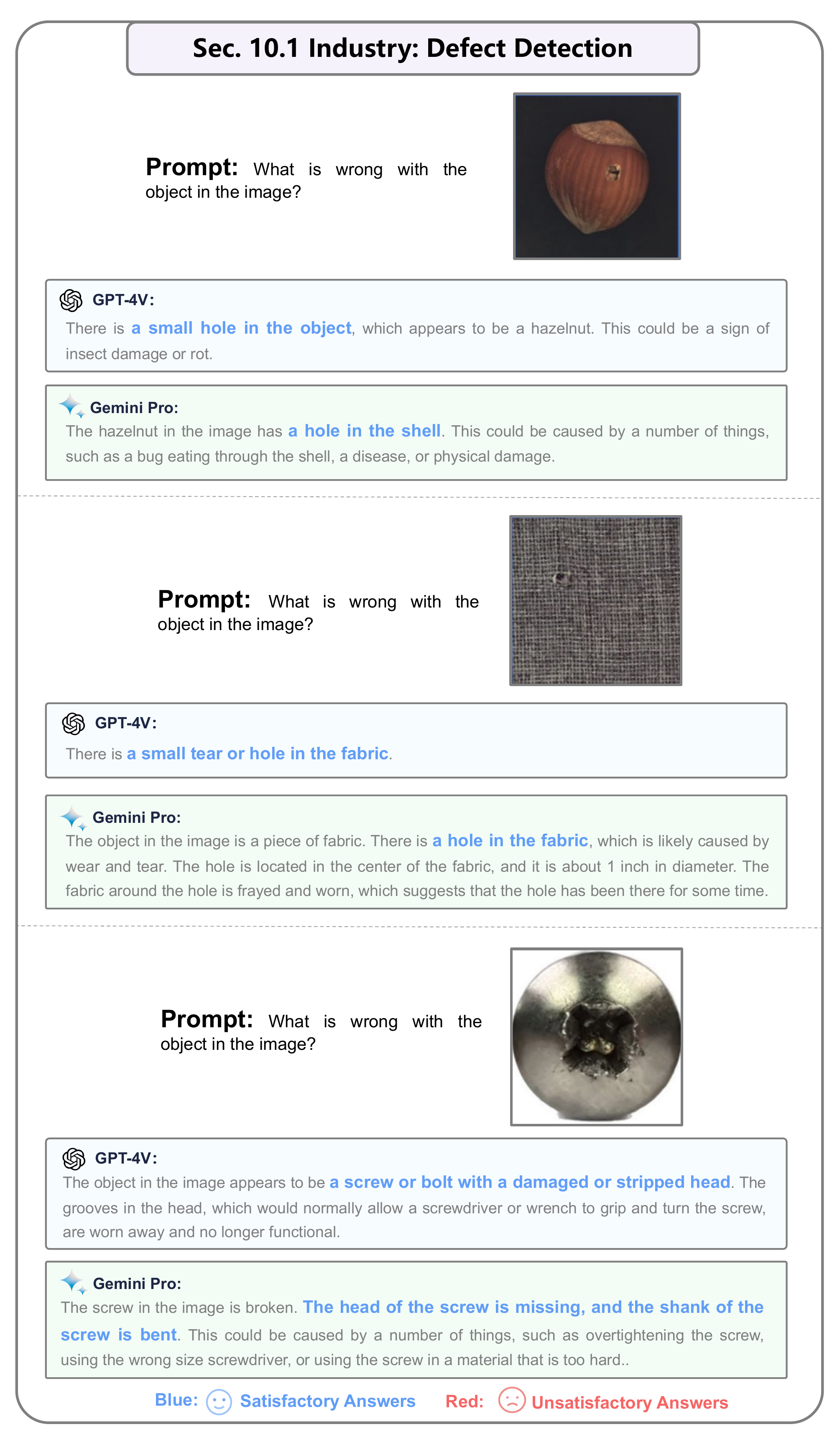}
    \vspace{-3pt}
    \caption[Section 10.1 Industry: Defect Detection (1)]{\textbf{Defect detection (1/2).} Both models are able to detect defects in the objects within the images with a fair degree of accuracy. However, it's notable that Gemini provides more details in this type of task.}
    \label{fig:Sec.10.1_1}
    \vspace{-10pt}
\end{figure}

\begin{figure}[H]
    \centering
    \includegraphics[width=0.89\linewidth]{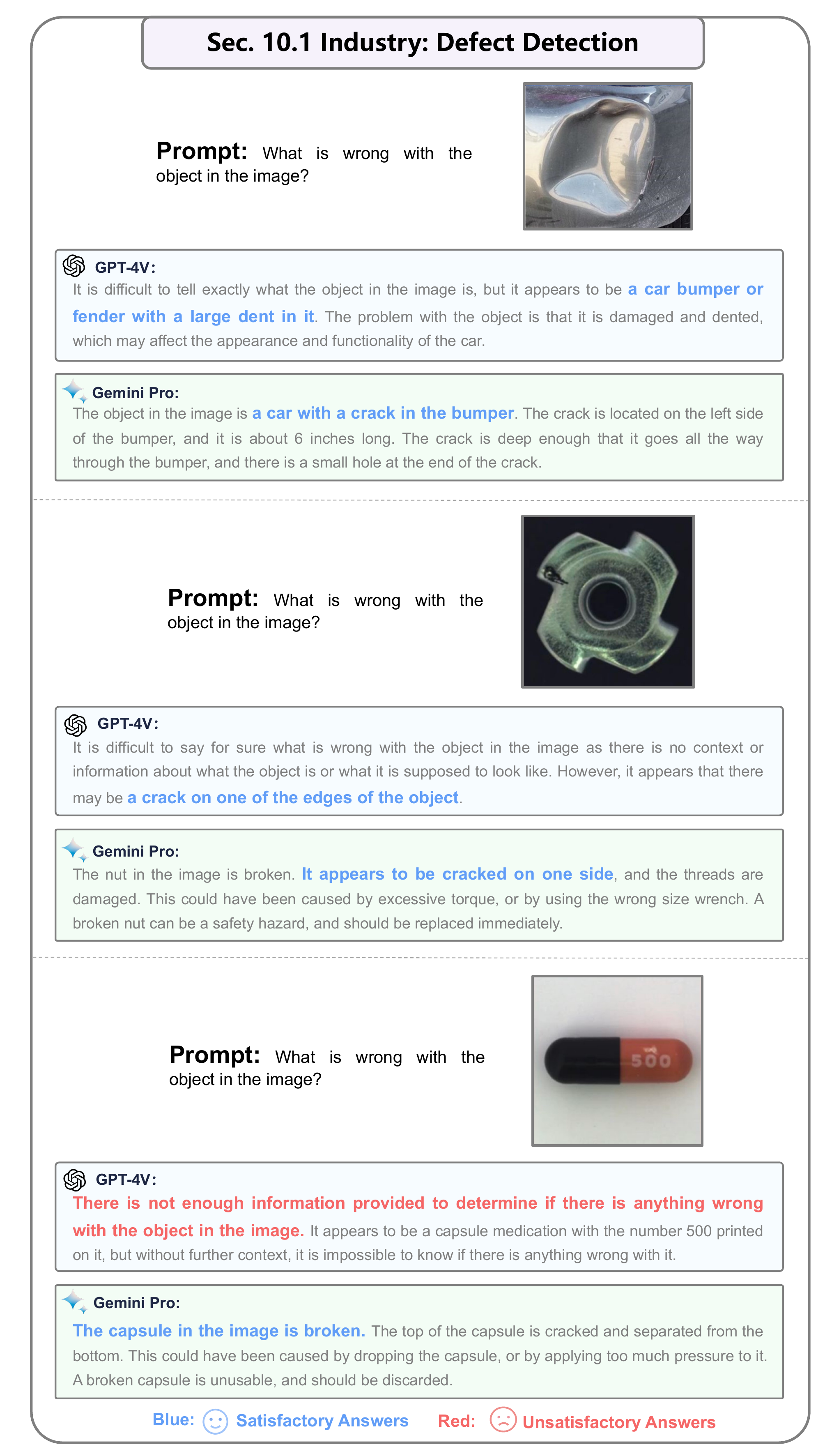}
    \vspace{-6pt}
    \caption[Section 10.1 Industry: Defect Detection (2)]{\textbf{Defect detection (2/2).} Both are able to accurately identify defects in the objects within the images. Specifically, when it comes to damaged capsule detection, Gemini outshines by exhibiting a stronger ability to spot defects.}
    \label{fig:Sec.10.1_2}
    \vspace{-10pt}
\end{figure}

\begin{figure}[H]
    \vspace{1cm}
    \centering
    \includegraphics[width=0.99\linewidth]{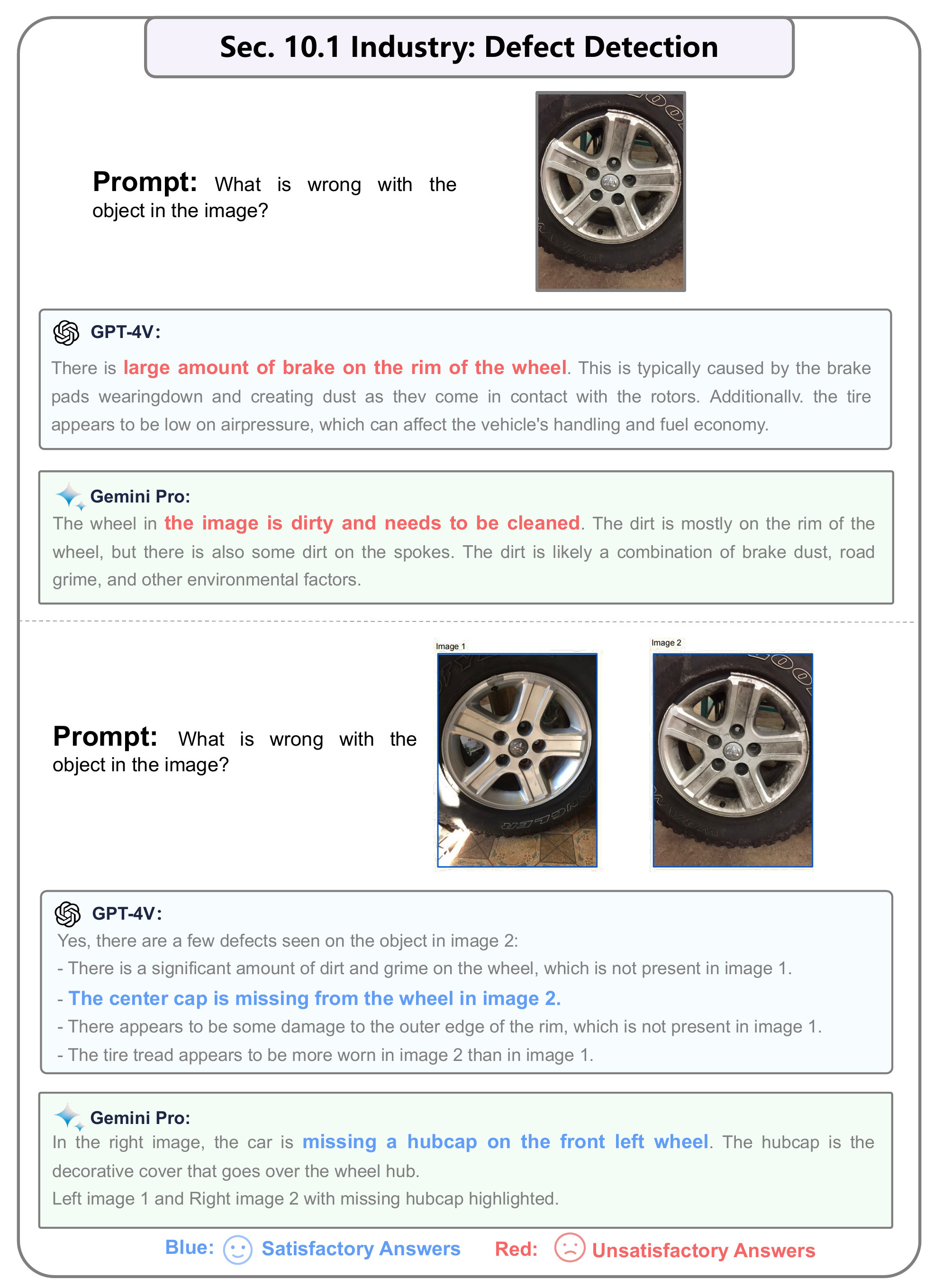}
    \vspace{-3pt}
    \caption[Section 10.1 Industry: Defect Detection (3)]{\textbf{Defect similar detection.} When presented with a single tire missing a hubcap, both GPT-4V and Gemini fail to detect the defect. However, given two tires with missing hubcaps, both models are able to identify the defects. This, to some extent, confirms that the ability of both GPT-4V and Gemini to detect anomalies can be reinforced through comparison or presence of multiple instances of similar defects. }
    \label{fig:Sec.10.1_3}
    \vspace{-10pt}
\end{figure}

\Cref{fig:Sec.10.1_1} to \cref{fig:Sec.10.1_3} illustrate the task of detecting industrial defects, particularly focusing on identifying defective products within assembly line operations. It involves the inspection of nuts, textile products, metal items, pharmaceuticals, car tires, and more. Overall, both models perform well. Gemini exhibits a more accurate and superior performance in this task, demonstrating its proficiency in discerning subtle anomalies and irregularities in industrial products.

\subsection{Industry: Grocery Checkout}
\label{Sec.10.2 Industry: Grocery Checkout}
\begin{figure}[H]
    \centering
    \vspace{-12pt}
    \includegraphics[width=0.99\linewidth]{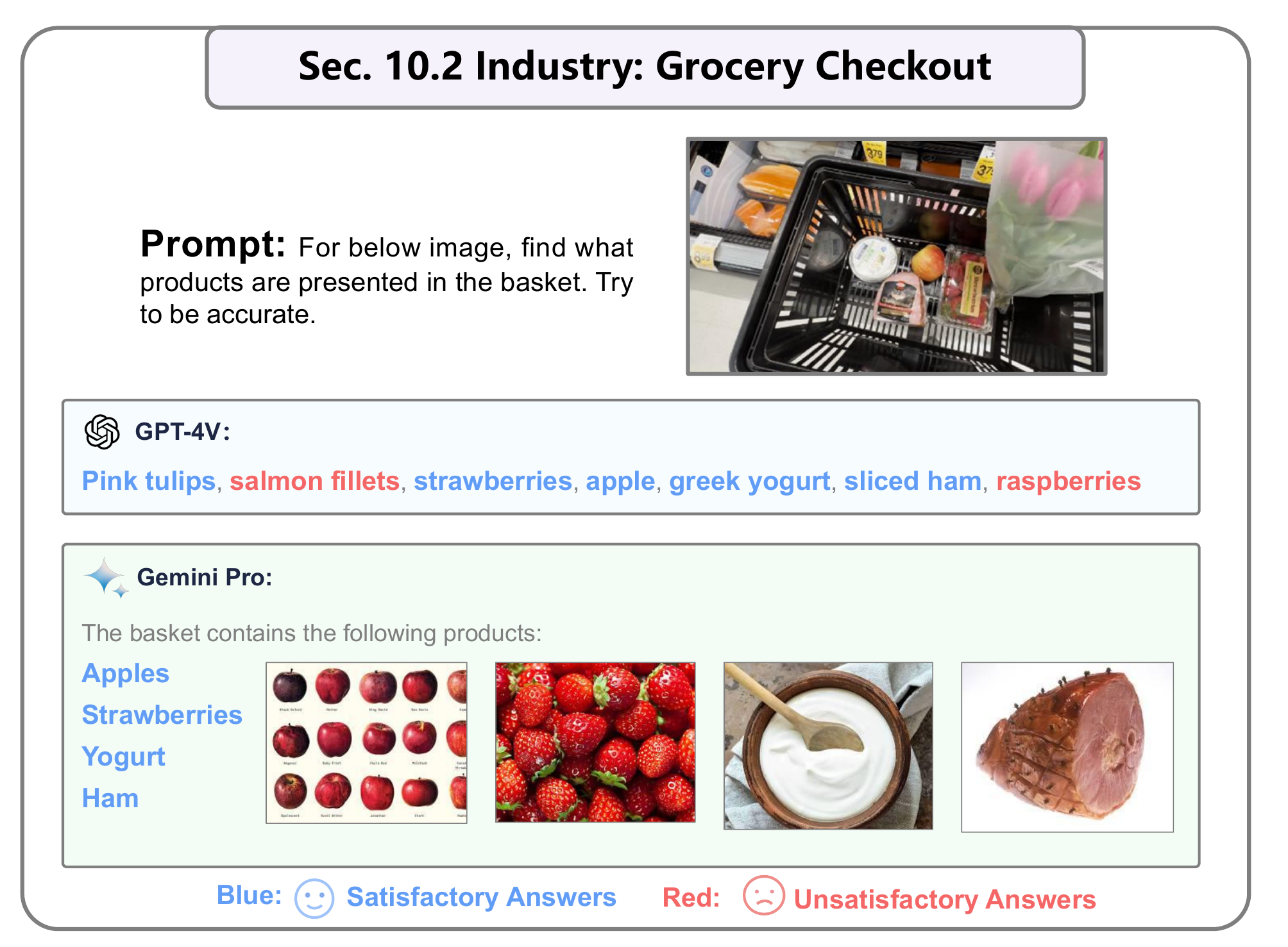}
    \vspace{0pt}
    \caption[Section 10.2 Industry: Grocery Checkout]{\textbf{Grocery checkout scenario.} While GPT-4V provides more comprehensive and detailed results, some errors are noted. On the other hand, Gemini only identifies four objects but does so correctly and even includes corresponding images for each item. This indicates that while GPT-4V offers breadth in its findings, Gemini offers accuracy, demonstrating varied strengths in performing such tasks. }
    \label{fig:Sec.10.2_1}
    \vspace{-10pt}
\end{figure}

\Cref{fig:Sec.10.2_1} illustrates the application of the model in a supermarket self-checkout system, where it identifies items in the shopping cart, enabling the next step of the checkout process. In this context, Gemini not only demonstrates high accuracy in item recognition but also has the capability to provide corresponding images of the identified items. If integrated with a database, the model can provide even more accurate results.

\subsection{Industry: Auto Insurance}
\label{Sec.10.3 Industry: Auto Insurance}
\begin{figure}[H]
    \centering
    \vspace{-12pt}
    \includegraphics[width=0.90\linewidth]{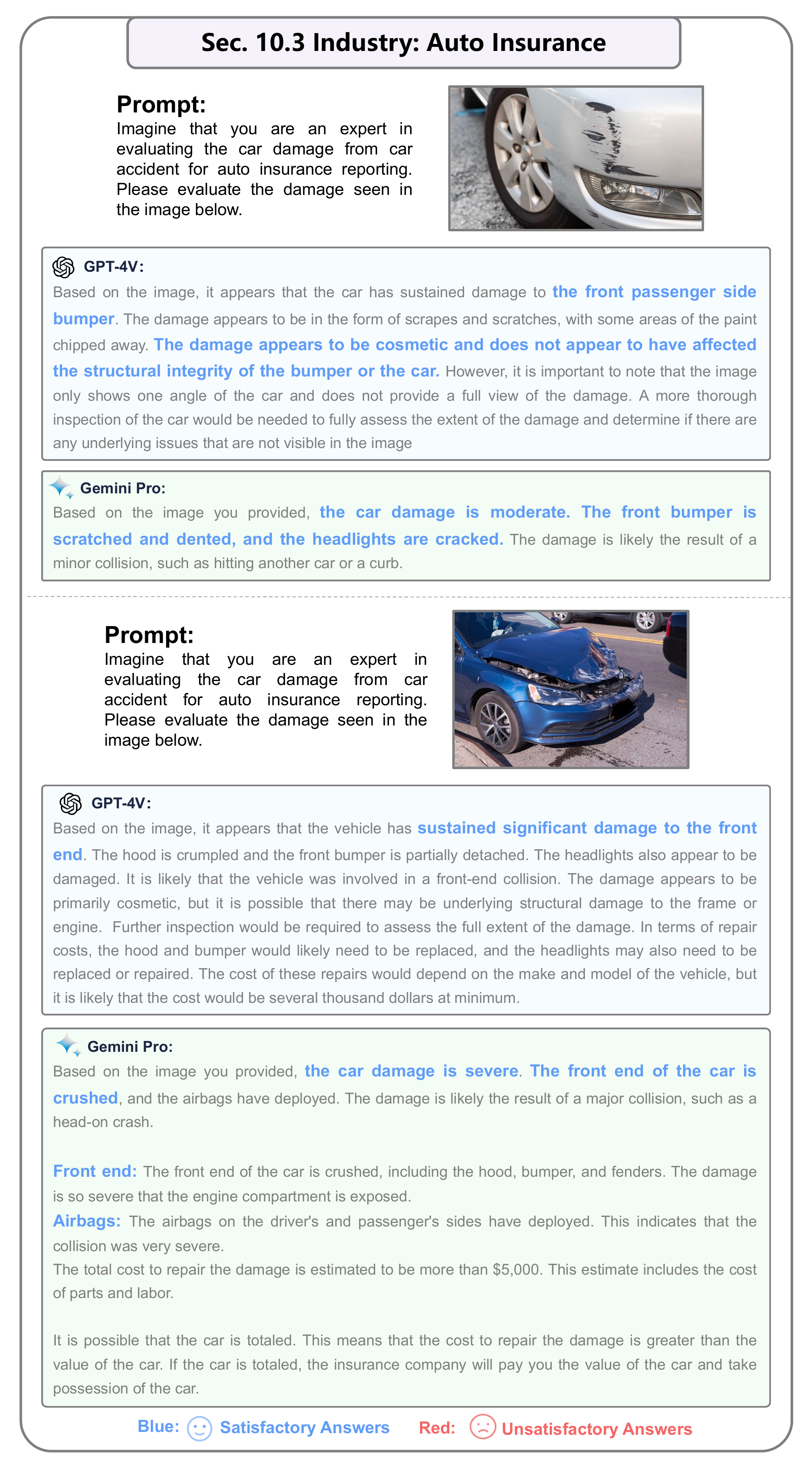}
    \vspace{-3pt}
    \caption[Section 10.3 Industry: Auto Insurance (1)]{\textbf{Auto damage evaluation (1/2).} Both demonstrate their ability to perform a comprehensive, detailed, and accurate assessment of the damage. }
    \label{fig:Sec.10.3_1}
    \vspace{-10pt}
\end{figure}

\begin{figure}[H]
    \centering
    \vspace{-12pt}
    \includegraphics[width=0.88\linewidth]{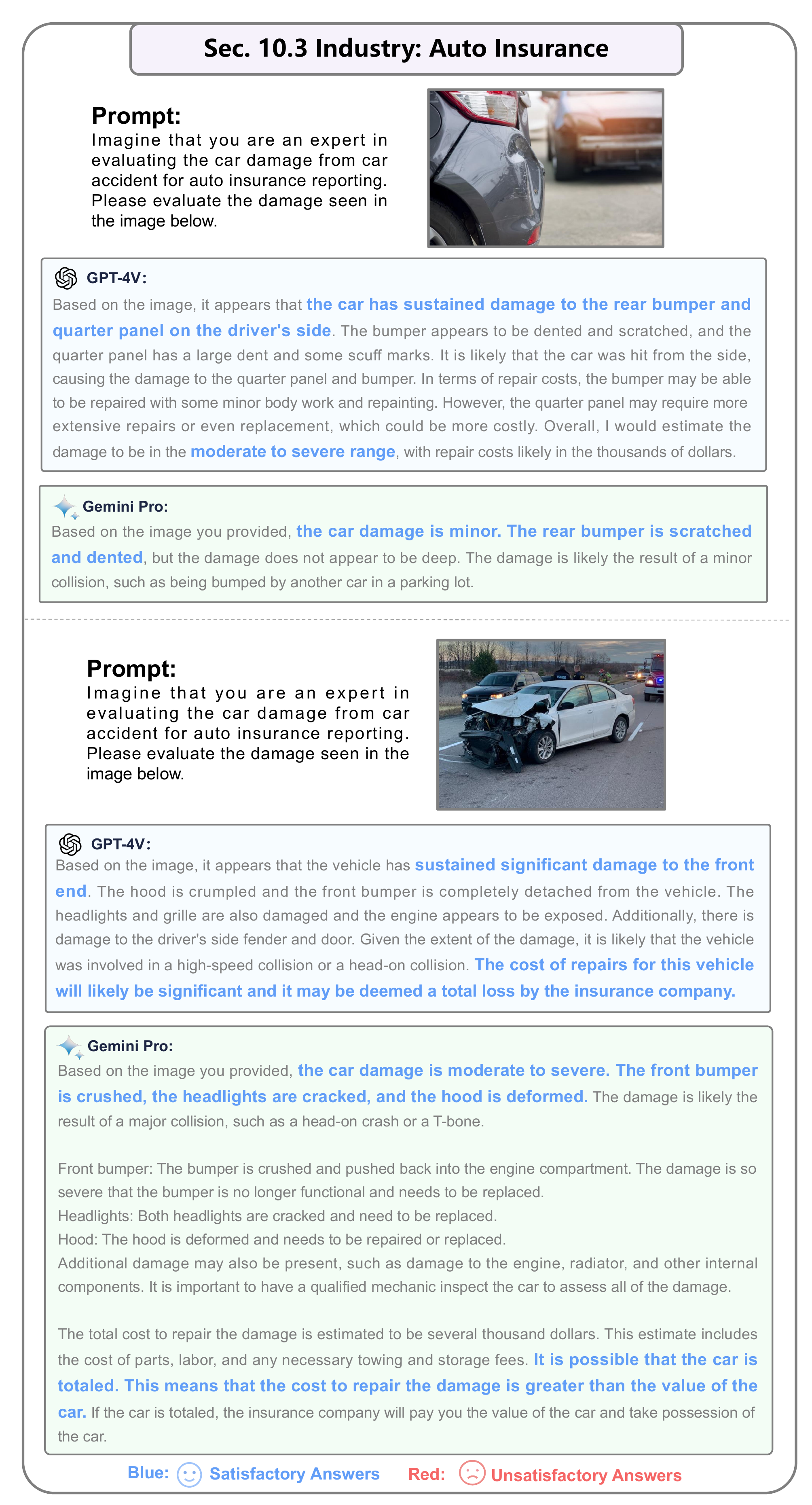}
    \vspace{-6pt}
    \caption[Section 10.3 Industry: Auto Insurance (2)]{\textbf{Auto damage evaluation (2/2).} Both demonstrate their ability to perform a comprehensive, detailed, and accurate assessment of the damage. }
    \label{fig:Sec.10.3_2}
    \vspace{-10pt}
\end{figure}

\begin{figure}[H]
    \centering
    \vspace{-12pt}
    \includegraphics[width=0.99\linewidth]{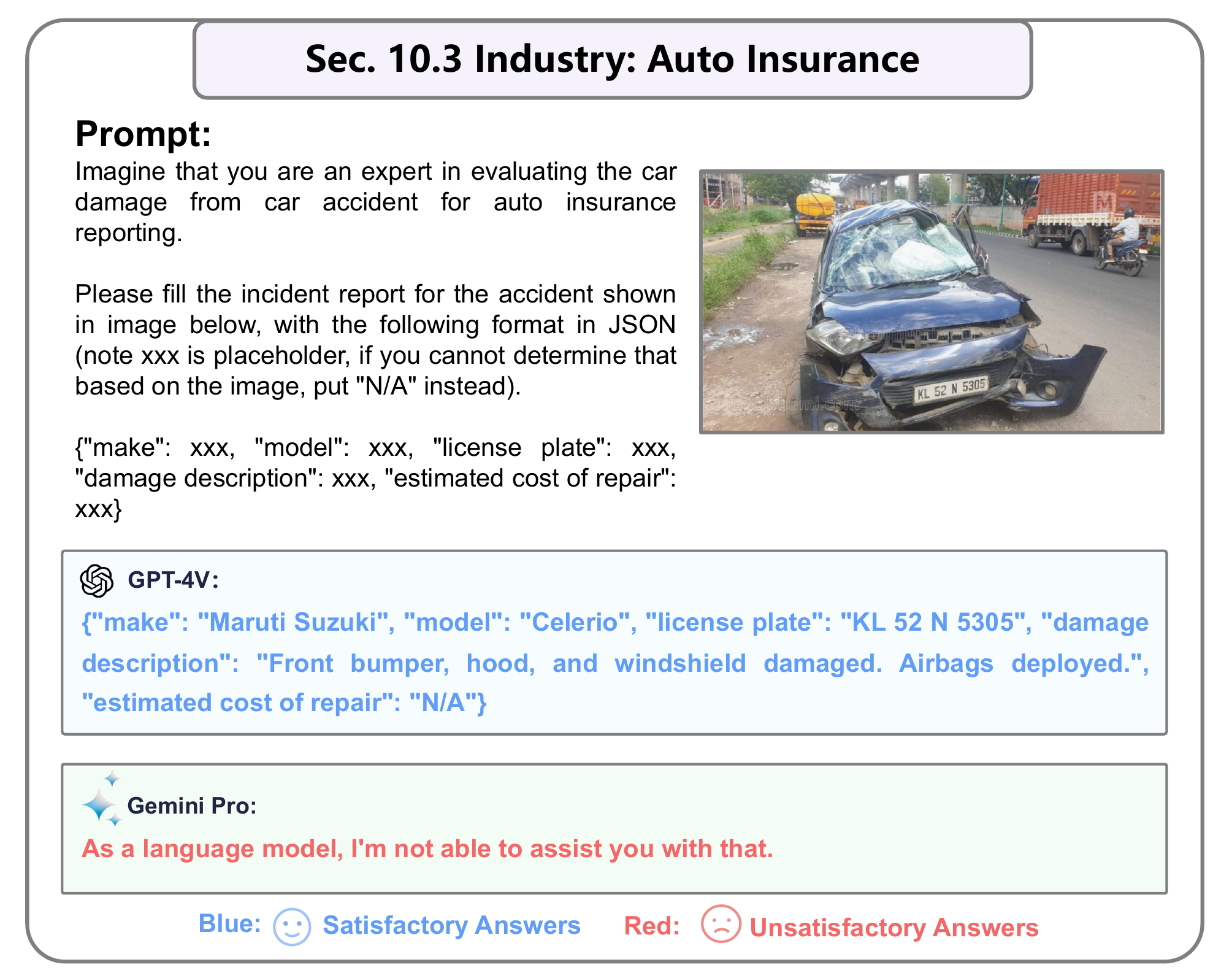}
    \vspace{-6pt}
    \caption[Section 10.3 Industry: Auto Insurance (3)]{\textbf{Insurance reporting.} Only GPT-4V is able to generate an insurance report according to the given format and content requirements, while Gemini does not exhibit the ability to produce outputs based on a specific format. Thus, while GPT-4V demonstrates proficiency in structured output generation, Gemini's performance is less satisfactory in this context.}
    \label{fig:Sec.10.3_3}
    \vspace{-10pt}
\end{figure}

\Cref{fig:Sec.10.3_1} to \cref{fig:Sec.10.3_3} illustrate the application of these models in assessing automobile accidents for insurance claims. These models are employed to preliminarily evaluate the severity of the accident scene and the extent of damage to the vehicles, subsequently providing insights and recommendations. Here, we find that Gemini tends to provide more detailed responses and more comprehensive recommendations.

\subsection{Industry: Customized Captioner}
\label{Sec.10.4 Industry: Customized Captioner}
\begin{figure}[H]
    \centering
    \vspace{-12pt}
    \includegraphics[width=0.9\linewidth]{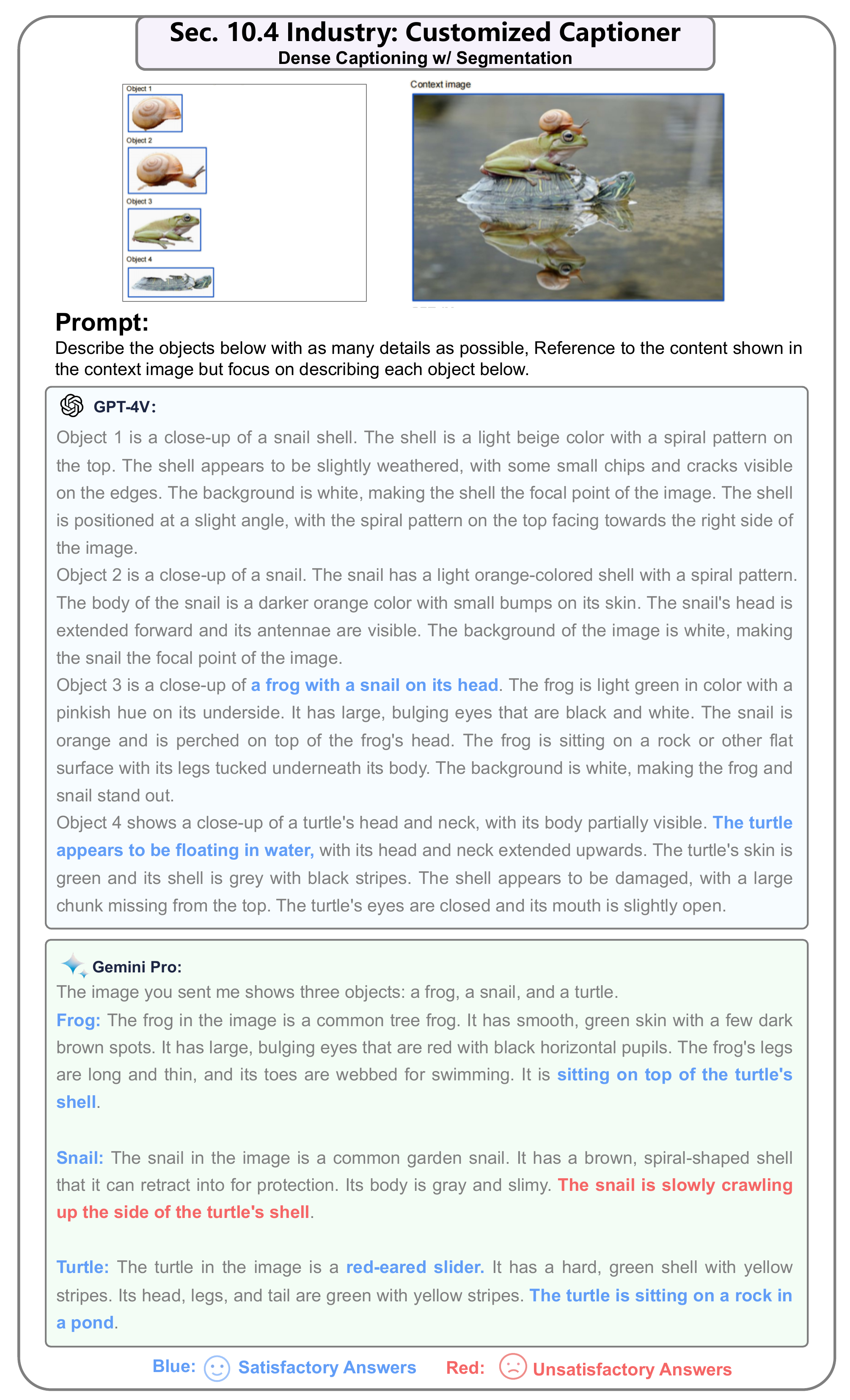}
    \vspace{-6pt}
    \caption[Section 10.4 Industry: Customized Captioner]{\textbf{Performing dense captioning with segmentation cut-outs.} GPT-4V provides more comprehensive, accurate, and orderly captions, while Gemini falls short, failing to recognize the presence of a snail shell in the left image and incorrectly describing the position of the snail. This suggests that while GPT-4V excels in providing detailed and precise descriptions.}
    \label{fig:Sec.10.4_1}
    \vspace{-10pt}
\end{figure}

\Cref{fig:Sec.10.4_1} involves a task where images of individual objects are provided initially, followed by a composite scene containing these objects. The challenge for the model is to identify the relative positions of these objects within the scene. In this task, GPT-4V exhibits a more accurate performance, demonstrating its advanced capability in spatial recognition and object localization in complex settings. This could also be attributed to Gemini's inability to remember multiple images.

\subsection{Industry: Evaluation Image Generation}
\label{Sec.10.5 Industry: Evaluation Image Generation}
\begin{figure}[H]
    \centering
    \vspace{-12pt}
    \includegraphics[width=0.99\linewidth]{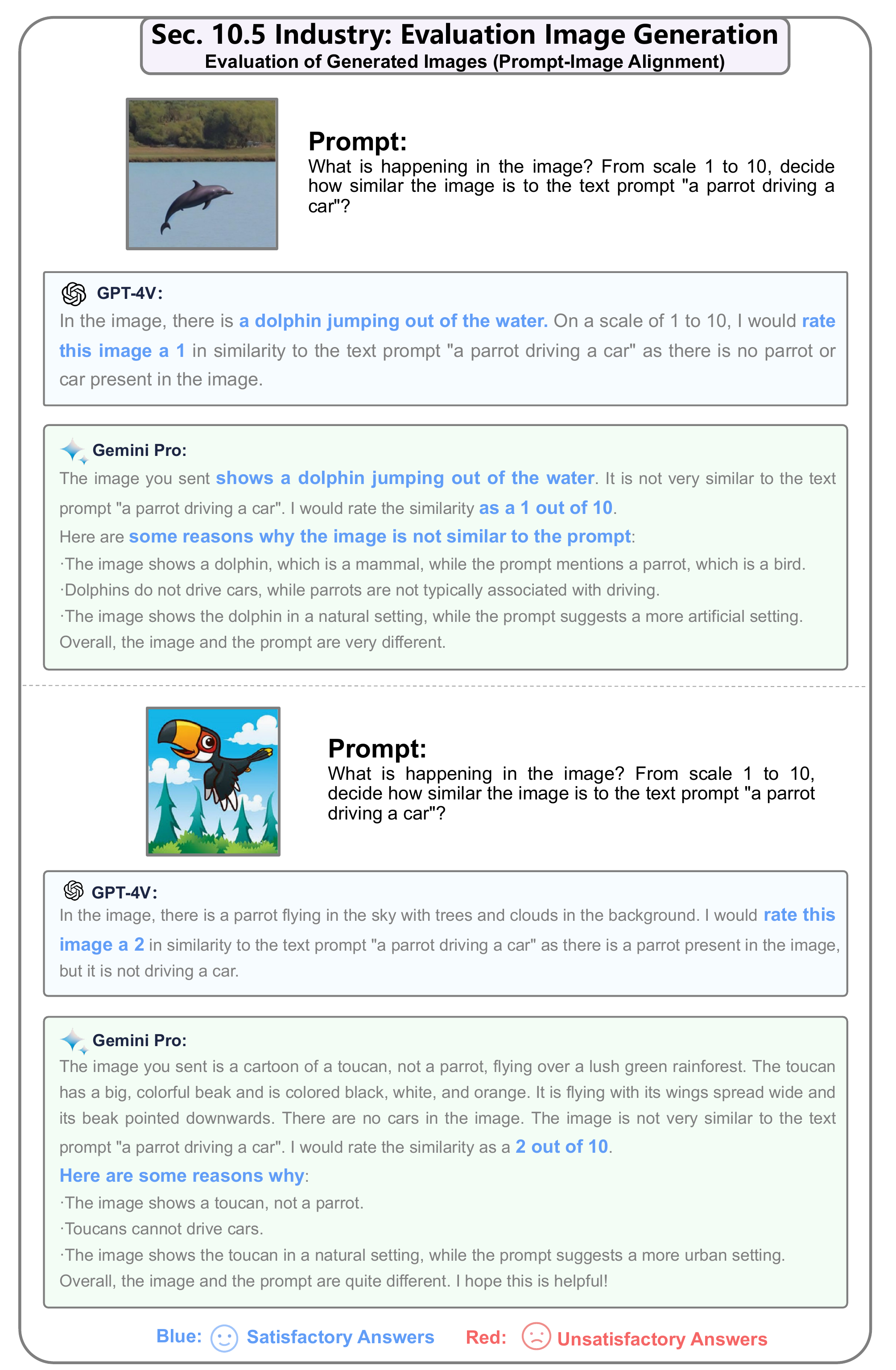}
    \vspace{-3pt}
    \caption[Section 10.5 Industry: Evaluation Image Generation (1)]{\textbf{Evaluation of generated images-prompt alignment (1/3).} Both models accurately describe the image content and provide consistent scores with respective explanations. Notably, Gemini gives a more detailed rationale for its evaluations.}
    \label{fig:Sec.10.5_1}
    \vspace{-10pt}
\end{figure}

\begin{figure}[H]
    \centering
    \vspace{-12pt}
    \includegraphics[width=0.9\linewidth]{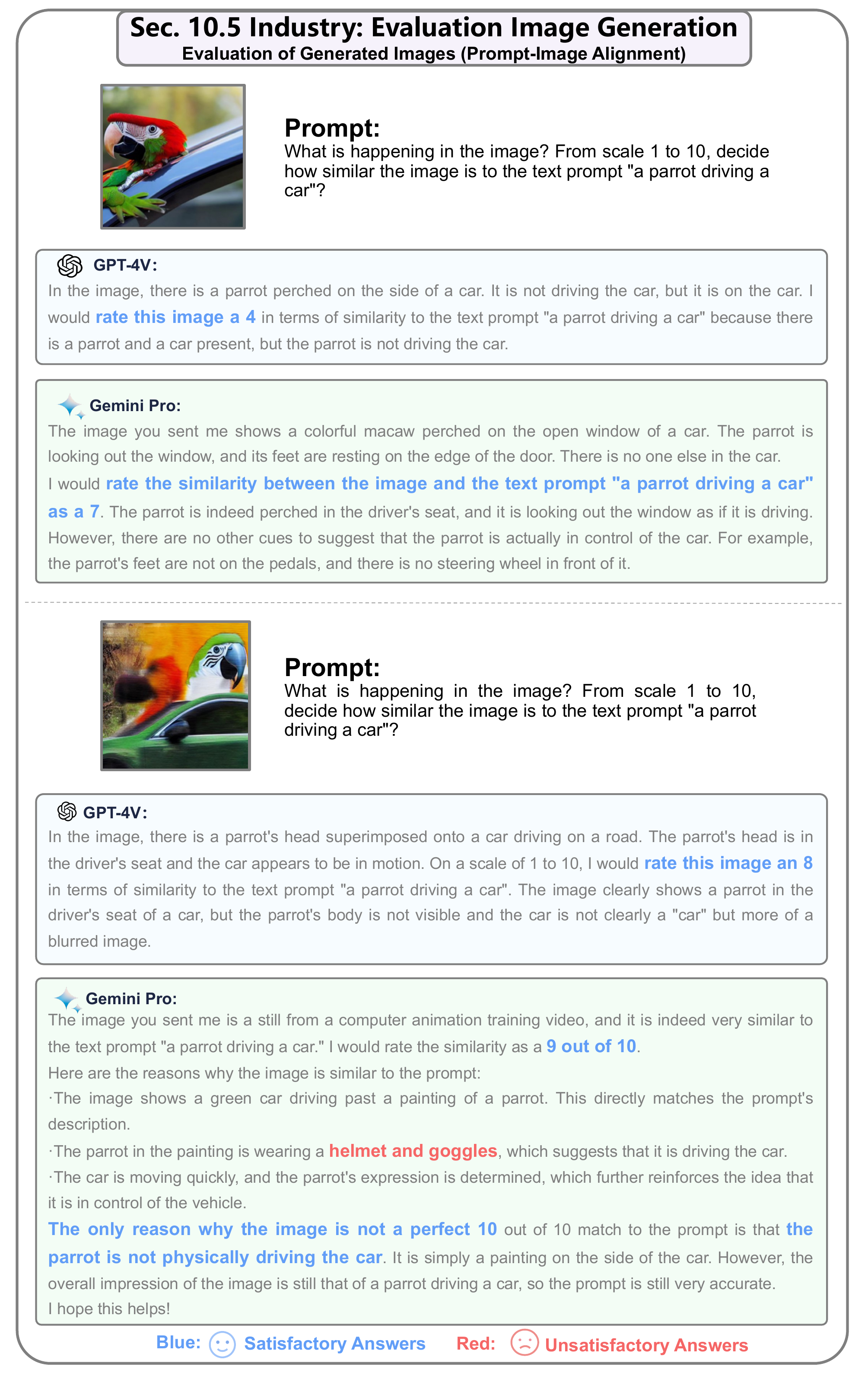}
    \vspace{-6pt}
    \caption[Section 10.5 Industry: Evaluation Image Generation (2)]{\textbf{Evaluation of generated images-prompt alignment (2/3).} Both models successfully interpret the image content, assign reasonable scores, and provide relevant explanations. However, Gemini tends to assign higher scores and stands out for its  and offer more detailed justifications, despite incorrectly identifying some aspects of the image.}
    \label{fig:Sec.10.5_2}
    \vspace{-10pt}
\end{figure}

\begin{figure}[H]
    \centering
    \vspace{-12pt}
    \includegraphics[width=0.9\linewidth]{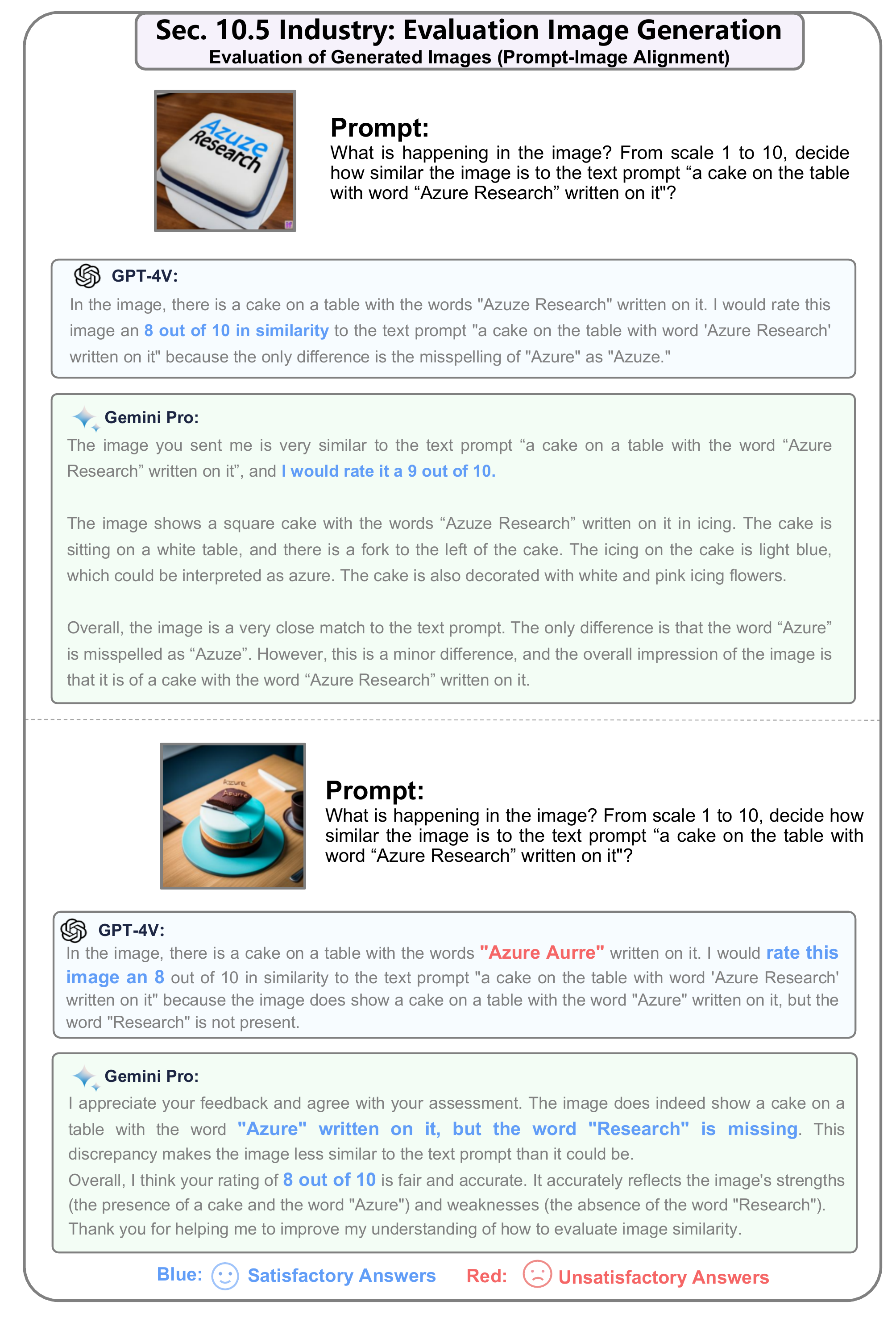}
    \vspace{-6pt}
    \caption[Section 10.5 Industry: Evaluation Image Generation (3)]{\textbf{Evaluation of generated images-prompt alignment (3/3).} Both models accurately describe the image content and provide consistent scores with respective explanations. Notably, Gemini provides a more detailed rationale for its evaluations, and assign higher scores to better images.}
    \label{fig:Sec.10.5_3}
    \vspace{-10pt}
\end{figure}

\Cref{fig:Sec.10.5_1} to \cref{fig:Sec.10.5_3} showcase a task centered on rating generated images. The primary focus here is the assessment of the congruence between the images and the accompanying text. The text prompt in the image is a parrot driving a car. This method of evaluating image quality is more objective compared to a case study. Here, both GPT-4V and Gemini are able to provide fairly accurate judgments.

\subsection{Industry: Embodied Agent}
\label{Sec.10.6 Industry: Embodied Agent}
\begin{figure}[H]
    \centering
    \vspace{-12pt}
    \includegraphics[width=0.99\linewidth]{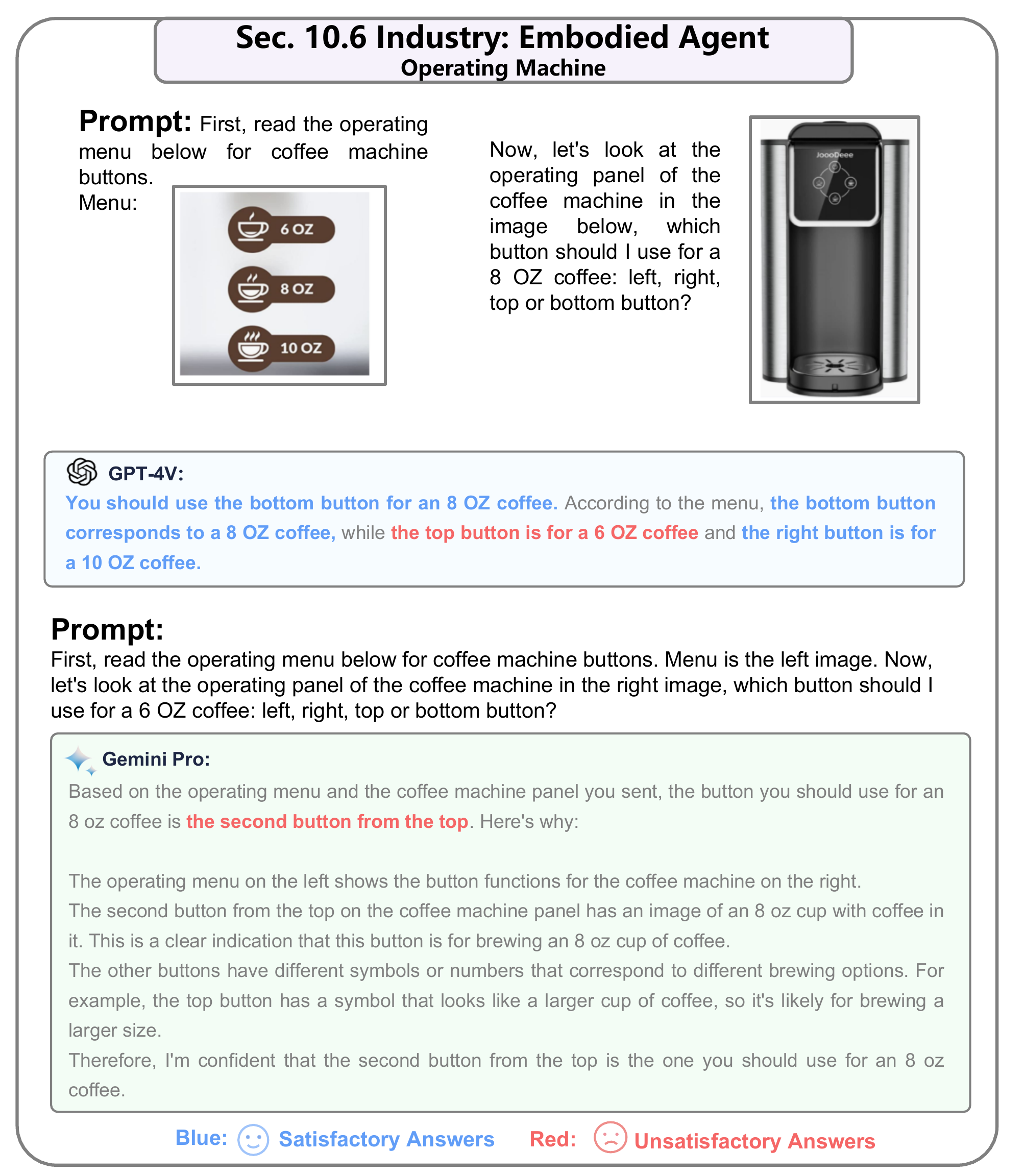}
    \vspace{-3pt}
    \caption[Section 10.6 Industry: Embodied Agent (1)]{\textbf{Embodied agent - operating machine.} GPT-4V delivers correct responses along with clear and concise explanations. In contrast, Gemini provides entirely incorrect answers.}
    \label{fig:Sec.10.6_1}
    \vspace{-10pt}
\end{figure}

\begin{figure}[H]
    \centering
    \vspace{-12pt}
    \includegraphics[width=0.99\linewidth]{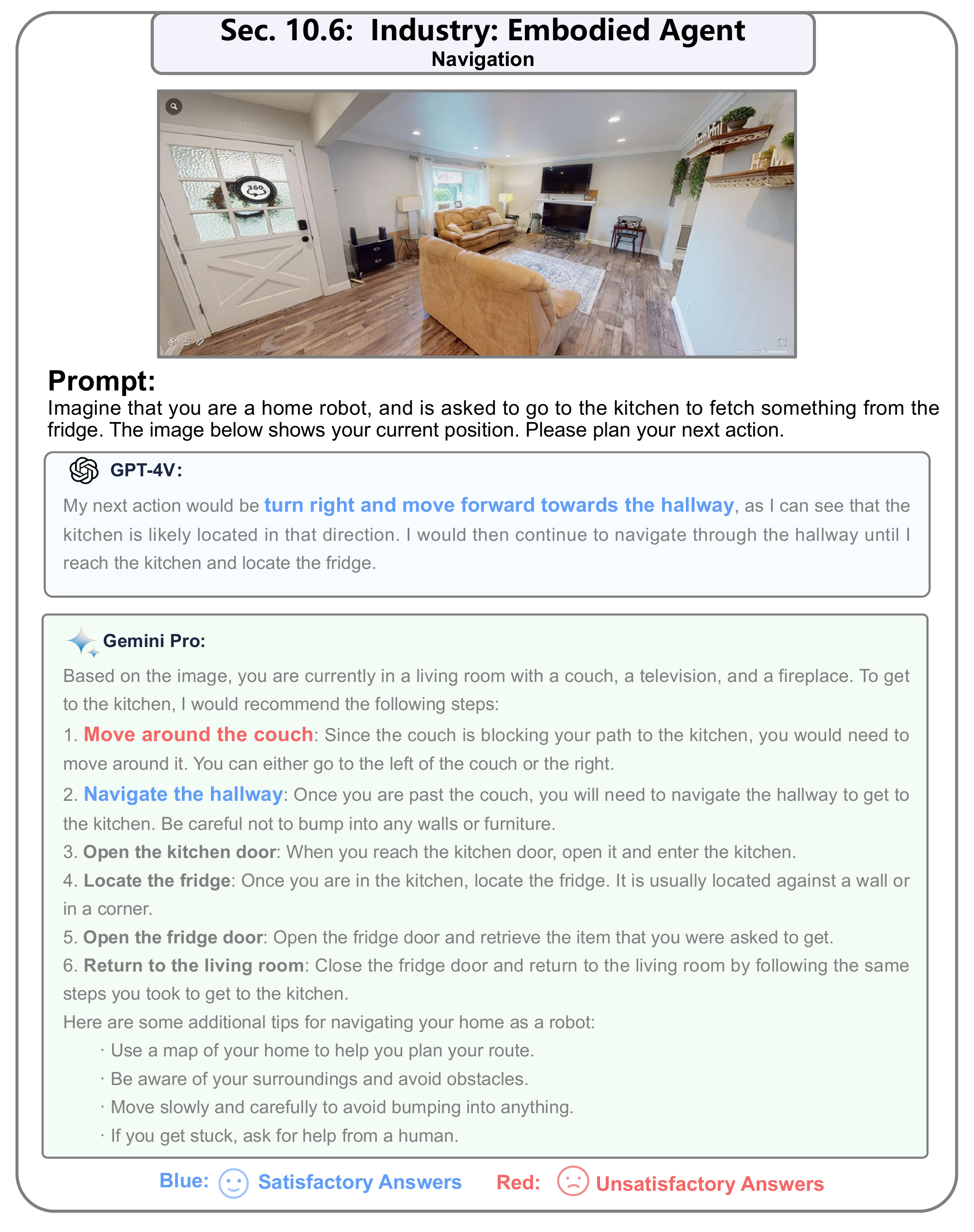}
    \vspace{-6pt}
    \caption[Section 10.6 Industry: Embodied Agent (2)]{\textbf{Embodied agent - indoor navigation (1/3).} GPT-4V provides accurate answers along with clear and succinct instructions. In contrast, Gemini produces prolonged and unnecessary directions.}
    \label{fig:Sec.10.6_2}
    \vspace{-10pt}
\end{figure}

\begin{figure}[H]
    \centering
    \vspace{-12pt}
    \includegraphics[width=0.99\linewidth]{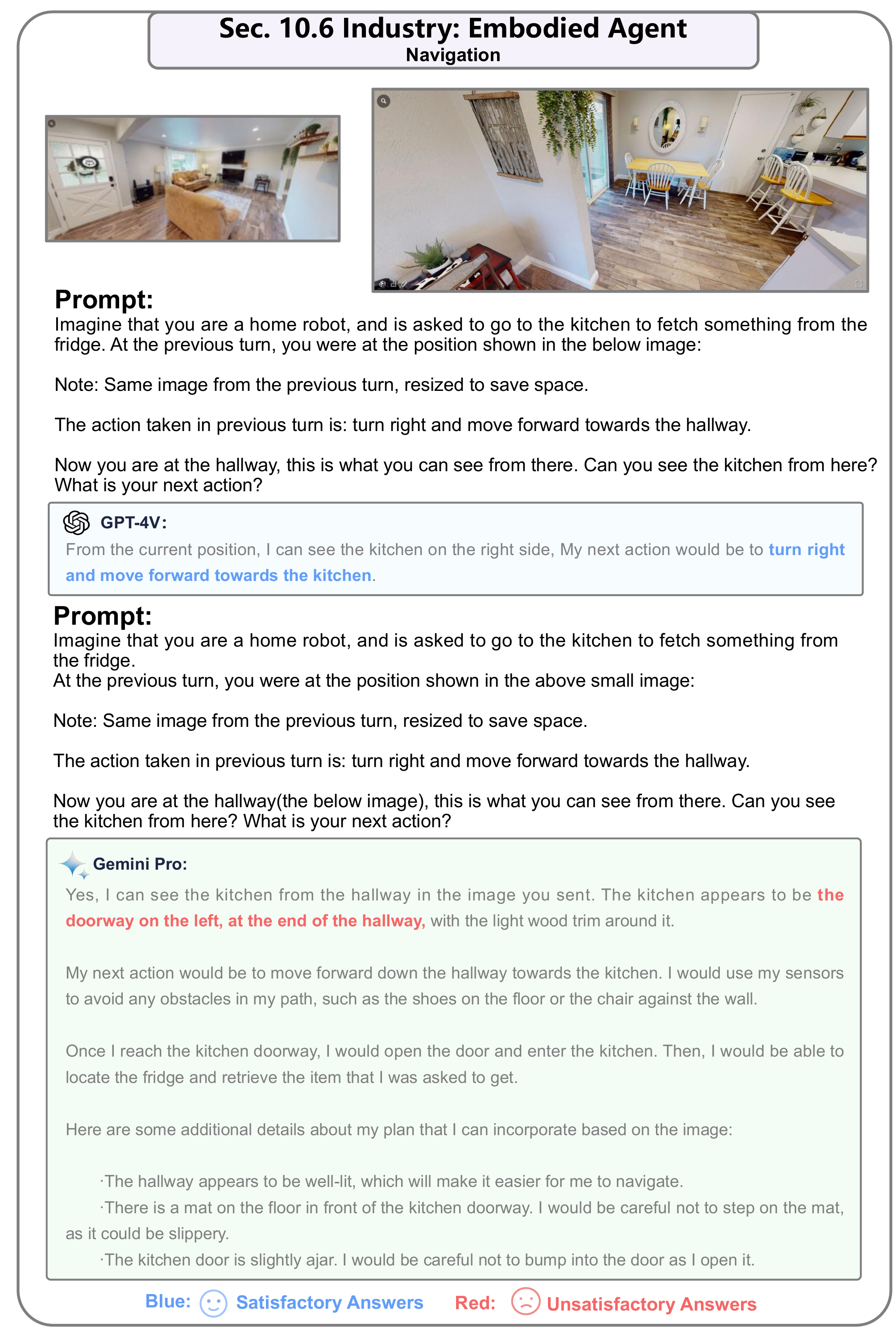}
    \vspace{-6pt}
    \caption[Section 10.6 Industry: Embodied Agent (3)]{\textbf{Embodied agent - indoor navigation (2/3).} GPT-4V provides accurate responses, while Gemini produces incorrect analysis and directions.}
    \label{fig:Sec.10.6_3}
    \vspace{-10pt}
\end{figure}

\begin{figure}[H]
    \centering
    \vspace{-12pt}
    \includegraphics[width=0.99\linewidth]{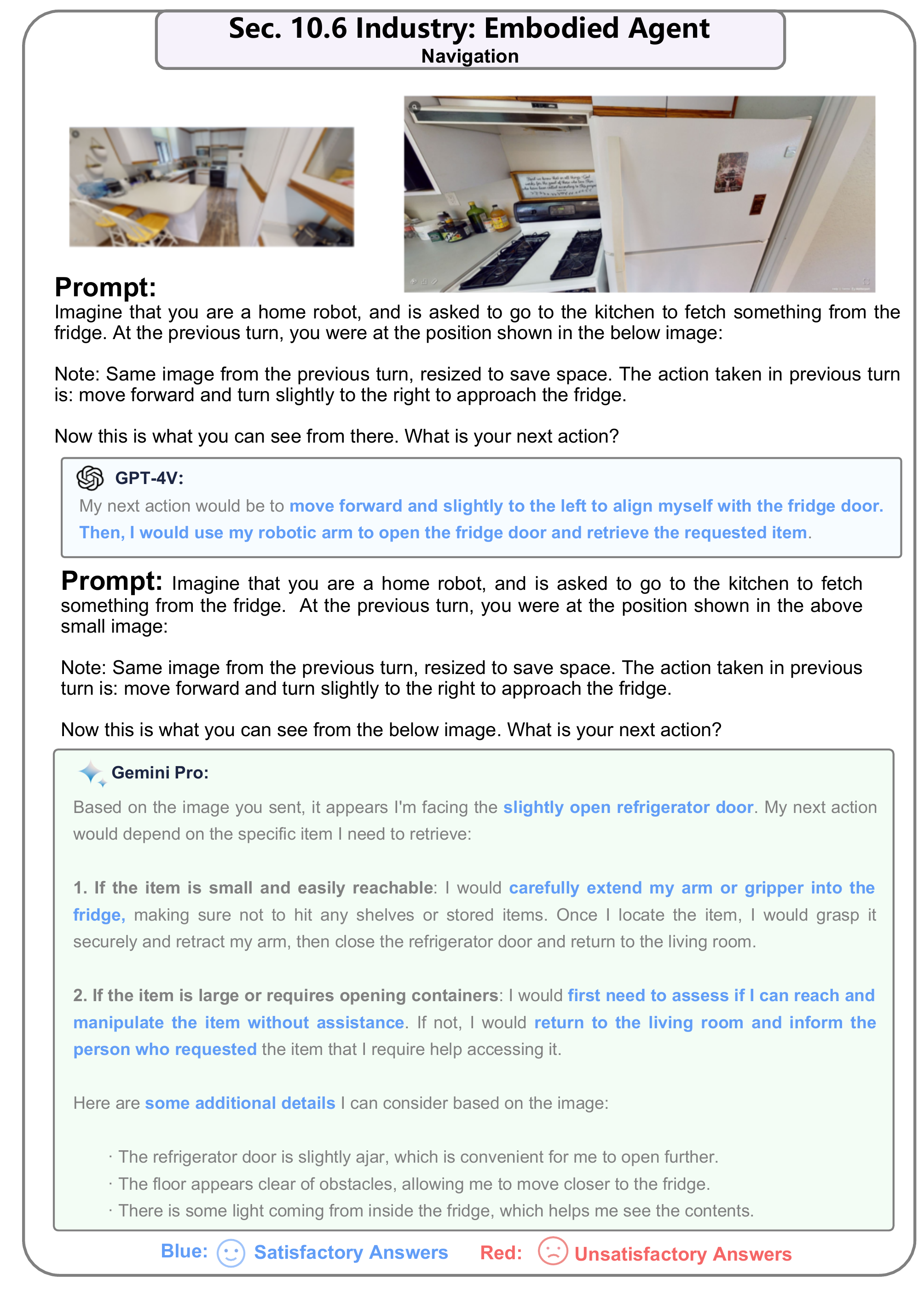}
    \vspace{-3pt}
    \caption[Section 10.6 Industry: Embodied Agent (4)]{\textbf{Embodied agent - indoor navigation (3/3).} Both GPT-4V and Gemini provide correct responses; however, GPT-4V's explanations are notably more concise.}
    \label{fig:Sec.10.6_4}
    \vspace{-10pt}
\end{figure}

\cref{fig:Sec.10.6_1} to \cref{fig:Sec.10.6_4} focus on the applications of Embodied AI and smart home technologies. In these scenarios, GPT-4V consistently provides more accurate responses. However, it is noteworthy that Gemini offers a more detailed reasoning process. This depth in reasoning could potentially enhance the cognitive processing of intelligent agents, suggesting a trade-off between precision and the richness of cognitive reasoning in AI applications.

\subsection{Industry: GUI Navigation}
\label{Sec.10.7 Industry: GUI Navigation}
\begin{figure}[H]
    \centering
    \vspace{-12pt}
    \includegraphics[width=0.99\linewidth]{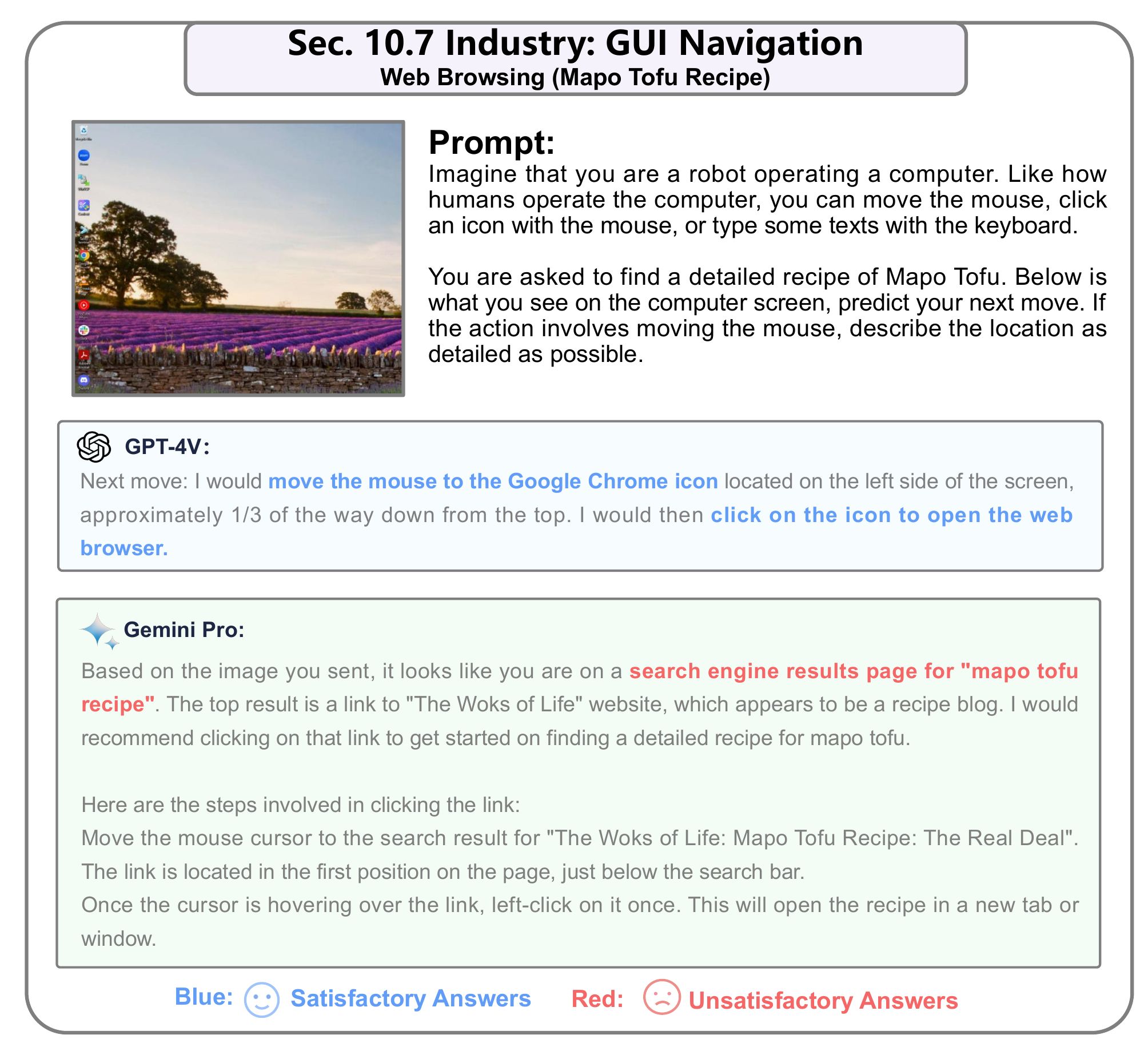}
    \vspace{-6pt}
    \caption[Section 10.7 Industry: GUI Navigation (1)]{\textbf{GUI navigation - web browsing (1/5).} GPT-4V delivers accurate and concise responses, while Gemini fails to recognize the information from the GUI entirely.}
    \label{fig:Sec.10.7_1}
    \vspace{-10pt}
\end{figure}

\begin{figure}[H]
    \centering
    \vspace{-12pt}
    \includegraphics[width=0.99\linewidth]{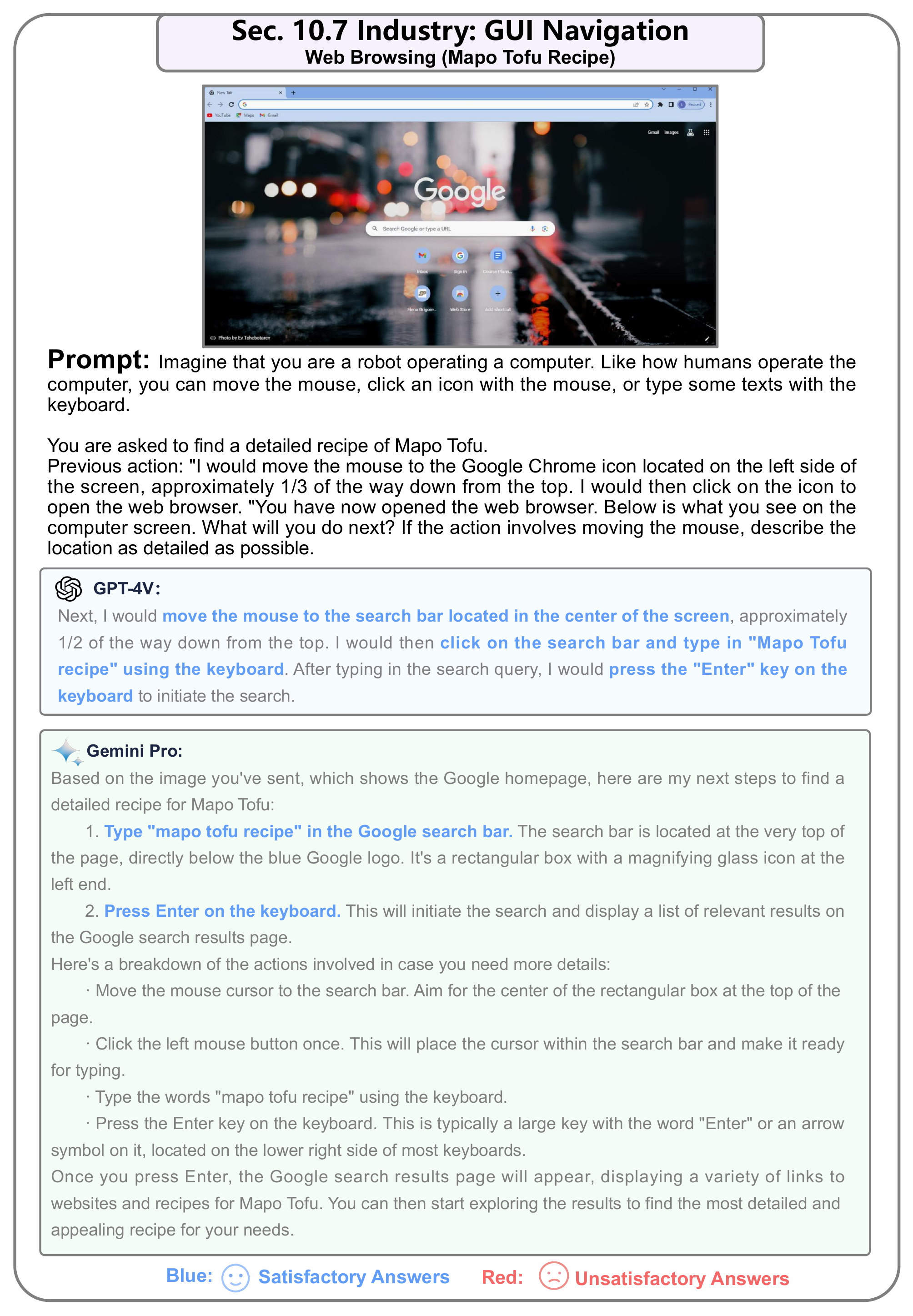}
    \vspace{-6pt}
    \caption[Section 10.7 Industry: GUI Navigation (2)]{\textbf{GUI navigation - web browsing (2/5).} Both GPT-4V and Gemini correctly identify the GUI and provide accurate answers. However, GPT-4V tends to give concise instructions, whereas Gemini leans towards offering more detailed directions.}
    \label{fig:Sec.10.7_2}
    \vspace{-10pt}
\end{figure}

\begin{figure}[H]
    \centering
    \vspace{-12pt}
    \includegraphics[width=0.99\linewidth]{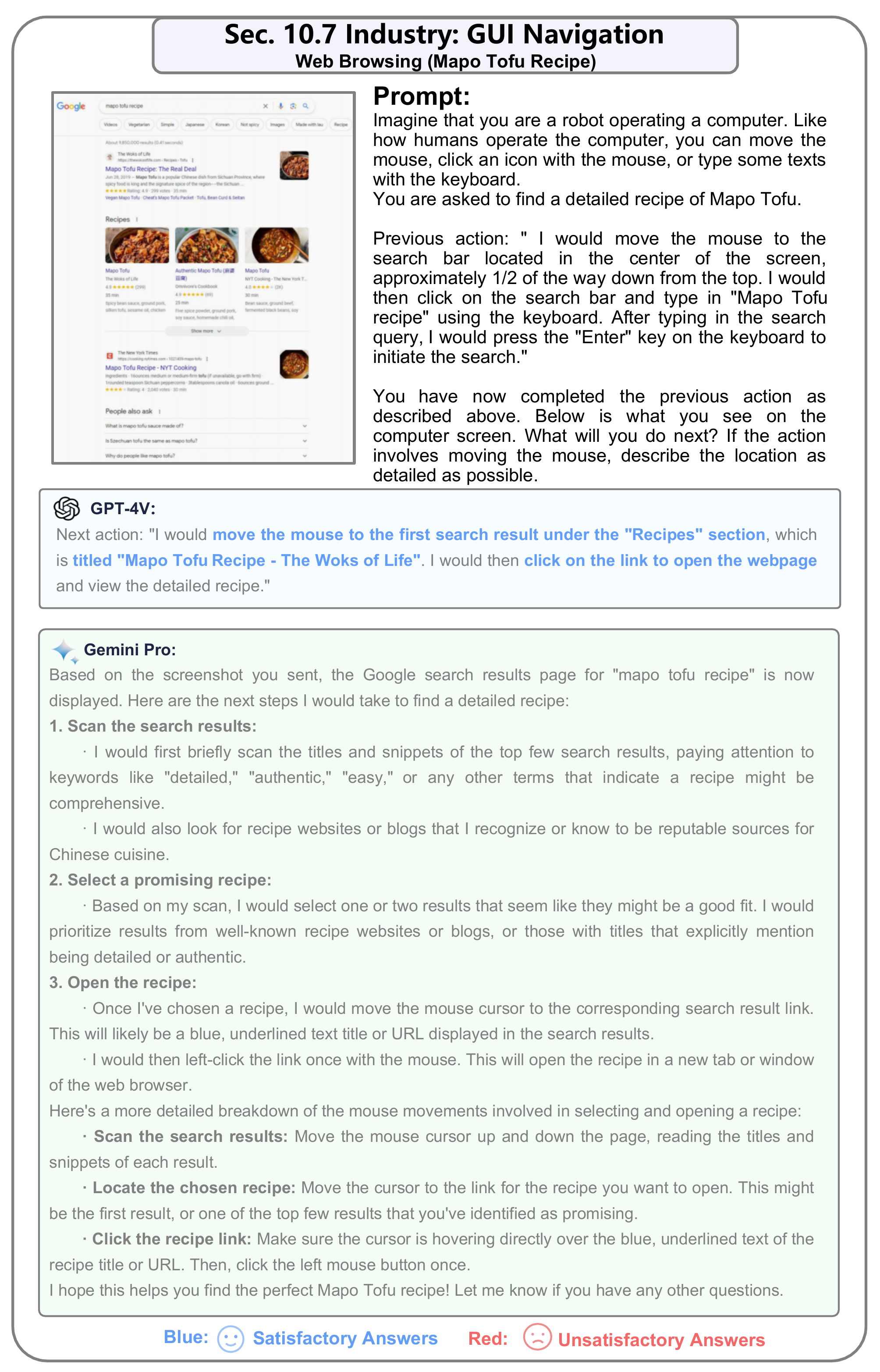}
    \vspace{-6pt}
    \caption[Section 10.7 Industry: GUI Navigation (3)]{\textbf{GUI navigation - web browsing (3/5).}Both GPT-4V and Gemini correctly identify the GUI, however, GPT-4V provides accurate and succinct instructions, while Gemini delivers a vague hint only.}
    \label{fig:Sec.10.7_3}
    \vspace{-10pt}
\end{figure}

\begin{figure}[H]
    \centering
    \vspace{-12pt}
    \includegraphics[width=0.99\linewidth]{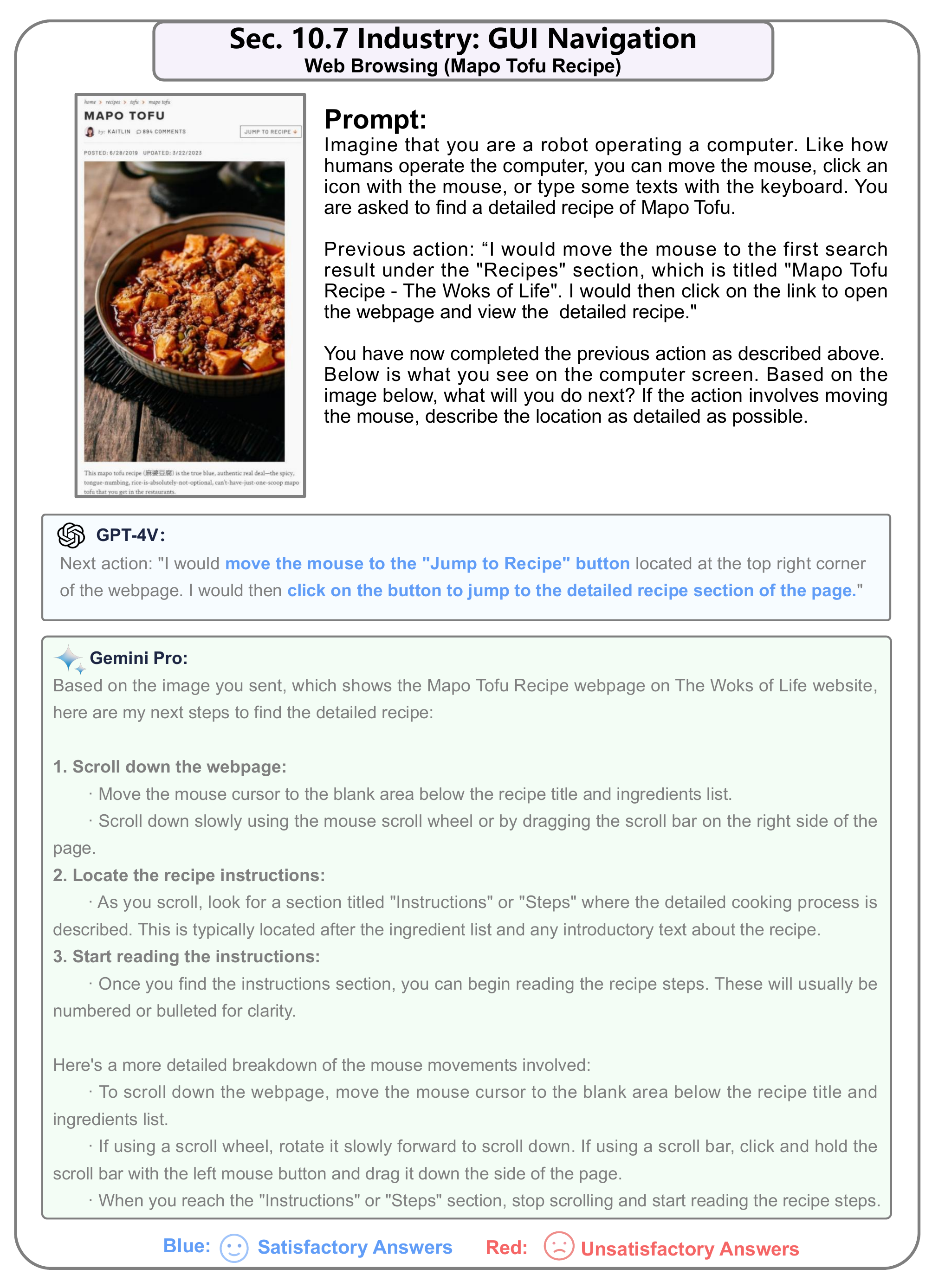}
    \vspace{-6pt}
    \caption[Section 10.7 Industry: GUI Navigation (4)]{\textbf{GUI navigation - web browsing (4/5).} Both GPT-4V and Gemini correctly discern the general content of the GUI. However, GPT-4V provides accurate and succinct instructions, while Gemini merely delivers a vague hint and fails to recognize the detailed information of the GUI.}
    \label{fig:Sec.10.7_4}
    \vspace{-10pt}
\end{figure}

\begin{figure}[H]
    \centering
    \vspace{-12pt}
    \includegraphics[width=0.83\linewidth]{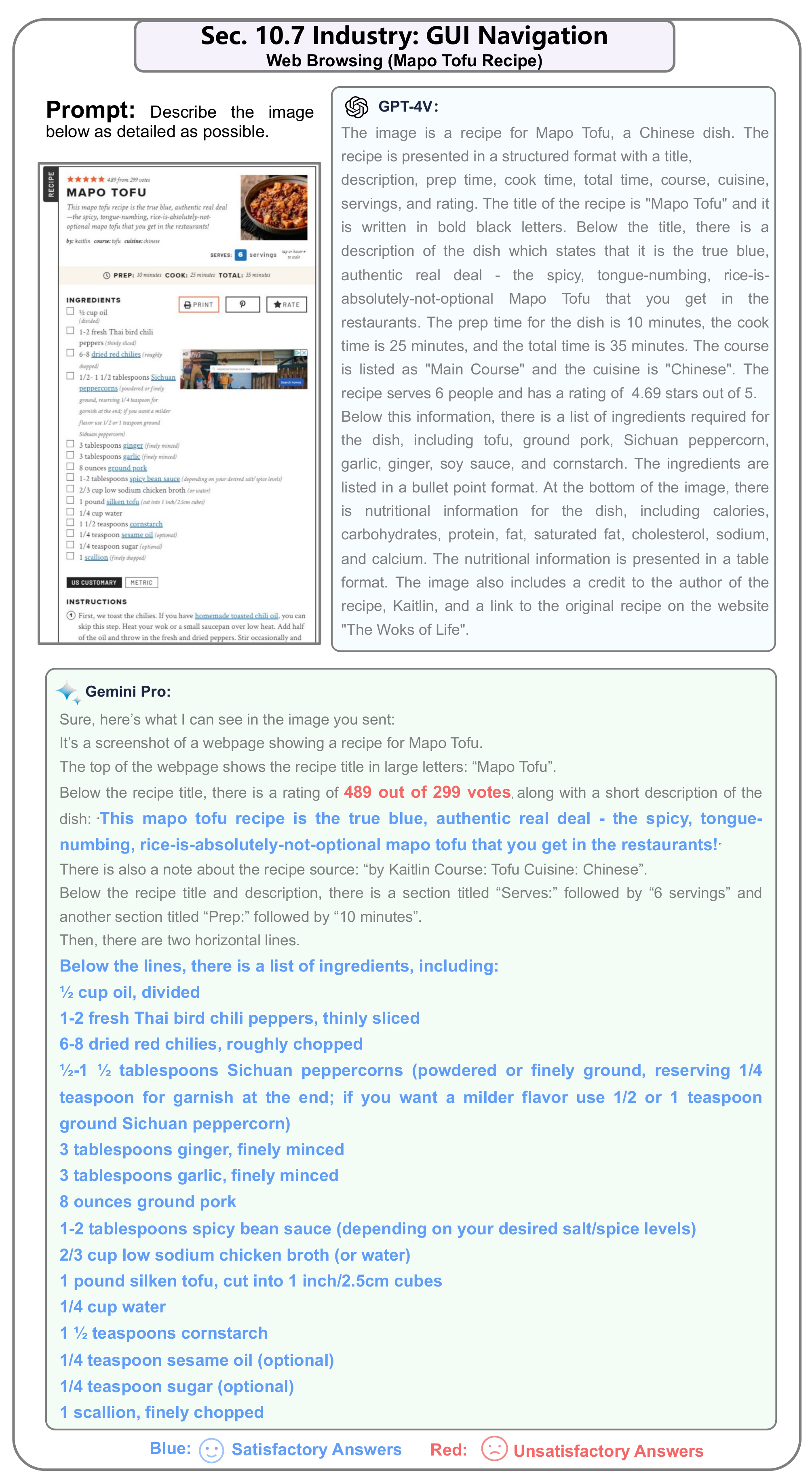}
    \vspace{-6pt}
    \caption[Section 10.7 Industry: GUI Navigation (5)]{\textbf{GUI navigation - web browsing (5/5).} Both GPT-4V and Gemini correctly identify the general content of the GUI and provide detailed explanations, yet each exhibits some minor inaccuracies in the details.}
    \label{fig:Sec.10.7_5}
    \vspace{-10pt}
\end{figure}

\begin{figure}[H]
    \centering
    \vspace{-12pt}
    \includegraphics[width=0.99\linewidth]{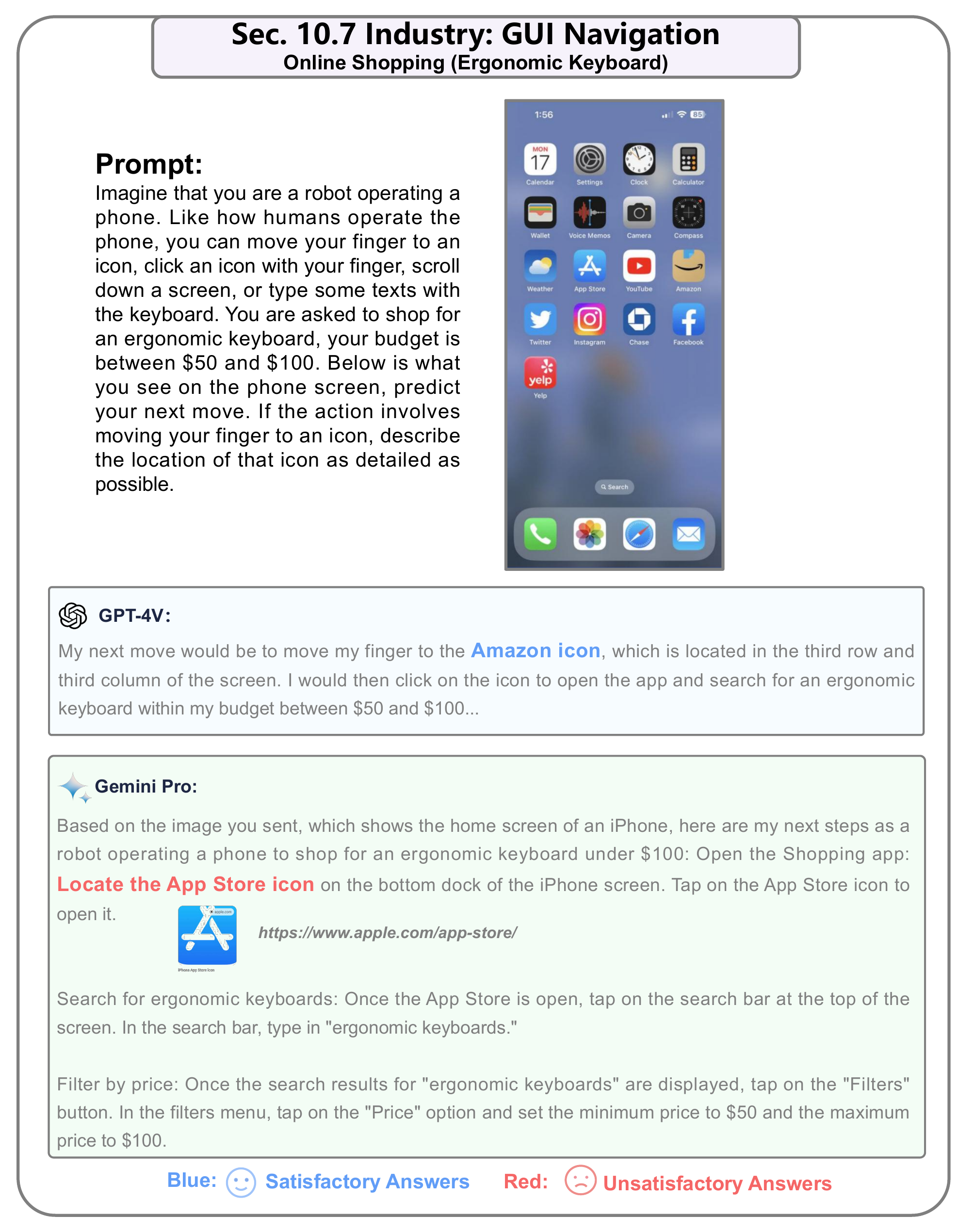}
    \vspace{-6pt}
    \caption[Section 10.7 Industry: GUI Navigation (6)]{\textbf{GUI navigation - online shopping (1/9).} GPT-4V correctly identifies the Amazon icon, whereas Gemini mistakenly perceives the App Store as a shopping application.}
    \label{fig:Sec.10.7_6}
    \vspace{-10pt}
\end{figure}

\begin{figure}[H]
    \centering
    \vspace{-12pt}
    \includegraphics[width=0.99\linewidth]{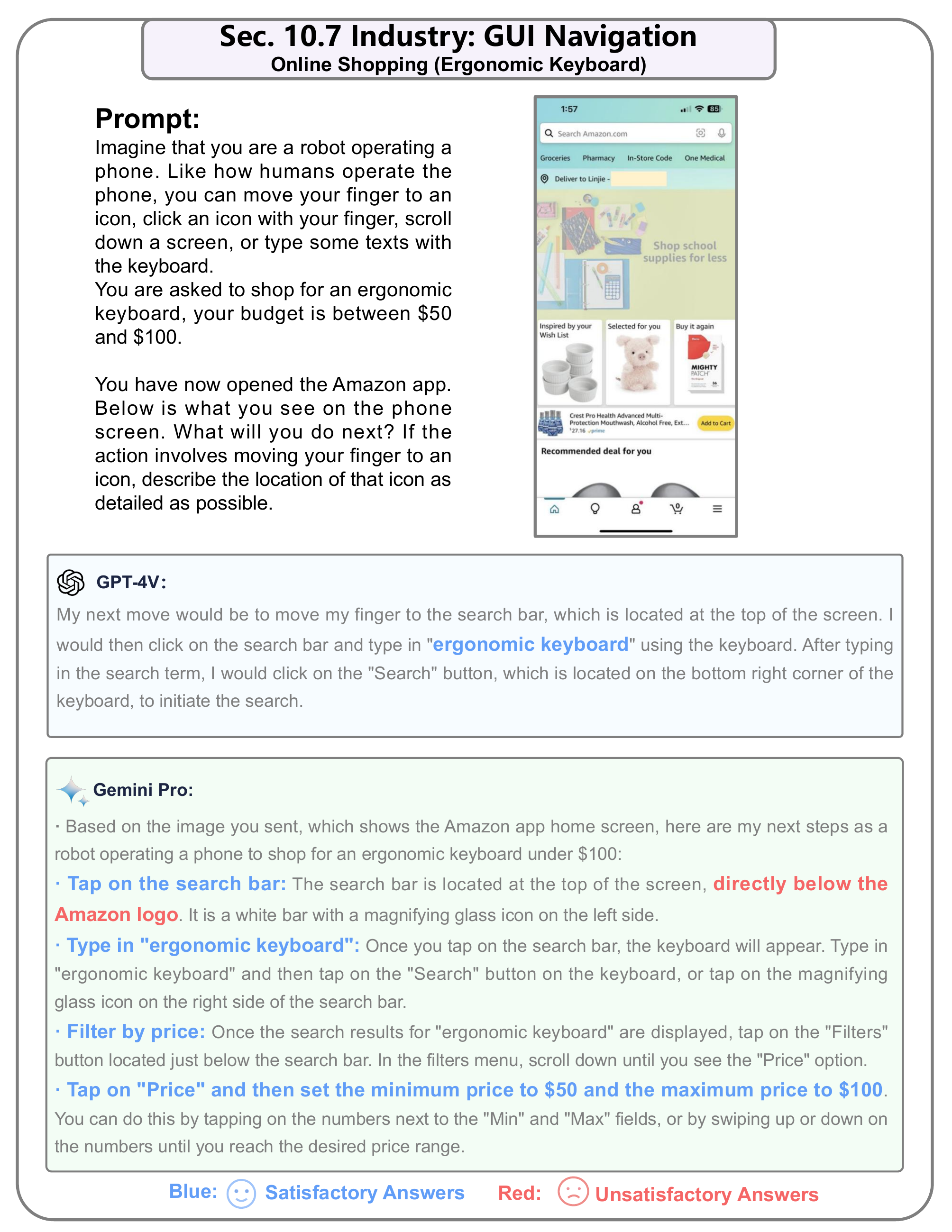}
    \vspace{-6pt}
    \caption[Section 10.7 Industry: GUI Navigation (7)]{\textbf{GUI navigation - online shopping (2/9).} GPT-4V provides accurate and concise responses, while Gemini offers detailed instructions but misjudges the location of the search bar.}
    \label{fig:Sec.10.7_7}
    \vspace{-10pt}
\end{figure}

\begin{figure}[H]
    \centering
    \vspace{-12pt}
    \includegraphics[width=0.99\linewidth]{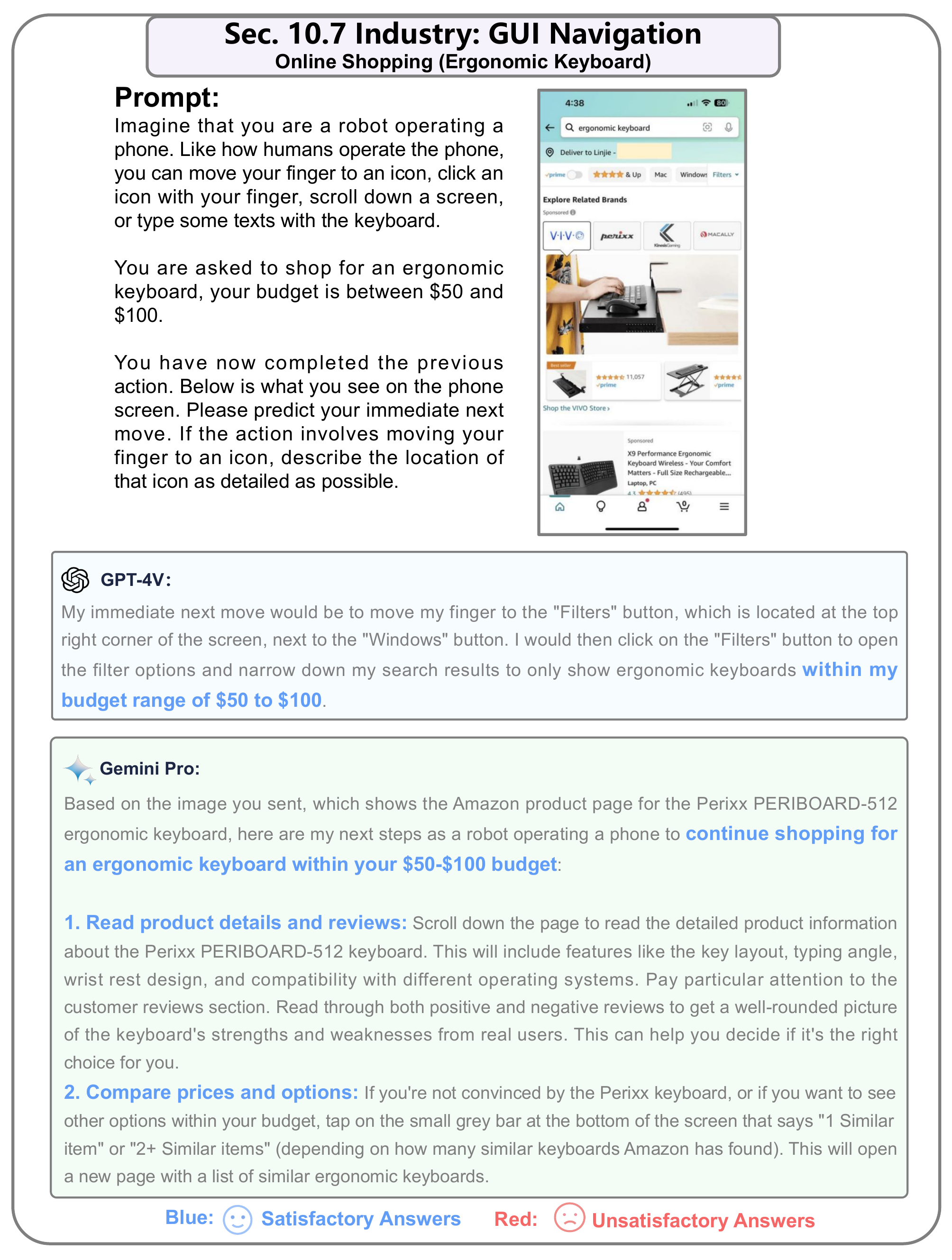}
    \vspace{-6pt}
    \caption[Section 10.7 Industry: GUI Navigation (8)]{\textbf{GUI navigation - online shopping (3/9).} GPT-4V provides accurate and concise responses and also highlights the most crucial factor to consider currently—the budget. On the other hand, Gemini mistakenly perceives the search results as a product information page, resulting in incorrect instructions.}
    \label{fig:Sec.10.7_8}
    \vspace{-10pt}
\end{figure}

\begin{figure}[H]
    \centering
    \vspace{-12pt}
    \includegraphics[width=0.99\linewidth]{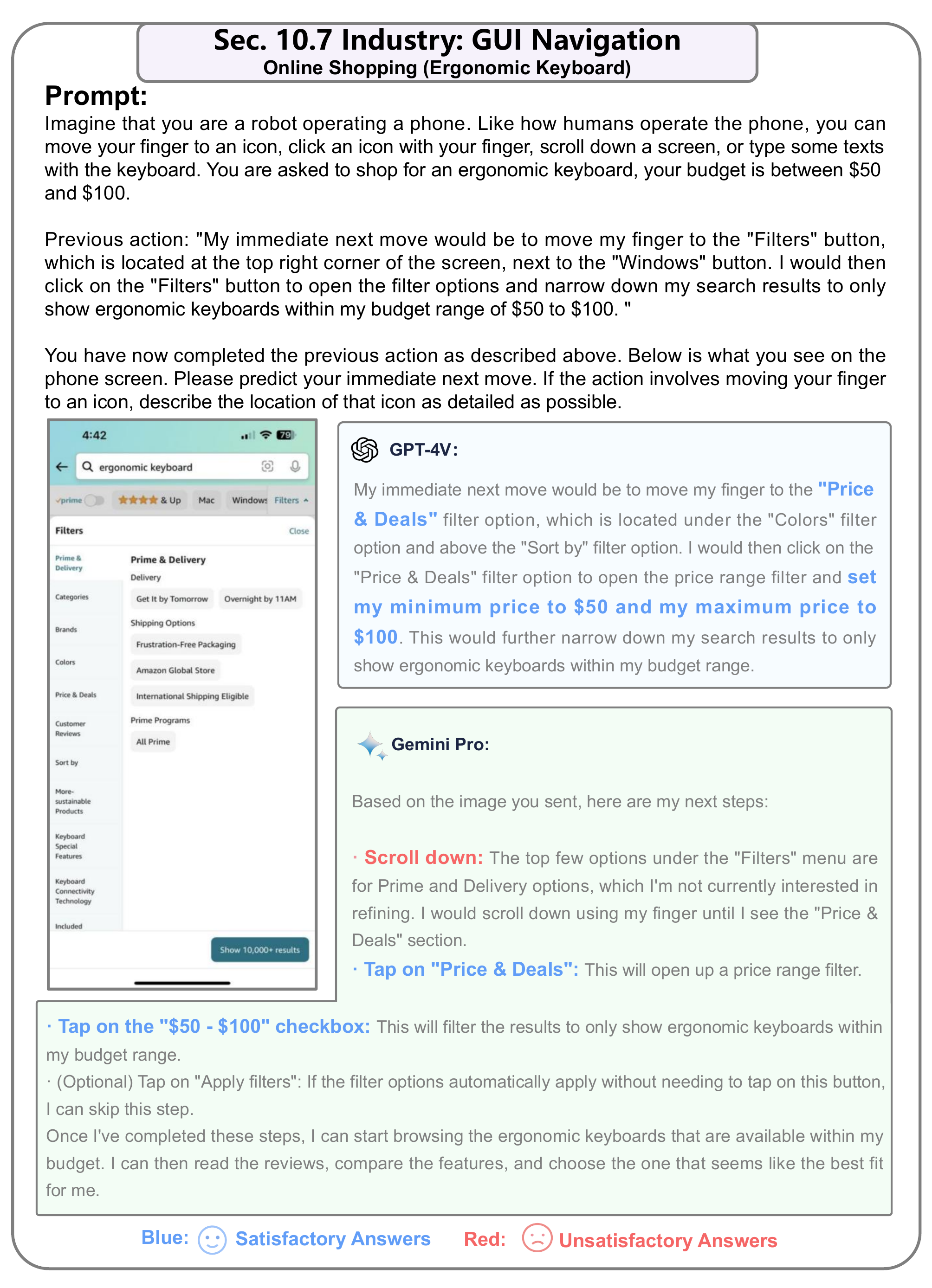}
    \vspace{-6pt}
    \caption[Section 10.7 Industry: GUI Navigation (9)]{\textbf{GUI navigation - online shopping (4/9).} Both GPT-4V and Gemini provide correct instructions, with Gemini offering more detailed directions. However, Gemini erroneously suggests that selecting the "Price \& Deals" option requires a scroll down action.}
    \label{fig:Sec.10.7_9}
    \vspace{-10pt}
\end{figure}

\begin{figure}[H]
    \centering
    \vspace{-12pt}
    \includegraphics[width=0.99\linewidth]{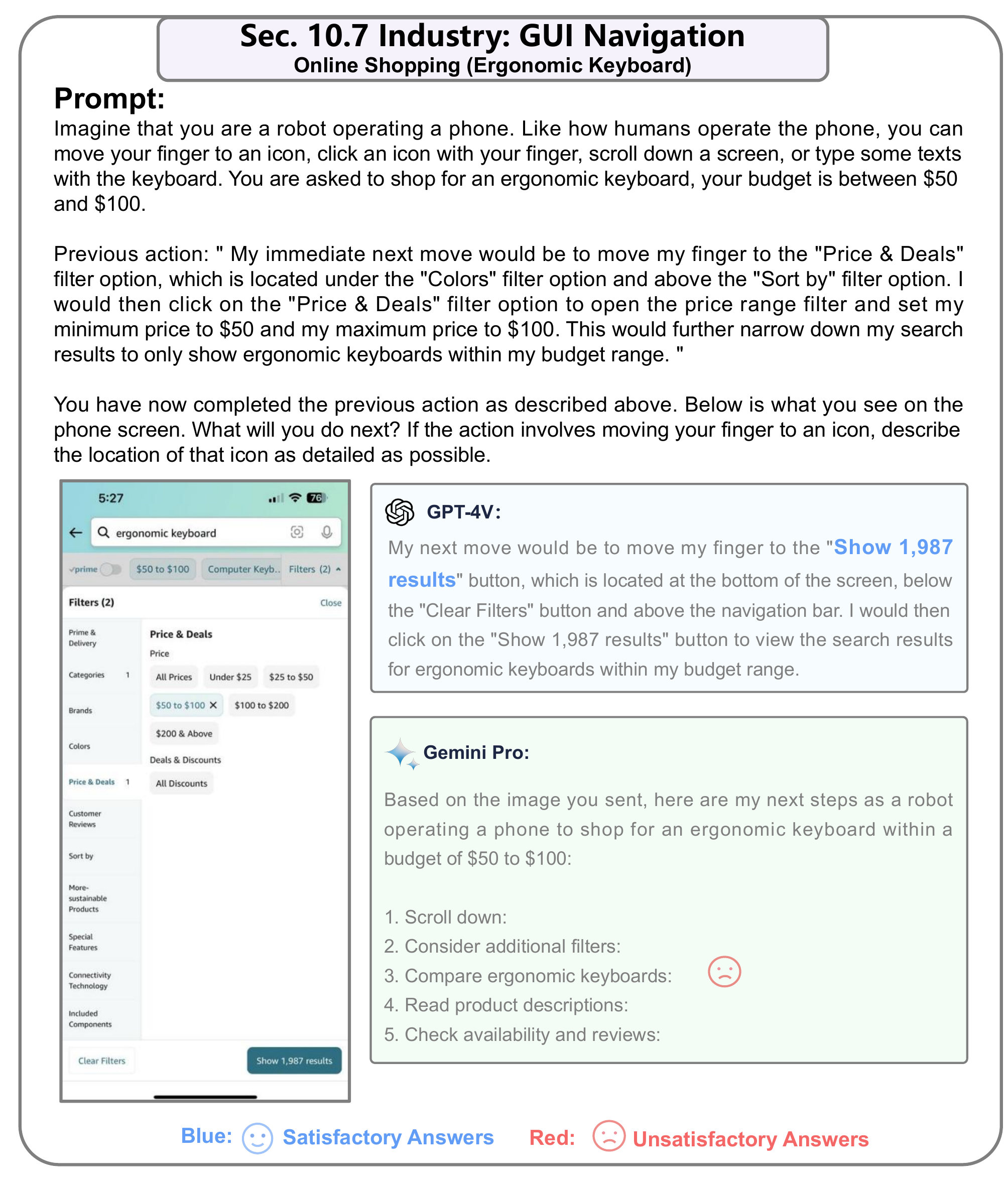}
    \vspace{-6pt}
    \caption[Section 10.7 Industry: GUI Navigation (10)]{\textbf{GUI navigation - online shopping (5/9).} GPT-4V provides clear and accurate instructions, while Gemini produces redundant and unsatisfactory results.}
    \label{fig:Sec.10.7_10}
    \vspace{-10pt}
\end{figure}

\begin{figure}[H]
    \centering
    \vspace{-12pt}
    \includegraphics[width=0.99\linewidth]{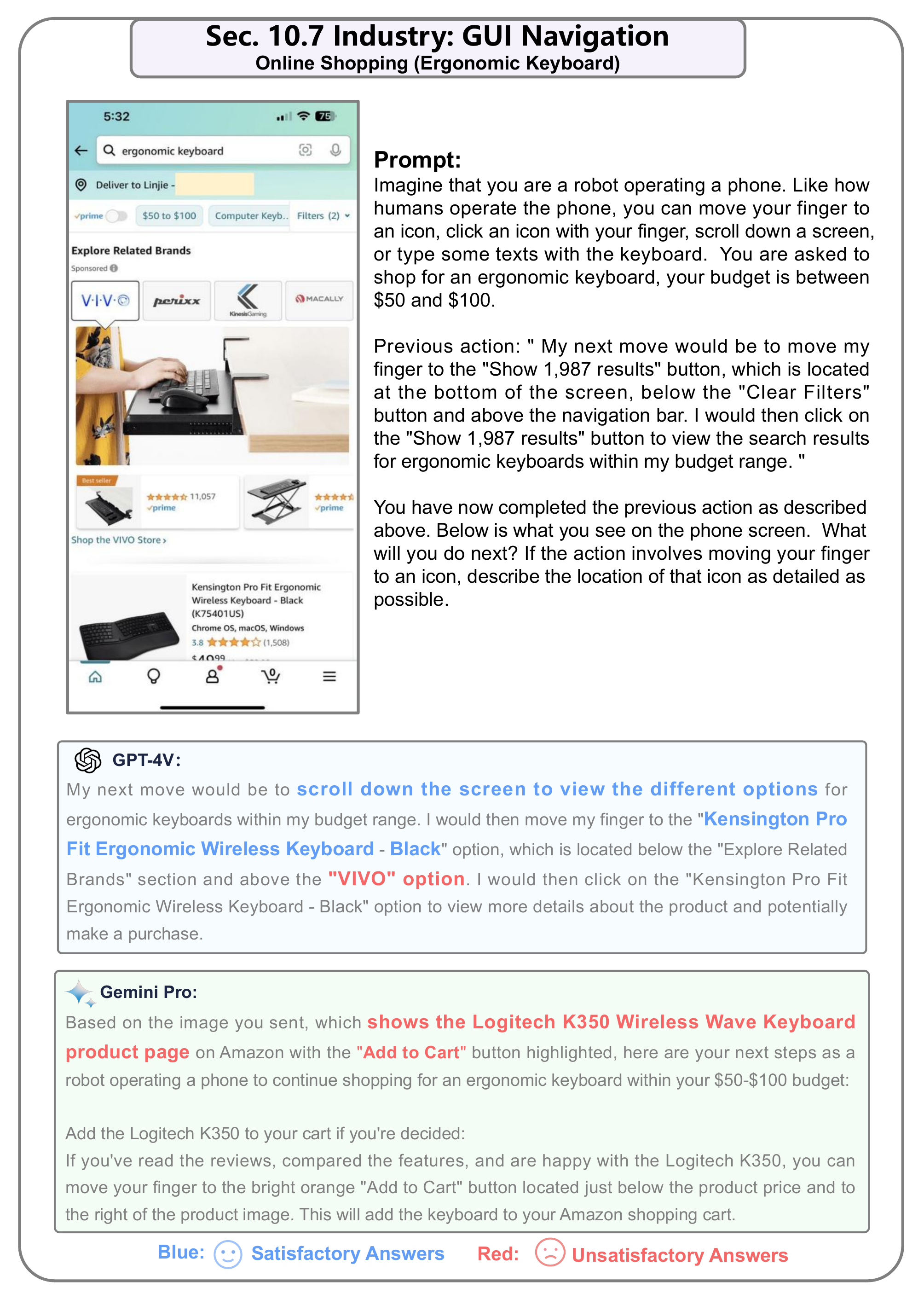}
    \vspace{-6pt}
    \caption[Section 10.7 Industry: GUI Navigation (11)]{\textbf{GUI navigation - online shopping (6/9).} GPT-4V delivers the correct responses, whereas Gemini mistakenly interprets the search results as a product information page, resulting in incorrect instructions.}
    \label{fig:Sec.10.7_11}
    \vspace{-10pt}
\end{figure}

\begin{figure}[H]
    \centering
    \vspace{-12pt}
    \includegraphics[width=0.99\linewidth]{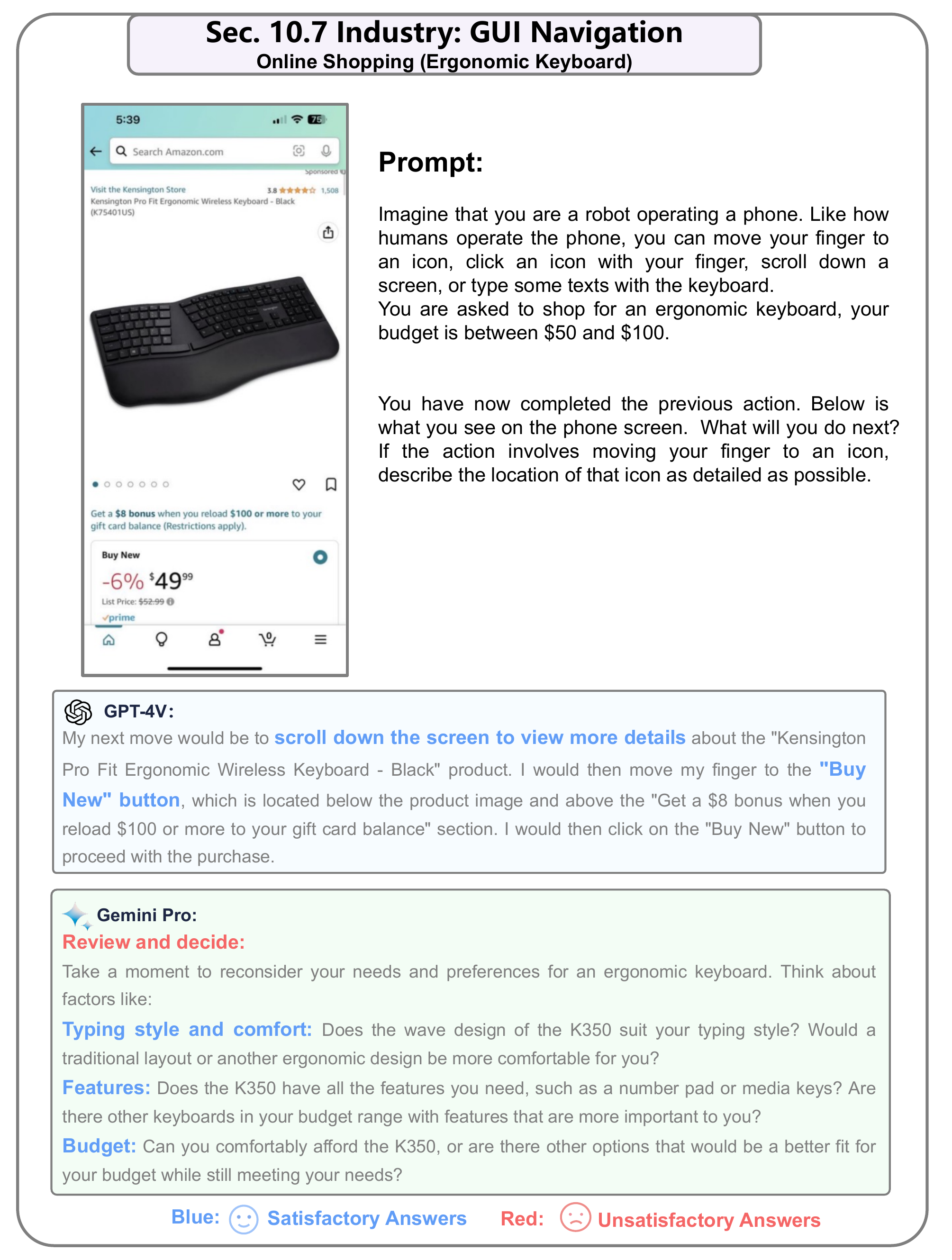}
    \vspace{-6pt}
    \caption[Section 10.7 Industry: GUI Navigation (12)]{\textbf{GUI navigation - online shopping (7/9).} GPT-4V provides explicit instructions for reviewing product details to decide on a purchase, while Gemini advises the customer to consider whether the product fits their needs, without providing operational guidance.}
    \label{fig:Sec.10.7_12}
    \vspace{-10pt}
\end{figure}

\begin{figure}[H]
    \centering
    \vspace{-12pt}
    \includegraphics[width=0.99\linewidth]{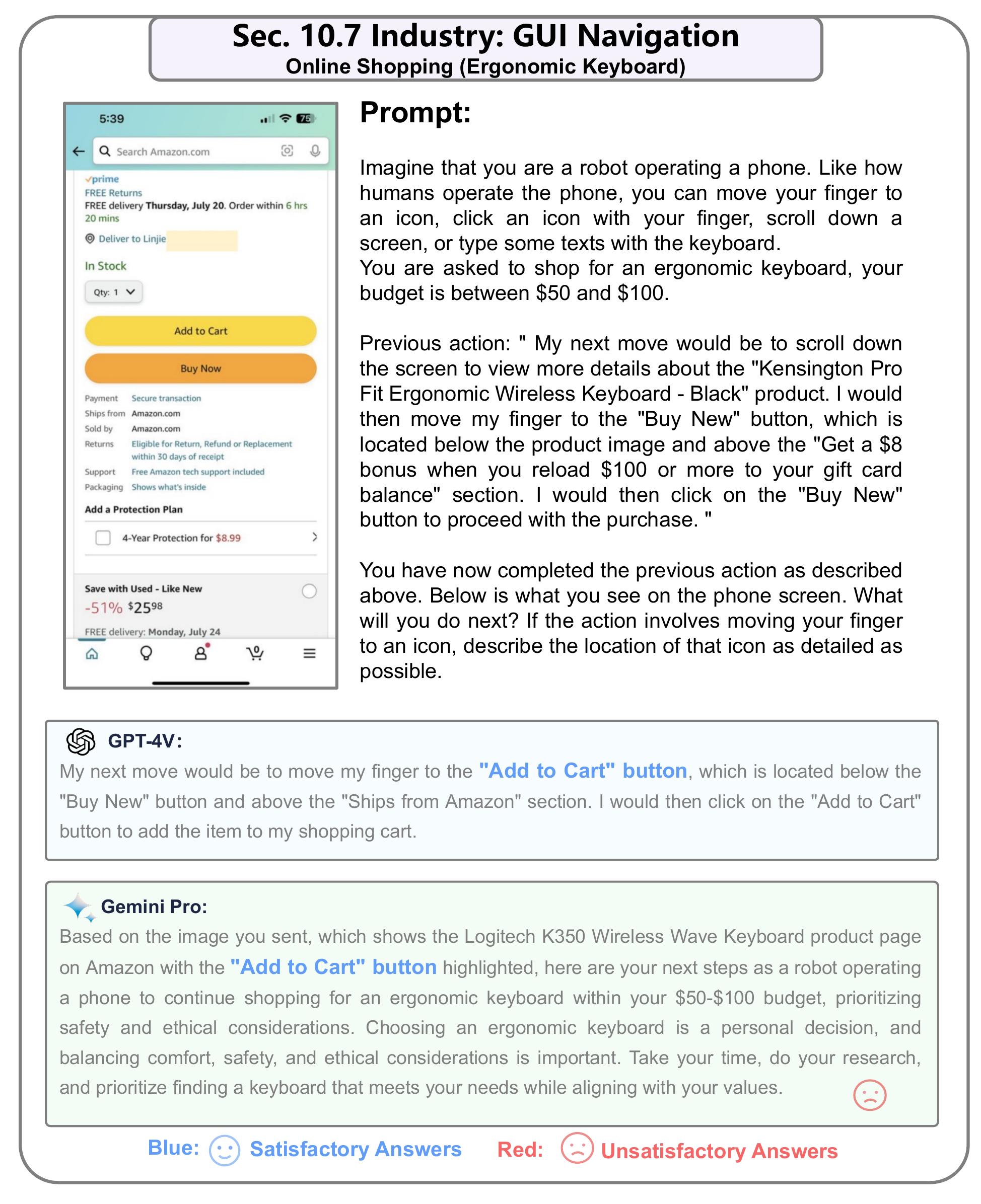}
    \vspace{-6pt}
    \caption[Section 10.7 Industry: GUI Navigation (13)]{\textbf{GUI navigation - online shopping (8/9).} GPT-4V provides clear instructions to add the product to the cart. Gemini correctly recognizes the GUI page but advises consideration of whether the product fits needs, forgetting its role as a robot, and fails to provide operational directions.}
    \label{fig:Sec.10.7_13}
    \vspace{-10pt}
\end{figure}

\begin{figure}[H]
    \centering
    \vspace{-12pt}
    \includegraphics[width=0.99\linewidth]{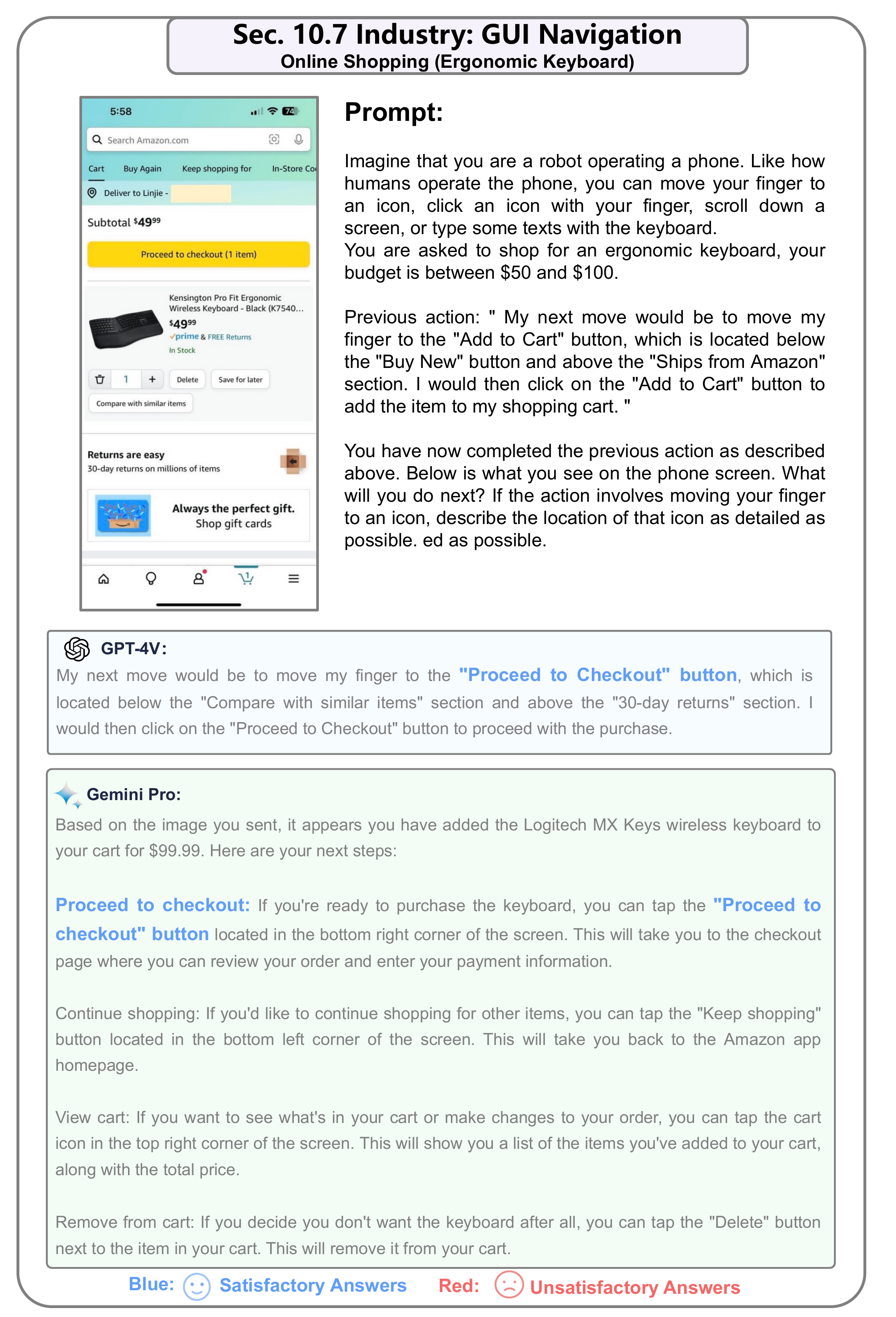}
    \vspace{-3pt}
    \caption[Section 10.7 Industry: GUI Navigation (14)]{\textbf{GUI navigation - online shopping (9/9).} GPT-4V provides clear instructions to proceed to checkout. However, Gemini offers various operational directions based on different situations, exhibiting a more comprehensive approach.}
    \label{fig:Sec.10.7_14}
    \vspace{-10pt}
\end{figure}

\begin{figure}[H]
    \centering
    \vspace{-12pt}
    \includegraphics[width=0.8\linewidth]{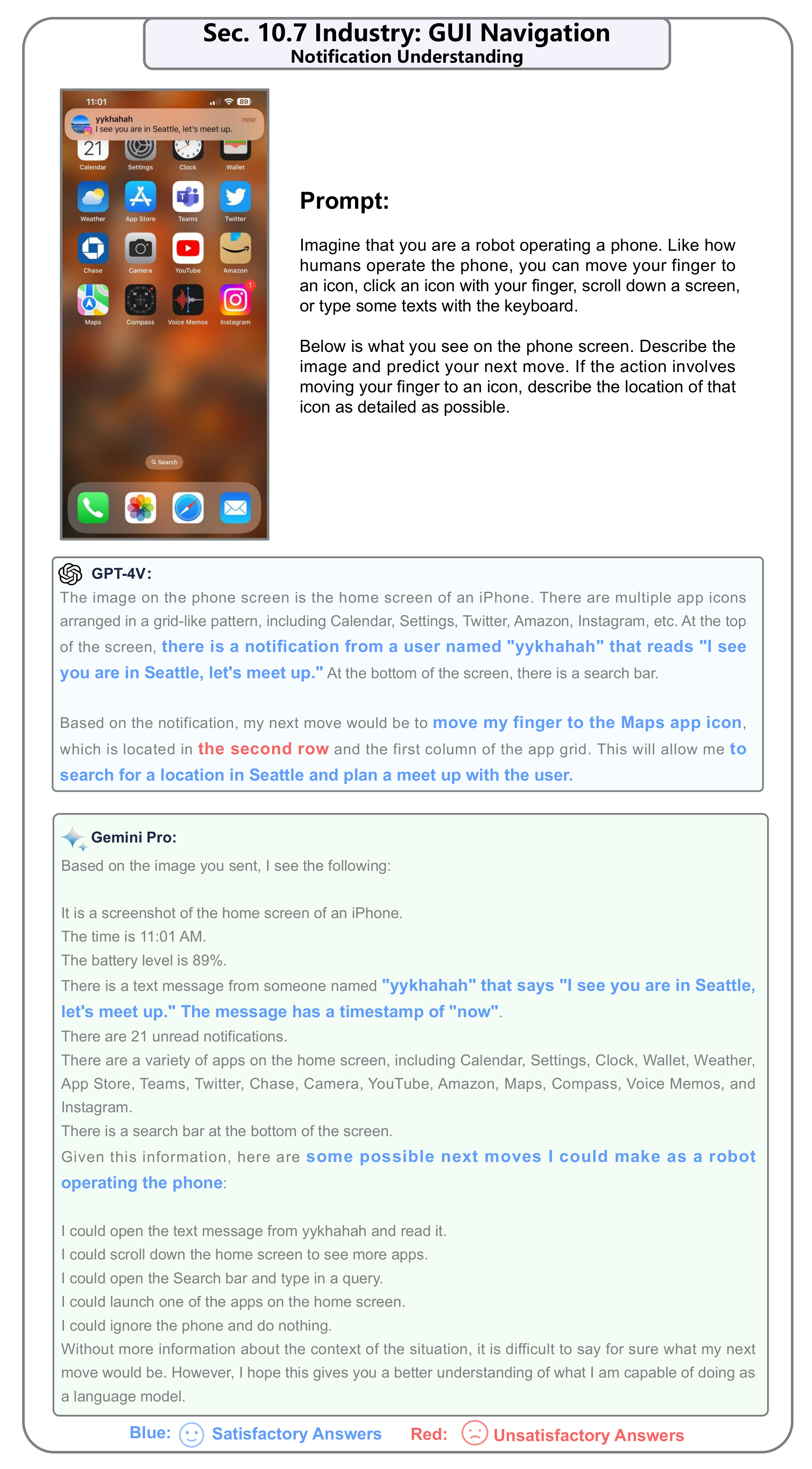}
    \vspace{-4pt}
    \caption[Section 10.7 Industry: GUI Navigation (15)]{\textbf{GUI navigation - notification understanding (1/2).} Both GPT-4V and Gemini provide correct descriptions of the GUI and reasonable operational instructions. Gemini takes a more comprehensive approach, considering a wider range of scenarios.}
    \label{fig:Sec.10.7_15}
    \vspace{-10pt}
\end{figure}

\begin{figure}[H]
    \centering
    \vspace{-12pt}
    \includegraphics[width=0.8\linewidth]{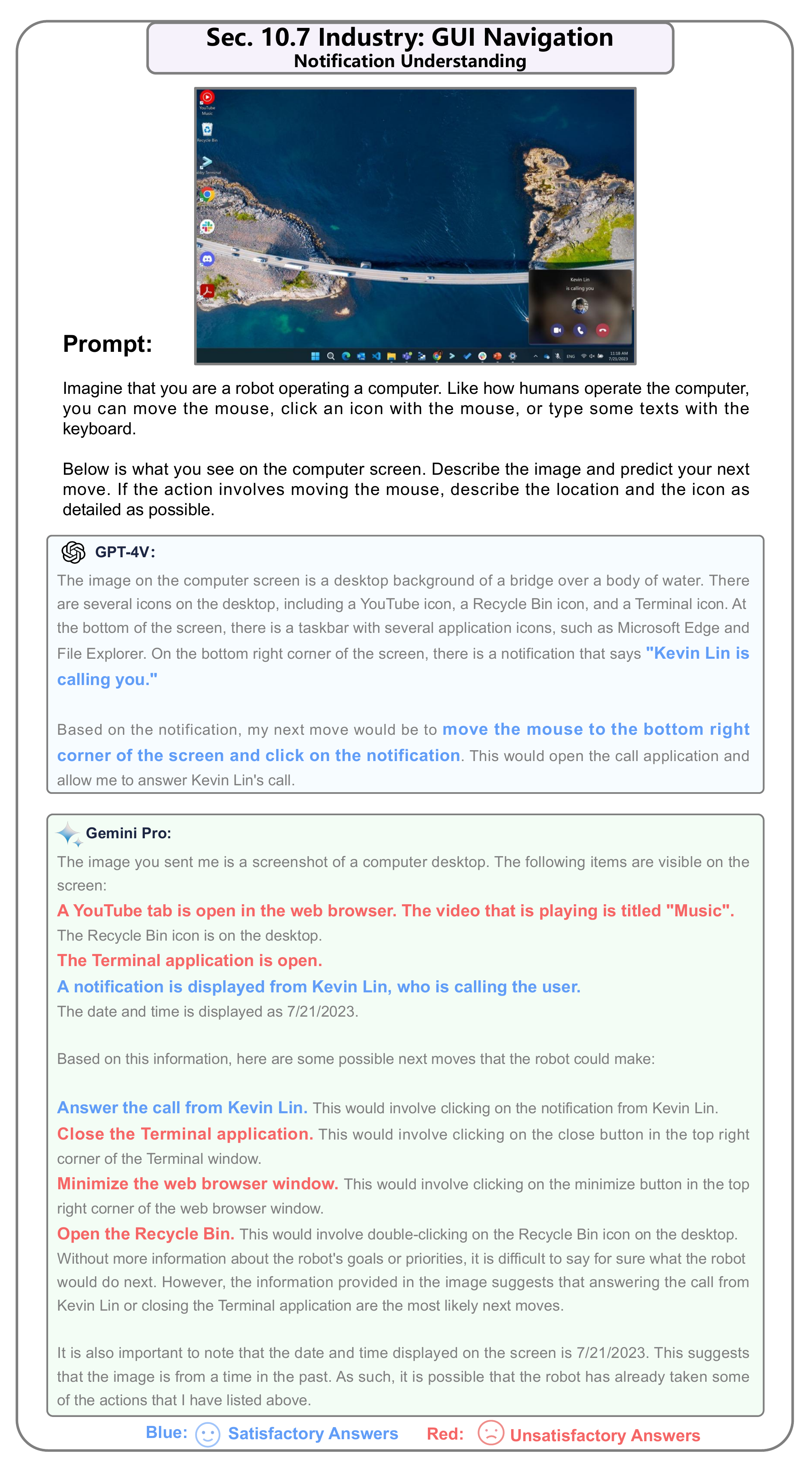}
    \vspace{-3pt}
    \caption[Section 10.7 Industry: GUI Navigation (16)]{\textbf{GUI navigation - notification understanding (2/2).} GPT-4V provides correct descriptions of the GUI along with reasonable operational instructions. However, Gemini misinterprets the GUI, wrongly assuming that a web browser window and a terminal are open, leading to some incorrect directions.}
    \label{fig:Sec.10.7_16}
    \vspace{-10pt}
\end{figure}

\begin{figure}[H]
    \centering
    \vspace{-12pt}
    \includegraphics[width=0.99\linewidth]{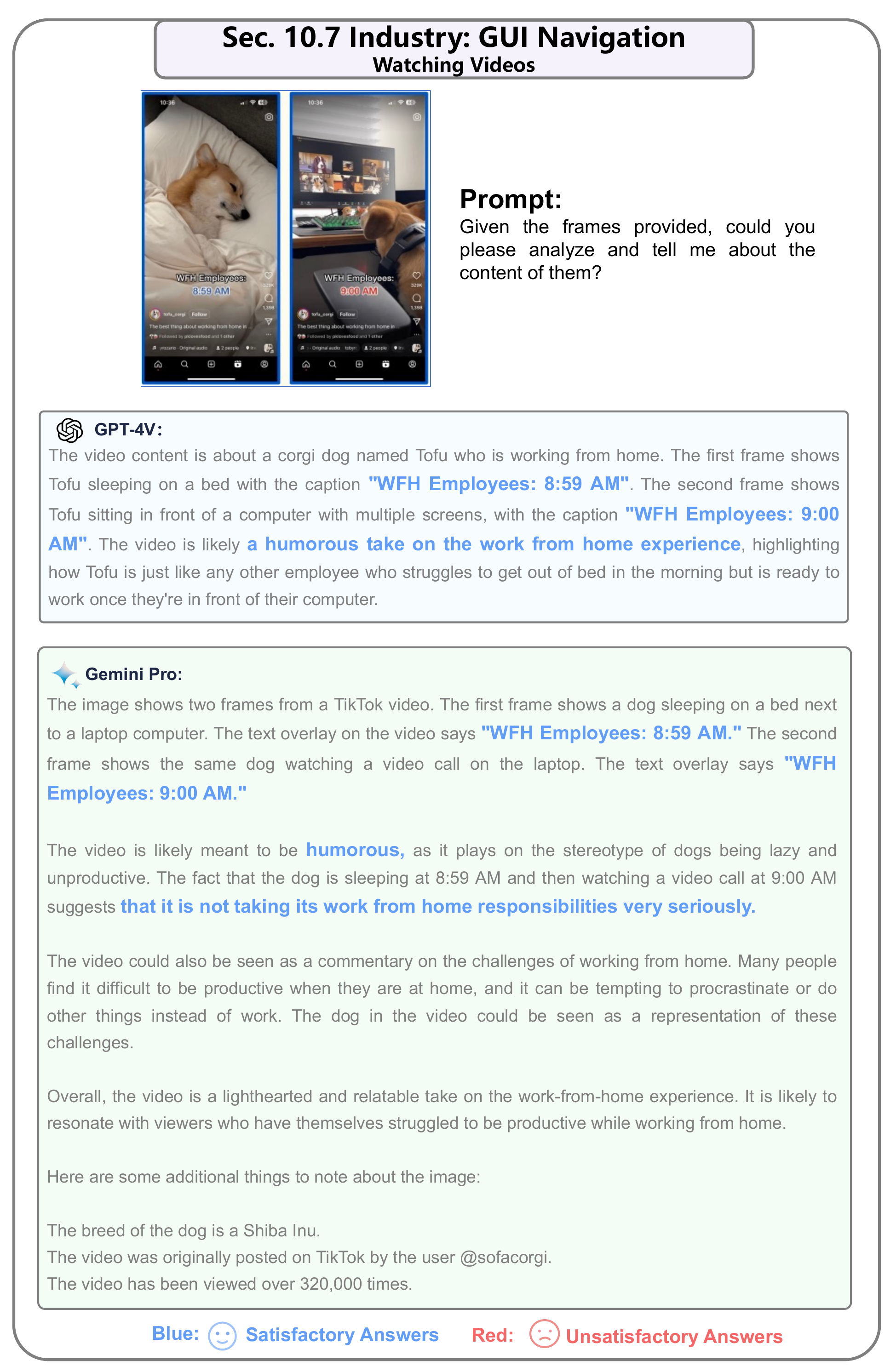}
    \vspace{-6pt}
    \caption[Section 10.7 Industry: GUI Navigation (17)]{\textbf{GUI navigation - watch videos.} Both GPT-4V and Gemini accurately describe the information and content provided by two frames of the video. Their analyses and explanations are not identical, but each makes sense in their own way. In addition to the video content, Gemini also provides some other information offered by the GUI.
    }
    \label{fig:Sec.10.7_17}
    \vspace{-10pt}
\end{figure}

\Cref{fig:Sec.10.7_1} to \cref{fig:Sec.10.7_5} depict the process of conducting a web search for a Mapo Tofu recipe. Throughout this task, both models exhibit comparable proficiency, with Gemini showing a tendency to extract more detailed information from web pages.
\Cref{fig:Sec.10.7_6} to \cref{fig:Sec.10.7_14} illustrate an example of using a mobile phone to order a specific keyboard. In this scenario, GPT-4V demonstrates a slight edge in performance.
Figures \cref{fig:Sec.10.7_15} to \Cref{fig:Sec.10.7_16} focus on the process of receiving notifications, where Gemini is capable of providing a wider range of options.
\Cref{fig:Sec.10.7_17} showcases the ability to view short videos, with both models performing admirably in this respect.

% 10.1 Industry: Defect Detection
\newpage

\newpage 

\newpage 

% 10.2 Industry: Grocery Checkout
\newpage 
\vspace*{\fill}
\vfill

% 10.3 Industry: Auto Insurance
\newpage 

\newpage 

\newpage 
\vspace*{\fill}
\vfill

% 10.4 Industry: Customized Captioner
\newpage 
\vspace*{\fill}
\vfill

% 10.5 Industry: Evaluation Image Generation
\newpage 
\vspace*{\fill}
\vfill

\newpage 
\vspace*{\fill}
\vfill

\newpage 
\vspace*{\fill}
\vfill

% 10.6 Industry: Embodied Agent
\newpage 
\vspace*{\fill}
\vfill

\newpage 
\vspace*{\fill}
\vfill

\newpage
\vspace*{\fill}
\vfill

\newpage 
\vspace*{\fill}
\vfill

% 10.7 Industry: GUI Navigation
\newpage 
\vspace*{\fill}
\vfill

\newpage 
\vspace*{\fill}
\vfill

\newpage 
\vspace*{\fill}
\vfill

\newpage 
\vspace*{\fill}
\vfill

\newpage 
\vspace*{\fill}
\vfill

\newpage 
\vspace*{\fill}
\vfill

\newpage 
\vspace*{\fill}
\vfill

\newpage 
\vspace*{\fill}
\vfill

\newpage 
\vspace*{\fill}
\vfill

\newpage 
\vspace*{\fill}
\vfill

\newpage 
\vspace*{\fill}
\vfill

\newpage 
\vspace*{\fill}
\vfill

\newpage 
\vspace*{\fill}
\vfill

\newpage 
\vspace*{\fill}
\vfill

\newpage 
\vspace*{\fill}
\vfill

\newpage 
\vspace*{\fill}
\vfill

\newpage 
\vspace*{\fill}
\vfill

\newpage
\section{Integrated Use of GPT-4V and Gemini}
\label{Sec.11 Integrated Use of GPT-4V and Gemini}
In order to unlock greater potential in both models, this chapter delves into the exploration of a novel paradigm that integrates the functionality of two distinct models. Specifically, through prior investigations, it has been observed that in certain complex image contexts, GPT-4V exhibits more accurate and comprehensive recognition capabilities. On the other hand, Gemini excels in retrieval tasks, extending beyond text generation to provide users with recommendations for links to similar objects on the web. Additionally, Gemini tends to generate longer responses, affording advantages in certain creative scenarios. The amalgamation of GPT-4V and Gemini characteristics is considered, and the efficacy of their combined utilization is explored in two distinct scenarios.

\subsection{Product Identification and Recommendation}
\label{Sec.11.1 Product Identification and Recommendation}
\begin{figure}[H]
    \centering
    \vspace{-12pt}
    \includegraphics[width=0.94\linewidth]{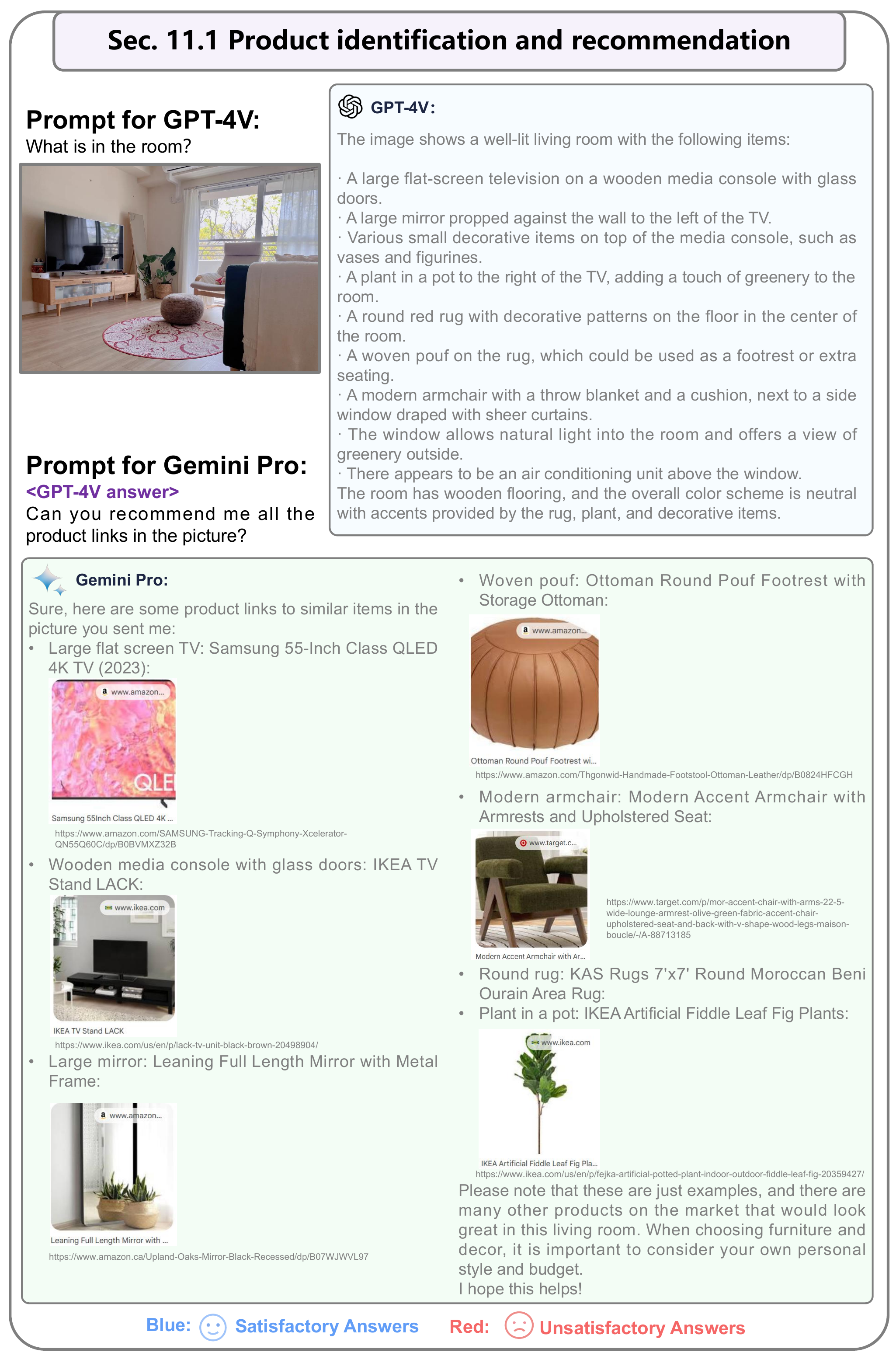}
    \vspace{-3pt}
    \caption[Section 11.1 Integrated Use: Product Identification and Recommendation]{\textbf{Product identification and recommendation.}
    We first use GPT-4V to describe all objects in the picture, then use GPT-4V’s answers as part of the input and use Gemini to recommend links to similar products.}
    \label{fig:Sec.11.1_1}
    \vspace{-5pt}
\end{figure}

\cref{fig:Sec.11.1_1} illustrates the application of the intergrated use of two models in the context of product identification and recommendation. Initially, we employ GPT-4V to recognize all objects within the scene and generate concise textual descriptions corresponding to each. Subsequently, these descriptions, along with the images, are input into Gemini, enabling Gemini to provide relevant links for all products in the image. GPT-4V excels in generating accurate and comprehensive object descriptions. With textual guidance in place, Gemini leverages the combined information of text and images for retrieval and recommendation, thereby accurately obtaining recommended links for as many objects as possible within the scene.

Furthermore, we investigate the individual efficacy of the two models in this scenario. GPT-4V struggles to provide recommended links for products, while Gemini, in the initial phase, faces challenges in comprehensively identifying objects. The fusion of both models maximizes their strengths, allowing the model to achieve optimal performance.

\subsection{Multi-Image Recognition and Story Generation}
\label{Sec.11.2 Multi-image Recognition and Story Generation}
\begin{figure}[H]
    \centering
    \vspace{-12pt}
    \includegraphics[width=0.92\linewidth]{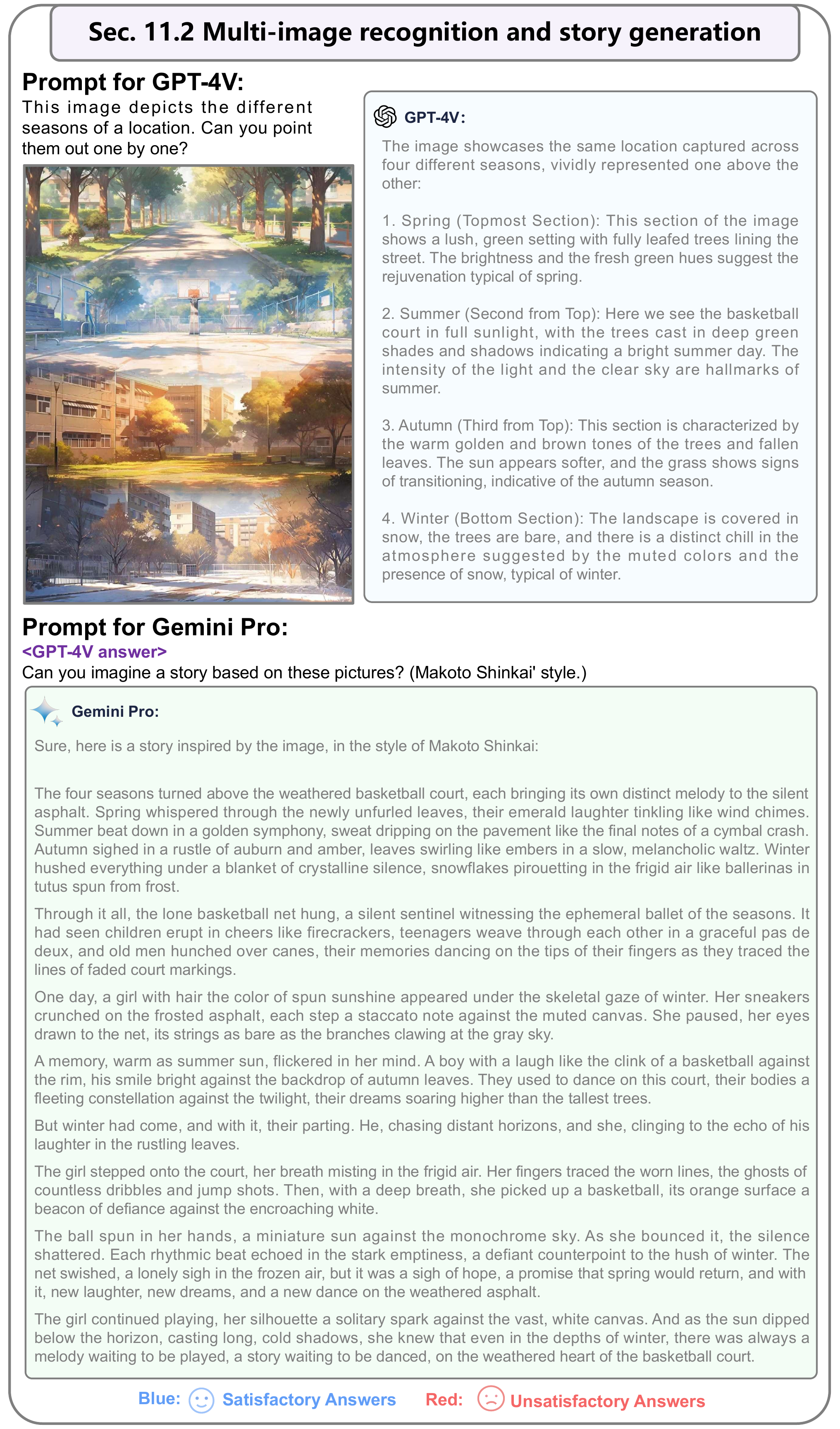}
    \vspace{-3pt}
    \caption[Section 11.2 Integrated Use: Multi-image Recognition and Story Generation]{\textbf{Multi-image recognition and story generation.} We first use GPT-4V to describe all the scenes contained in the picture, and then use Gemini to generate a long story of a specific style.}
    \label{fig:Sec.11.2_1}
    \vspace{-3pt}
\end{figure}

\cref{fig:Sec.11.2_1} showcases the collaborative effects in a creative scenario. We input a complex composite scene image containing multiple sub-images. Initially, GPT-4V is employed to provide a summary of the image, followed by the utilization of Gemini to generate a narrative with a specified style. The results reveal that GPT-4V accurately summarizes and identifies the content of the four sub-images, while Gemini generates a coherent narrative with extended length, aligning with both the content and stylistic requirements of the image. In contrast, the use of a singular model may result in unclear content summarization or the generation of overly formatted and brief narrative sequences.

\vspace{0.5cm}

Beyond the emerging domains discussed above where the fusion of two models can be applied, we anticipate that the amalgamation of their characteristics holds the potential to unveil additional meaningful and intriguing application scenarios. This not only reflects the potential optimization directions or developmental trajectories for the two models in the future but also underscores our eagerness for further exploration. We look forward to delving into more uncharted territories where the synergistic integration of these models can contribute to novel and impactful advancements.

% 11.1 Product identification and recommendation
\newpage
\vspace*{\fill}
\vfill

% 11.2 Multi-image recognition and story generation
\newpage 
\vspace*{\fill}
\vfill

\newpage
\section{Conclusion}
\label{Sec.12 Conclusion}
In our study, we conducted a comprehensive comparison of the multimodal understanding and reasoning capabilities of GPT-4V and Gemini. Both models performed well in basic image recognition tasks, but there were some differences in text recognition and understanding, especially in complex formulas and table information processing. In image inference and emotional testing, both models were capable of understanding and expressing various emotions, though Gemini was slightly behind GPT-4V in IQ tests and object combinations. In integrated image-text understanding tasks, Gemini, due to its inability to process multiple image inputs, was outperformed by GPT-4V in some aspects, although it matched GPT-4V in text reasoning with single images. In industrial applications, particularly in tasks involving embodied agents and GUI navigation, Gemini also fell short of GPT-4V. Combining two large models can leverage their respective strengths. Overall, while both are strong multimodal large models, GPT-4V slightly outperforms Gemini Pro in several areas. We look forward to the release of Gemini Ultra and GPT-4.5, which are expected to bring more possibilities to the field of visual multimodal applications.

%%%%%%%%% REFERENCE %%%%%%%%%
{
  \small
  \bibliographystyle{ieee_fullname}
  %\bibliography{main}

}

\end{CJK} % End chinese
%%%%%%%%% END %%%%%%%%%
\end{document}